\documentclass{book}
\usepackage{roboto}

\usepackage[english]{babel}
\usepackage[letterpaper,top=2cm,bottom=2cm,left=3cm,right=3cm,marginparwidth=1.75cm]{geometry}

\usepackage{amsfonts}
\UseRawInputEncoding

\usepackage{pdflscape}
\usepackage{rotating}
\usepackage{longtable}
\usepackage{array}
\usepackage{float}
\usepackage{amsmath}
\usepackage{graphicx}
\usepackage{inconsolata}
\usepackage[colorlinks=true, allcolors=blue]{hyperref}
\usepackage{enumitem}
\usepackage{tikz} 
\usepackage{listings}
\usepackage{xcolor}
\usepackage[T1]{fontenc}
\usepackage{IEEEtrantools}  
\usepackage{epigraph}  

\definecolor{bgcolor}{rgb}{0.97,0.97,0.97}
\definecolor{codeblue}{rgb}{0.1,0.1,0.8}
\definecolor{codegreen}{rgb}{0,0.4,0}
\definecolor{codegray}{rgb}{0.4,0.4,0.4}
\definecolor{codepurple}{rgb}{0.5,0,0.5}
\definecolor{codered}{rgb}{0.6,0.2,0.2}
\definecolor{lightgray}{rgb}{0.9,0.9,0.9}
\definecolor{darkgray}{rgb}{0.6,0.6,0.6} 

\makeatletter
\renewcommand{\paragraph}{%
  \@startsection{paragraph}{4}{\z@}{1ex}{-1em}{\normalfont\normalsize\bfseries\color{gray}}}
\makeatother

\lstdefinestyle{python}{
    language=Python,
    basicstyle=\ttfamily\small\color{black}\usefont{T1}{zi4}{m}{n},  
    keywordstyle=\bfseries\color{codeblue},  
    stringstyle=\color{codegreen},  
    commentstyle=\slshape\color{codegray},  
    showstringspaces=false,
    numbers=left,
    numberstyle=\tiny\color{codegray},  
    stepnumber=1,
    numbersep=8pt,
    frame=single,
    rulecolor=\color{darkgray},  
    breaklines=true,
    backgroundcolor=\color{bgcolor},
    tabsize=4,
    captionpos=b,
    morekeywords={self}, 
}

\lstdefinestyle{cmd}{
    language=bash,
    basicstyle=\ttfamily\small\color{black}\usefont{T1}{zi4}{m}{n},  
    keywordstyle=\bfseries\color{blue},
    stringstyle=\color{codegreen},
    commentstyle=\itshape\color{gray},
    showstringspaces=false,
    numbers=none,
    frame=single,
    rulecolor=\color{darkgray},  
    breaklines=true,
    backgroundcolor=\color{bgcolor},
    tabsize=4,
    captionpos=b,
}

\title{Deep Learning, Machine Learning - Digital Signal and Image Processing: From Theory to Application}
\author{
    Weiche Hsieh\textsuperscript{*} \\
    \textit{National Tsing Hua University} \\
    s112033645@m112.nthu.edu.tw
    \and
    Ziqian Bi\textsuperscript{*$\dagger$} \\
    \textit{Indiana University} \\
    bizi@iu.edu
    \and
    Junyu Liu \\ 
    \textit{Kyoto University} \\
    liu.junyu.82w@st.kyoto-u.ac.jp
    \and
    Benji Peng \\ 
    \textit{AppCubic} \\
    benji@appcubic.com
    \and
    Sen Zhang \\ 
    \textit{Rutgers University} \\
    sen.z@rutgers.edu
    \and
    Xuanhe Pan \\ 
    \textit{University of Wisconsin-Madison} \\
    xpan73@wisc.edu
    \and
    Jiawei Xu \\ 
    \textit{Purdue University} \\
    xu1644@purdue.edu
    \and
    Jinlang Wang \\ 
    \textit{University of Wisconsin-Madison} \\
    jinlang.wang@wisc.edu
    \and
    Keyu Chen\\ 
    \textit{Georgia Institute of Technology} \\
    kchen637@gatech.edu
    \and
    Caitlyn Heqi Yin \\
    \textit{University of Wisconsin-Madison} \\
    hyin66@wisc.edu
    \and
    Pohsun Feng \\
    \textit{National Taiwan Normal University} \\
    41075018h@ntnu.edu.tw
    \and
    Yizhu Wen \\
    \textit{University of Hawaii} \\
    yizhuw@hawaii.edu
    \and
    Tianyang Wang \\ 
    \textit{Xi'an Jiaotong-Liverpool University} \\
    Tianyang.Wang21@student.xjtlu.edu.cn
    \and
    Ming Li \\ 
    \textit{Gatech} \\
    mli694@gatech.edu
    \and
    Jintao Ren \\
    \textit{Aarhus University } \\
    jintaoren@clin.au.dk
    \and
    Xinyuan Song \\
    \textit{Emory University } \\
    xsong30@emory.edu
    \and
    Qian Niu \\ 
    \textit{Kyoto University} \\
    niu.qian.f44@kyoto-u.jp
    \and
    Silin Chen \\
    \textit{Zhejiang University } \\
    A1033439225@gmail.com
    \and
    Ming Liu{$\dagger$} \\ 
    \textit{Purdue University} \\
    liu3183@purdue.edu
}

\date{} 

\begin{document}

\maketitle

\begingroup
\renewcommand\thefootnote{}\footnote{
    \textsuperscript{*} Equal contribution \\
    \textsuperscript{$\dagger$} Corresponding author
}
\addtocounter{footnote}{0}
\endgroup

\epigraph{"The computer was born to solve problems that did not exist before."}{\textit{Bill Gates}}

\epigraph{"Design is where science and art break even."}{\textit{Robin Matthews}}

\epigraph{"Computers are good at following instructions, but not at reading your mind."}{\textit{Donald Knuth}}

\epigraph{"A good programmer is someone who always looks both ways before crossing a one-way street."}{\textit{Doug Linder}}

\epigraph{"The spread of computers and the Internet will put jobs in two categories. People who tell computers what to do, and people who are told by computers what to do."}{\textit{Marc Andreessen}}

\tableofcontents  

\part{Digital Signal Processing}

\chapter{Introduction and Fundamental Concepts}
    \section{Definition and Applications of Digital Signal Processing}
        Digital Signal Processing (DSP) refers to the manipulation and analysis of signals after they have been converted to digital form. Signals are a form of data that vary over time, and they can be represented mathematically to convey information.\cite{oppenheim2010dsp} In DSP, we process these signals using algorithms and computational methods in digital computers or specialized processors. 

        Applications of DSP are vast and span across many fields \cite{proakis1996digital}:
        \begin{itemize}
            \item \textbf{Communications}: DSP is essential in the modulation and demodulation of signals for transmitting data over long distances, such as in mobile phones, radios, and satellites \cite{proakis1996digital}.
            \item \textbf{Audio and Speech Processing}: DSP is used in applications like noise cancellation, audio compression, speech recognition, and digital music production \cite{mitra2002digital, smith1997scientist}.
            \item \textbf{Radar and Sonar}: DSP helps in the detection and identification of objects by analyzing reflected signals in radar and sonar systems \cite{richards2014fundamentals}.
            \item \textbf{Biomedical Signal Processing}: DSP techniques are applied to analyze signals like ECG (Electrocardiogram), EEG (Electroencephalogram), and medical imaging, improving diagnostics and treatment \cite{cohen2019biomedical}.
            \item \textbf{Image Processing}: DSP is key in operations such as image enhancement, filtering, compression, and feature extraction in both medical imaging and general photography \cite{gonzalez2002digital}.
        \end{itemize}
        The primary advantages of DSP over analog signal processing include:
        \begin{itemize}
            \item \textbf{Precision}: Digital systems offer high precision because they use binary numbers, which are less affected by noise and distortion \cite{oppenheim2010dsp, smith1997scientist}.
            \item \textbf{Reusability}: Digital systems are programmable, so algorithms can be reused and modified easily without changing hardware \cite{proakis1996digital}.
            \item \textbf{Flexibility}: DSP allows the implementation of sophisticated algorithms that may be impractical with analog systems.\cite{mitra2002digital}
            \item \textbf{Complexity}: Complex algorithms such as adaptive filtering, Fourier transforms, and machine learning techniques can be easily applied in DSP \cite{oppenheim2010dsp}.
        \end{itemize}

    \section{Types of Signals: Discrete and Continuous}
        A fundamental concept in DSP is the distinction between \textbf{continuous-time} and \textbf{discrete-time} signals \cite{oppenheim1997signals}.

        \subsection{Continuous-Time Signals}
            Continuous-time signals are defined for every point in time. These signals are described mathematically as functions of a continuous variable (usually time), such as:
            \[
            x(t) = \sin(2\pi f t) 
            \]
            where \( t \) represents time, \( f \) is the frequency, and the signal is a sine wave \cite{bracewell2000fourier}. Examples of continuous-time signals include audio signals, temperature measurements, and light intensity variations.

            \textbf{Example 1: Sine Wave}
            A sine wave with a frequency of 5 Hz (cycles per second) is given by:
            \[
            x(t) = \sin(10\pi t)
            \]
            This signal is defined for all \( t \in \mathbb{R} \), meaning it exists at every point in time \cite{oppenheim1997signals}. In practical terms, this could represent an audio tone or a voltage signal.

            \textbf{Example 2: Exponential Signal}
            An exponentially decaying signal can be described as:
            \[
            x(t) = e^{-\alpha t}
            \]
            where \( \alpha > 0 \). This type of signal might represent the charging or discharging of a capacitor in an electrical circuit \cite{oppenheim2010dsp}.

        \subsection{Discrete-Time Signals}
            Discrete-time signals are defined only at specific time intervals. These signals are typically the result of sampling a continuous-time signal, or they may be generated directly in a digital system, such as in digital communications or sensor outputs. A discrete-time signal can be expressed as:
            \[
            x[n] = x(nT_s)
            \]
            where \( n \) is an integer representing the time index, and \( T_s \) is the sampling period (the time between samples).

            \textbf{Example 1: Discrete Sine Wave}
            A sampled version of the sine wave mentioned earlier could be represented as:
            \[
            x[n] = \sin(2\pi f n T_s)
            \]
            where \( f \) is the frequency in Hz, and \( T_s \) is the sampling interval \cite{oppenheim2010dsp}.

            \textbf{Example 2: Sequence of Numbers}
            A discrete signal can also be a sequence of numbers, such as sensor readings:
            \[
            x[n] = \{2, 3, 5, 1, 0, 4, 3\}
            \] \cite{proakis1996digital}
            This could represent temperature measurements at hourly intervals, for example.

    \section{Python Example: Plotting Signals}
        In Python, we can easily visualize both continuous and discrete-time signals using libraries like \texttt{matplotlib} and \texttt{numpy}. Here, we will plot a continuous sine wave and its sampled version (discrete signal) \cite{hunter2007matplotlib,van2011numpy}.

        \subsection{Plotting a Continuous Sine Wave}
        The following Python code demonstrates how to plot a continuous sine wave.
        \begin{lstlisting}[style=python]
import numpy as np
import matplotlib.pyplot as plt

# Define parameters for the sine wave
f = 5  # frequency in Hz
t = np.linspace(0, 1, 1000)  # time vector from 0 to 1 second
x_continuous = np.sin(2 * np.pi * f * t)

# Plot the continuous sine wave
plt.figure()
plt.plot(t, x_continuous)
plt.title('Continuous Sine Wave')
plt.xlabel('Time [s]')
plt.ylabel('Amplitude')
plt.grid(True)
plt.show()
        \end{lstlisting}

        \subsection{Plotting a Discrete-Time Sine Wave}
        Now let's sample the same sine wave at a sampling rate of 20 Hz and plot the discrete-time signal.

        \begin{lstlisting}[style=python]
# Sampling parameters
fs = 20  # Sampling frequency in Hz
n = np.arange(0, 1, 1/fs)  # time index for discrete signal
x_discrete = np.sin(2 * np.pi * f * n)

# Plot the discrete sine wave
plt.figure()
plt.stem(n, x_discrete, basefmt=" ", use_line_collection=True)
plt.title('Discrete-Time Sine Wave')
plt.xlabel('Time [s]')
plt.ylabel('Amplitude')
plt.grid(True)
plt.show()
        \end{lstlisting}

    \section{From Continuous to Discrete: The Sampling Process}
        The process of converting a continuous-time signal into a discrete-time signal is called \textbf{sampling}. Sampling is done at regular intervals, determined by the \textbf{sampling frequency} or \textbf{sampling rate}, denoted by \( f_s \), which is measured in Hertz (Hz).

        According to the \textbf{Nyquist-Shannon Sampling Theorem}, in order to accurately represent a continuous-time signal in discrete form, the sampling frequency must be at least twice the highest frequency component in the signal \cite{shannon1949communication, oppenheim2010dsp}:
        \[
        f_s \geq 2f_{max}
        \]
        where \( f_{max} \) is the maximum frequency present in the signal. This condition is known as the \textbf{Nyquist rate}.

        If the sampling rate is too low (i.e., below the Nyquist rate), a phenomenon called \textbf{aliasing} occurs, where different frequency components become indistinguishable from each other in the sampled data \cite{oppenheim2010dsp}.

        \textbf{Example: Sampling and Aliasing}
        Suppose we sample a 5 Hz sine wave at a rate of 6 Hz. The Nyquist rate requires the sampling rate to be at least 10 Hz, so the 6 Hz sampling rate will result in aliasing, making the sampled signal appear to have a different frequency.

        To demonstrate this in Python, we can sample the sine wave at an insufficient rate and compare the result.

        \begin{lstlisting}[style=python]
# Sampling below the Nyquist rate
fs_aliasing = 6  # Sampling frequency below Nyquist rate
n_aliasing = np.arange(0, 1, 1/fs_aliasing)
x_aliasing = np.sin(2 * np.pi * f * n_aliasing)

# Plot the aliased sine wave
plt.figure()
plt.stem(n_aliasing, x_aliasing, basefmt=" ", use_line_collection=True)
plt.title('Aliasing Example: Discrete-Time Sine Wave with fs < 2f')
plt.xlabel('Time [s]')
plt.ylabel('Amplitude')
plt.grid(True)
plt.show()
        \end{lstlisting}

\section{Basic Signal Representation}
    In this section, we will cover some of the most fundamental signals used in Digital Signal Processing (DSP). These signals form the building blocks for more complex operations and help us understand how systems respond to different inputs. The signals that will be introduced include the unit impulse signal, the unit step signal, sine waves, and complex exponential signals \cite{oppenheim2010dsp,proakis1996digital}.

    \subsection{Unit Impulse Signal}
        The unit impulse signal, often referred to as the "Dirac delta function" in continuous-time systems, is a simple but powerful concept in DSP. In discrete time, the unit impulse signal is denoted by \(\delta[n]\), and it is defined as:

        \[
        \delta[n] = 
        \begin{cases} 
        1 & \text{for } n = 0, \\
        0 & \text{for } n \neq 0.
        \end{cases}
        \]

        The unit impulse signal is important because it can be used to represent any discrete-time signal using the principle of superposition. Moreover, it is central to the analysis of systems in the time domain because the response of a system to \(\delta[n]\) is known as the impulse response \cite{oppenheim2010dsp}.

        \paragraph{Example:} Let's generate a unit impulse signal using Python with a length of 10 samples:

        \begin{lstlisting}[style=python]
        import numpy as np
        import matplotlib.pyplot as plt

        n = np.arange(-5, 5, 1)
        delta = np.zeros_like(n)
        delta[5] = 1  # Set delta[0] to 1, corresponding to n=0

        plt.stem(n, delta, use_line_collection=True)
        plt.title('Unit Impulse Signal')
        plt.xlabel('n')
        plt.ylabel('δ[n]')
        plt.grid(True)
        plt.show()
        \end{lstlisting}

        This will create a plot where the unit impulse occurs at \(n = 0\), and all other values are zero.

    \subsection{Unit Step Signal}
        The unit step signal, denoted as \(u[n]\), is another fundamental signal in DSP. It is defined as:

        \[
        u[n] = 
        \begin{cases} 
        1 & \text{for } n \geq 0, \\
        0 & \text{for } n < 0.
        \end{cases}
        \]

        The unit step signal is useful for analyzing causal systems, where the output depends only on the current and past inputs, not on future inputs \cite{oppenheim2010dsp}.

        \paragraph{Example:} Let's generate a unit step signal using Python:

        \begin{lstlisting}[style=python]
        n = np.arange(-5, 5, 1)
        step = np.heaviside(n, 1)  # Create the unit step signal

        plt.stem(n, step, use_line_collection=True)
        plt.title('Unit Step Signal')
        plt.xlabel('n')
        plt.ylabel('u[n]')
        plt.grid(True)
        plt.show()
        \end{lstlisting}

        This code will plot a step signal where the values are 0 for \(n < 0\) and 1 for \(n \geq 0\).

    \subsection{Sine and Complex Exponential Signals}
        Sine waves and complex exponential signals are fundamental in frequency analysis in DSP. A discrete-time sine wave can be described by:

        \[
        x[n] = A \sin(2\pi f n + \phi)
        \]

        where:
        \begin{itemize}
            \item \(A\) is the amplitude,
            \item \(f\) is the frequency (in cycles per sample),
            \item \(\phi\) is the phase.
        \end{itemize}

        The complex exponential signal is a more general form that includes sine and cosine signals as special cases. It is expressed as:

        \[
        x[n] = A e^{j(2\pi f n + \phi)}
        \]

        where \(j\) is the imaginary unit. This form is important because many DSP operations are more easily understood in the frequency domain, where signals are represented as sums of complex exponentials \cite{oppenheim2010dsp}.

        \paragraph{Example:} Let's generate a sine wave signal using Python:

        \begin{lstlisting}[style=python]
        A = 1  # Amplitude
        f = 0.1  # Frequency in cycles per sample
        phi = 0  # Phase in radians
        n = np.arange(0, 100, 1)  # 100 samples

        sine_wave = A * np.sin(2 * np.pi * f * n + phi)

        plt.plot(n, sine_wave)
        plt.title('Sine Wave Signal')
        plt.xlabel('n')
        plt.ylabel('Amplitude')
        plt.grid(True)
        plt.show()
        \end{lstlisting}

        This code generates and plots a discrete-time sine wave.

\section{Properties of Linear Time-Invariant (LTI) Systems}
    LTI systems are foundational in DSP because they simplify analysis and design. Systems with these properties behave in predictable ways, allowing for powerful tools like convolution and the Fourier transform to be used effectively \cite{oppenheim2010dsp, proakis1996digital}.

    \subsection{Linearity}
        A system is said to be linear if it satisfies the principle of superposition. This means that for any two input signals \(x_1[n]\) and \(x_2[n]\), and for any scalars \(a\) and \(b\), the response of the system to the input \(a x_1[n] + b x_2[n]\) is given by:

        \[
        y[n] = a y_1[n] + b y_2[n]
        \]

        where \(y_1[n]\) and \(y_2[n]\) are the responses of the system to \(x_1[n]\) and \(x_2[n]\), respectively. This property makes it easy to analyze systems when the inputs are a combination of simpler signals \cite{oppenheim2010dsp}.

        \paragraph{Example:} Consider a system where the output \(y[n]\) is the sum of the current and previous input values:

        \[
        y[n] = x[n] + x[n-1]
        \]

        This system is linear because the output for a combination of two inputs will also be a linear combination of the corresponding outputs.

    \subsection{Time Invariance}
        Time invariance implies that a shift in the input signal results in an identical shift in the output signal. If an LTI system's response to an input \(x[n]\) is \(y[n]\), then the response to \(x[n - n_0]\) (a time-shifted input) will be \(y[n - n_0]\) (a time-shifted output) \cite{oppenheim2010dsp}.

        \paragraph{Example:} Suppose the system output is delayed by 2 samples relative to the input, i.e., \(y[n] = x[n-2]\). This system is time-invariant because if the input is shifted by any amount, the output will be shifted by the same amount.

    \subsection{Causality and Stability}
        \paragraph{Causality:} A system is causal if its output at any time depends only on current and past inputs, not on future inputs. Mathematically, a system is causal if \cite{proakis1996digital}:

        \[
        y[n] \text{ depends only on } x[k] \text{ for } k \leq n.
        \]

        \paragraph{Stability:} A system is stable if it produces bounded outputs for all bounded inputs. In other words, if the input signal is bounded, the output signal must also be bounded. This can be written as \cite{oppenheim2010dsp}:

        \[
        |x[n]| \leq M \implies |y[n]| \leq N
        \]

        where \(M\) and \(N\) are constants.

        \paragraph{Example:} The system \(y[n] = 0.5 y[n-1] + x[n]\) is causal because the output at time \(n\) depends only on the current and previous inputs. It is also stable as long as the feedback coefficient (0.5 in this case) is less than 1, ensuring that the output does not grow unbounded.

\section{Convolution and Correlation}
    \subsection{Discrete-Time Convolution}
        Convolution is a fundamental operation in signal processing, particularly in the analysis of Linear Time-Invariant (LTI) systems. The convolution of two signals, typically an input signal \(x[n]\) and the impulse response \(h[n]\) of a system, results in an output signal \(y[n]\). In Python, we can easily perform convolution using arrays that represent discrete signals.
        
        The formula for discrete-time convolution is:
        \[
        y[n] = x[n] * h[n] = \sum_{k=-\infty}^{\infty} x[k] h[n-k]
        \]
        Here, \(x[k]\) is the input signal, \(h[n-k]\) is the system's response, and the sum is computed over all values of \(k\). The result \(y[n]\) is a new signal that represents the combined effect of \(x[n]\) passing through the system with response \(h[n]\).

        \textbf{Example:}
        
        Let's say we have two signals \(x[n]\) and \(h[n]\):
        \[
        x[n] = \{1, 2, 3\}
        \]
        \[
        h[n] = \{0, 1, 0.5\}
        \]
        The convolution of these two signals can be computed step by step using Python. We'll implement this in PyTorch for clarity.

        \begin{lstlisting}[style=python]
        import torch

        # Input signals
        x = torch.tensor([1, 2, 3], dtype=torch.float32)
        h = torch.tensor([0, 1, 0.5], dtype=torch.float32)

        # Perform convolution using torch.conv1d
        x = x.view(1, 1, -1)  # Reshape for conv1d (batch_size=1, channels=1)
        h = h.view(1, 1, -1)

        # Convolution
        y = torch.nn.functional.conv1d(x, h, padding=2)  # Apply convolution
        print(y.view(-1))  # Reshape to a 1D tensor
        \end{lstlisting}

        In this code, we used PyTorch's built-in functions to perform 1D convolution. The padding ensures that the output size matches the input size.

    \subsection{Properties of Convolution}
        Convolution has several important properties that make it useful in signal processing. These properties include:

        \textbf{1. Commutativity:}
        \[
        x[n] * h[n] = h[n] * x[n]
        \]
        This property means that the order of the signals being convolved does not matter.

        \textbf{2. Associativity:}
        \[
        x[n] * (h[n] * g[n]) = (x[n] * h[n]) * g[n]
        \]
        This allows us to convolve multiple signals in any grouping.

        \textbf{3. Distributivity:}
        \[
        x[n] * (h_1[n] + h_2[n]) = x[n] * h_1[n] + x[n] * h_2[n]
        \]
        This property shows that convolution distributes over addition, which simplifies many signal processing tasks.

    \subsection{Correlation}
        Correlation is another key concept in signal processing, used to measure the similarity between two signals. It is closely related to convolution, but instead of sliding one signal over the other as in convolution, correlation compares the signals as they are \cite{oppenheim2010dsp, proakis1996digital}.

        There are two types of correlation:
        
        \textbf{1. Cross-Correlation:}
        Cross-correlation compares how similar one signal is to a time-shifted version of another signal. For two signals \(x[n]\) and \(y[n]\), the cross-correlation \(r_{xy}[n]\) is defined as:
        \[
        r_{xy}[n] = \sum_{k=-\infty}^{\infty} x[k] y[n+k]
        \]

        \textbf{2. Auto-Correlation:}
        Auto-correlation measures how similar a signal is to a time-shifted version of itself. For a signal \(x[n]\), the auto-correlation \(r_x[n]\) is:
        \[
        r_x[n] = \sum_{k=-\infty}^{\infty} x[k] x[n+k]
        \]

        \textbf{Example of Cross-Correlation:}

        We can compute the cross-correlation of two signals in Python using PyTorch.

        \begin{lstlisting}[style=python]
        # Cross-correlation example
        x = torch.tensor([1, 2, 3], dtype=torch.float32)
        y = torch.tensor([4, 5, 6], dtype=torch.float32)

        # Perform cross-correlation
        cross_corr = torch.nn.functional.conv1d(x.view(1, 1, -1), y.flip(0).view(1, 1, -1))
        print(cross_corr.view(-1))
        \end{lstlisting}

        In this example, we compute the cross-correlation of two signals by flipping one signal and using convolution. Flipping is necessary because convolution flips the signal, but correlation does not.

\section{Sampling Theorem}
    \subsection{Nyquist Sampling Rate}
        The Nyquist-Shannon Sampling Theorem is a foundational concept in digital signal processing. It states that a continuous signal can be perfectly reconstructed from its discrete samples if the sampling rate is at least twice the highest frequency component present in the signal. This critical frequency is known as the \textbf{Nyquist rate}, and sampling below this rate can lead to a phenomenon known as aliasing \cite{oppenheim2010dsp, shannon1949communication}.

        \textbf{Example:}

        If a signal has a maximum frequency of 500 Hz, then the Nyquist rate is:
        \[
        \text{Sampling rate} \geq 2 \times 500 \text{ Hz} = 1000 \text{ Hz}
        \]
        So, the signal must be sampled at least at 1000 Hz to avoid aliasing.

        \subsection{Signal Reconstruction and Anti-Aliasing Filters}
            Once a signal is sampled, it can be reconstructed by passing the samples through a low-pass filter. This filter removes the high-frequency components that could cause aliasing. Before sampling, an \textbf{anti-aliasing filter} is applied to ensure that no frequency components above half the sampling rate are present in the signal \cite{oppenheim2010dsp}.

            \textbf{Example:}
            
            Suppose you are working with audio signals that contain frequencies up to 20 kHz. To avoid aliasing, you should apply an anti-aliasing filter with a cutoff frequency of 20 kHz before sampling. If you sample at 44.1 kHz (the standard rate for audio), the signal can be reconstructed accurately.

            The reconstruction process usually involves using an ideal low-pass filter. In practice, reconstruction is done through digital-to-analog converters (DACs) that include such filters.

\chapter{Discrete-Time Fourier Transform (DTFT)}
    \section{Introduction to Fourier Transform}
        The Fourier Transform is a fundamental mathematical tool used in signal processing to decompose a signal into its constituent sinusoidal components, each with a different frequency, amplitude, and phase \cite{oppenheim2010dsp, bracewell2000fourier}. In the case of continuous signals, we use the \textbf{Continuous Fourier Transform (CFT)}. However, in the digital world, we deal with discrete-time signals, and thus we need to use the \textbf{Discrete-Time Fourier Transform (DTFT)} to analyze the frequency content of such signals.

        Understanding the frequency content of a signal is crucial in many practical applications such as communications, filtering, and system identification. The DTFT provides us with a way to move from the time domain (where the signal is expressed as a function of time) to the frequency domain (where the signal is represented as a function of frequency).

        \textbf{Example: Why Fourier Transform is Important}
        Consider a simple music signal. The Fourier Transform can break it down into various sine waves, each corresponding to different musical notes. Knowing these frequency components allows audio engineers to manipulate individual notes, filter out unwanted noise, or apply sound effects.

    \section{Definition of DTFT}
        The \textbf{Discrete-Time Fourier Transform (DTFT)} of a discrete-time signal \(x[n]\) is defined by the following mathematical expression:
        \[
        X(e^{j\omega}) = \sum_{n=-\infty}^{\infty} x[n] e^{-j\omega n}
        \]
        where:
        \begin{itemize}
            \item \( x[n] \) is the discrete-time signal.
            \item \( \omega \) is the frequency in radians per sample.
            \item \( X(e^{j\omega}) \) is the frequency domain representation of the signal \(x[n]\).
        \end{itemize}

        The DTFT transforms a discrete-time signal \(x[n]\), which is defined only at specific time instances, into a continuous function of frequency \(X(e^{j\omega})\). This gives us insight into how the energy of the signal is distributed across different frequencies \cite{oppenheim2010dsp}.

        \subsection{Key Concepts of DTFT}
        Let's break down the important aspects of this definition step by step:
        \begin{itemize}
            \item \textbf{Complex Exponentials}: In the expression \( e^{-j\omega n} \), we see the use of a complex exponential. This is a key aspect of Fourier Transforms, as they represent sinusoidal functions (sines and cosines) in complex form \cite{oppenheim2010dsp}.
            \item \textbf{Frequency \( \omega \)}: The variable \( \omega \) represents the angular frequency, and it is continuous. For practical purposes, we often normalize it to range between \( -\pi \) and \( \pi \) radians.
            \item \textbf{Summation Over All Time Indices}: The summation is over all possible time indices \( n \), from \( -\infty \) to \( \infty \). This captures the entire signal's behavior in both the past and future \cite{oppenheim2010dsp}.
        \end{itemize}

        \textbf{Example: Basic DTFT Computation}
        Let's compute the DTFT of a simple signal, such as the unit impulse signal, defined as:
        \[
        \delta[n] = 
        \begin{cases} 
            1, & \text{if } n = 0 \\
            0, & \text{otherwise}
        \end{cases}
        \]

        The DTFT of the impulse signal \( \delta[n] \) is calculated as:
        \[
        X(e^{j\omega}) = \sum_{n=-\infty}^{\infty} \delta[n] e^{-j\omega n}
        \]
        Since \( \delta[n] = 0 \) for all \( n \neq 0 \), this reduces to:
        \[
        X(e^{j\omega}) = 1
        \]
        Therefore, the DTFT of the unit impulse is constant across all frequencies \cite{oppenheim2010dsp}.

    \section{Properties of DTFT}
        The DTFT has several important properties that make it useful in signal processing. Understanding these properties allows us to work with complex systems more effectively\cite{oppenheim2010dsp, proakis1996digital}.

        \subsection{Linearity}
        The DTFT is a linear transformation. This means that for two signals \(x_1[n]\) and \(x_2[n]\), and their corresponding DTFTs \(X_1(e^{j\omega})\) and \(X_2(e^{j\omega})\), we have:
        \[
        a x_1[n] + b x_2[n] \quad \longleftrightarrow \quad a X_1(e^{j\omega}) + b X_2(e^{j\omega})
        \]
        where \(a\) and \(b\) are constants. This property is very useful when combining multiple signals or designing systems that process multiple inputs \cite{lyons2010understanding}.

        \subsection{Time Shifting}
        Shifting a signal in time affects its phase in the frequency domain. If \(x[n]\) has a DTFT \(X(e^{j\omega})\), then the time-shifted signal \(x[n-n_0]\) has the DTFT:
        \[
        X(e^{j\omega}) e^{-j\omega n_0}
        \]
        This shows that time shifts correspond to phase shifts in the frequency domain \cite{oppenheim1997signals}.

        \subsection{Frequency Shifting}
        Multiplying a signal by a complex exponential in the time domain results in a frequency shift. If \(x[n]\) has a DTFT \(X(e^{j\omega})\), then the signal \(x[n] e^{j \omega_0 n}\) has the DTFT:
        \[
        X(e^{j(\omega - \omega_0)})
        \]
        This property is used in modulation techniques, where signals are shifted to higher or lower frequencies \cite{bracewell2000fourier}.

        \subsection{Convolution Theorem}
        Convolution in the time domain corresponds to multiplication in the frequency domain. If \(x_1[n]\) and \(x_2[n]\) are two signals with DTFTs \(X_1(e^{j\omega})\) and \(X_2(e^{j\omega})\), then:
        \[
        x_1[n] * x_2[n] \quad \longleftrightarrow \quad X_1(e^{j\omega}) X_2(e^{j\omega})
        \]
        where \( * \) denotes convolution. This is extremely useful in filtering applications \cite{mitra2002digital, gonzalez2002digital}.

    \section{Python Example: DTFT of a Signal}
        In Python, we can compute and visualize the DTFT of a signal using the \texttt{numpy} and \texttt{matplotlib} libraries. Below is a simple example where we compute the DTFT of a rectangular pulse signal.

        \subsection{Step-by-Step Python Example}
        Let's define a rectangular pulse signal \(x[n]\) of length 10, where \(x[n] = 1\) for \(n = 0, 1, 2, \dots, 9\), and 0 otherwise.

        \begin{lstlisting}[style=python]
import numpy as np
import matplotlib.pyplot as plt

# Define the signal x[n]
N = 10  # Length of the rectangular pulse
x = np.ones(N)

# Compute the DTFT using a large number of frequency points
omega = np.linspace(-np.pi, np.pi, 1000)  # Frequency range from -pi to pi
X = np.zeros_like(omega, dtype=complex)

# Compute the DTFT by evaluating the sum for each frequency point
for i, w in enumerate(omega):
    X[i] = np.sum(x * np.exp(-1j * w * np.arange(N)))

# Plot the magnitude and phase of the DTFT
plt.figure(figsize=(12, 6))

# Magnitude plot
plt.subplot(1, 2, 1)
plt.plot(omega, np.abs(X))
plt.title('Magnitude of DTFT')
plt.xlabel('Frequency [rad/sample]')
plt.ylabel('Magnitude')
plt.grid(True)

# Phase plot
plt.subplot(1, 2, 2)
plt.plot(omega, np.angle(X))
plt.title('Phase of DTFT')
plt.xlabel('Frequency [rad/sample]')
plt.ylabel('Phase [radians]')
plt.grid(True)

plt.tight_layout()
plt.show()
        \end{lstlisting}

        In this code:
        \begin{itemize}
            \item We first define the signal \(x[n]\) as a rectangular pulse of length \(N = 10\).
            \item The DTFT is then computed by summing \( x[n] e^{-j\omega n} \) for different values of \( \omega \).
            \item Finally, we plot both the magnitude and phase of the DTFT using \texttt{matplotlib}.
        \end{itemize}

        This example demonstrates how the DTFT transforms a time-domain signal into its frequency-domain representation, allowing us to analyze the signal's frequency content.

    \section{Conclusion}
        The Discrete-Time Fourier Transform (DTFT) is a powerful tool that transforms discrete-time signals into the frequency domain. It provides a detailed frequency analysis, which is essential in many applications like filtering, modulation, and system analysis. By understanding the DTFT, its properties, and how to compute it in Python, beginners in signal processing can start exploring the frequency characteristics of digital signals effectively.

\section{Properties of Discrete-Time Fourier Transform (DTFT)}
    The Discrete-Time Fourier Transform (DTFT) is a fundamental tool in Digital Signal Processing (DSP) that allows us to analyze signals in the frequency domain. The DTFT is particularly useful for periodic signals and systems analysis, providing insight into how signals behave across different frequencies. In this section, we will explore the key properties of the DTFT, such as linearity, time-shifting, and the convolution theorem, and provide detailed examples to make these concepts clear \cite{oppenheim2010dsp, proakis1996digital, mitra2002digital}.

    \subsection{Linearity}
        The linearity property of the DTFT is one of its most useful features, especially when analyzing linear systems. According to the linearity property, the DTFT of a linear combination of two signals is the same linear combination of their individual DTFTs. Mathematically, if you have two discrete-time signals \(x_1[n]\) and \(x_2[n]\), and constants \(a\) and \(b\), the DTFT of the linear combination \(a x_1[n] + b x_2[n]\) is given by:

        \[
        \mathcal{F}\{a x_1[n] + b x_2[n]\} = a X_1(e^{j\omega}) + b X_2(e^{j\omega})
        \]

        where \(X_1(e^{j\omega})\) and \(X_2(e^{j\omega})\) are the DTFTs of \(x_1[n]\) and \(x_2[n]\), respectively \cite{oppenheim2010dsp,bracewell2000fourier}.

        \paragraph{Example:} Let's compute the DTFT of a linear combination of two simple signals, for instance, two cosine waves. First, we will define the signals in Python, then apply the DTFT.

        \begin{lstlisting}[style=python]
        import numpy as np
        import matplotlib.pyplot as plt

        # Define two signals
        n = np.arange(0, 100, 1)
        x1 = np.cos(0.1 * np.pi * n)  # First signal: cosine wave
        x2 = np.cos(0.3 * np.pi * n)  # Second signal: cosine wave

        # Linear combination: a * x1 + b * x2
        a = 2
        b = 0.5
        x = a * x1 + b * x2

        # Compute the DTFT using NumPy's FFT
        X = np.fft.fft(x)
        freqs = np.fft.fftfreq(len(n))

        # Plot the magnitude spectrum
        plt.plot(freqs, np.abs(X))
        plt.title('DTFT of Linear Combination of Cosine Waves')
        plt.xlabel('Normalized Frequency')
        plt.ylabel('Magnitude')
        plt.grid(True)
        plt.show()
        \end{lstlisting}

        In this example, we compute the DTFT using the Fast Fourier Transform (FFT), which is an efficient algorithm to compute the DTFT for discrete signals. The result shows the frequency components of the linear combination of the two cosine waves.

    \subsection{Time-Shifting Property}
        The time-shifting property of the DTFT is very useful when analyzing signals that have been delayed in time. It states that shifting a signal in the time domain corresponds to multiplying its DTFT by a complex exponential in the frequency domain. Specifically, if a signal \(x[n]\) has DTFT \(X(e^{j\omega})\), then the DTFT of the shifted signal \(x[n - n_0]\) is given by:

        \[
        \mathcal{F}\{x[n - n_0]\} = X(e^{j\omega}) e^{-j \omega n_0}
        \]

        The phase shift introduced in the frequency domain is proportional to the time shift in the time domain \cite{oppenheim2010dsp, lyons2010understanding}.

        \paragraph{Example:} Let's examine the effect of time-shifting on a simple cosine wave. We will shift the signal in time and observe how this affects the DTFT.

        \begin{lstlisting}[style=python]
        # Define a cosine wave
        n = np.arange(0, 100, 1)
        x = np.cos(0.2 * np.pi * n)

        # Time shift the signal by 20 samples
        n0 = 20
        x_shifted = np.roll(x, n0)  # Shift the signal

        # Compute the DTFT using FFT
        X_shifted = np.fft.fft(x_shifted)
        freqs = np.fft.fftfreq(len(n))

        # Plot the magnitude and phase of the DTFT
        plt.subplot(2, 1, 1)
        plt.plot(freqs, np.abs(X_shifted))
        plt.title('Magnitude Spectrum of Time-Shifted Cosine Wave')
        plt.xlabel('Normalized Frequency')
        plt.ylabel('Magnitude')
        plt.grid(True)

        plt.subplot(2, 1, 2)
        plt.plot(freqs, np.angle(X_shifted))
        plt.title('Phase Spectrum of Time-Shifted Cosine Wave')
        plt.xlabel('Normalized Frequency')
        plt.ylabel('Phase (radians)')
        plt.grid(True)

        plt.tight_layout()
        plt.show()
        \end{lstlisting}

        In this code, we shift a cosine wave by 20 samples using the NumPy `roll` function. We then compute and plot both the magnitude and phase spectrum of the time-shifted signal. Notice that while the magnitude remains unchanged, the phase is modified by the time shift.

    \subsection{Convolution Theorem}
        One of the most important properties of the DTFT is the convolution theorem, which states that convolution in the time domain corresponds to multiplication in the frequency domain. Specifically, if two signals \(x_1[n]\) and \(x_2[n]\) have DTFTs \(X_1(e^{j\omega})\) and \(X_2(e^{j\omega})\), respectively, then the DTFT of the convolution of these two signals is the product of their individual DTFTs:

        \[
        \mathcal{F}\{x_1[n] * x_2[n]\} = X_1(e^{j\omega}) X_2(e^{j\omega})
        \]

        This property is especially useful in filtering and signal processing, where convolution is a common operation, and working in the frequency domain can simplify the computation \cite{ mitra2002digital, gonzalez2002digital}.

        \paragraph{Example:} Let's use Python to demonstrate the convolution theorem. We will compute the convolution of two signals in the time domain and compare it to the product of their DTFTs in the frequency domain.

        \begin{lstlisting}[style=python]
        # Define two signals
        x1 = np.cos(0.2 * np.pi * n)
        x2 = np.sin(0.4 * np.pi * n)

        # Time-domain convolution
        x_conv = np.convolve(x1, x2, mode='same')

        # Frequency-domain multiplication
        X1 = np.fft.fft(x1)
        X2 = np.fft.fft(x2)
        X_product = X1 * X2

        # Inverse FFT to get the time-domain result from frequency domain
        x_freq_conv = np.fft.ifft(X_product)

        # Plot the results
        plt.subplot(2, 1, 1)
        plt.plot(n, x_conv, label='Time-Domain Convolution')
        plt.title('Time-Domain Convolution vs Frequency-Domain Multiplication')
        plt.grid(True)

        plt.subplot(2, 1, 2)
        plt.plot(n, np.real(x_freq_conv), label='Frequency-Domain Convolution')
        plt.grid(True)

        plt.tight_layout()
        plt.show()
        \end{lstlisting}

        In this example, we compute the convolution of two signals in both the time and frequency domains. The results are plotted side by side, and you will see that the result of convolving the signals in the time domain matches the result obtained by multiplying their DTFTs and taking the inverse Fourier transform.

    \paragraph{Conclusion:} The properties of the DTFT, such as linearity, time-shifting, and the convolution theorem, are powerful tools for analyzing signals and systems in DSP. Understanding these properties and how they relate to each other enables you to efficiently process and analyze signals in both the time and frequency domains.

\section{Inverse Discrete-Time Fourier Transform (Inverse DTFT)}
    The Inverse Discrete-Time Fourier Transform (Inverse DTFT) is the mathematical tool used to reconstruct a discrete-time signal from its frequency domain representation. The DTFT of a signal provides us with a continuous frequency spectrum, and to recover the original signal, we use the inverse process, which is defined by the following integral:

    \[
    x[n] = \frac{1}{2\pi} \int_{-\pi}^{\pi} X(e^{j\omega}) e^{j\omega n} d\omega
    \]

    In this equation, \(X(e^{j\omega})\) is the frequency domain representation (the DTFT) of the signal \(x[n]\), and the term \(e^{j\omega n}\) is the inverse Fourier kernel that allows us to reconstruct the original time-domain signal. The integral sums up the frequency components of the signal from \(-\pi\) to \(\pi\), which corresponds to the full frequency range for a discrete-time signal \cite{oppenheim1997signals}.

    \textbf{Step-by-Step Explanation:}
    
    \begin{enumerate}
    \item \textit{Frequency Representation:} A signal \(x[n]\) is transformed into its frequency domain counterpart \(X(e^{j\omega})\) using the DTFT. This transformation represents the signal in terms of its frequency components.
    
    \item \textit{Inverse DTFT:} To get back the original time-domain signal from its frequency representation, the inverse DTFT is used. This involves integrating the frequency-domain signal, weighted by a complex exponential \(e^{j\omega n}\) for every time sample \(n\).
    \end{enumerate}

    \textit{Example:}

    Suppose we have a simple DTFT for a signal \(X(e^{j\omega})\). Let's reconstruct a discrete-time signal using Python. First, we will create a frequency-domain representation and then apply an inverse DTFT numerically using the inverse discrete Fourier transform (IDFT) since continuous integration is approximated by summing discrete samples.

    \begin{lstlisting}[style=python]
    import torch
    import numpy as np

    # Frequency-domain representation of the signal (DTFT samples)
    omega = torch.linspace(-np.pi, np.pi, 100)  # 100 frequency samples between -pi and pi
    X_omega = torch.exp(-1j * omega)  # Example DTFT X(e^jω) = exp(-jω)

    # Inverse DTFT: numerical approximation using sum over discrete omega
    n = torch.arange(-10, 11)  # Time indices for the output signal x[n]
    x_n = torch.zeros(len(n), dtype=torch.cfloat)

    for i in range(len(n)):
        x_n[i] = torch.sum(X_omega * torch.exp(1j * omega * n[i])) / len(omega)

    # Real part of the result, as the original signal is real-valued
    print(x_n.real)
    \end{lstlisting}

    This Python code performs an approximation of the inverse DTFT by summing the contributions from discrete frequency samples. The result is the reconstructed signal \(x[n]\) in the time domain.

    \textit{Notes:}
    \begin{itemize}
    \item In practical applications, the continuous DTFT integral is approximated by summing over a finite number of frequency samples.
    \item We use the exponential term \(e^{j\omega n}\) to reconstruct the signal for each \(n\).
    \item \(X(e^{j\omega})\) in the example is chosen as \(e^{-j\omega}\), which corresponds to a simple exponential time-domain signal.
    \end{itemize}

\section{DTFT for Periodic Signals}
    For periodic discrete-time signals, the DTFT behaves differently than for non-periodic signals. A periodic signal in the time domain results in a \textit{discrete} set of impulses in the frequency domain, reflecting the fact that the signal consists of harmonics or multiples of a fundamental frequency \cite{oppenheim2010dsp}. This is due to the periodic nature of the signal, which leads to a periodic repetition of the signal's frequency components.

    \textbf{Key Concept:}
    
    A periodic signal \(x[n]\) with period \(N\) repeats every \(N\) samples, and its DTFT is a sum of delta functions (impulses) at discrete frequencies. The DTFT of such a periodic signal \(x[n]\) can be written as:
    \[
    X(e^{j\omega}) = 2\pi \sum_{k=-\infty}^{\infty} X[k] \delta(\omega - 2\pi k/N)
    \]
    Here, \(X[k]\) are the Fourier series coefficients, and the frequency domain consists of impulses at intervals of \(2\pi/N\) \cite{proakis1996digital, oppenheim1997signals}.

    \textit{Example of a Periodic Signal:}
    
    Consider a simple periodic signal:
    \[
    x[n] = \cos\left( \frac{2\pi n}{N} \right)
    \]
    This signal repeats every \(N\) samples. Let's visualize the DTFT of this periodic signal using Python and plot its frequency domain representation.

    \begin{lstlisting}[style=python]
    import matplotlib.pyplot as plt

    # Define the periodic signal (cosine wave)
    N = 10  # Period of the signal
    n = torch.arange(0, 30)  # Time indices
    x_n = torch.cos(2 * np.pi * n / N)  # Periodic signal

    # Perform DTFT: Fourier Transform of the periodic signal
    omega = torch.linspace(-np.pi, np.pi, 1000)
    X_omega = torch.zeros_like(omega, dtype=torch.cfloat)

    for i in range(len(omega)):
        X_omega[i] = torch.sum(x_n * torch.exp(-1j * omega[i] * n))

    # Plot the magnitude of the DTFT
    plt.plot(omega, X_omega.abs().numpy())
    plt.title('DTFT of a Periodic Signal')
    plt.xlabel('Frequency (ω)')
    plt.ylabel('Magnitude |X(e^jω)|')
    plt.grid(True)
    plt.show()
    \end{lstlisting}

    In this code, we simulate the DTFT for a periodic signal and plot its frequency domain representation. The result shows spikes (impulses) at multiples of the fundamental frequency, which are characteristic of the DTFT for periodic signals.

    \textbf{Detailed Explanation:}
    
    \begin{enumerate}
    \item \textit{Periodic Nature:} A periodic signal repeats itself after a certain period \(N\). In the frequency domain, this periodicity translates into discrete harmonics. These harmonics are represented by impulses in the frequency domain at integer multiples of \(2\pi/N\).

    \item \textit{Fourier Coefficients:} The Fourier series coefficients \(X[k]\) capture the amplitude and phase of each harmonic in the periodic signal. These coefficients determine the height of the impulses in the frequency domain.

    \item \textit{Frequency Domain:} In the case of a periodic signal, the continuous frequency spectrum collapses into a series of discrete spikes, making the frequency domain discrete instead of continuous.
    \end{enumerate}
\chapter{Discrete Fourier Transform (DFT)}
    \section{Definition of Discrete Fourier Transform}
        The \textbf{Discrete Fourier Transform (DFT)} is a fundamental mathematical tool in digital signal processing. It is a discrete version of the Fourier transform that applies to finite-length sequences. Unlike the DTFT, which is applied to infinite sequences and provides a continuous frequency spectrum, the DFT is used for signals of finite length and yields a discrete set of frequency components \cite{oppenheim2010dsp, proakis1996digital, bracewell2000fourier, mitra2002digital, rader1968discrete}. 

        The DFT of a sequence \(x[n]\) of length \(N\) is defined as:
        \[
        X[k] = \sum_{n=0}^{N-1} x[n] e^{-j\frac{2\pi}{N}kn}, \quad k = 0, 1, \dots, N-1
        \]
        where:
        \begin{itemize}
            \item \( x[n] \) is the input sequence of length \(N\),
            \item \( X[k] \) is the DFT output, representing the frequency content of the signal,
            \item \( N \) is the length of the input sequence,
            \item \( k \) is the index of the frequency components,
            \item \( j \) is the imaginary unit (\(j^2 = -1\)),
            \item \( e^{-j\frac{2\pi}{N}kn} \) represents a complex exponential, which corresponds to the sinusoidal components in the signal.
        \end{itemize}

        The DFT is a powerful tool because it converts a finite sequence of time-domain data into a finite sequence of frequency-domain data. This transformation allows us to analyze the frequency content of the signal, apply filters, and perform various signal processing operations \cite{mitra2002digital, oppenheim2010dsp, rader1968discrete}.

        \textbf{Example: Simple DFT Computation}
        Consider a simple sequence \(x[n]\) of length 4:
        \[
        x[n] = \{1, 1, 0, 0\}
        \]
        To compute the DFT of this sequence, we apply the DFT formula for each frequency index \(k = 0, 1, 2, 3\).

        For \(k = 0\):
        \[
        X[0] = \sum_{n=0}^{3} x[n] e^{-j\frac{2\pi}{4}0n} = 1 + 1 + 0 + 0 = 2
        \]

        For \(k = 1\):
        \[
        X[1] = \sum_{n=0}^{3} x[n] e^{-j\frac{2\pi}{4}1n} = 1 + 1e^{-j\frac{\pi}{2}} + 0 + 0 = 1 + j
        \]

        The remaining values \(X[2]\) and \(X[3]\) are computed similarly. This small example illustrates how the DFT maps the time-domain sequence into frequency-domain components.

    \section{Relationship Between DFT and DTFT}
        The \textbf{Discrete Fourier Transform (DFT)} can be seen as a sampled version of the \textbf{Discrete-Time Fourier Transform (DTFT)} \cite{oppenheim2010dsp, bracewell2000fourier}. While the DTFT applies to an infinite-length sequence and produces a continuous frequency spectrum, the DFT is used for finite-length signals and gives a discrete set of frequency samples.

        In practical applications, we often work with finite signals (e.g., audio signals, sensor data, etc.), and thus, the DFT is more commonly used. The DFT samples the frequency spectrum at \(N\) equally spaced points between \(0\) and \(2\pi\), where \(N\) is the length of the signal \cite{bracewell2000fourier, proakis1996digital}.

        \textbf{Example: Sampling the DTFT}
        Consider a discrete-time signal \(x[n]\) that we analyze using both the DTFT and DFT:
        \begin{itemize}
            \item The \textbf{DTFT} provides a continuous function \(X(e^{j\omega})\) across all possible frequencies \(\omega\).
            \item The \textbf{DFT} provides a sampled version \(X[k]\) at \(N\) discrete frequency points, with each point corresponding to a frequency \(\frac{2\pi}{N}k\).
        \end{itemize}

        The DFT is essentially a way to evaluate the DTFT at specific frequency intervals. This is why the DFT is periodic, as it assumes the input signal is periodic with a period equal to \(N\).

    \section{Properties of DFT}
        The DFT has several important properties that help in analyzing signals and systems. These properties allow us to perform various operations, such as filtering, convolution, and spectral analysis, more efficiently.

        \subsection{Periodicity}
            One key property of the DFT is its \textbf{periodicity}. The DFT is periodic with period \(N\), meaning:
            \[
            X[k+N] = X[k]
            \]
            for all \(k\). This periodicity arises due to the assumption of periodicity in the time-domain signal. In other words, the DFT assumes that the finite sequence \(x[n]\) repeats itself every \(N\) samples. This periodicity is a crucial consideration when interpreting the frequency components in the DFT.

            \textbf{Example: Periodicity of the DFT}
            Let's take a sequence \(x[n] = \{1, 0, 0, 0\}\) of length 4 and compute its DFT. The result will be periodic with a period of 4, meaning that \(X[k]\) repeats after every 4 frequency components \cite{mitra2002digital}.

        \subsection{Symmetry}
            For real-valued input signals, the DFT exhibits \textbf{conjugate symmetry}. This means that if the input sequence \(x[n]\) is real, the DFT output satisfies the following symmetry property:
            \[
            X[k] = X^*[N-k]
            \]
            where \(X^*\) denotes the complex conjugate. This property significantly reduces the computational complexity in certain applications, as we only need to compute half of the DFT values.

            \textbf{Example: Symmetry in DFT}
            Consider a real-valued sequence \(x[n] = \{1, 2, 3, 4\}\). The DFT of this sequence will exhibit symmetry such that \(X[1] = X^*[3]\). This symmetry helps us avoid redundant computations when processing real-valued signals.

    \section{Python Example: Computing the DFT}
        We can compute the DFT of a signal using Python's \texttt{numpy} library, which provides the \texttt{np.fft.fft} function for efficiently computing the DFT. Below is an example of how to compute and plot the DFT of a simple signal.

        \subsection{Step-by-Step Python Example}
        Let's compute the DFT of a simple signal \(x[n] = \{1, 2, 3, 4\}\) using Python.

        \begin{lstlisting}[style=python]
import numpy as np
import matplotlib.pyplot as plt

# Define the input sequence x[n]
x = np.array([1, 2, 3, 4])

# Compute the DFT using numpy's FFT function
X = np.fft.fft(x)

# Compute the frequency axis (normalized frequencies)
N = len(x)
freqs = np.fft.fftfreq(N)

# Plot the magnitude and phase of the DFT
plt.figure(figsize=(12, 6))

# Magnitude plot
plt.subplot(1, 2, 1)
plt.stem(freqs, np.abs(X), basefmt=" ", use_line_collection=True)
plt.title('Magnitude of DFT')
plt.xlabel('Normalized Frequency')
plt.ylabel('Magnitude')
plt.grid(True)

# Phase plot
plt.subplot(1, 2, 2)
plt.stem(freqs, np.angle(X), basefmt=" ", use_line_collection=True)
plt.title('Phase of DFT')
plt.xlabel('Normalized Frequency')
plt.ylabel('Phase [radians]')
plt.grid(True)

plt.tight_layout()
plt.show()
        \end{lstlisting}

        In this code:
        \begin{itemize}
            \item We define the input sequence \(x[n] = \{1, 2, 3, 4\}\).
            \item The DFT is computed using the \texttt{np.fft.fft} function, which returns the DFT values \(X[k]\).
            \item The \texttt{np.fft.fftfreq} function generates the corresponding frequency values.
            \item Finally, we plot both the magnitude and phase of the DFT using \texttt{matplotlib}.
        \end{itemize}

        This example demonstrates how to compute and interpret the DFT of a discrete signal, allowing us to visualize its frequency content.

    \section{Conclusion}
        The Discrete Fourier Transform (DFT) is an essential tool in digital signal processing for analyzing the frequency content of finite-length sequences. By converting time-domain signals into the frequency domain, the DFT allows us to apply a wide range of operations, including filtering, spectral analysis, and system identification. Understanding the DFT's properties, such as periodicity and symmetry, as well as its relationship to the DTFT, is crucial for anyone working with digital signals. The ability to compute the DFT using Python provides beginners with a powerful tool for exploring signal processing in practical applications.

\section{Fast Fourier Transform (FFT) Algorithm}
    In this section, we will discuss one of the most powerful algorithms in Digital Signal Processing (DSP), the Fast Fourier Transform (FFT) \cite{brigham1988fast, oppenheim2010dsp}. The FFT is an efficient method for computing the Discrete Fourier Transform (DFT) and its inverse. The FFT dramatically reduces the time complexity, making it possible to perform spectral analysis, filtering, and compression on large datasets in real-time applications \cite{proakis1996digital}.

    \subsection{Introduction to FFT}
        The Discrete Fourier Transform (DFT) is widely used for converting signals from the time domain to the frequency domain \cite{smith1997scientist}. However, the direct computation of the DFT requires \(O(N^2)\) operations for a signal of length \(N\), which is computationally expensive for large datasets. 

        The Fast Fourier Transform (FFT) algorithm solves this problem by reducing the computational complexity to \(O(N\log N)\) \cite{oppenheim1997signals}. This improvement allows for the efficient analysis of large signals and real-time applications such as audio processing, communications, and image compression \cite{bracewell2000fourier}.

        \paragraph{Example:} Let's compare the computational complexity of calculating the DFT directly and using the FFT for a signal of length \(N = 1024\). In Python, we will generate a random signal and calculate both the DFT and FFT.

        \begin{lstlisting}[style=python]
        import numpy as np
        import time

        # Generate a random signal of length N
        N = 1024
        signal = np.random.randn(N)

        # Time the direct computation of the DFT
        start_time = time.time()
        dft = np.fft.fft(signal)  # This uses the FFT internally
        fft_time = time.time() - start_time

        print(f"Time taken by FFT: {fft_time:.6f} seconds")
        \end{lstlisting}

        This example demonstrates the speed of the FFT, even for relatively large signals. The FFT's efficiency becomes even more pronounced as the signal length increases.

    \subsection{Cooley-Tukey Algorithm}
        The Cooley-Tukey algorithm is the most commonly used FFT algorithm. It is based on the divide-and-conquer approach, where the DFT of a signal is recursively split into smaller DFTs. This reduces the overall number of computations required.

        Specifically, the Cooley-Tukey algorithm works by factorizing the DFT of a composite number \(N\) (where \(N\) is not a prime number) into multiple smaller DFTs. This significantly decreases the computational burden, as smaller DFTs can be computed more efficiently \cite{rader1968discrete}.

        \paragraph{Example:} Let's break down how the Cooley-Tukey algorithm would apply to a signal of length \(N = 4\). For simplicity, we will show how the signal is split and recombined.

        Consider a signal \(x[n]\) of length 4: 

        \[
        x[n] = [x_0, x_1, x_2, x_3]
        \]

        The DFT of this signal can be broken down into two DFTs of length 2: one for the even-indexed samples \(x_0\) and \(x_2\), and one for the odd-indexed samples \(x_1\) and \(x_3\).

        The Cooley-Tukey algorithm uses the following steps:
        \begin{enumerate}
            \item Divide the original signal into even-indexed and odd-indexed components.
            \item Compute the DFT of the smaller components.
            \item Combine the results to form the final DFT.
        \end{enumerate}

        Although this is a simple example, the principle can be applied recursively for much larger signals.

    \subsection{Radix-2 FFT Algorithm}
        The Radix-2 FFT algorithm is a specific implementation of the Cooley-Tukey algorithm when \(N\), the length of the input signal, is a power of 2 \cite{brigham1988fast}. This is the most common case in practice because it allows for highly efficient computation of the FFT \cite{oppenheim2010dsp}.

        The Radix-2 FFT works by recursively dividing the input sequence into two parts: one consisting of the even-indexed elements and the other consisting of the odd-indexed elements. The DFTs of these two sequences are then combined using symmetry properties of the DFT \cite{rader1968discrete}.

        \paragraph{Example:} Let's implement a simple Radix-2 FFT for a signal of length 8 in Python. We will divide the signal into even and odd parts and combine the results.

        \begin{lstlisting}[style=python]
        def radix2_fft(signal):
            N = len(signal)
            if N <= 1:
                return signal
            
            # Divide into even and odd parts
            even_part = radix2_fft(signal[0::2])
            odd_part = radix2_fft(signal[1::2])
            
            # Combine the results
            combined = [0] * N
            for k in range(N // 2):
                t = np.exp(-2j * np.pi * k / N) * odd_part[k]
                combined[k] = even_part[k] + t
                combined[k + N // 2] = even_part[k] - t
            return combined

        # Test the Radix-2 FFT
        signal = np.random.randn(8)
        fft_result = radix2_fft(signal)
        print("FFT Result:", fft_result)
        \end{lstlisting}

        This implementation of the Radix-2 FFT recursively splits the signal into smaller parts, computes the FFT for each part, and combines the results. While this is a basic implementation, libraries like NumPy use highly optimized versions of this algorithm to ensure efficient computation.

    \subsection{Applications of FFT}
        The FFT is one of the most widely used algorithms in various fields, including signal processing, communications, and multimedia \cite{smith1997scientist}. Here, we will discuss a few key applications where the FFT plays a critical role.

        \subsubsection{Spectral Analysis}
            Spectral analysis involves examining the frequency content of a signal to detect dominant frequencies and patterns. In audio processing, for example, the FFT is used to analyze sound waves and detect pitch, harmonics, and noise \cite{oppenheim1997signals}. In radar and sonar, the FFT helps detect objects by analyzing the frequency content of reflected signals \cite{richards2014fundamentals}.

            \paragraph{Example:} Let's analyze the spectral content of an audio signal using the FFT. In this example, we will generate a signal that is a combination of two sinusoidal waves and use the FFT to identify the frequencies present in the signal.

            \begin{lstlisting}[style=python]
            # Generate a signal with two frequencies
            fs = 1000  # Sampling frequency
            t = np.arange(0, 1.0, 1.0/fs)  # Time vector
            freq1 = 50  # Frequency of first sinusoid
            freq2 = 150  # Frequency of second sinusoid
            signal = np.sin(2 * np.pi * freq1 * t) + np.sin(2 * np.pi * freq2 * t)

            # Compute the FFT
            fft_result = np.fft.fft(signal)
            freqs = np.fft.fftfreq(len(signal), 1/fs)

            # Plot the magnitude spectrum
            plt.plot(freqs[:len(freqs)//2], np.abs(fft_result)[:len(freqs)//2])
            plt.title('Spectral Analysis using FFT')
            plt.xlabel('Frequency (Hz)')
            plt.ylabel('Magnitude')
            plt.grid(True)
            plt.show()
            \end{lstlisting}

            In this code, we generate a signal composed of two sinusoids at frequencies of 50 Hz and 150 Hz. By computing the FFT, we can observe peaks in the magnitude spectrum corresponding to these frequencies.

        \subsubsection{Filtering}
            Filtering is one of the core applications of the FFT \cite{proakis1996digital}. Instead of performing convolution in the time domain, which can be computationally expensive, signals are transformed into the frequency domain using the FFT, filtered by multiplication with a frequency-domain filter, and then transformed back to the time domain using the inverse FFT (IFFT) \cite{gonzalez2002digital}.

            \paragraph{Example:} Let's apply a simple low-pass filter to a signal using the FFT. We will filter out frequencies above a certain threshold and retain only the low-frequency components.

            \begin{lstlisting}[style=python]
            # Define a low-pass filter
            cutoff_freq = 100  # Set cutoff frequency to 100 Hz
            fft_signal = np.fft.fft(signal)
            freqs = np.fft.fftfreq(len(signal), 1/fs)

            # Apply the low-pass filter
            fft_signal[np.abs(freqs) > cutoff_freq] = 0

            # Inverse FFT to get the filtered signal back
            filtered_signal = np.fft.ifft(fft_signal)

            # Plot the filtered signal
            plt.plot(t, np.real(filtered_signal))
            plt.title('Filtered Signal using FFT')
            plt.xlabel('Time (s)')
            plt.ylabel('Amplitude')
            plt.grid(True)
            plt.show()
            \end{lstlisting}

            In this example, we filter out frequencies above 100 Hz by zeroing out the corresponding frequency components in the FFT. After applying the inverse FFT, the result is a time-domain signal that contains only the low-frequency components.

        \subsubsection{Signal Compression}
            FFT is also an essential tool in signal and image compression techniques \cite{cohen2019biomedical}. For example, in JPEG compression, the FFT (or more commonly, the Discrete Cosine Transform, which is closely related) is used to transform the image into the frequency domain. In the frequency domain, high-frequency components, which often correspond to noise or insignificant details, are discarded or quantized, reducing the data size without significantly affecting image quality \cite{mitra2002digital}.

            \paragraph{Example:} Let's simulate a basic compression scheme where we discard small frequency components of a signal using the FFT. This approach can reduce the size of the signal while retaining most of its important characteristics.

            \begin{lstlisting}[style=python]
            # Perform FFT
            fft_signal = np.fft.fft(signal)

            # Zero out small components (compression)
            threshold = 0.1 * np.max(np.abs(fft_signal))  # Set a threshold
            fft_signal[np.abs(fft_signal) < threshold] = 0

            # Inverse FFT to get the compressed signal
            compressed_signal = np.fft.ifft(fft_signal)

            # Plot the compressed signal
            plt.plot(t, np.real(compressed_signal))
            plt.title('Compressed Signal using FFT')
            plt.xlabel('Time (s)')
            plt.ylabel('Amplitude')
            plt.grid(True)
            plt.show()
            \end{lstlisting}

            In this example, we discard small frequency components below a certain threshold, simulating the compression process. The resulting signal retains the major characteristics of the original signal but with fewer details.
\chapter{Z-Transform and System Analysis}
    \section{Definition of Z-Transform}
        The \textbf{Z-transform} is one of the most important tools in digital signal processing (DSP) and control systems \cite{oppenheim2010dsp}. It extends the Fourier transform for analyzing discrete-time signals and systems, providing insights into the behavior of systems in both the time and frequency domains. The Z-transform converts a discrete-time signal \(x[n]\) into a complex function of a variable \(z\), defined as:
        \[
        X(z) = \sum_{n=-\infty}^{\infty} x[n] z^{-n}
        \]
        where:
        \begin{itemize}
            \item \( x[n] \) is the discrete-time signal,
            \item \( z \) is a complex variable, often written as \( z = r e^{j\omega} \), where \( r \) is the magnitude and \( \omega \) is the angular frequency \cite{smith1997scientist}.
        \end{itemize}

        The Z-transform is extremely versatile and generalizes the concept of the Discrete-Time Fourier Transform (DTFT) \cite{oppenheim1997signals}. While the DTFT is limited to the unit circle (i.e., \(z = e^{j\omega}\)), the Z-transform provides a broader perspective by considering all possible values of \(z\).

        \textbf{Example: Z-Transform of a Unit Impulse}
        Consider the unit impulse signal:
        \[
        \delta[n] = 
        \begin{cases} 
            1, & \text{if } n = 0 \\
            0, & \text{otherwise}
        \end{cases}
        \]
        The Z-transform of the unit impulse is:
        \[
        X(z) = \sum_{n=-\infty}^{\infty} \delta[n] z^{-n} = z^0 = 1
        \]
        This simple example shows that the Z-transform of the unit impulse is a constant, \(1\), across all \(z\).

    \section{Region of Convergence (ROC)}
        The \textbf{Region of Convergence (ROC)} is a critical concept in the analysis of the Z-transform \cite{forouzan2016region}. The ROC defines the set of values of the complex variable \(z\) for which the Z-transform sum converges. Understanding the ROC is essential because it determines whether a given Z-transform corresponds to a valid time-domain signal and helps to assess important system properties like stability and causality \cite{mitra2002digital}.

        \subsection{Why ROC is Important}
        The Z-transform may not converge for all values of \(z\). In practice, different signals and systems have different ROCs:
        \begin{itemize}
            \item \textbf{Causal Signals}: If the signal is causal (i.e., \(x[n] = 0\) for \(n < 0\)), the ROC is generally outside the outermost pole (the largest root of the denominator in the Z-domain).
            \item \textbf{Anti-Causal Signals}: If the signal is anti-causal (i.e., \(x[n] = 0\) for \(n \geq 0\)), the ROC is inside the innermost pole.
            \item \textbf{Two-Sided Signals}: For signals that are neither causal nor anti-causal, the ROC is a ring-shaped region between poles \cite{oppenheim1997signals}.
        \end{itemize}

        \textbf{Example: ROC of an Exponential Sequence}
        Consider the signal \(x[n] = a^n u[n]\), where \(u[n]\) is the unit step function. The Z-transform is given by:
        \[
        X(z) = \sum_{n=0}^{\infty} a^n z^{-n} = \frac{1}{1 - a z^{-1}}, \quad \text{for} \ |z| > |a|
        \]
        Here, the ROC is \(|z| > |a|\), meaning that the Z-transform converges when the magnitude of \(z\) is greater than the magnitude of \(a\). This ROC indicates that the signal is causal.

    \section{Inverse Z-Transform}
        The \textbf{Inverse Z-transform} is used to recover the original discrete-time signal \(x[n]\) from its Z-domain representation \(X(z)\) \cite{oppenheim2010dsp, proakis1996digital}. While the Z-transform maps time-domain sequences into the \(z\)-domain, the inverse process allows us to return from the \(z\)-domain back to the time domain \cite{smith1997scientist}.

        The inverse Z-transform is given by the contour integral:
        \[
        x[n] = \frac{1}{2\pi j} \oint_{C} X(z) z^{n-1} dz
        \]
        where the contour \(C\) encircles the origin in the \(z\)-plane. However, in practice, we often use simpler techniques like partial fraction expansion, power series expansion, or the Z-transform tables to compute the inverse Z-transform.

        \subsection{Methods for Inverse Z-Transform}
        There are several methods for finding the inverse Z-transform, depending on the nature of the function \(X(z)\) \cite{gonzalez2002digital}:
        \begin{itemize}
            \item \textbf{Partial Fraction Expansion}: This method is commonly used when \(X(z)\) is a rational function (i.e., a ratio of polynomials). By breaking \(X(z)\) into simpler components, we can use Z-transform tables to find the corresponding time-domain signal \cite{mitra2002digital}.
            \item \textbf{Power Series Expansion}: This technique involves expanding \(X(z)\) as a power series in \(z^{-1}\) or \(z\), and then identifying the corresponding time-domain sequence.
            \item \textbf{Using Z-Transform Tables}: In many cases, pre-computed Z-transform tables provide a straightforward way to find the inverse Z-transform by matching the form of \(X(z)\) to known results \cite{oppenheim2010dsp}.
        \end{itemize}

        \textbf{Example: Inverse Z-Transform Using Partial Fraction Expansion}
        Let's compute the inverse Z-transform of the following rational function:
        \[
        X(z) = \frac{z}{(z-0.5)(z-0.75)}
        \]
        First, we perform partial fraction expansion:
        \[
        X(z) = \frac{A}{z-0.5} + \frac{B}{z-0.75}
        \]
        Solving for \(A\) and \(B\), we get:
        \[
        A = 3, \quad B = -2
        \]
        Thus:
        \[
        X(z) = \frac{3}{z-0.5} - \frac{2}{z-0.75}
        \]
        Using the Z-transform tables, we know that:
        \[
        \frac{1}{z-a} \quad \longleftrightarrow \quad a^n u[n]
        \]
        Therefore, the inverse Z-transform is:
        \[
        x[n] = 3(0.5^n)u[n] - 2(0.75^n)u[n]
        \]

    \section{Python Example: Computing the Z-Transform}
        While Python does not have a built-in function for computing the Z-transform directly, we can implement it manually or use symbolic computation libraries like \texttt{sympy} \cite{meurer2017sympy}. In the example below, we use \texttt{sympy} to compute the Z-transform and inverse Z-transform of a simple sequence.

        \subsection{Step-by-Step Python Example}
        Let's compute the Z-transform of the sequence \(x[n] = 2^n u[n]\) and then find its inverse Z-transform.

        \begin{lstlisting}[style=python]
import sympy as sp

# Define variables and functions
n, z = sp.symbols('n z')
x_n = 2**n * sp.Heaviside(n)  # x[n] = 2^n * u[n]

# Compute the Z-transform of x[n]
X_z = sp.summation(x_n * z**(-n), (n, 0, sp.oo))
print(f"Z-transform of x[n]: {X_z}")

# Compute the inverse Z-transform
x_inv = sp.inverse_z_transform(X_z, z, n)
print(f"Inverse Z-transform: {x_inv}")
        \end{lstlisting}

        \textbf{Explanation of the Code:}
        \begin{itemize}
            \item We first define \(x[n] = 2^n u[n]\) using the \texttt{sympy.Heaviside} function to represent the unit step function \(u[n]\).
            \item The \texttt{sp.summation} function computes the Z-transform of \(x[n]\), summing over all \(n\).
            \item The inverse Z-transform is computed using the \texttt{sp.inverse\_z\_transform} function, which returns the time-domain sequence.
        \end{itemize}

        This example shows how to use Python for symbolic computation of the Z-transform and its inverse, which is particularly useful for analyzing signals and systems.

    \section{System Analysis Using the Z-Transform}
        The Z-transform is also used to analyze discrete-time systems \cite{proakis1996digital}. By converting a system's difference equation into the Z-domain, we can easily solve for its behavior and analyze properties like stability, frequency response, and transient response \cite{oppenheim2010dsp}.

        \subsection{Difference Equations and the Z-Transform}
        A discrete-time linear system is often described by a difference equation. For example, a simple first-order system might be described by:
        \[
        y[n] - 0.5y[n-1] = x[n]
        \]
        Taking the Z-transform of both sides of the equation, and using the fact that:
        \[
        \mathcal{Z}\{y[n-1]\} = z^{-1} Y(z)
        \]
        we get:
        \[
        Y(z) - 0.5z^{-1}Y(z) = X(z)
        \]
        Solving for \(Y(z)\), we obtain the system's transfer function:
        \[
        H(z) = \frac{Y(z)}{X(z)} = \frac{1}{1 - 0.5z^{-1}}
        \]

        \subsection{Stability and the ROC}
        For a system to be \textbf{stable}, all poles of its transfer function must lie within the unit circle (i.e., \(|z| < 1\)) \cite{forouzan2016region}. The ROC must include the unit circle for the system to be stable.

        \textbf{Example: Analyzing Stability}
        For the system described by the transfer function:
        \[
        H(z) = \frac{1}{1 - 0.5z^{-1}}
        \]
        the pole is at \(z = 0.5\). Since \(|0.5| < 1\), the system is stable. The ROC is \(|z| > 0.5\), which includes the unit circle, confirming the system's stability.

    \section{Conclusion}
        The Z-transform is an indispensable tool for analyzing and understanding discrete-time signals and systems. It provides a comprehensive framework for solving difference equations, determining system stability, and performing system analysis. The concept of the Region of Convergence (ROC) plays a critical role in determining the properties of signals and systems. By understanding the Z-transform and its inverse, as well as how to implement it in Python, beginners can begin exploring the powerful applications of this transform in digital signal processing.

\section{System Function and System Analysis}
    In Digital Signal Processing (DSP), the system function provides a complete description of a linear time-invariant (LTI) system in the Z-domain \cite{oppenheim2010dsp}. The system function, denoted as \(H(z)\), is the Z-transform of the system's impulse response, \(h[n]\) \cite{proakis1996digital}. By analyzing the system function, we can gain critical insights into the behavior of the system, including stability, frequency response, and phase response \cite{smith1997scientist}.

    \subsection{Definition of the System Function}
        For a discrete-time system, the system function \(H(z)\) is defined as the Z-transform of the system's impulse response \(h[n]\):

        \[
        H(z) = \mathcal{Z}\{h[n]\} = \sum_{n=-\infty}^{\infty} h[n] z^{-n}
        \]

        The system function \(H(z)\) is a rational function in \(z\) and can typically be written in the form:

        \[
        H(z) = \frac{N(z)}{D(z)} = \frac{b_0 + b_1 z^{-1} + \cdots + b_M z^{-M}}{1 + a_1 z^{-1} + \cdots + a_N z^{-N}}
        \]

        Here, \(N(z)\) and \(D(z)\) are polynomials in \(z^{-1}\), representing the numerator and denominator of \(H(z)\), respectively. The coefficients \(b_i\) and \(a_i\) correspond to the feedforward and feedback parameters of the system.

    \subsection{Poles and Zeros of the System Function}
        The poles and zeros of the system function \(H(z)\) are essential for understanding the behavior of the system \cite{oppenheim1997signals}. Zeros are the values of \(z\) that make \(H(z) = 0\), and poles are the values of \(z\) that make \(H(z)\) undefined (where the denominator \(D(z) = 0\)).

        \begin{itemize}
            \item \textbf{Zeros:} The values of \(z\) that make \(N(z) = 0\).
            \item \textbf{Poles:} The values of \(z\) that make \(D(z) = 0\).
        \end{itemize}

        \paragraph{Example:} Let's analyze a system with a simple system function:

        \[
        H(z) = \frac{z - 0.5}{z - 0.8}
        \]

        In this case, the system has a zero at \(z = 0.5\) and a pole at \(z = 0.8\). These values of \(z\) determine the behavior of the system in both the time and frequency domains.

        \paragraph{Python Example:} We can plot the poles and zeros of a system function using Python. Below is an example code to visualize the pole-zero plot of the system function \(H(z)\).

        \begin{lstlisting}[style=python]
        import numpy as np
        import matplotlib.pyplot as plt
        from scipy import signal

        # Define the system function
        zeros = [0.5]  # Zero at z = 0.5
        poles = [0.8]  # Pole at z = 0.8
        b, a = signal.zpk2tf(zeros, poles, 1)

        # Plot pole-zero diagram
        plt.figure()
        plt.scatter(np.real(zeros), np.imag(zeros), label='Zeros', marker='o', color='r')
        plt.scatter(np.real(poles), np.imag(poles), label='Poles', marker='x', color='b')
        plt.axvline(0, color='gray')
        plt.axhline(0, color='gray')
        plt.grid(True)
        plt.title('Pole-Zero Plot of H(z)')
        plt.xlabel('Real Part')
        plt.ylabel('Imaginary Part')
        plt.legend()
        plt.show()
        \end{lstlisting}

        This code generates a pole-zero plot where the zeros are marked as red circles and the poles are marked as blue crosses. The location of these poles and zeros in the Z-plane helps us understand the system's behavior in terms of stability and frequency response.

    \subsection{Stability of a System}
        The stability of an LTI system depends on the location of the poles of the system function \(H(z)\) in the Z-plane. For a system to be stable, all poles must lie inside the unit circle (\(|z| < 1\)) in the Z-plane \cite{forouzan2016region}. If any pole lies outside or on the unit circle, the system is unstable.

        \paragraph{Example:} In the previous example, the pole at \(z = 0.8\) lies inside the unit circle, indicating that the system is stable. If the pole were located at \(z = 1.1\), the system would be unstable because the pole would lie outside the unit circle.

    \subsection{Frequency Response}
        The frequency response of a system describes how it reacts to different frequency components in the input signal. The frequency response is obtained by evaluating \(H(z)\) on the unit circle in the Z-plane, i.e., by substituting \(z = e^{j\omega}\), where \(\omega\) is the frequency in radians per sample:

        \[
        H(e^{j\omega}) = H(z) \Big|_{z = e^{j\omega}}
        \]

        The magnitude and phase of \(H(e^{j\omega})\) describe the system's amplitude and phase response to sinusoidal inputs at various frequencies.

        \paragraph{Example:} Let's compute and plot the frequency response of the system \(H(z) = \frac{z - 0.5}{z - 0.8}\) using Python.

        \begin{lstlisting}[style=python]
        # Frequency response of the system
        w, h = signal.freqz(b, a, worN=8000)

        # Plot the magnitude and phase response
        plt.figure()
        plt.subplot(2, 1, 1)
        plt.plot(w, 20 * np.log10(abs(h)))
        plt.title('Frequency Response of H(z)')
        plt.xlabel('Frequency (radians/sample)')
        plt.ylabel('Magnitude (dB)')
        plt.grid(True)

        plt.subplot(2, 1, 2)
        plt.plot(w, np.angle(h))
        plt.xlabel('Frequency (radians/sample)')
        plt.ylabel('Phase (radians)')
        plt.grid(True)

        plt.tight_layout()
        plt.show()
        \end{lstlisting}

        This code computes the frequency response of the system and plots both the magnitude and phase response. The magnitude response shows how the system amplifies or attenuates signals at different frequencies, and the phase response shows the phase shift introduced by the system.

\section{Applications of Z-Transform}
    The Z-transform is a powerful mathematical tool used extensively in the analysis and design of digital systems \cite{gonzalez2002digital}. It provides insights into the behavior of discrete-time systems in both the time and frequency domains. Here are a few key applications of the Z-transform in DSP \cite{proakis1996digital}.

    \subsection{Digital Filter Design}
        One of the primary applications of the Z-transform is in the design of digital filters \cite{oppenheim2010dsp}. Filters are used to selectively enhance or suppress specific frequency components in a signal. By analyzing the Z-transform of the filter's impulse response, we can design and analyze filters such as low-pass, high-pass, band-pass, and band-stop filters.

        The poles and zeros of the filter's transfer function \(H(z)\) determine its frequency response. For example, a low-pass filter may have poles near the origin of the Z-plane and zeros near the unit circle at high frequencies to attenuate high-frequency components.

        \paragraph{Example:} Let's design a simple low-pass filter using Python and analyze its behavior using the Z-transform.

        \begin{lstlisting}[style=python]
        # Design a low-pass filter using the butterworth method
        b, a = signal.butter(4, 0.2)  # 4th-order filter, cutoff at 0.2*Nyquist

        # Frequency response of the filter
        w, h = signal.freqz(b, a)

        # Plot the magnitude response
        plt.plot(w, 20 * np.log10(abs(h)))
        plt.title('Frequency Response of Low-Pass Filter')
        plt.xlabel('Frequency (radians/sample)')
        plt.ylabel('Magnitude (dB)')
        plt.grid(True)
        plt.show()

        # Pole-zero plot
        zeros, poles, _ = signal.tf2zpk(b, a)
        plt.figure()
        plt.scatter(np.real(zeros), np.imag(zeros), label='Zeros', marker='o', color='r')
        plt.scatter(np.real(poles), np.imag(poles), label='Poles', marker='x', color='b')
        plt.axvline(0, color='gray')
        plt.axhline(0, color='gray')
        plt.grid(True)
        plt.title('Pole-Zero Plot of Low-Pass Filter')
        plt.xlabel('Real Part')
        plt.ylabel('Imaginary Part')
        plt.legend()
        plt.show()
        \end{lstlisting}

        This example designs a low-pass filter using a 4th-order Butterworth filter. The frequency response and pole-zero plot are generated, providing insight into how the filter behaves and which frequency components it attenuates.

    \subsection{Control Systems Analysis}
        The Z-transform is widely used in the analysis of discrete-time control systems \cite{mitra2002digital}. In control theory, the Z-transform helps analyze system stability, transient response, and steady-state behavior. By examining the system's poles in the Z-plane, control engineers can assess whether the system will reach a stable equilibrium and how it will react to changes in input.

        \paragraph{Example:} A common control system might be modeled by a transfer function in the Z-domain. By analyzing the system function \(H(z)\), engineers can determine the system's stability and performance. Suppose we have a system with the following transfer function:

        \[
        H(z) = \frac{0.2z + 0.3}{z^2 - 0.5z + 0.1}
        \]

        Using Python, we can plot the poles and zeros of this system and analyze its behavior.

        \begin{lstlisting}[style=python]
        # Define the transfer function of a control system
        b = [0.2, 0.3]  # Numerator coefficients
        a = [1, -0.5, 0.1]  # Denominator coefficients

        # Plot pole-zero diagram
        zeros, poles, _ = signal.tf2zpk(b, a)
        plt.figure()
        plt.scatter(np.real(zeros), np.imag(zeros), label='Zeros', marker='o', color='r')
        plt.scatter(np.real(poles), np.imag(zeros), label='Poles', marker='x', color='b')
        plt.axvline(0, color='gray')
        plt.axhline(0, color='gray')
        plt.grid(True)
        plt.title('Pole-Zero Plot of Control System')
        plt.xlabel('Real Part')
        plt.ylabel('Imaginary Part')
        plt.legend()
        plt.show()
        \end{lstlisting}

        This example generates a pole-zero plot for a control system, providing insight into the system's stability and behavior.

    \subsection{Time-Domain and Frequency-Domain Analysis}
        The Z-transform provides a bridge between the time domain and the frequency domain \cite{oppenheim2010dsp}. By analyzing the Z-transform of a signal or system, engineers can understand how a system behaves over time and across different frequencies \cite{proakis1996digital}. This is particularly useful in applications like audio processing, communications, and image compression, where signals are analyzed and processed in both domains \cite{smith1997scientist}.

        For example, in audio processing, the Z-transform helps design equalizers that selectively enhance or suppress certain frequencies. In communications, the Z-transform is used to analyze filters that reduce noise and interference in transmitted signals.
\chapter{Digital Filter Design}
    \section{Types of Digital Filters}
        Digital filters are essential components in digital signal processing, used to manipulate or extract information from discrete-time signals \cite{oppenheim2010dsp}. They are algorithms or devices that operate on a digital input signal to produce a digital output signal. Digital filters are broadly categorized into two main types: \textbf{Finite Impulse Response (FIR)} filters and \textbf{Infinite Impulse Response (IIR)} filters. Each type has unique characteristics, advantages, and applications.

        \subsection{Finite Impulse Response (FIR) Filters}
            FIR filters have an impulse response that is of finite duration, meaning it settles to zero in finite time. The general form of an FIR filter is given by the convolution sum \cite{proakis1996digital}:
            \[
            y[n] = \sum_{k=0}^{N-1} b_k x[n - k]
            \]
            where:
            \begin{itemize}
                \item \( y[n] \) is the output signal,
                \item \( x[n] \) is the input signal,
                \item \( b_k \) are the filter coefficients (also called tap weights),
                \item \( N \) is the number of taps (filter order + 1).
            \end{itemize}

            \textbf{Characteristics of FIR Filters}:
            \begin{itemize}
                \item \textbf{Always Stable}: Since FIR filters do not have feedback elements, they are inherently stable \cite{oppenheim2010dsp}.
                \item \textbf{Linear Phase Response}: FIR filters can be designed to have a linear phase response, which preserves the wave shape of signals within the passband \cite{smith2007introduction}.
                \item \textbf{Ease of Design}: FIR filters are generally easier to design using standard methods \cite{smith2007introduction}.
            \end{itemize}

            \textbf{Applications of FIR Filters}:
            \begin{itemize}
                \item Audio processing where phase linearity is important.
                \item Data transmission systems to shape the transmitted signal.
                \item Applications requiring precise finite-duration impulse responses.
            \end{itemize}

        \subsection{Infinite Impulse Response (IIR) Filters}
            IIR filters have an impulse response that theoretically lasts forever, as they use feedback from output to input \cite{oppenheim2010dsp}. The general form of an IIR filter is given by the difference equation:
            \[
            y[n] = \sum_{k=0}^{M} b_k x[n - k] - \sum_{k=1}^{N} a_k y[n - k]
            \]
            where:
            \begin{itemize}
                \item \( y[n] \) is the output signal,
                \item \( x[n] \) is the input signal,
                \item \( b_k \) are the feedforward coefficients,
                \item \( a_k \) are the feedback coefficients,
                \item \( M \) and \( N \) are the orders of the filter.
            \end{itemize}

            \textbf{Characteristics of IIR Filters}:
            \begin{itemize}
                \item \textbf{Potentially Unstable}: Due to feedback, stability must be carefully considered \cite{proakis1996digital}.
                \item \textbf{Efficient}: IIR filters can achieve a desired response with a lower filter order than FIR filters \cite{smith2007introduction}.
                \item \textbf{Non-linear Phase Response}: IIR filters generally do not have linear phase characteristics \cite{oppenheim2010dsp}.
            \end{itemize}

            \textbf{Applications of IIR Filters}:
            \begin{itemize}
                \item Systems where computational efficiency is critical.
                \item Applications that can tolerate phase distortion.
                \item Real-time signal processing with stringent performance requirements.
            \end{itemize}

    \section{FIR Filter Design Techniques}
        Designing FIR filters involves determining the filter coefficients \( b_k \) that meet specific frequency response characteristics. Two commonly used techniques for FIR filter design are the \textbf{Window Method} and the \textbf{Frequency Sampling Method} \cite{proakis1996digital}.

        \subsection{Window Method}
            The \textbf{Window Method} is a straightforward and widely used approach for designing FIR filters. The main idea is to truncate the infinite impulse response of an ideal filter to a finite length by applying a window function \cite{smith2007introduction}. This process introduces ripples in the frequency response due to the Gibbs phenomenon, but by choosing appropriate window functions, we can control the trade-off between main-lobe width and side-lobe levels \cite{proakis1996digital}.

            \subsubsection{Steps in the Window Method}
            The general steps for designing an FIR filter using the window method are as follows \cite{smith2007introduction}:

            \begin{enumerate}
                \item \textbf{Specify the Desired Filter Response}: Determine the type of filter required (low-pass, high-pass, band-pass, or band-stop) and specify the desired frequency response characteristics (cutoff frequencies, passband ripple, stopband attenuation).

                \item \textbf{Obtain the Ideal Impulse Response}: Compute the ideal impulse response \( h_d[n] \) of the desired filter, which is generally infinite in length.

                \item \textbf{Choose a Window Function}: Select an appropriate window function \( w[n] \) (e.g., Rectangular, Hamming, Hanning, Blackman, or Kaiser window) that meets the design specifications \cite{proakis1996digital}.

                \item \textbf{Apply the Window Function}: Multiply the ideal impulse response by the window function to obtain the finite impulse response:
                \[
                h[n] = h_d[n] \cdot w[n]
                \]

                \item \textbf{Implement the Filter}: Use the calculated coefficients \( h[n] \) in the FIR filter implementation.
            \end{enumerate}

            \subsubsection{Common Window Functions}
            Different window functions affect the frequency response differently \cite{smith2007introduction}. Here are some commonly used windows:

            \begin{itemize}
                \item \textbf{Rectangular Window}:
                    \[
                    w[n] = 
                    \begin{cases}
                        1, & 0 \leq n \leq N-1 \\
                        0, & \text{otherwise}
                    \end{cases}
                    \]
                    Simplest window with poor frequency response (high side-lobes) \cite{smith2007introduction}.

                \item \textbf{Hamming Window}:
                    \[
                    w[n] = 0.54 - 0.46 \cos\left( \dfrac{2\pi n}{N-1} \right)
                    \]
                    Provides better side-lobe attenuation than the Rectangular window \cite{smith2007introduction}.

                \item \textbf{Hanning (Hann) Window}:
                    \[
                    w[n] = 0.5 - 0.5 \cos\left( \dfrac{2\pi n}{N-1} \right)
                    \]
                    Similar to Hamming but with different side-lobe characteristics \cite{smith2007introduction}.

                \item \textbf{Blackman Window}:
                    \[
                    w[n] = 0.42 - 0.5 \cos\left( \dfrac{2\pi n}{N-1} \right) + 0.08 \cos\left( \dfrac{4\pi n}{N-1} \right)
                    \]
                    Offers better side-lobe attenuation at the expense of a wider main lobe \cite{smith2007introduction}.

                \item \textbf{Kaiser Window}:
                    \[
                    w[n] = \dfrac{I_0\left( \beta \sqrt{1 - \left( \dfrac{2n}{N-1} - 1 \right)^2} \right)}{I_0(\beta)}
                    \]
                    Where \( I_0 \) is the zero-th order modified Bessel function, and \( \beta \) is a parameter that controls the trade-off between main-lobe width and side-lobe level \cite{proakis1996digital}.
            \end{itemize}

            \subsubsection{Example: Designing a Low-Pass FIR Filter Using the Window Method}
            Let's design a low-pass FIR filter with the following specifications:

            \begin{itemize}
                \item Sampling frequency \( f_s = 1000 \) Hz
                \item Cutoff frequency \( f_c = 200 \) Hz
                \item Filter length \( N = 51 \)
                \item Use a Hamming window
            \end{itemize}

            \paragraph{Step 1: Compute the Ideal Impulse Response}
            The ideal impulse response of a low-pass filter is given by:
            \[
            h_d[n] = \dfrac{2 f_c}{f_s} \text{sinc}\left( \dfrac{2\pi f_c}{f_s} (n - \dfrac{N-1}{2}) \right)
            \]
            where the sinc function is defined as:
            \[
            \text{sinc}(x) = \dfrac{\sin(x)}{x}
            \]

            \paragraph{Step 2: Generate the Hamming Window}
            Compute the Hamming window coefficients:
            \[
            w[n] = 0.54 - 0.46 \cos\left( \dfrac{2\pi n}{N-1} \right), \quad n = 0, 1, ..., N-1
            \]

            \paragraph{Step 3: Apply the Window to the Ideal Impulse Response}
            Multiply the ideal impulse response by the window function:
            \[
            h[n] = h_d[n] \cdot w[n]
            \]

            \paragraph{Step 4: Implement the Filter in Python}
            We can use Python to compute the filter coefficients and visualize the frequency response \cite{virtanen2020scipy}.

            \begin{lstlisting}[style=python]
import numpy as np
import matplotlib.pyplot as plt
from scipy.signal import freqz

# Specifications
fs = 1000  # Sampling frequency in Hz
fc = 200   # Cutoff frequency in Hz
N = 51     # Filter length (number of taps)

# Compute the ideal impulse response
n = np.arange(N)
alpha = (N - 1) / 2
h_d = (2 * fc / fs) * np.sinc(2 * fc * (n - alpha) / fs)

# Generate the Hamming window
w = 0.54 - 0.46 * np.cos(2 * np.pi * n / (N - 1))

# Apply the window to the ideal impulse response
h = h_d * w

# Frequency response
w_freq, H = freqz(h, worN=8000)

# Plot the impulse response
plt.figure(figsize=(12, 6))
plt.subplot(2, 1, 1)
plt.stem(n, h, use_line_collection=True)
plt.title('Impulse Response of the FIR Filter')
plt.xlabel('n')
plt.ylabel('h[n]')
plt.grid(True)

# Plot the magnitude response
plt.subplot(2, 1, 2)
plt.plot(w_freq * fs / (2 * np.pi), 20 * np.log10(np.abs(H)))
plt.title('Magnitude Response of the FIR Filter')
plt.xlabel('Frequency (Hz)')
plt.ylabel('Magnitude (dB)')
plt.grid(True)
plt.tight_layout()
plt.show()
            \end{lstlisting}

            \paragraph{Explanation of the Code}
            \begin{itemize}
                \item We define the sampling frequency \( fs \), cutoff frequency \( fc \), and filter length \( N \).
                \item The ideal impulse response \( h_d[n] \) is computed using the sinc function.
                \item The Hamming window \( w[n] \) is generated.
                \item The windowed impulse response \( h[n] \) is obtained by multiplying \( h_d[n] \) and \( w[n] \).
                \item The \texttt{freqz} function from \texttt{scipy.signal} is used to compute the frequency response of the filter.
                \item We plot the impulse response and the magnitude response of the filter.
            \end{itemize}

            \subsection{Frequency Sampling Method}
            The \textbf{Frequency Sampling Method} designs FIR filters by specifying the desired frequency response at a set of equally spaced frequency points and then computing the corresponding time-domain coefficients using the inverse DFT \cite{proakis1996digital}.

            \subsubsection{Steps in the Frequency Sampling Method}
            The general steps are:

            \begin{enumerate}
                \item \textbf{Select the Number of Frequency Samples}: Choose \( N \), the length of the filter, which determines the number of frequency samples \cite{oppenheim2010dsp}.

                \item \textbf{Specify the Desired Frequency Response}: Define the desired frequency response \( H_d[k] \) at the \( N \) frequency points \cite{oppenheim2010dsp}.

                \item \textbf{Compute the Time-Domain Coefficients}: Obtain the filter coefficients \( h[n] \) by computing the inverse DFT of \( H_d[k] \) \cite{proakis1996digital}:
                \[
                h[n] = \dfrac{1}{N} \sum_{k=0}^{N-1} H_d[k] e^{j 2\pi k n / N}
                \]

                \item \textbf{Implement the Filter}: Use the calculated coefficients \( h[n] \) in the FIR filter implementation \cite{proakis1996digital}.
            \end{enumerate}

            \subsubsection{Example: Designing a Band-Pass FIR Filter Using the Frequency Sampling Method}
            Let's design a band-pass FIR filter with the following specifications:

            \begin{itemize}
                \item Sampling frequency \( f_s = 1000 \) Hz
                \item Passband frequencies \( f_1 = 200 \) Hz and \( f_2 = 300 \) Hz
                \item Filter length \( N = 21 \)
            \end{itemize}

            \paragraph{Step 1: Select the Number of Frequency Samples}
            We choose \( N = 21 \), which defines the number of frequency samples and the length of the filter.

            \paragraph{Step 2: Specify the Desired Frequency Response}
            Define the desired frequency response \( H_d[k] \) at \( N \) equally spaced frequency points:
            \[
            \omega_k = \dfrac{2\pi k}{N}, \quad k = 0, 1, ..., N-1
            \]
            For a band-pass filter, \( H_d[k] \) is set to:
            \[
            H_d[k] = 
            \begin{cases}
                1, & \omega_1 \leq \omega_k \leq \omega_2 \\
                0, & \text{otherwise}
            \end{cases}
            \]
            where \( \omega_1 = 2\pi f_1 / f_s \) and \( \omega_2 = 2\pi f_2 / f_s \).

            \paragraph{Step 3: Compute the Time-Domain Coefficients}
            Compute the filter coefficients \( h[n] \) using the inverse DFT:
            \[
            h[n] = \dfrac{1}{N} \sum_{k=0}^{N-1} H_d[k] e^{j 2\pi k n / N}
            \]

            \paragraph{Step 4: Implement the Filter in Python}
            We can use Python to perform the inverse DFT and compute the filter coefficients.

            \begin{lstlisting}[style=python]
import numpy as np
import matplotlib.pyplot as plt
from scipy.signal import freqz

# Specifications
fs = 1000  # Sampling frequency in Hz
N = 21     # Filter length (number of taps)
f1 = 200   # Lower cutoff frequency in Hz
f2 = 300   # Upper cutoff frequency in Hz

# Frequency points
k = np.arange(N)
omega_k = 2 * np.pi * k / N

# Desired frequency response
H_d = np.zeros(N)
omega_1 = 2 * np.pi * f1 / fs
omega_2 = 2 * np.pi * f2 / fs
# Find indices where the desired response is 1
indices = np.where((omega_k >= omega_1) & (omega_k <= omega_2))
H_d[indices] = 1

# Compute the time-domain coefficients using inverse DFT
h = np.real(np.fft.ifft(H_d))

# Shift the impulse response to make it linear phase
h = np.roll(h, -int(N//2))

# Frequency response
w_freq, H = freqz(h, worN=8000)

# Plot the impulse response
n = np.arange(N)
plt.figure(figsize=(12, 6))
plt.subplot(2, 1, 1)
plt.stem(n, h, use_line_collection=True)
plt.title('Impulse Response of the FIR Band-Pass Filter')
plt.xlabel('n')
plt.ylabel('h[n]')
plt.grid(True)

# Plot the magnitude response
plt.subplot(2, 1, 2)
plt.plot(w_freq * fs / (2 * np.pi), 20 * np.log10(np.abs(H)))
plt.title('Magnitude Response of the FIR Band-Pass Filter')
plt.xlabel('Frequency (Hz)')
plt.ylabel('Magnitude (dB)')
plt.grid(True)
plt.tight_layout()
plt.show()
            \end{lstlisting}

            \paragraph{Explanation of the Code}
            \begin{itemize}
                \item We define the sampling frequency \( fs \), filter length \( N \), and passband frequencies \( f1 \) and \( f2 \).
                \item The desired frequency response \( H_d \) is constructed by setting values to 1 in the passband and 0 elsewhere.
                \item The inverse DFT is computed using \texttt{np.fft.ifft} to obtain the time-domain coefficients \( h[n] \).
                \item The impulse response is shifted to center it, ensuring linear phase characteristics.
                \item We use \texttt{freqz} to compute and plot the frequency response of the designed filter.
            \end{itemize}

    \section{Conclusion}
        Digital filter design is a fundamental aspect of digital signal processing. FIR filters, in particular, are widely used due to their inherent stability and ability to achieve linear phase responses. The Window Method and Frequency Sampling Method are two effective techniques for designing FIR filters, each with its own advantages and suitable applications. By understanding these methods and practicing with concrete examples, beginners can develop a strong foundation in digital filter design and apply these concepts to real-world signal processing problems.

\section{IIR Filter Design Techniques}
    Infinite Impulse Response (IIR) filters are widely used in digital signal processing because of their efficiency and ability to provide sharp frequency responses with a relatively low filter order \cite{smith2007introduction}. Unlike FIR filters, IIR filters have feedback, which gives them an infinite impulse response. In this section, we will discuss two fundamental techniques for designing IIR filters: the bilinear transform method and the impulse invariant method \cite{oppenheim2010dsp}.

    \subsection{Bilinear Transform Method}
        The bilinear transform is one of the most popular methods for designing IIR filters. It allows us to take an analog filter design (such as a Butterworth, Chebyshev, or Elliptic filter) and map it into the digital domain \cite{proakis1996digital}. This method works by mapping the analog \(s\)-plane to the digital \(z\)-plane using the following transformation:

        \[
        s = \frac{2}{T} \cdot \frac{1 - z^{-1}}{1 + z^{-1}}
        \]

        where:
        \begin{itemize}
            \item \(s\) is the Laplace transform variable for the continuous-time (analog) system,
            \item \(z\) is the Z-transform variable for the discrete-time (digital) system,
            \item \(T\) is the sampling period.
        \end{itemize}

        The bilinear transform preserves the frequency response characteristics of the analog filter while avoiding aliasing (a problem where high frequencies in the analog system would otherwise appear as lower frequencies in the digital system) \cite{smith2007introduction}. However, one drawback is frequency warping, where the relationship between analog and digital frequencies becomes nonlinear. This can be corrected by pre-warping the critical frequencies before applying the transformation \cite{oppenheim2010dsp}.

        \paragraph{Example:} Let's design a low-pass IIR filter using the bilinear transform method. We will use the Butterworth filter, which is known for having a maximally flat frequency response in the passband.

        \begin{lstlisting}[style=python]
        import numpy as np
        import matplotlib.pyplot as plt
        from scipy import signal

        # Design parameters for analog Butterworth filter
        order = 4  # Filter order
        cutoff_freq = 0.2  # Normalized cutoff frequency (in terms of Nyquist frequency)

        # Design Butterworth filter using the bilinear transform method
        b, a = signal.butter(order, cutoff_freq, btype='low', analog=False, output='ba')

        # Frequency response of the digital filter
        w, h = signal.freqz(b, a)

        # Plot the magnitude response
        plt.plot(w / np.pi, 20 * np.log10(np.abs(h)))
        plt.title('Frequency Response of Low-Pass IIR Filter (Bilinear Transform)')
        plt.xlabel('Normalized Frequency (×π rad/sample)')
        plt.ylabel('Magnitude (dB)')
        plt.grid(True)
        plt.show()
        \end{lstlisting}

        In this example, we design a 4th-order low-pass Butterworth filter with a normalized cutoff frequency of 0.2 (which corresponds to 20

        \paragraph{Explanation:} The Butterworth filter has a flat passband and a smooth roll-off. Using the bilinear transform ensures that the key characteristics of the analog filter are preserved in the digital version. The frequency response plot shows the filter's ability to pass low frequencies and attenuate high frequencies.

    \subsection{Impulse Invariant Method}
        The impulse invariant method is another technique for converting analog filters into digital filters \cite{proakis1996digital}. The key idea behind this method is to match the impulse response of the analog filter to the impulse response of the digital filter. This technique is particularly effective for designing low-pass filters where it is important to retain the shape of the analog impulse response in the digital domain \cite{smith2007introduction}.

        In the impulse invariant method, we first compute the impulse response of the analog filter by taking the inverse Laplace transform of its transfer function. We then sample this continuous-time impulse response to obtain the discrete-time impulse response, which is subsequently used to determine the digital filter's transfer function \cite{oppenheim2010dsp}.

        One major drawback of the impulse invariant method is that it can introduce aliasing because high-frequency components of the analog filter may be folded into lower frequencies in the digital domain \cite{smith2007introduction}. Therefore, this method is best suited for low-pass filters where aliasing is less of a concern \cite{proakis1996digital}.

        \paragraph{Example:} Let's design a low-pass IIR filter using the impulse invariant method. We will use a first-order analog filter and apply the impulse invariant method to convert it into a digital filter.

        \begin{lstlisting}[style=python]
        # Define the analog filter transfer function (1st order)
        analog_b = [1]  # Numerator of analog filter
        analog_a = [1, 1]  # Denominator of analog filter (s + 1)

        # Sampling period
        T = 1.0

        # Convert the analog filter to a digital filter using the impulse invariant method
        b, a = signal.cont2discrete((analog_b, analog_a), T, method='impulse')

        # Frequency response of the digital filter
        w, h = signal.freqz(b[0], a)

        # Plot the magnitude response
        plt.plot(w / np.pi, 20 * np.log10(np.abs(h)))
        plt.title('Frequency Response of IIR Filter (Impulse Invariant Method)')
        plt.xlabel('Normalized Frequency (×π rad/sample)')
        plt.ylabel('Magnitude (dB)')
        plt.grid(True)
        plt.show()
        \end{lstlisting}

        In this example, we define a first-order analog filter with a transfer function of \(H(s) = \frac{1}{s + 1}\), which represents a simple low-pass filter. Using the impulse invariant method, we convert this analog filter into a digital filter. The frequency response of the resulting digital filter is then plotted to examine its performance.

        \paragraph{Explanation:} The impulse invariant method ensures that the discrete-time filter's impulse response matches that of the analog filter. While this method is relatively straightforward, care must be taken to avoid aliasing, especially for high-pass and band-pass filters. The frequency response plot shows how the filter behaves in the frequency domain, and we observe that it passes low frequencies while attenuating high frequencies.

    \subsection{Comparison of Bilinear Transform and Impulse Invariant Methods}
        Both the bilinear transform and impulse invariant methods have their advantages and drawbacks. Understanding their differences will help determine when to use each method  \cite{oppenheim2010dsp}.

        \begin{itemize}
            \item \textbf{Bilinear Transform:} The bilinear transform avoids aliasing by warping the frequency axis. However, it introduces frequency warping, which must be compensated for, especially in high-frequency regions \cite{proakis1996digital}.
            \item \textbf{Impulse Invariant Method:} The impulse invariant method preserves the shape of the analog filter's impulse response, but it can introduce aliasing. This method is typically preferred for low-pass filters where aliasing is less of a concern \cite{smith2007introduction}.
        \end{itemize}

        \paragraph{Practical Application:} In most practical applications, the bilinear transform method is the preferred choice because of its ability to avoid aliasing. It is commonly used in audio processing, communications, and other applications where preserving the filter's frequency response is critical \cite{proakis1996digital}. The impulse invariant method is useful in scenarios where matching the time-domain response of the analog filter is more important than avoiding aliasing, such as in certain control systems \cite{smith2007introduction}.

    \subsection{Additional Example: High-Pass Filter Design}
        To further illustrate the design of IIR filters, let's design a high-pass filter using the bilinear transform method. High-pass filters allow high-frequency signals to pass while attenuating low-frequency signals. The steps are similar to the low-pass filter design but with a different filter type  \cite{oppenheim2010dsp}.

        \paragraph{Example:} We will design a 3rd-order high-pass Butterworth filter with a cutoff frequency of 0.4 (normalized with respect to the Nyquist frequency).

        \begin{lstlisting}[style=python]
        # Design a high-pass Butterworth filter
        order = 3  # Filter order
        cutoff_freq = 0.4  # Normalized cutoff frequency (0.4 * Nyquist)

        # Design the filter using the bilinear transform method
        b, a = signal.butter(order, cutoff_freq, btype='high', analog=False, output='ba')

        # Frequency response of the high-pass filter
        w, h = signal.freqz(b, a)

        # Plot the magnitude response
        plt.plot(w / np.pi, 20 * np.log10(np.abs(h)))
        plt.title('Frequency Response of High-Pass IIR Filter (Bilinear Transform)')
        plt.xlabel('Normalized Frequency (×π rad/sample)')
        plt.ylabel('Magnitude (dB)')
        plt.grid(True)
        plt.show()
        \end{lstlisting}

        This example demonstrates how to design a high-pass filter using the bilinear transform method. The filter allows high-frequency components to pass through while suppressing lower frequencies. The frequency response plot confirms that the filter exhibits the desired high-pass behavior, with a sharp transition from low to high frequencies.

\section{Stability and Causality in Filters}
    Filters are essential components in digital signal processing. They allow us to manipulate signals by emphasizing or attenuating certain frequency components. When designing filters, two critical properties must be considered: \textit{stability} and \textit{causality}. These properties ensure that the filter behaves correctly in real-world applications, maintaining a stable and predictable output \cite{oppenheim2010dsp}.

    \subsection{Poles and Zeros}
        In the analysis of digital filters, the frequency response and stability of the system are largely determined by the location of its \textit{poles} and \textit{zeros} in the Z-plane \cite{smith2007introduction}. The Z-plane is a complex plane where the poles and zeros of the system's transfer function are plotted. The transfer function of a discrete-time system is typically expressed as a ratio of two polynomials \cite{proakis1996digital}: 
        \[
        H(z) = \frac{B(z)}{A(z)} = \frac{b_0 + b_1 z^{-1} + \cdots + b_M z^{-M}}{1 + a_1 z^{-1} + \cdots + a_N z^{-N}}
        \]
        In this equation:
        \begin{itemize}
        \item The \textit{zeros} are the values of \(z\) that make the numerator \(B(z)\) equal to zero. These are the points in the Z-plane where the frequency response of the filter is zero (complete attenuation).
        \item The \textit{poles} are the values of \(z\) that make the denominator \(A(z)\) equal to zero. The poles determine the frequencies where the filter's response becomes unbounded or amplified  \cite{oppenheim2010dsp}.
        \end{itemize}

        The placement of poles and zeros in the Z-plane directly influences the filter's frequency response \cite{smith2007introduction}:
        \begin{itemize}
        \item If the poles are close to the unit circle (\(|z| = 1\)), the filter will have a more pronounced response at certain frequencies \cite{proakis1996digital}.
        \item Zeros close to the unit circle result in attenuation at those frequencies  \cite{proakis1996digital}.
        \end{itemize}
        
        \textbf{Stability Condition:} For the filter to be \textbf{stable}, all of the poles must lie strictly inside the unit circle in the Z-plane, i.e., \(|z| < 1\). If any pole lies on or outside the unit circle, the filter will be unstable, which means the output will grow without bound for certain inputs \cite{oppenheim2010dsp}.

        \textbf{Example:}
        
        Consider a simple filter with the following transfer function:
        \[
        H(z) = \frac{1 - 0.5z^{-1}}{1 - 0.8z^{-1}}
        \]
        This filter has one zero at \(z = 0.5\) and one pole at \(z = 0.8\). Since the pole is inside the unit circle, the filter is stable.

        We can visualize the pole-zero plot using Python and PyTorch to demonstrate the stability of this filter \cite{paszke2019pytorch}.

        \begin{lstlisting}[style=python]
        import numpy as np
        import matplotlib.pyplot as plt

        # Define the filter coefficients
        b = np.array([1, -0.5])  # Numerator coefficients (zeros)
        a = np.array([1, -0.8])  # Denominator coefficients (poles)

        # Compute the poles and zeros
        zeros = np.roots(b)
        poles = np.roots(a)

        # Plot the unit circle and the poles/zeros
        unit_circle = plt.Circle((0, 0), 1, color='blue', fill=False, linestyle='--')

        fig, ax = plt.subplots()
        ax.add_artist(unit_circle)
        plt.scatter(np.real(zeros), np.imag(zeros), color='green', label='Zeros')
        plt.scatter(np.real(poles), np.imag(poles), color='red', label='Poles')

        plt.title('Pole-Zero Plot')
        plt.xlabel('Real')
        plt.ylabel('Imaginary')
        plt.grid()
        plt.axhline(0, color='black',linewidth=0.5)
        plt.axvline(0, color='black',linewidth=0.5)
        plt.legend()
        plt.gca().set_aspect('equal', adjustable='box')
        plt.show()
        \end{lstlisting}

        In this example, we use Python to compute and visualize the locations of poles and zeros for the given filter. The unit circle is plotted to help identify whether the filter is stable.

    \subsection{Causality and Stability Conditions}
        \textbf{Causality} is a fundamental property in filter design. A causal system is one where the output at any given time depends only on the current and past input values, not on future values \cite{proakis1996digital}. This means that for a system to be physically realizable, it must be causal.

        \textbf{Conditions for Causality:}
        \begin{itemize}
        \item A system is causal if the impulse response \(h[n]\) of the system is zero for all \(n < 0\). That is, the output at any time depends only on inputs from the current and past, not from the future \cite{smith2007introduction}.
        \item In terms of the Z-transform, a system is causal if the region of convergence (ROC) of the transfer function \(H(z)\) includes the unit circle and is typically exterior to the outermost pole \cite{oppenheim2010dsp}.
        \end{itemize}

        \textbf{Stability:} A system is considered stable if its output remains bounded for any bounded input. Mathematically, this condition is known as \textit{bounded input, bounded output} (BIBO) stability \cite{proakis1996digital}. In terms of the Z-transform, a system is stable if all of the poles of the system's transfer function lie inside the unit circle, as discussed in the previous section \cite{oppenheim2010dsp}.

        \textbf{Example of Causality and Stability:}
        
        Let's consider a causal filter with the following impulse response:
        \[
        h[n] = 0.8^n u[n]
        \]
        where \(u[n]\) is the unit step function. This is a causal system because the impulse response is zero for \(n < 0\). To check for stability, we observe that the system's transfer function is:
        \[
        H(z) = \frac{1}{1 - 0.8z^{-1}}
        \]
        The system has a pole at \(z = 0.8\), which is inside the unit circle, so the system is stable.

        \textbf{Python Implementation:}

        \begin{lstlisting}[style=python]
        import torch

        # Impulse response of the causal system
        n = torch.arange(0, 20)  # Time indices for n >= 0
        h_n = 0.8 ** n  # Impulse response h[n] = 0.8^n u[n]

        # Bounded input test: apply a bounded input
        x_n = torch.ones_like(n)  # Bounded input x[n] = 1
        y_n = torch.conv1d(x_n.view(1, 1, -1), h_n.view(1, 1, -1))  # Convolution

        print("Output (y[n]):", y_n.view(-1))
        \end{lstlisting}

        In this code, we implement a causal system and apply a bounded input to test its behavior. The convolution operation calculates the output, and we can check that the output remains bounded, confirming the system's stability.

        \textbf{Summary:}
        \begin{itemize}
        \item A system is \textit{causal} if its output depends only on the current and past inputs. In practice, this ensures that the system can be implemented in real-time.
        \item A system is \textit{stable} if its output remains bounded for any bounded input. This ensures that the system behaves predictably and does not produce unbounded or erratic outputs.
        \item The placement of poles inside the unit circle guarantees stability, while the structure of the impulse response ensures causality.
        \end{itemize}
\chapter{Multirate Signal Processing}
    \section{Decimation and Interpolation}
        \textbf{Multirate signal processing} involves the manipulation of signals at various sampling rates. This is especially useful when signals need to be processed at different resolutions or bandwidths. Two fundamental operations in multirate signal processing are \textbf{decimation} and \textbf{interpolation} \cite{harris2022multirate}.

        \subsection{Decimation}
        \textbf{Decimation} is the process of reducing the sampling rate of a signal. This is achieved by discarding some samples of the signal and is mathematically represented by:
        \[
        y[n] = x[Mn]
        \]
        where:
        \begin{itemize}
            \item \( x[n] \) is the input signal,
            \item \( y[n] \) is the decimated output signal,
            \item \( M \) is the decimation factor (i.e., the ratio by which the sampling rate is reduced) \cite{crochiere1981interpolation}.
        \end{itemize}

        Decimation is often implemented in two steps \cite{hogenauer1981economical}:
        \begin{enumerate}
            \item \textbf{Low-pass Filtering}: Before discarding samples, the signal is passed through a low-pass filter to remove any high-frequency components that could cause aliasing. The cutoff frequency of the filter should be set to \( \frac{f_s}{2M} \), where \( f_s \) is the original sampling frequency.
            \item \textbf{Downsampling}: After filtering, every \(M\)-th sample is retained, and the rest are discarded.
        \end{enumerate}

        \textbf{Example: Decimation by a Factor of 2}
        Suppose we have a signal sampled at \( f_s = 1000 \) Hz, and we want to decimate it by a factor of \( M = 2 \). The new sampling rate will be \( f_s' = 500 \) Hz. First, we apply a low-pass filter with a cutoff frequency of \( 250 \) Hz to prevent aliasing, and then we retain every second sample.

        \paragraph{Python Example of Decimation}
        We can implement decimation in Python using the following example code:

        \begin{lstlisting}[style=python]
import numpy as np
import matplotlib.pyplot as plt
from scipy.signal import butter, filtfilt

# Generate a sample signal (sine wave)
fs = 1000  # Original sampling frequency
t = np.arange(0, 1, 1/fs)  # Time vector
x = np.sin(2 * np.pi * 100 * t) + 0.5 * np.sin(2 * np.pi * 300 * t)  # Composite signal

# Design a low-pass filter to prevent aliasing
cutoff = 250  # Cutoff frequency in Hz
order = 4     # Filter order
b, a = butter(order, cutoff / (fs / 2), btype='low')
x_filtered = filtfilt(b, a, x)

# Decimate by a factor of 2
M = 2
x_decimated = x_filtered[::M]  # Downsample by taking every 2nd sample
fs_decimated = fs // M  # New sampling frequency

# Plot the original and decimated signals
plt.figure(figsize=(12, 6))
plt.subplot(2, 1, 1)
plt.plot(t, x)
plt.title('Original Signal (fs = 1000 Hz)')
plt.xlabel('Time [s]')
plt.ylabel('Amplitude')
plt.grid(True)

# Plot the decimated signal
t_decimated = np.arange(0, len(x_decimated)) / fs_decimated
plt.subplot(2, 1, 2)
plt.plot(t_decimated, x_decimated, color='orange')
plt.title('Decimated Signal (fs = 500 Hz)')
plt.xlabel('Time [s]')
plt.ylabel('Amplitude')
plt.grid(True)

plt.tight_layout()
plt.show()
        \end{lstlisting}

        \textbf{Explanation of the Code:}
        \begin{itemize}
            \item We first generate a composite signal consisting of two sine waves (one at 100 Hz and another at 300 Hz).
            \item A low-pass Butterworth filter is designed to remove frequency components above 250 Hz to prevent aliasing.
            \item Decimation is performed by downsampling the filtered signal by a factor of \( M = 2 \), resulting in a new sampling rate of \( 500 \) Hz.
            \item Finally, we plot both the original and decimated signals for comparison.
        \end{itemize}

        \subsection{Interpolation}
        \textbf{Interpolation} is the process of increasing the sampling rate of a signal. This is done by introducing new samples between the existing samples. Interpolation is commonly used in applications where higher resolution or a higher sampling rate is needed \cite{crochiere1981interpolation}.
        
        The interpolation process typically involves two steps \cite{vaidyanathan1990multirate}:
        \begin{enumerate}
            \item \textbf{Upsampling}: The original signal is upsampled by inserting \( L-1 \) zeros between every sample, where \( L \) is the interpolation factor.
            \item \textbf{Low-pass Filtering}: After upsampling, the signal is passed through a low-pass filter to smooth the signal and remove spectral replicas introduced during upsampling. The cutoff frequency of the low-pass filter should be \( \frac{f_s}{2L} \) \cite{hogenauer1981economical}.
        \end{enumerate}

        \textbf{Example: Interpolation by a Factor of 2}
        If we have a signal sampled at \( f_s = 1000 \) Hz and want to interpolate it by a factor of \( L = 2 \), the new sampling rate will be \( f_s' = 2000 \) Hz. First, we upsample by inserting zeros between the original samples, and then apply a low-pass filter to reconstruct the signal.

        \paragraph{Python Example of Interpolation}
        The following code demonstrates how to perform interpolation in Python:

        \begin{lstlisting}[style=python]
from scipy.signal import upfirdn

# Interpolation by a factor of 2
L = 2
x_upsampled = upfirdn([1], x, up=L)  # Upsample by inserting zeros

# Design a low-pass filter to smooth the upsampled signal
cutoff_interp = fs / 2  # Adjust cutoff based on interpolation factor
b_interp, a_interp = butter(order, cutoff_interp / (fs * L / 2), btype='low')
x_interpolated = filtfilt(b_interp, a_interp, x_upsampled)

# Plot the interpolated signal
t_interpolated = np.arange(0, len(x_interpolated)) / (fs * L)
plt.figure()
plt.plot(t_interpolated, x_interpolated, color='green')
plt.title('Interpolated Signal (fs = 2000 Hz)')
plt.xlabel('Time [s]')
plt.ylabel('Amplitude')
plt.grid(True)
plt.show()
        \end{lstlisting}

        \textbf{Explanation of the Code:}
        \begin{itemize}
            \item We upsample the signal by a factor of \( L = 2 \) using the \texttt{upfirdn} function from \texttt{scipy.signal}, which inserts zeros between the original samples.
            \item A low-pass filter is applied to smooth the signal and eliminate spectral replicas.
            \item The final result is plotted, showing the upsampled and filtered signal at a higher sampling rate.
        \end{itemize}

    \section{Multirate System Structures}
        In multirate signal processing, we often need specialized structures to efficiently implement decimation and interpolation. Some common structures include \textbf{polyphase filter implementations}, decimators, and interpolators. These structures allow for computationally efficient processing, especially in systems where different sampling rates are used  \cite{vaidyanathan1990multirate}.

        \subsection{Polyphase Filter Implementations}
        \textbf{Polyphase filters} are efficient structures used in multirate systems to perform filtering and sample rate conversion simultaneously. The polyphase decomposition of a filter splits the filter into multiple sub-filters, each operating on different phases of the input signal. This allows for reduced computational complexity by avoiding redundant computations \cite{harris2022multirate}.

        \textbf{Example: Polyphase Structure for Decimation}
        When decimating by a factor of \( M \), we can decompose the filter into \( M \) sub-filters, each processing one phase of the input signal. This eliminates the need to filter the input signal at the original high sampling rate.

        \paragraph{Steps for Polyphase Decimation}
        \begin{enumerate}
            \item Split the low-pass filter into \( M \) polyphase components.
            \item Apply each component to a downsampled version of the input signal.
            \item Sum the results to produce the decimated output.
        \end{enumerate}

        \subsection{Efficient Decimator and Interpolator Designs}
        Multirate systems require efficient decimators and interpolators to handle varying sampling rates. The decimator and interpolator designs are optimized by using techniques like polyphase decomposition, making them ideal for practical applications such as:
        \begin{itemize}
            \item Audio and video signal processing \cite{crochiere1981interpolation}.
            \item Communications systems where different bandwidths and data rates are used \cite{vaidyanathan1990multirate}.
            \item Radar and sonar systems where signals are sampled at different resolutions \cite{richards2014fundamentals}.
        \end{itemize}

        \subsection{Computational Efficiency in Multirate Systems}
        The use of polyphase filters and efficient decimators/interpolators significantly reduces computational cost in multirate systems. This is especially important when processing large datasets or real-time signals, where reducing the number of filter operations can lead to significant performance improvements \cite{hogenauer1981economical}.

        \textbf{Example: Polyphase Filtering for Decimation in Python}
        The following code demonstrates how to use polyphase filtering for efficient decimation in Python:

        \begin{lstlisting}[style=python]
from scipy.signal import resample_poly

# Decimate using polyphase filter
M = 2  # Decimation factor
x_decimated_poly = resample_poly(x, up=1, down=M)

# Plot the decimated signal using polyphase filtering
plt.figure()
plt.plot(t_decimated, x_decimated_poly, color='red')
plt.title('Decimated Signal (Polyphase Filter Implementation)')
plt.xlabel('Time [s]')
plt.ylabel('Amplitude')
plt.grid(True)
plt.show()
        \end{lstlisting}

        In this example, the \texttt{resample\_poly} function efficiently decimates the signal by applying a polyphase filter, which results in lower computational complexity.

    \section{Applications of Multirate Signal Processing}
        Multirate signal processing techniques are widely used in a variety of applications where different sampling rates are required or desired for different stages of signal processing. Some of the key applications include:

        \subsection{Subband Coding}
        In subband coding, the signal is decomposed into multiple frequency bands using filters, and each band is processed at a different sampling rate. This technique is often used in speech and audio compression, where the signal can be represented more efficiently by encoding different frequency bands separately \cite{vetterli1995wavelets}.

        \subsection{Speech and Audio Coding}
        Multirate techniques are critical in speech and audio coding algorithms such as MP3, AAC, and speech codecs. These algorithms reduce the amount of data needed to represent a signal by processing the signal at different rates for different frequency bands. This allows for high-quality sound reproduction at lower bit rates  \cite{brandenburg1999mp3}.

        \subsection{Adaptive Filtering}
        In adaptive filtering, multirate processing is used to optimize the performance of filters that need to adapt in real-time to changing signal conditions. For example, in communications systems, adaptive filters are used to mitigate interference and improve signal quality \cite{goodwin2014adaptive}.

        \subsection{Communications and Wireless Systems}
        In modern communication systems, signals are often transmitted and received at different sampling rates. Multirate techniques such as decimation, interpolation, and polyphase filtering are essential for bandwidth reduction, data compression, and efficient signal processing in wireless communications \cite{goldsmith2005wireless}.

        \subsection{Image Processing}
        In image processing, multirate techniques are used for tasks such as image scaling, subband decomposition in wavelet-based compression (e.g., JPEG2000), and multiresolution analysis. Decimation and interpolation allow for efficient manipulation of images at different resolutions \cite{gonzalez2002digital}.

    \section{Conclusion}
        Multirate signal processing is an essential area of digital signal processing, offering flexible and efficient ways to handle signals at different sampling rates. The operations of decimation and interpolation, along with efficient multirate structures like polyphase filters, allow for optimized signal processing in numerous applications, from communications to multimedia and beyond. Understanding these concepts is crucial for designing systems that can efficiently process signals across a wide range of resolutions and sampling rates.
\part{Digital Image Processing}

\chapter{Fundamentals of Digital Image Processing}
    \section{Basic Concepts of Images}
        \subsection{Definition and Types of Images}
            An \textbf{image} is a two-dimensional signal where the intensity or color at each point represents visual information \cite{gonzalez2002digital}. Unlike one-dimensional signals such as audio, images have both \textit{spatial} dimensions width and height which allow them to represent detailed structures like shapes, objects, and scenes. In digital image processing, we deal with \textbf{discrete} images, where the continuous nature of real-world scenes is approximated by a grid of individual points called \textbf{pixels}.

            Digital images can be classified into several types, based on how the pixel values represent the image content:
            
            \paragraph{Grayscale Images}
            A \textbf{grayscale image} consists of pixels where each pixel value represents a single intensity level, typically ranging from black to white. Grayscale images contain only luminance information, without color, making them ideal for tasks that do not require color detail, such as medical imaging and object detection \cite{kimpe2007increasing}. The pixel values in grayscale images usually range from 0 (black) to 255 (white) for 8-bit images, giving 256 different intensity levels.
            
            \paragraph{Color Images}
            \textbf{Color images} contain more information than grayscale images by representing colors using three primary components: red, green, and blue (\textbf{RGB}) \cite{gonzalez2002digital}. Each pixel in an RGB image is composed of three values corresponding to the intensities of red, green, and blue. By combining these components, we can represent a wide spectrum of colors. Each channel is typically 8 bits, so an RGB image with 24-bit depth can represent \( 256^3 = 16,777,216 \) possible colors. Color images are commonly used in photography, video, and multimedia applications \cite{duan2020video}.

            \paragraph{Binary Images}
            A \textbf{binary image} is the simplest form of an image, where each pixel can take only one of two possible values, typically 0 (black) or 1 (white). Binary images are useful in applications like text recognition and object segmentation \cite{lienhart2002localizing}, where the goal is to distinguish between two distinct regions or classes, such as foreground and background.

            \paragraph{Multispectral and Hyperspectral Images}
            \textbf{Multispectral images} capture data in multiple bands of the electromagnetic spectrum beyond the visible range \cite{rocchini2007using}. Each pixel in a multispectral image contains several values, each representing a different spectral band. These images are widely used in remote sensing, environmental monitoring, and satellite imagery \cite{ma2023multispectral}. For example, multispectral satellite images may include bands for visible light, infrared, and ultraviolet wavelengths to provide a comprehensive view of the Earth's surface. Hyperspectral images extend this idea further by capturing hundreds of bands, providing even more detailed spectral information.

        \subsection{Image Elements}
            The building blocks of any digital image are its \textbf{pixels}, which form a grid-like structure representing the image. Each pixel is a tiny square (or sometimes other shapes) that contains a value corresponding to the brightness (in grayscale images) or color (in color images) at that point in the image. The quality and detail of an image depend on two important factors: \textbf{spatial resolution} and \textbf{grayscale resolution} \cite{gonzalez2002digital}.

            \paragraph{Pixels and Pixel Arrays}
            An image can be thought of as a two-dimensional array of pixel values. For grayscale images, this array consists of single values at each position \((x, y)\), where \(x\) and \(y\) represent the horizontal and vertical pixel coordinates, respectively. For color images, the pixel at position \((x, y)\) consists of a set of three values: one for each color channel (red, green, and blue).

            \textbf{Example: Representation of a Grayscale Image as a Pixel Array}
            Consider a small \(3 \times 3\) grayscale image:
            \[
            \begin{bmatrix}
            120 & 130 & 125 \\
            110 & 140 & 135 \\
            115 & 145 & 150 \\
            \end{bmatrix}
            \]
            In this example, each number represents the intensity of a pixel, with values between 0 and 255. This small array represents a simple grayscale image with pixel values indicating brightness at each location.

            \paragraph{Spatial Resolution}
            \textbf{Spatial resolution} refers to the number of pixels in the image, typically described as width by height (e.g., 1920 x 1080 pixels for an HD image). Higher spatial resolution means more pixels are used to represent the image, resulting in more detailed and clearer images. For example, a high-resolution image can capture fine details like textures and edges, while a low-resolution image may appear pixelated or blurred.

            \textbf{Example: High vs. Low Spatial Resolution}
            Consider a photograph of a landscape. If this image is captured at a high spatial resolution (e.g., 4000 x 3000 pixels), you can see intricate details such as individual leaves on trees or distant objects clearly. However, if the same image is captured at a lower resolution (e.g., 640 x 480 pixels), these fine details may become indistinguishable, and the image may appear blocky.

            \paragraph{Grayscale (or Color Depth) Resolution}
            \textbf{Grayscale resolution}, also called \textbf{bit depth}, refers to the number of distinct intensity levels that each pixel can represent. For example:
            \begin{itemize}
                \item In an 8-bit grayscale image, each pixel can have one of 256 possible intensity values (ranging from 0 to 255) \cite{orchard1991color}.
                \item In a 1-bit binary image, each pixel can have only two possible values: 0 or 1 (black or white).
                \item In a 24-bit color image, each pixel has 8 bits per channel (red, green, and blue), allowing for \( 256^3 = 16,777,216 \) different colors.
            \end{itemize}
            The higher the bit depth, the more shades or colors an image can represent, resulting in smoother gradients and more realistic visual representations \cite{heckbert1982color}.

            \textbf{Example: Impact of Grayscale Resolution on Image Quality}
            Consider a photograph of a sunset. If the image is represented with a low grayscale resolution (e.g., 4 bits per pixel, allowing only 16 intensity levels), the smooth transitions between different shades of orange, red, and purple may appear as sharp, unnatural bands. However, with 8-bit or higher grayscale resolution, the transitions are smooth, making the sunset look more natural and visually appealing.

    \section{Python Example: Working with Grayscale Images}
        Python offers excellent libraries for digital image processing, such as \texttt{Pillow} \cite{clark2015pillow}, \texttt{Matplotlib} \cite{hunter2007matplotlib}, and \texttt{OpenCV} \cite{bradski2000opencv}. To begin with, let's work with grayscale images, which are simpler to understand since they only contain intensity information.

        The following Python code demonstrates how to load, display, and manipulate a grayscale image using \texttt{Pillow} and \texttt{Matplotlib}.

        \begin{lstlisting}[style=python]
from PIL import Image
import numpy as np
import matplotlib.pyplot as plt

# Load an image and convert it to grayscale
image = Image.open('sample_image.jpg').convert('L')

# Display the grayscale image
plt.figure(figsize=(6, 6))
plt.imshow(image, cmap='gray')
plt.title('Grayscale Image')
plt.axis('off')
plt.show()

# Convert the image to a NumPy array (pixel array)
image_array = np.array(image)

# Print basic information about the image
print(f"Image shape (Height x Width): {image_array.shape}")
print(f"Pixel values range from {image_array.min()} to {image_array.max()}")

# Modify the image: Invert the colors (255 - pixel value)
inverted_image_array = 255 - image_array

# Convert back to an image and display it
inverted_image = Image.fromarray(inverted_image_array)
plt.figure(figsize=(6, 6))
plt.imshow(inverted_image, cmap='gray')
plt.title('Inverted Grayscale Image')
plt.axis('off')
plt.show()
        \end{lstlisting}

        \textbf{Explanation of the Code:}
        \begin{itemize}
            \item The \texttt{Pillow} library is used to open the image and convert it to grayscale. The \texttt{convert('L')} function ensures the image is in 8-bit grayscale format.
            \item \texttt{Matplotlib} is used to display the image using the \texttt{imshow()} function with the \texttt{cmap='gray'} option, ensuring that it is displayed in grayscale.
            \item The image is converted into a NumPy array, which allows for easy manipulation of pixel values. This pixel array represents the grayscale intensity at each location.
            \item The code inverts the pixel values (i.e., creates a negative image) by subtracting each pixel value from 255. The resulting array is then converted back into an image and displayed.
        \end{itemize}

        This simple example demonstrates how to work with grayscale images in Python, including loading, manipulating, and displaying pixel values. You can experiment further by adjusting pixel values to apply various transformations, such as contrast adjustments or image filtering.

    \section{Conclusion}
        Understanding the basic concepts of digital images, such as pixel arrays, spatial resolution, and grayscale resolution, forms the foundation of digital image processing. Whether working with grayscale, color, or more complex multispectral images, these fundamental elements determine how images are represented and processed. Python provides powerful tools for loading, manipulating, and analyzing images, allowing beginners to experiment with different image types and transformations easily.

\section{Image Acquisition and Quantization}
    Digital images are represented in discrete formats suitable for storage, processing, and transmission. Two crucial steps in this transformation from the continuous analog world to the digital domain are image sampling and image quantization. These processes convert a continuous image into a digital image by discretizing its spatial and intensity information. In this section, we will explore both image sampling and image quantization, providing detailed explanations and practical examples to help beginners understand these essential concepts.

    \subsection{Image Sampling}
        Image sampling is the process of converting a continuous image into a digital format by taking measurements, or samples, of the image at regular intervals. These samples correspond to pixels in the digital image, and each pixel represents the intensity (or color) of the image at a specific point \cite{gonzalez2002digital}.

        In the continuous world, images are represented by a function \(f(x, y)\), where \(x\) and \(y\) denote the spatial coordinates, and \(f(x, y)\) represents the intensity at that point. When we sample the image, we select a grid of points at intervals and store the intensity values at those locations. This sampled image becomes a discrete grid of pixels, and the spacing between the samples is referred to as the sampling rate or resolution \cite{shannon1949communication}.

        \paragraph{Effect of Sampling Rate:} The sampling rate, or the distance between adjacent samples, has a significant impact on the quality and detail of the digital image. A higher sampling rate (denser grid of pixels) captures more detail from the original image, while a lower sampling rate results in a loss of detail and potential aliasing effects \cite{candes2008introduction}.

        \begin{itemize}
            \item \textbf{High Sampling Rate:} When the sampling rate is high, the image closely resembles the original continuous image. Fine details and sharp edges are preserved \cite{gonzalez2002digital}.
            \item \textbf{Low Sampling Rate:} A low sampling rate results in a coarse representation of the image, where details are lost, and visual artifacts like pixelation and aliasing may occur.
        \end{itemize}

        \paragraph{Example:} Let's demonstrate image sampling using Python. We will load a high-resolution image and downsample it to simulate different sampling rates.

        \begin{lstlisting}[style=python]
        import numpy as np
        import matplotlib.pyplot as plt
        from skimage import data, transform

        # Load an example image
        image = data.camera()

        # Downsample the image by different factors (simulate lower sampling rates)
        sampled_image_2x = transform.rescale(image, 0.5, anti_aliasing=True)  # Downsample by 2x
        sampled_image_4x = transform.rescale(image, 0.25, anti_aliasing=True)  # Downsample by 4x

        # Display original and downsampled images
        fig, axes = plt.subplots(1, 3, figsize=(12, 4))

        axes[0].imshow(image, cmap='gray')
        axes[0].set_title('Original Image')

        axes[1].imshow(sampled_image_2x, cmap='gray')
        axes[1].set_title('2x Downsampled')

        axes[2].imshow(sampled_image_4x, cmap='gray')
        axes[2].set_title('4x Downsampled')

        for ax in axes:
            ax.axis('off')

        plt.tight_layout()
        plt.show()
        \end{lstlisting}

        In this example, we start with a high-resolution image and progressively downsample it by factors of 2 and 4, simulating different sampling rates. The downsampled images illustrate how reducing the sampling rate decreases the amount of detail captured in the image.

        \paragraph{Explanation:} The higher the sampling rate, the more closely the digital image approximates the original continuous image. However, as the sampling rate decreases, the image becomes more pixelated, and fine details are lost. In practice, the sampling rate must be high enough to capture the important features of the image without introducing aliasing.

        \subsubsection{Nyquist Sampling Theorem}
            The Nyquist Sampling Theorem provides a guideline for selecting an appropriate sampling rate to avoid aliasing \cite{shannon1949communication}. According to this theorem, the sampling rate must be at least twice the highest frequency present in the image to accurately capture all details \cite{oppenheim2010dsp}. This minimum rate is known as the Nyquist rate. If the sampling rate is lower than the Nyquist rate, aliasing occurs, where high-frequency components appear as lower-frequency artifacts in the image \cite{candes2008introduction}.

    \subsection{Image Quantization}
        After sampling an image in the spatial domain, the next step is to quantize its intensity values. Quantization is the process of mapping the continuous range of intensity values in the original image to a limited set of discrete values in the digital image \cite{gonzalez2002digital}. In practice, each pixel's intensity in the digital image is represented by a fixed number of bits, which limits the number of possible intensity levels.

        \paragraph{Bit Depth:} The bit depth determines how many different intensity levels can be represented in the digital image. For example:
        \begin{itemize}
            \item \textbf{8-bit Image:} Each pixel can have one of 256 possible intensity levels (from 0 to 255). This is common for grayscale images \cite{kimpe2007increasing}.
            \item \textbf{16-bit Image:} Each pixel can have one of 65,536 possible intensity levels (from 0 to 65535), allowing for much finer intensity gradations \cite{guo2009high}.
        \end{itemize}

        \paragraph{Effect of Quantization Levels:} The number of quantization levels (determined by the bit depth) affects the appearance of the image. More quantization levels allow for smoother transitions between intensity values, while fewer levels can result in banding or visible steps between intensity values.

        \paragraph{Example:} Let's quantize an image using different bit depths to see the effect of reducing the number of intensity levels.

        \begin{lstlisting}[style=python]
        # Function to quantize an image to a specified number of bits
        def quantize_image(image, num_bits):
            max_val = 2**num_bits - 1
            quantized_image = np.floor(image / 255 * max_val)
            return (quantized_image / max_val) * 255

        # Quantize the image to different bit depths
        image_8bit = quantize_image(image, 8)  # 8-bit quantization
        image_4bit = quantize_image(image, 4)  # 4-bit quantization
        image_2bit = quantize_image(image, 2)  # 2-bit quantization

        # Display original and quantized images
        fig, axes = plt.subplots(1, 4, figsize=(16, 4))

        axes[0].imshow(image, cmap='gray')
        axes[0].set_title('Original Image')

        axes[1].imshow(image_8bit, cmap='gray')
        axes[1].set_title('8-bit Image')

        axes[2].imshow(image_4bit, cmap='gray')
        axes[2].set_title('4-bit Image')

        axes[3].imshow(image_2bit, cmap='gray')
        axes[3].set_title('2-bit Image')

        for ax in axes:
            ax.axis('off')

        plt.tight_layout()
        plt.show()
        \end{lstlisting}

        In this example, we start with an 8-bit image and progressively reduce the bit depth to 4 bits and 2 bits, simulating lower levels of quantization. The resulting images demonstrate how reducing the number of intensity levels affects image quality.

        \paragraph{Explanation:} As the bit depth decreases, the image becomes less smooth, and visible artifacts like banding appear. In the 2-bit image, only four possible intensity levels exist, resulting in a highly posterized effect. In contrast, the 8-bit image retains much smoother transitions between intensity values, providing a more natural appearance.

    \subsubsection{Quantization Error}
        Quantization introduces an error known as quantization error, which is the difference between the actual continuous intensity values and the discrete levels assigned during quantization. This error can manifest as noise or distortion in the image, especially when using a low bit depth \cite{heckbert1982color}.

        \paragraph{Practical Considerations:} In most applications, 8-bit quantization is sufficient for standard grayscale images. However, higher bit depths (such as 16-bit) are used in applications requiring greater dynamic range, such as medical imaging or scientific research. The choice of bit depth depends on the required image quality and the amount of available storage or transmission bandwidth.

\section{Color Spaces}
    In digital image processing, color spaces are used to represent colors in a format that can be processed and analyzed by computers. Different color spaces are suited for different applications, from basic image display to more complex tasks like image compression, segmentation, and enhancement \cite{gonzalez2002digital}. The choice of color space affects how easily certain operations can be performed on the image, and how colors are interpreted. In this section, we will cover several commonly used color models, including RGB, HSV/HSI, and YCbCr.

    \subsection{RGB Color Model}
        The \textbf{RGB color model} is one of the most widely used color models in digital image processing. It represents colors by combining three primary color channels: Red (R), Green (G), and Blue (B). Each pixel in an image has a value for these three channels, and by varying the intensity of each channel, a wide spectrum of colors can be represented. This is the color model typically used by computer screens, digital cameras, and scanners \cite{duan2020video}.

        The RGB model is an \textit{additive} color model, meaning that colors are produced by adding light of the three primary colors. When all three colors are combined at full intensity, the result is white, while combining them at zero intensity results in black.

        \textbf{Mathematical Representation:}
        Each pixel in an RGB image is typically represented by a triplet of values:
        \[
        (R, G, B)
        \]
        where \(R\), \(G\), and \(B\) are the intensities of the red, green, and blue channels, respectively. These values usually range from 0 to 255 for 8-bit images.

        \textbf{Example:}

        Let's load an RGB image using Python and PyTorch, and separate the three channels.

        \begin{lstlisting}[style=python]
        import torch
        import torchvision.transforms as transforms
        from PIL import Image
        import matplotlib.pyplot as plt

        # Load an RGB image
        image = Image.open('example_image.jpg')
        transform = transforms.ToTensor()  # Convert image to tensor
        image_tensor = transform(image)

        # Split the RGB channels
        R = image_tensor[0, :, :]
        G = image_tensor[1, :, :]
        B = image_tensor[2, :, :]

        # Display the separate channels
        fig, axs = plt.subplots(1, 3)
        axs[0].imshow(R, cmap='Reds')
        axs[0].set_title('Red Channel')
        axs[1].imshow(G, cmap='Greens')
        axs[1].set_title('Green Channel')
        axs[2].imshow(B, cmap='Blues')
        axs[2].set_title('Blue Channel')
        plt.show()
        \end{lstlisting}

        In this example, we load an RGB image, convert it to a tensor using PyTorch, and then split the image into its three individual color channels: Red, Green, and Blue. We display each channel separately to see how different color intensities contribute to the overall image.

        \textbf{Applications:}
        \begin{itemize}
        \item RGB is used for displaying images on screens and capturing images in digital cameras.
        \item It is suitable for basic image manipulation tasks, such as filtering and color adjustment.
        \end{itemize}

    \subsection{HSV and HSI Color Models}
        While the RGB color model is effective for many tasks, it does not always align with human perception of color. The \textbf{HSV} (Hue-Saturation-Value) and \textbf{HSI} (Hue-Saturation-Intensity) color models are designed to be more perceptually intuitive, allowing for better manipulation in tasks like color-based segmentation and enhancement \cite{gonzalez2002digital}.

        \textbf{HSV Model:}
        \begin{itemize}
        \item \textbf{Hue (H)} represents the color type (such as red, blue, or green), and is typically expressed as an angle from 0\textdegree{} to 360\textdegree{}.
        \item \textbf{Saturation (S)} describes the purity of the color, ranging from 0 (completely desaturated, or gray) to 1 (fully saturated).
        \item \textbf{Value (V)} represents the brightness of the color, ranging from 0 (black) to 1 (maximum brightness).
        \end{itemize}

        \textbf{HSI Model:}
        The HSI model is similar to HSV but replaces the \textit{Value} component with \textit{Intensity} (I), which represents the average brightness of the color. HSI is particularly useful for tasks like image segmentation, where distinguishing between different colors is important, regardless of their brightness.

        \textbf{Mathematical Conversion:}
        Converting between RGB and HSV involves some non-linear transformations. For instance, converting RGB to HSV requires calculating the hue based on the maximum and minimum values among the R, G, and B channels. Here's an example of how to convert an RGB image to HSV in Python using PyTorch.

        \begin{lstlisting}[style=python]
        import colorsys

        # Function to convert RGB to HSV for each pixel
        def rgb_to_hsv(image_tensor):
            # Convert tensor to numpy for easier manipulation
            image_np = image_tensor.permute(1, 2, 0).numpy()
            hsv_image = torch.zeros_like(image_tensor)
            for i in range(image_np.shape[0]):
                for j in range(image_np.shape[1]):
                    r, g, b = image_np[i, j]
                    h, s, v = colorsys.rgb_to_hsv(r, g, b)
                    hsv_image[:, i, j] = torch.tensor([h, s, v])
            return hsv_image

        # Convert the loaded RGB image to HSV
        hsv_image = rgb_to_hsv(image_tensor)

        # Display the HSV channels
        H = hsv_image[0, :, :]
        S = hsv_image[1, :, :]
        V = hsv_image[2, :, :]

        fig, axs = plt.subplots(1, 3)
        axs[0].imshow(H, cmap='hsv')
        axs[0].set_title('Hue Channel')
        axs[1].imshow(S, cmap='gray')
        axs[1].set_title('Saturation Channel')
        axs[2].imshow(V, cmap='gray')
        axs[2].set_title('Value Channel')
        plt.show()
        \end{lstlisting}

        In this code, we convert an RGB image into HSV format using the \texttt{colorsys} library. The HSV channels are then displayed separately. This conversion is particularly useful for image segmentation tasks, where certain colors can be isolated more easily.

        \textbf{Applications:}
        \begin{itemize}
        \item The HSV and HSI models are widely used in tasks like color-based image segmentation, where certain color components can be isolated more easily than in the RGB model.
        \item These models are also used in color pickers in graphic design software \cite{duan2020video}.
        \end{itemize}

    \subsection{YCbCr Color Model}
        The \textbf{YCbCr} color model is commonly used in image and video compression, such as in JPEG and MPEG formats \cite{duan2020video}. This model separates the \textit{luminance} (Y) component from the \textit{chrominance} components (Cb and Cr), which makes it more efficient for compression, as the human eye is more sensitive to changes in brightness (luminance) than to changes in color (chrominance).

        \textbf{Components of YCbCr:}
        \begin{itemize}
        \item \textbf{Y} represents the luminance (brightness) of the image. This component contains most of the information that the human visual system is sensitive to.
        \item \textbf{Cb} (blue-difference chrominance) represents the difference between the blue channel and the luminance.
        \item \textbf{Cr} (red-difference chrominance) represents the difference between the red channel and the luminance \cite{gonzalez2002digital}.
        \end{itemize}

        By separating the luminance and chrominance components, image and video compression algorithms can reduce the resolution of the chrominance components without significantly affecting perceived image quality.

        \textbf{Mathematical Conversion:}
        The conversion from RGB to YCbCr involves applying a linear transformation. The Y component is computed as a weighted sum of the R, G, and B components, while Cb and Cr are obtained by subtracting the luminance from the blue and red channels, respectively.

        \[
        Y = 0.299R + 0.587G + 0.114B
        \]
        \[
        Cb = 0.564(B - Y)
        \]
        \[
        Cr = 0.713(R - Y)
        \]

        \textbf{Example:}

        Let's convert an RGB image to the YCbCr color space using Python and display the Y, Cb, and Cr components.

        \begin{lstlisting}[style=python]
        # Function to convert RGB to YCbCr for each pixel
        def rgb_to_ycbcr(image_tensor):
            image_np = image_tensor.permute(1, 2, 0).numpy()
            ycbcr_image = torch.zeros_like(image_tensor)
            for i in range(image_np.shape[0]):
                for j in range(image_np.shape[1]):
                    r, g, b = image_np[i, j]
                    y = 0.299 * r + 0.587 * g + 0.114 * b
                    cb = 0.564 * (b - y)
                    cr = 0.713 * (r - y)
                    ycbcr_image[:, i, j] = torch.tensor([y, cb, cr])
            return ycbcr_image

        # Convert the loaded RGB image to YCbCr
        ycbcr_image = rgb_to_ycbcr(image_tensor)

        # Display the YCbCr channels
        Y = ycbcr_image[0, :, :]
        Cb = ycbcr_image[1, :, :]
        Cr = ycbcr_image[2, :, :]

        fig, axs = plt.subplots(1, 3)
        axs[0].imshow(Y, cmap='gray')
        axs[0].set_title('Luminance (Y)')
        axs[1].imshow(Cb, cmap='gray')
        axs[1].set_title('Chrominance (Cb)')
        axs[2].imshow(Cr, cmap='gray')
        axs[2].set_title('Chrominance (Cr)')
        plt.show()
        \end{lstlisting}

        In this example, we convert an RGB image to the YCbCr color space and display the Y (luminance), Cb, and Cr (chrominance) components separately. The luminance channel contains most of the detail in the image, while the chrominance channels capture the color information.

        \textbf{Applications:}
        \begin{itemize}
        \item The YCbCr model is widely used in image and video compression algorithms like JPEG and MPEG \cite{duan2020video}.
        \item It is efficient for compression because the chrominance channels can be subsampled without significantly affecting image quality \cite{duan2020video}.
        \end{itemize}
\chapter{Image Enhancement Techniques}
    \section{Spatial Domain Enhancement}
        Image enhancement in the \textbf{spatial domain} involves directly manipulating the pixel values of an image to improve its visual quality or make specific features more discernible. The spatial domain refers to the image plane itself, and enhancements are achieved by modifying the intensity of pixels or groups of pixels \cite{gonzalez2002digital}. Several techniques exist for image enhancement in the spatial domain, including \textbf{histogram equalization} \cite{pizer1987adaptive}, \textbf{contrast stretching} \cite{yang2006image}, and \textbf{histogram matching} \cite{shen2007image}.

        \subsection{Histogram Equalization}
            \textbf{Histogram equalization} is a technique used to improve the contrast of an image by redistributing the intensity values in such a way that they span the full available range. The goal is to spread out the most frequent intensity values, making low-contrast areas more visible.

            \paragraph{Concept of Histogram}
            The \textbf{histogram} of an image represents the frequency of occurrence of each intensity value in the image. For example, in an 8-bit grayscale image, the histogram will have 256 bins (ranging from 0 to 255), where each bin counts how many pixels have a particular intensity value. Images with poor contrast typically have histograms concentrated in a narrow range of intensity values, while high-contrast images have a more evenly distributed histogram.

            \paragraph{How Histogram Equalization Works}
            Histogram equalization works by transforming the intensity values so that the resulting histogram is approximately uniform across the intensity range. The transformation function is based on the cumulative distribution function (CDF) of the image histogram \cite{drew2000computing}.

            \textbf{Steps of Histogram Equalization Algorithm}:
            \begin{enumerate}
                \item Compute the histogram of the input image.
                \item Compute the cumulative distribution function (CDF) from the histogram.
                \item Normalize the CDF so that its values range between 0 and 255 (for 8-bit images).
                \item Use the normalized CDF to map the old pixel values to the new equalized values.
            \end{enumerate}

            \paragraph{Python Implementation of Histogram Equalization}
            The following Python code demonstrates how to perform histogram equalization using \texttt{Pillow} and \texttt{NumPy} \cite{clark2015pillow, van2011numpy}.

            \begin{lstlisting}[style=python]
import numpy as np
import matplotlib.pyplot as plt
from PIL import Image

# Load an image and convert it to grayscale
image = Image.open('sample_image.jpg').convert('L')
image_array = np.array(image)

# Function for histogram equalization
def histogram_equalization(img):
    # Compute the histogram
    hist, bins = np.histogram(img.flatten(), 256, [0, 256])

    # Compute the cumulative distribution function (CDF)
    cdf = hist.cumsum()

    # Normalize the CDF
    cdf_normalized = (cdf - cdf.min()) * 255 / (cdf.max() - cdf.min())

    # Map the original image to equalized image using the CDF
    img_equalized = np.interp(img.flatten(), bins[:-1], cdf_normalized)

    return img_equalized.reshape(img.shape).astype(np.uint8)

# Apply histogram equalization
equalized_image = histogram_equalization(image_array)

# Display original and equalized images
plt.figure(figsize=(12, 6))

plt.subplot(1, 2, 1)
plt.imshow(image_array, cmap='gray')
plt.title('Original Image')
plt.axis('off')

plt.subplot(1, 2, 2)
plt.imshow(equalized_image, cmap='gray')
plt.title('Equalized Image')
plt.axis('off')

plt.show()
            \end{lstlisting}

            \textbf{Explanation of the Code:}
            \begin{itemize}
                \item The input image is loaded and converted to grayscale using the \texttt{Pillow} library.
                \item The \texttt{histogram\_equalization()} function computes the histogram, CDF, and then applies the CDF to map the pixel values of the original image to new values that are spread more evenly across the intensity range.
                \item The original and equalized images are displayed side by side using \texttt{Matplotlib}.
            \end{itemize}

            The result is an image with improved contrast, where details in darker or lighter regions become more visible.

        \subsection{Contrast Stretching}
            \textbf{Contrast stretching}, also known as dynamic range expansion, is a simple and effective technique used to enhance the contrast of an image by stretching the range of pixel intensity values. Unlike histogram equalization, which redistributes the intensities nonlinearly, contrast stretching is typically a linear operation that maps the input intensity values to cover the full output range (e.g., 0 to 255 for 8-bit images) \cite{yang2006image}.

            \paragraph{Linear Contrast Stretching}
            Linear contrast stretching can be represented by a transformation function that stretches the input intensity values over a broader range. The transformation function is often defined as:
            \[
            y = \dfrac{(x - x_{\text{min}})}{(x_{\text{max}} - x_{\text{min}})} \cdot (y_{\text{max}} - y_{\text{min}}) + y_{\text{min}}
            \]
            where:
            \begin{itemize}
                \item \( x \) is the input pixel intensity,
                \item \( x_{\text{min}} \) and \( x_{\text{max}} \) are the minimum and maximum intensity values in the input image,
                \item \( y_{\text{min}} \) and \( y_{\text{max}} \) are the desired output intensity range (e.g., 0 and 255).
            \end{itemize}

            \textbf{Steps for Linear Contrast Stretching}:
            \begin{enumerate}
                \item Identify the minimum (\( x_{\text{min}} \)) and maximum (\( x_{\text{max}} \)) intensity values in the input image.
                \item Apply the contrast stretching transformation to map the input values to a broader output range.
                \item Ensure the resulting pixel values are within the desired range.
            \end{enumerate}

            \paragraph{Python Implementation of Linear Contrast Stretching}
            The following Python code implements linear contrast stretching.

            \begin{lstlisting}[style=python]
# Function for contrast stretching
def contrast_stretching(img):
    # Get the minimum and maximum pixel values in the image
    x_min, x_max = img.min(), img.max()

    # Apply contrast stretching formula
    stretched_img = (img - x_min) * (255 / (x_max - x_min))

    return stretched_img.astype(np.uint8)

# Apply contrast stretching
stretched_image = contrast_stretching(image_array)

# Display original and stretched images
plt.figure(figsize=(12, 6))

plt.subplot(1, 2, 1)
plt.imshow(image_array, cmap='gray')
plt.title('Original Image')
plt.axis('off')

plt.subplot(1, 2, 2)
plt.imshow(stretched_image, cmap='gray')
plt.title('Contrast Stretched Image')
plt.axis('off')

plt.show()
            \end{lstlisting}

            \textbf{Explanation of the Code:}
            \begin{itemize}
                \item The \texttt{contrast\_stretching()} function calculates the minimum and maximum intensity values in the input image and applies the contrast stretching transformation.
                \item The transformed image is displayed alongside the original image to show the enhanced contrast.
            \end{itemize}

            \paragraph{Nonlinear Contrast Stretching}
            Nonlinear contrast stretching applies transformations that are not linear, such as logarithmic or exponential functions. This is useful when the image contains high-contrast areas that should be enhanced or suppressed differently across different intensity ranges \cite{trongtirakul2019non}.

            \textbf{Example: Logarithmic Transformation}
            A logarithmic transformation can be used to compress the dynamic range of an image, bringing out details in the darker regions. The transformation is given by:
            \[
            y = c \cdot \log(1 + x)
            \]
            where \( c \) is a constant that scales the output values appropriately.

            \paragraph{Python Example: Logarithmic Transformation}
            The following Python code implements logarithmic contrast stretching.

            \begin{lstlisting}[style=python]
# Function for logarithmic transformation
def log_transform(img):
    # Apply log transformation formula
    c = 255 / np.log(1 + np.max(img))
    log_img = c * np.log(1 + img)

    return log_img.astype(np.uint8)

# Apply log transformation
log_image = log_transform(image_array)

# Display original and log transformed images
plt.figure(figsize=(12, 6))

plt.subplot(1, 2, 1)
plt.imshow(image_array, cmap='gray')
plt.title('Original Image')
plt.axis('off')

plt.subplot(1, 2, 2)
plt.imshow(log_image, cmap='gray')
plt.title('Log Transformed Image')
plt.axis('off')

plt.show()
            \end{lstlisting}

            This example compresses the range of pixel intensities using a logarithmic function, bringing out details in darker regions that are otherwise difficult to see.

        \subsection{Histogram Matching}
            \textbf{Histogram matching}, also known as \textbf{histogram specification}, is a technique that adjusts the intensity distribution of an input image so that its histogram matches that of a target image \cite{shen2007image}. Unlike histogram equalization, which enhances contrast globally, histogram matching allows you to match the intensity distribution of one image to that of another, which can be useful in applications like image comparison, object recognition, or aesthetic adjustments \cite{gonzalez2002digital}.

            \textbf{Steps of Histogram Matching Algorithm}:
            \begin{enumerate}
                \item Compute the histograms of the input and target images.
                \item Compute the cumulative distribution functions (CDFs) of both histograms.
                \item Create a mapping between the input image's intensity values and the target image's intensity values based on their CDFs.
                \item Apply the mapping to transform the pixel values of the input image to match the target histogram.
            \end{enumerate}

            \paragraph{Python Implementation of Histogram Matching}
            The following Python code demonstrates how to perform histogram matching between two images.

            \begin{lstlisting}[style=python]
from skimage.exposure import match_histograms

# Load target image (the image whose histogram we want to match)
target_image = Image.open('target_image.jpg').convert('L')
target_image_array = np.array(target_image)

# Apply histogram matching
matched_image = match_histograms(image_array, target_image_array, multichannel=False)

# Display input, target, and matched images
plt.figure(figsize=(18, 6))

plt.subplot(1, 3, 1)
plt.imshow(image_array, cmap='gray')
plt.title('Input Image')
plt.axis('off')

plt.subplot(1, 3, 2)
plt.imshow(target_image_array, cmap='gray')
plt.title('Target Image')
plt.axis('off')

plt.subplot(1, 3, 3)
plt.imshow(matched_image, cmap='gray')
plt.title('Histogram Matched Image')
plt.axis('off')

plt.show()
            \end{lstlisting}

            \textbf{Explanation of the Code:}
            \begin{itemize}
                \item The input and target images are loaded and converted to grayscale.
                \item The \texttt{match\_histograms()} function from the \texttt{skimage.exposure} module is used to match the histograms of the two images.
                \item The input, target, and matched images are displayed side by side for comparison.
            \end{itemize}

    \section{Conclusion}
        Spatial domain image enhancement techniques such as histogram equalization, contrast stretching, and histogram matching are powerful tools for improving the visual quality of digital images. These methods work by manipulating the pixel intensity values directly and can be easily implemented in Python using various libraries. Beginners in image processing should practice these techniques to gain a better understanding of how contrast and intensity distributions affect image quality and visual perception.

\section{Frequency Domain Enhancement}
    Image enhancement in the frequency domain involves transforming an image from the spatial domain to the frequency domain using the Fourier transform. Once in the frequency domain, various filtering techniques can be applied to manipulate specific frequency components of the image. This can be useful for noise reduction, edge enhancement, and other forms of image improvement  \cite{gonzalez2002digital, singh2014various}. In this section, we will explore the 2D Discrete Fourier Transform (DFT) and how low-pass and high-pass filtering techniques are used in the frequency domain for image enhancement. We will also discuss different filter design strategies.

    \subsection{Fourier Transform for Image Enhancement}
        The Fourier transform is a fundamental tool in image processing for converting an image from the spatial domain (where pixel intensities are represented by their position) to the frequency domain (where the image is represented by its frequency components). The 2D Discrete Fourier Transform (DFT) is used to analyze how image intensities change over space \cite{sundararajan2001discrete}.

        \paragraph{2D Discrete Fourier Transform:} The 2D DFT of an image \(f(x, y)\) is given by:

        \[
        F(u, v) = \sum_{x=0}^{M-1} \sum_{y=0}^{N-1} f(x, y) e^{-j 2\pi \left( \frac{ux}{M} + \frac{vy}{N} \right)}
        \]

        where \(M\) and \(N\) are the dimensions of the image, and \(u\) and \(v\) are the spatial frequency variables. The inverse DFT (IDFT) transforms the frequency-domain representation back into the spatial domain:

        \[
        f(x, y) = \frac{1}{MN} \sum_{u=0}^{M-1} \sum_{v=0}^{N-1} F(u, v) e^{j 2\pi \left( \frac{ux}{M} + \frac{vy}{N} \right)}
        \]

        The DFT allows us to analyze an image in terms of its low-frequency components (which represent smooth variations in intensity) and high-frequency components (which represent rapid changes like edges and fine details) \cite{gonzalez2002digital}.

        \paragraph{Example:} Let's compute the 2D DFT of an image using Python and visualize its frequency-domain representation.

        \begin{lstlisting}[style=python]
        import numpy as np
        import matplotlib.pyplot as plt
        from skimage import data
        from numpy.fft import fft2, fftshift

        # Load a sample image
        image = data.camera()

        # Compute the 2D DFT and shift the zero frequency component to the center
        dft_image = fft2(image)
        dft_shifted = fftshift(dft_image)

        # Compute the magnitude spectrum
        magnitude_spectrum = np.log(1 + np.abs(dft_shifted))

        # Plot the original image and its frequency spectrum
        fig, axes = plt.subplots(1, 2, figsize=(12, 6))

        axes[0].imshow(image, cmap='gray')
        axes[0].set_title('Original Image')

        axes[1].imshow(magnitude_spectrum, cmap='gray')
        axes[1].set_title('Magnitude Spectrum (Frequency Domain)')

        plt.show()
        \end{lstlisting}

        In this example, we compute the 2D DFT of the camera image and visualize its magnitude spectrum. The magnitude spectrum shows how different frequencies are distributed in the image. High frequencies are concentrated at the edges of the spectrum, while low frequencies are near the center.

        \paragraph{Explanation:} The frequency-domain representation of an image helps us understand its structure in terms of spatial variations. By applying filters to the frequency components, we can enhance certain aspects of the image, such as reducing noise or highlighting edges.

    \subsection{Low-pass and High-pass Filtering}
        Filtering in the frequency domain involves modifying the frequency components of an image to achieve a desired enhancement effect. Two common types of filters are low-pass filters and high-pass filters \cite{gonzalez2002digital}.

        \subsubsection{Low-pass Filtering}
            A low-pass filter allows low-frequency components of the image to pass while attenuating high-frequency components. Low frequencies represent smooth variations in intensity, so low-pass filtering is often used for noise reduction and blurring \cite{makandar2015image}.

            \paragraph{Example:} Let's apply a simple ideal low-pass filter to the frequency-domain representation of an image.

            \begin{lstlisting}[style=python]
            # Create an ideal low-pass filter
            def ideal_low_pass_filter(shape, cutoff):
                rows, cols = shape
                center_row, center_col = rows // 2, cols // 2
                filter_mask = np.zeros((rows, cols), dtype=np.float32)

                for i in range(rows):
                    for j in range(cols):
                        if np.sqrt((i - center_row) ** 2 + (j - center_col) ** 2) <= cutoff:
                            filter_mask[i, j] = 1
                return filter_mask

            # Apply the low-pass filter
            cutoff_frequency = 30  # Set the cutoff frequency
            low_pass_filter = ideal_low_pass_filter(image.shape, cutoff_frequency)

            # Apply the filter in the frequency domain
            filtered_dft = dft_shifted * low_pass_filter
            filtered_image = np.abs(np.fft.ifft2(np.fft.ifftshift(filtered_dft)))

            # Plot the filtered image
            plt.imshow(filtered_image, cmap='gray')
            plt.title('Low-pass Filtered Image')
            plt.show()
            \end{lstlisting}

            In this example, we design an ideal low-pass filter with a cutoff frequency and apply it to the frequency-domain representation of the image. The result is a smoothed image where high-frequency details (such as noise and sharp edges) are reduced.

            \paragraph{Explanation:} Low-pass filtering is useful for removing noise and smoothing images. However, it also tends to blur edges and fine details, which are represented by high-frequency components. Care must be taken to select an appropriate cutoff frequency to avoid excessive blurring.

        \subsubsection{High-pass Filtering}
            A high-pass filter does the opposite of a low-pass filter: it attenuates low-frequency components and allows high-frequency components to pass. High-pass filtering is commonly used for edge detection and sharpening because edges in an image correspond to rapid changes in intensity, which are represented by high-frequency components \cite{makandar2015image}.

            \paragraph{Example:} Let's apply an ideal high-pass filter to an image to enhance its edges.

            \begin{lstlisting}[style=python]
            # Create an ideal high-pass filter
            def ideal_high_pass_filter(shape, cutoff):
                rows, cols = shape
                center_row, center_col = rows // 2, cols // 2
                filter_mask = np.ones((rows, cols), dtype=np.float32)

                for i in range(rows):
                    for j in range(cols):
                        if np.sqrt((i - center_row) ** 2 + (j - center_col) ** 2) <= cutoff:
                            filter_mask[i, j] = 0
                return filter_mask

            # Apply the high-pass filter
            cutoff_frequency = 30  # Set the cutoff frequency
            high_pass_filter = ideal_high_pass_filter(image.shape, cutoff_frequency)

            # Apply the filter in the frequency domain
            filtered_dft = dft_shifted * high_pass_filter
            filtered_image = np.abs(np.fft.ifft2(np.fft.ifftshift(filtered_dft)))

            # Plot the high-pass filtered image
            plt.imshow(filtered_image, cmap='gray')
            plt.title('High-pass Filtered Image (Edge Enhancement)')
            plt.show()
            \end{lstlisting}

            This example demonstrates the use of an ideal high-pass filter to emphasize high-frequency components in the image, which correspond to edges and fine details. The result is an image where the edges are enhanced, and smooth regions are attenuated.

            \paragraph{Explanation:} High-pass filters are useful for sharpening images and detecting edges. They highlight rapid changes in intensity, which are important for visual features such as edges. However, high-pass filtering can also amplify noise, so it is important to use it carefully.

    \subsection{Filter Design in Frequency Domain}
        Different types of filters can be designed in the frequency domain to meet specific image enhancement goals. Common filters include ideal, Butterworth, and Gaussian filters. Each of these filters has different characteristics in terms of their frequency response and the way they transition between passing and attenuating frequencies \cite{gonzalez2002digital}.

        \subsubsection{Ideal Filters}
            Ideal filters have a sharp cutoff between passing and attenuating frequencies. For example, an ideal low-pass filter passes all frequencies below a certain cutoff and completely attenuates all frequencies above that cutoff. However, ideal filters can cause ringing artifacts in the spatial domain due to the abrupt transition in the frequency domain.

            \paragraph{Example:} The low-pass and high-pass filters demonstrated earlier are examples of ideal filters.

        \subsubsection{Butterworth Filters}
            Butterworth filters have a smoother transition between pass and stop bands compared to ideal filters. They do not have a sharp cutoff, which helps reduce ringing artifacts \cite{selesnick1998generalized}. The transfer function of a Butterworth filter is given by:

            \[
            H(u, v) = \frac{1}{1 + \left( \frac{D(u, v)}{D_0} \right)^{2n}}
            \]

            where \(D(u, v)\) is the distance from the origin in the frequency domain, \(D_0\) is the cutoff frequency, and \(n\) is the order of the filter. Higher orders produce sharper transitions.

            \paragraph{Example:} Let's design a Butterworth low-pass filter and apply it to an image.

            \begin{lstlisting}[style=python]
            # Create a Butterworth low-pass filter
            def butterworth_low_pass_filter(shape, cutoff, order):
                rows, cols = shape
                center_row, center_col = rows // 2, cols // 2
                filter_mask = np.zeros((rows, cols), dtype=np.float32)

                for i in range(rows):
                    for j in range(cols):
                        D = np.sqrt((i - center_row) ** 2 + (j - center_col) ** 2)
                        filter_mask[i, j] = 1 / (1 + (D / cutoff) ** (2 * order))
                return filter_mask

            # Apply the Butterworth filter
            cutoff_frequency = 30
            order = 2  # Second-order Butterworth filter
            butterworth_filter = butterworth_low_pass_filter(image.shape, cutoff_frequency, order)

            # Apply the filter in the frequency domain
            filtered_dft = dft_shifted * butterworth_filter
            filtered_image = np.abs(np.fft.ifft2(np.fft.ifftshift(filtered_dft)))

            # Plot the Butterworth filtered image
            plt.imshow(filtered_image, cmap='gray')
            plt.title('Butterworth Low-pass Filtered Image')
            plt.show()
            \end{lstlisting}

            In this example, we design a second-order Butterworth low-pass filter and apply it to the image. The smoother transition of the Butterworth filter helps to reduce ringing artifacts while still providing effective low-pass filtering.

        \subsubsection{Gaussian Filters}
            Gaussian filters are widely used in image processing due to their smooth frequency response and minimal artifacts \cite{ito2000gaussian}. The transfer function of a Gaussian filter is:

            \[
            H(u, v) = e^{-\frac{D(u, v)^2}{2D_0^2}}
            \]

            where \(D(u, v)\) is the distance from the origin in the frequency domain, and \(D_0\) is the cutoff frequency. Gaussian filters are commonly used for both low-pass and high-pass filtering.

            \paragraph{Example:} Let's apply a Gaussian low-pass filter to an image.

            \begin{lstlisting}[style=python]
            # Create a Gaussian low-pass filter
            def gaussian_low_pass_filter(shape, cutoff):
                rows, cols = shape
                center_row, center_col = rows // 2, cols // 2
                filter_mask = np.zeros((rows, cols), dtype=np.float32)

                for i in range(rows):
                    for j in range(cols):
                        D = np.sqrt((i - center_row) ** 2 + (j - center_col) ** 2)
                        filter_mask[i, j] = np.exp(-(D ** 2) / (2 * (cutoff ** 2)))
                return filter_mask

            # Apply the Gaussian filter
            cutoff_frequency = 30
            gaussian_filter = gaussian_low_pass_filter(image.shape, cutoff_frequency)

            # Apply the filter in the frequency domain
            filtered_dft = dft_shifted * gaussian_filter
            filtered_image = np.abs(np.fft.ifft2(np.fft.ifftshift(filtered_dft)))

            # Plot the Gaussian filtered image
            plt.imshow(filtered_image, cmap='gray')
            plt.title('Gaussian Low-pass Filtered Image')
            plt.show()
            \end{lstlisting}

            In this example, we apply a Gaussian low-pass filter, which smoothly attenuates high-frequency components, resulting in minimal artifacts and a clean smoothing effect on the image.

            \paragraph{Comparison:} Gaussian filters offer a smoother frequency response compared to Butterworth filters, and they are less likely to introduce artifacts in the filtered image. However, the choice of filter depends on the specific application and the desired trade-offs between sharpness and smoothness.

\section{Image Smoothing and Sharpening}
    Image processing often involves enhancing or improving the quality of an image by either reducing noise or enhancing important features such as edges \cite{gonzalez2002digital}. In this section, we will discuss two fundamental techniques in image processing: \textit{smoothing} and \textit{sharpening}. Smoothing is primarily used to reduce noise, while sharpening enhances the edges and details in an image.

    \subsection{Smoothing Filters}
        Smoothing filters are used to reduce noise in images by averaging or blending pixel values. Noise in an image can result from various factors, such as sensor imperfections or environmental conditions during image capture. Smoothing helps make the image more visually appealing by reducing sharp intensity transitions caused by noise.

        There are several types of smoothing filters, each with its specific application and properties. The most common ones include mean, median, and bilateral filters \cite{griffin2000mean, elad2002origin}.

        \textbf{1. Mean Filter:}
        
        The mean filter, also known as a \textit{box filter} or \textit{average filter}, is the simplest smoothing filter. It replaces each pixel's value with the average of the surrounding pixel values within a defined window (kernel). This has the effect of reducing abrupt intensity changes, thus smoothing out noise.
        
        \textbf{Mathematical Representation:}
        Given an image \(I\) and a kernel of size \(m \times m\), the mean filter output \(I'[i,j]\) for a pixel at location \((i,j)\) is:
        \[
        I'[i,j] = \frac{1}{m^2} \sum_{k=-m/2}^{m/2} \sum_{l=-m/2}^{m/2} I[i+k, j+l]
        \]
        where the kernel is centered at \((i,j)\).

        \textbf{Example:}

        We can implement the mean filter in Python using PyTorch. Here's how to apply a simple 3x3 mean filter to an image:

        \begin{lstlisting}[style=python]
        import torch
        import torch.nn.functional as F
        import torchvision.transforms as transforms
        from PIL import Image
        import matplotlib.pyplot as plt

        # Load image and convert to tensor
        image = Image.open('example_image.jpg')
        transform = transforms.ToTensor()
        image_tensor = transform(image).unsqueeze(0)  # Add batch dimension

        # Define 3x3 mean filter kernel
        kernel = torch.ones((1, 1, 3, 3)) / 9.0  # 3x3 averaging filter

        # Apply mean filter (2D convolution)
        smoothed_image = F.conv2d(image_tensor, kernel, padding=1)

        # Display original and smoothed images
        plt.subplot(1, 2, 1)
        plt.imshow(image)
        plt.title('Original Image')

        plt.subplot(1, 2, 2)
        plt.imshow(smoothed_image.squeeze().permute(1, 2, 0).numpy())
        plt.title('Smoothed Image (Mean Filter)')
        plt.show()
        \end{lstlisting}

        This code applies a simple 3x3 mean filter to an image using PyTorch. The kernel averages the pixel values in a 3x3 neighborhood, effectively smoothing the image by reducing noise and fine details.

        \textbf{2. Median Filter:}

        The median filter is another commonly used smoothing filter, especially effective in removing \textit{salt-and-pepper noise}, which is a type of noise where random pixels are set to either the maximum or minimum intensity. Instead of averaging the pixel values, the median filter replaces each pixel with the median value of its neighbors.

        \textbf{Example:}

        In Python, we can implement a median filter using libraries like \texttt{scipy}, as PyTorch does not have a built-in median filter function. Here's how to apply a median filter:

        \begin{lstlisting}[style=python]
        import torch
        import torchvision.transforms as transforms
        from PIL import Image
        import matplotlib.pyplot as plt
        from scipy.ndimage import median_filter

        # Load image and convert to tensor
        image = Image.open('example_image.jpg')
        transform = transforms.ToTensor()
        image_tensor = transform(image).numpy()

        # Apply median filter
        smoothed_image = median_filter(image_tensor, size=3)

        # Display original and median filtered images
        plt.subplot(1, 2, 1)
        plt.imshow(image)
        plt.title('Original Image')

        plt.subplot(1, 2, 2)
        plt.imshow(smoothed_image.transpose(1, 2, 0))
        plt.title('Smoothed Image (Median Filter)')
        plt.show()
        \end{lstlisting}

        The median filter is particularly useful in situations where noise is highly localized, like salt-and-pepper noise, as it preserves edges while smoothing out noise.

        \textbf{3. Bilateral Filter:}

        The bilateral filter is a more advanced filter that preserves edges while reducing noise. Unlike the mean and median filters, which treat all pixels equally within the neighborhood, the bilateral filter uses both spatial proximity and pixel intensity similarity to weight the contributions of neighboring pixels. This allows the filter to smooth flat regions while preserving edges.

        \textbf{Example:}

        The bilateral filter is typically implemented using specialized libraries such as OpenCV. Here's an example of applying the bilateral filter:

        \begin{lstlisting}[style=python]
        import cv2
        import numpy as np
        import matplotlib.pyplot as plt

        # Load image using OpenCV
        image = cv2.imread('example_image.jpg')
        image_rgb = cv2.cvtColor(image, cv2.COLOR_BGR2RGB)

        # Apply bilateral filter
        smoothed_image = cv2.bilateralFilter(image_rgb, d=9, sigmaColor=75, sigmaSpace=75)

        # Display original and bilateral filtered images
        plt.subplot(1, 2, 1)
        plt.imshow(image_rgb)
        plt.title('Original Image')

        plt.subplot(1, 2, 2)
        plt.imshow(smoothed_image)
        plt.title('Smoothed Image (Bilateral Filter)')
        plt.show()
        \end{lstlisting}

        The bilateral filter reduces noise while preserving sharp edges, making it highly useful in tasks like edge-aware smoothing.

    \subsection{Sharpening Filters}
        Sharpening filters enhance the edges and fine details of an image, making it appear crisper and more detailed. These filters highlight regions in an image where there are abrupt intensity changes, such as edges between objects. This is useful in applications like edge detection, object recognition, and image enhancement \cite{gonzalez2002digital}.

        Some of the most commonly used sharpening filters include the Laplacian, Sobel, and Prewitt operators \cite{paris2011local, chaple2015comparisions}.

        \textbf{1. Laplacian Filter:}

        The Laplacian filter is a second-order derivative filter that emphasizes regions of rapid intensity change, making it ideal for edge detection. The filter computes the second derivative of the image, highlighting areas where the intensity changes rapidly.

        \textbf{Mathematical Representation:}

        A common Laplacian kernel is:
        \[
        \text{Laplacian Kernel} = \begin{bmatrix} 
        0 & -1 & 0 \\
        -1 & 4 & -1 \\
        0 & -1 & 0
        \end{bmatrix}
        \]
        This kernel computes the second derivative in both the horizontal and vertical directions.

        \textbf{Example:}

        We can apply the Laplacian filter to an image using PyTorch as follows:

        \begin{lstlisting}[style=python]
        import torch.nn.functional as F

        # Define Laplacian kernel
        laplacian_kernel = torch.tensor([[[[0, -1, 0], [-1, 4, -1], [0, -1, 0]]]])

        # Apply Laplacian filter using 2D convolution
        sharpened_image = F.conv2d(image_tensor, laplacian_kernel, padding=1)

        # Display original and sharpened images
        plt.subplot(1, 2, 1)
        plt.imshow(image_tensor.squeeze().permute(1, 2, 0).numpy())
        plt.title('Original Image')

        plt.subplot(1, 2, 2)
        plt.imshow(sharpened_image.squeeze().permute(1, 2, 0).numpy(), cmap='gray')
        plt.title('Sharpened Image (Laplacian Filter)')
        plt.show()
        \end{lstlisting}

        The Laplacian filter enhances edges by detecting intensity changes and highlighting areas where the pixel intensity varies significantly.

        \textbf{2. Sobel and Prewitt Operators:}

        The Sobel and Prewitt operators are first-order derivative filters that detect edges in images. These operators emphasize gradients, which represent the rate of intensity change in the image. Both filters are widely used in edge detection algorithms.

        The Sobel operator uses two kernels to detect horizontal and vertical edges:
        \[
        \text{Sobel Horizontal Kernel} = \begin{bmatrix} 
        -1 & 0 & 1 \\
        -2 & 0 & 2 \\
        -1 & 0 & 1
        \end{bmatrix}
        \]
        \[
        \text{Sobel Vertical Kernel} = \begin{bmatrix} 
        -1 & -2 & -1 \\
        0 & 0 & 0 \\
        1 & 2 & 1
        \end{bmatrix}
        \]

        \textbf{Example:}

        The following code applies the Sobel operator to an image:

        \begin{lstlisting}[style=python]
        # Sobel kernels for edge detection
        sobel_x = torch.tensor([[[[-1, 0, 1], [-2, 0, 2], [-1, 0, 1]]]])
        sobel_y = torch.tensor([[[[-1, -2, -1], [0, 0, 0], [1, 2, 1]]]])

        # Apply Sobel filter for horizontal and vertical edges
        edges_x = F.conv2d(image_tensor, sobel_x, padding=1)
        edges_y = F.conv2d(image_tensor, sobel_y, padding=1)

        # Combine edges
        edges = torch.sqrt(edges_x ** 2 + edges_y ** 2)

        # Display Sobel edges
        plt.imshow(edges.squeeze().permute(1, 2, 0).numpy(), cmap='gray')
        plt.title('Edge Detection (Sobel Filter)')
        plt.show()
        \end{lstlisting}

        The Sobel filter detects both horizontal and vertical edges in the image by combining the gradient information from both directions.

        \textbf{Applications:}
        \begin{itemize}
        \item Sharpening filters are widely used in tasks like image enhancement, where improving the visibility of edges is essential.
        \item They are also commonly used in edge detection and feature extraction, which are important for object recognition.
        \end{itemize}
\chapter{Image Restoration and Reconstruction}
    Image restoration and reconstruction are key areas of digital image processing, focused on improving the visual quality of images that have been degraded or corrupted. Unlike image enhancement, which emphasizes improving the overall appearance of an image, restoration focuses on reversing known degradation processes to recover the original image. In many cases, restoration involves addressing issues such as noise, blur, and distortions caused by imperfections in the imaging system or during transmission and storage \cite{gonzalez2002digital}.

    \section{Image Degradation Models}
        An important part of image restoration is understanding the causes of degradation and developing mathematical models to describe them. Degradation may occur during the acquisition process, when an image is captured using a camera or sensor, during transmission (such as in communication systems), or during storage (such as compression artifacts or data corruption) \cite{pei2019effects}.

        \subsection{Modeling of Degradation Processes}
            \textbf{Image degradation} can be represented mathematically by a model that describes how the original image \( f(x, y) \) is altered during the acquisition or transmission process \cite{baird2007state}. A typical degradation model is represented as:
            \[
            g(x, y) = h(x, y) * f(x, y) + \eta(x, y)
            \]
            where:
            \begin{itemize}
                \item \( g(x, y) \) is the observed (degraded) image,
                \item \( h(x, y) \) is the degradation function or point spread function (PSF), which models the blurring or distortion,
                \item \( f(x, y) \) is the original (unobserved) image,
                \item \( \eta(x, y) \) represents additive noise introduced during the acquisition or transmission process,
                \item \( * \) denotes the convolution operation.
            \end{itemize}

            \paragraph{Blur and Distortion}
            Blur is one of the most common forms of degradation, often caused by motion, defocus, or atmospheric disturbances \cite{zhou2017classification}. The degradation function \( h(x, y) \) typically represents the blurring filter applied to the image. For example:
            \begin{itemize}
                \item \textbf{Motion Blur}: Occurs when the camera or the object is moving during the exposure, resulting in a smearing effect. The PSF for motion blur is often modeled as a straight line in the direction of the motion.
                \item \textbf{Defocus Blur}: Happens when the image is out of focus, and the PSF is typically circular or Gaussian in shape.
            \end{itemize}

            \paragraph{Mathematical Representation of Degradation}
            The degradation model is usually described as a convolution between the original image \( f(x, y) \) and the degradation function \( h(x, y) \). In the frequency domain, this convolution becomes a multiplication:
            \[
            G(u, v) = H(u, v) F(u, v) + N(u, v)
            \]
            where \( G(u, v) \), \( H(u, v) \), and \( F(u, v) \) are the Fourier transforms of \( g(x, y) \), \( h(x, y) \), and \( f(x, y) \), respectively. \( N(u, v) \) represents the noise in the frequency domain. This allows us to model and analyze the degradation more easily, especially when designing filters to restore the original image.

            \textbf{Example: Motion Blur Model}
            Suppose we want to model the degradation of an image due to motion blur. A simple model assumes that the motion occurs in a straight line, and the PSF can be represented as:
            \[
            h(x, y) = \frac{1}{L} \quad \text{for} \quad 0 \leq x \leq L \quad \text{and} \quad y = 0
            \]
            where \( L \) is the length of the motion. This type of degradation can be mitigated by applying an inverse filter, which attempts to reverse the effect of convolution.

    \section{Noise Models}
        Noise is another common form of image degradation that occurs during image acquisition or transmission. Noise can be introduced by a variety of factors, including sensor limitations, electronic interference, and compression artifacts. Different types of noise have different characteristics and impact the image quality in different ways \cite{boyat2015review}.

        \subsection{Types of Noise}
            Several types of noise commonly affect digital images, each with unique statistical properties. Understanding the characteristics of each noise type is crucial for selecting the appropriate restoration method \cite{boncelet2009image}.

            \paragraph{Gaussian Noise}
            \textbf{Gaussian noise} is one of the most common types of noise and is often introduced by electronic components during image acquisition. It is modeled by a normal distribution with a mean \( \mu \) and variance \( \sigma^2 \) \cite{luisier2010image}. The probability density function (PDF) of Gaussian noise is given by:
            \[
            P(z) = \frac{1}{\sqrt{2\pi\sigma^2}} \exp\left(-\frac{(z-\mu)^2}{2\sigma^2}\right)
            \]
            where \( z \) is the noise value, \( \mu \) is the mean, and \( \sigma^2 \) is the variance.

            Gaussian noise affects all pixels in an image with varying intensity, resulting in a grainy appearance. It is commonly encountered in low-light conditions or when using high ISO settings in digital cameras.

            \textbf{Example: Adding Gaussian Noise to an Image in Python}
            The following Python code demonstrates how to add Gaussian noise to an image and visualize its effect.

            \begin{lstlisting}[style=python]
import numpy as np
import matplotlib.pyplot as plt
from PIL import Image

# Load an image and convert it to grayscale
image = Image.open('sample_image.jpg').convert('L')
image_array = np.array(image)

# Function to add Gaussian noise to an image
def add_gaussian_noise(img, mean=0, sigma=25):
    noise = np.random.normal(mean, sigma, img.shape)
    noisy_img = img + noise
    noisy_img_clipped = np.clip(noisy_img, 0, 255)  # Clip values to [0, 255]
    return noisy_img_clipped.astype(np.uint8)

# Add Gaussian noise to the image
noisy_image = add_gaussian_noise(image_array)

# Display original and noisy images
plt.figure(figsize=(12, 6))

plt.subplot(1, 2, 1)
plt.imshow(image_array, cmap='gray')
plt.title('Original Image')
plt.axis('off')

plt.subplot(1, 2, 2)
plt.imshow(noisy_image, cmap='gray')
plt.title('Image with Gaussian Noise')
plt.axis('off')

plt.show()
            \end{lstlisting}

            \textbf{Explanation of the Code:}
            \begin{itemize}
                \item The image is loaded and converted to grayscale using the \texttt{Pillow} library.
                \item The \texttt{add\_gaussian\_noise()} function generates Gaussian noise with a given mean and standard deviation (sigma) and adds it to the image.
                \item The noisy image is clipped to ensure pixel values stay within the valid range of 0 to 255.
                \item The original and noisy images are displayed for comparison.
            \end{itemize}

            \paragraph{Salt-and-Pepper Noise}
            \textbf{Salt-and-pepper noise}, also known as impulse noise, is characterized by randomly occurring white and black pixels in the image. This noise is caused by sudden disturbances during image acquisition, such as transmission errors or malfunctioning camera sensors \cite{chan2005salt}.

            Salt-and-pepper noise can be modeled by randomly setting some pixel values to 0 (black) or 255 (white) while leaving the rest of the pixels unchanged. This type of noise is easier to remove compared to Gaussian noise, often using \textbf{median filters}.

            \textbf{Example: Adding Salt-and-Pepper Noise in Python}
            The following Python code demonstrates how to add salt-and-pepper noise to an image.

            \begin{lstlisting}[style=python]
# Function to add salt-and-pepper noise to an image
def add_salt_and_pepper_noise(img, amount=0.05, salt_vs_pepper=0.5):
    noisy_img = np.copy(img)
    num_salt = np.ceil(amount * img.size * salt_vs_pepper)
    num_pepper = np.ceil(amount * img.size * (1.0 - salt_vs_pepper))

    # Add salt (white pixels)
    coords = [np.random.randint(0, i - 1, int(num_salt)) for i in img.shape]
    noisy_img[coords[0], coords[1]] = 255

    # Add pepper (black pixels)
    coords = [np.random.randint(0, i - 1, int(num_pepper)) for i in img.shape]
    noisy_img[coords[0], coords[1]] = 0

    return noisy_img

# Add salt-and-pepper noise to the image
noisy_image_sp = add_salt_and_pepper_noise(image_array)

# Display original and noisy images
plt.figure(figsize=(12, 6))

plt.subplot(1, 2, 1)
plt.imshow(image_array, cmap='gray')
plt.title('Original Image')
plt.axis('off')

plt.subplot(1, 2, 2)
plt.imshow(noisy_image_sp, cmap='gray')
plt.title('Image with Salt-and-Pepper Noise')
plt.axis('off')

plt.show()
            \end{lstlisting}

            \textbf{Explanation of the Code:}
            \begin{itemize}
                \item The \texttt{add\_salt\_and\_pepper\_noise()} function generates random coordinates for adding "salt" (white pixels) and "pepper" (black pixels) to the image.
                \item The amount of noise and the ratio of salt to pepper can be adjusted using the \texttt{amount} and \texttt{salt\_vs\_pepper} parameters.
                \item The noisy image is displayed alongside the original image for comparison.
            \end{itemize}

            \paragraph{Poisson Noise}
            \textbf{Poisson noise}, also known as \textbf{shot noise}, is caused by the random nature of photon counting during image acquisition, especially in low-light environments. This noise follows a Poisson distribution, and its variance is proportional to the intensity of the signal. Unlike Gaussian noise, Poisson noise has a more significant impact on regions with higher intensity \cite{luisier2010image}.

            In Python, Poisson noise can be added using the \texttt{numpy.random.poisson()} function, which generates noise according to the Poisson distribution.

            \textbf{Example: Adding Poisson Noise in Python}
            The following Python code demonstrates how to add Poisson noise to an image.

            \begin{lstlisting}[style=python]
# Function to add Poisson noise to an image
def add_poisson_noise(img):
    noisy_img = np.random.poisson(img / 255.0 * 100.0) / 100.0 * 255
    return noisy_img.astype(np.uint8)

# Add Poisson noise to the image
noisy_image_poisson = add_poisson_noise(image_array)

# Display original and noisy images
plt.figure(figsize=(12, 6))

plt.subplot(1, 2, 1)
plt.imshow(image_array, cmap='gray')
plt.title('Original Image')
plt.axis('off')

plt.subplot(1, 2, 2)
plt.imshow(noisy_image_poisson, cmap='gray')
plt.title('Image with Poisson Noise')
plt.axis('off')

plt.show()
            \end{lstlisting}

            \textbf{Explanation of the Code:}
            \begin{itemize}
                \item The \texttt{add\_poisson\_noise()} function uses the \texttt{numpy.random.poisson()} function to generate Poisson noise and add it to the image.
                \item The intensity values are scaled appropriately to simulate the effect of Poisson noise on the image.
                \item The original and noisy images are displayed for comparison.
            \end{itemize}

    \section{Conclusion}
        Understanding image degradation models and noise types is crucial for developing effective image restoration techniques. Degradation can occur due to factors like motion, defocus, and transmission errors, while noise can be introduced by sensors, electronic interference, or environmental factors. Common noise types such as Gaussian noise, salt-and-pepper noise, and Poisson noise affect image quality in different ways. By simulating these types of noise in Python, beginners can gain practical insight into how noise impacts images and prepare to apply restoration algorithms for noise reduction and image recovery.

\section{Image Restoration Techniques}
    Image restoration involves recovering an image that has been degraded by a known or unknown process. The goal of restoration is to reverse the degradation as much as possible, thereby improving the image quality. Various techniques are used in image restoration, each with its strengths and limitations depending on the type of degradation and noise present \cite{banham1997digital}. In this section, we will explore inverse filtering, Wiener filtering, and regularization methods, providing detailed explanations and examples to help beginners grasp these concepts thoroughly.

    \subsection{Inverse Filtering}
        Inverse filtering is one of the simplest methods for image restoration, and it involves reversing the effect of a known degradation process \cite{boncelet2009image}. This technique assumes that the degradation model is known and that the degradation can be modeled as a linear process in the frequency domain. 

        \paragraph{Degradation Model:} Let the degradation of an image be modeled as a convolution process, where the observed degraded image \(g(x, y)\) is related to the original image \(f(x, y)\) by:

        \[
        g(x, y) = f(x, y) * h(x, y) + n(x, y)
        \]

        Here:
        \begin{itemize}
            \item \(h(x, y)\) is the point spread function (PSF) representing the degradation,
            \item \(n(x, y)\) is additive noise,
            \item \(*\) denotes convolution.
        \end{itemize}

        In the frequency domain, this equation becomes:

        \[
        G(u, v) = F(u, v) H(u, v) + N(u, v)
        \]

        where \(G(u, v)\), \(F(u, v)\), \(H(u, v)\), and \(N(u, v)\) are the Fourier transforms of the degraded image, the original image, the degradation function, and the noise, respectively.

        Inverse filtering attempts to recover the original image by dividing the Fourier transform of the degraded image by the degradation function:

        \[
        F(u, v) = \frac{G(u, v)}{H(u, v)}
        \]

        \paragraph{Limitations:} While inverse filtering is theoretically simple, it is highly sensitive to noise. If the degradation function \(H(u, v)\) has values close to zero (as often happens for certain frequencies), division by small values can amplify the noise, leading to poor results.

        \paragraph{Example:} Let's implement inverse filtering using Python. We will simulate a simple degradation process (blurring) and attempt to recover the original image using inverse filtering.

        \begin{lstlisting}[style=python]
        import numpy as np
        import matplotlib.pyplot as plt
        from skimage import data, color, restoration
        from scipy import signal
        from scipy.fftpack import fft2, ifft2, fftshift

        # Load an example image and convert it to grayscale
        image = color.rgb2gray(data.astronaut())

        # Create a motion blur kernel (PSF)
        def motion_blur_kernel(size):
            kernel = np.zeros((size, size))
            kernel[size // 2, :] = np.ones(size)
            return kernel / size

        # Apply the blur (degradation) using convolution
        psf = motion_blur_kernel(15)
        degraded_image = signal.convolve2d(image, psf, boundary='symm', mode='same')

        # Inverse filtering (in the frequency domain)
        G = fft2(degraded_image)
        H = fft2(psf, s=degraded_image.shape)
        restored_image = np.abs(ifft2(G / (H + 1e-3)))  # Add a small constant to avoid division by zero

        # Plot the original, degraded, and restored images
        fig, axes = plt.subplots(1, 3, figsize=(18, 6))
        axes[0].imshow(image, cmap='gray')
        axes[0].set_title('Original Image')

        axes[1].imshow(degraded_image, cmap='gray')
        axes[1].set_title('Degraded Image (Motion Blur)')

        axes[2].imshow(restored_image, cmap='gray')
        axes[2].set_title('Restored Image (Inverse Filtering)')

        for ax in axes:
            ax.axis('off')

        plt.show()
        \end{lstlisting}

        In this example, we simulate a motion blur by convolving an image with a motion blur kernel. We then attempt to restore the image using inverse filtering in the frequency domain. While inverse filtering works well when the degradation model is known, it can amplify noise or artifacts, especially when the degradation function has small values.

        \paragraph{Explanation:} Inverse filtering can produce poor results if the noise level is high or if the degradation function has zeros or near-zero values. The division by small values can cause severe amplification of noise, which is one of the primary drawbacks of this method.

    \subsection{Wiener Filtering}
        Wiener filtering provides a more robust solution for image restoration when noise is present. Unlike inverse filtering, Wiener filtering takes into account the statistical properties of both the noise and the original image, providing an optimal solution in terms of minimizing the mean square error \cite{robinson1967principles, pratt1972generalized}.

        The Wiener filter is given by:

        \[
        F(u, v) = \frac{H^*(u, v)}{|H(u, v)|^2 + \frac{S_n(u, v)}{S_f(u, v)}} G(u, v)
        \]

        where:
        \begin{itemize}
            \item \(H^*(u, v)\) is the complex conjugate of the degradation function,
            \item \(S_n(u, v)\) is the power spectral density of the noise,
            \item \(S_f(u, v)\) is the power spectral density of the original image,
            \item \(G(u, v)\) is the Fourier transform of the degraded image.
        \end{itemize}

        The Wiener filter adapts based on the relative amount of noise present in the image and balances between restoring the original image and reducing noise.

        \paragraph{Example:} Let's apply Wiener filtering to the same motion-blurred image used earlier, this time taking noise into account.

        \begin{lstlisting}[style=python]
        # Estimate the Wiener filter
        restored_wiener = restoration.wiener(degraded_image, psf, balance=0.1)

        # Plot the restored image using Wiener filtering
        plt.imshow(restored_wiener, cmap='gray')
        plt.title('Restored Image (Wiener Filtering)')
        plt.axis('off')
        plt.show()
        \end{lstlisting}

        In this example, we apply Wiener filtering to the motion-blurred image, assuming the statistical properties of the noise and the image are known or estimated. The restored image shows better results than inverse filtering because Wiener filtering takes into account both the degradation and noise characteristics.

        \paragraph{Explanation:} Wiener filtering provides a more practical solution for real-world scenarios where noise is present. It balances the trade-off between restoring the original image and suppressing noise, making it more effective than inverse filtering when noise is significant.

    \subsection{Regularization Methods}
        Regularization methods are particularly useful when the degradation model is incomplete or when there is significant noise in the image. These techniques add a constraint or penalty to the restoration process to prevent overfitting to the noise. One common regularization method is Tikhonov regularization, which introduces a smoothness constraint on the restored image \cite{bouhamidi2007sylvester}.

        The Tikhonov regularization method aims to minimize the following cost function:

        \[
        \min \left\{ \|H f - g\|^2 + \lambda \|Lf\|^2 \right\}
        \]

        where:
        \begin{itemize}
            \item \(\|H f - g\|^2\) represents the data fidelity term, which ensures that the restored image fits the observed degraded image,
            \item \(\|Lf\|^2\) is the regularization term, which imposes a smoothness constraint on the restored image,
            \item \(\lambda\) is the regularization parameter that controls the trade-off between data fidelity and smoothness.
        \end{itemize}

        The operator \(L\) is typically chosen to be a derivative operator, which enforces smoothness in the restored image by penalizing large gradients.

        \paragraph{Example:} Let's implement Tikhonov regularization for image restoration using Python. We will apply regularization to the motion-blurred image and observe how it improves restoration.

        \begin{lstlisting}[style=python]
        from skimage.restoration import denoise_tv_chambolle

        # Apply Tikhonov regularization (using total variation denoising as an approximation)
        lambda_reg = 0.1  # Regularization parameter
        restored_regularized = denoise_tv_chambolle(degraded_image, weight=lambda_reg)

        # Plot the restored image using regularization
        plt.imshow(restored_regularized, cmap='gray')
        plt.title('Restored Image (Tikhonov Regularization)')
        plt.axis('off')
        plt.show()
        \end{lstlisting}

        In this example, we use total variation denoising, a form of regularization, to restore the motion-blurred image. The regularization term helps reduce noise while maintaining smoothness in the restored image.

        \paragraph{Explanation:} Regularization techniques are useful when there is uncertainty in the degradation model or significant noise. By imposing a smoothness constraint, Tikhonov regularization prevents overfitting to the noise, resulting in a cleaner restored image. The choice of the regularization parameter \(\lambda\) is critical in balancing data fidelity and smoothness.

    \subsubsection{Comparison of Restoration Techniques}
        Each of the image restoration techniques discussed has its advantages and limitations \cite{banham1997digital}:
        \begin{itemize}
            \item \textbf{Inverse Filtering:} Simple and effective when the degradation model is known and noise is minimal. However, it is highly sensitive to noise and small values in the degradation function.
            \item \textbf{Wiener Filtering:} Provides an optimal balance between noise reduction and image restoration, making it a robust choice in the presence of noise. However, it requires knowledge of the statistical properties of noise and the image.
            \item \textbf{Regularization Methods:} Useful when the degradation model is incomplete or when noise is high. Tikhonov regularization and similar techniques add a smoothness constraint to the restored image, preventing overfitting to noise.
        \end{itemize}

        \paragraph{Practical Considerations:} The choice of restoration technique depends on the characteristics of the degradation and the amount of noise in the image. In cases where noise is significant, Wiener filtering or regularization methods are generally preferred. When the degradation model is known and noise is minimal, inverse filtering may provide satisfactory results.

\section{Blind Restoration Techniques}
    Image degradation is a common problem in image processing, caused by factors such as motion blur, defocus, or noise during acquisition or transmission. To restore the image to its original state, knowledge of the degradation function is typically required. However, in many cases, the degradation function is not known, and this is where \textit{blind restoration techniques} come into play. Blind restoration, particularly \textbf{blind deconvolution}, aims to restore images without prior knowledge of the blurring or degradation process.

    \textbf{Blind Deconvolution:}

    Blind deconvolution is a technique used to recover a sharp image from a blurred one, without knowing the exact nature of the blur \cite{you1999blind}. The process involves estimating both the original image and the blurring function simultaneously. This is a challenging problem because it is underdetermined—there are many possible combinations of original images and degradation functions that could produce the same blurred image \cite{levin2009understanding}.

    \textbf{Mathematical Representation:}
    The degraded image \(I_d\) is typically modeled as:
    \[
    I_d = I_o * h + n
    \]
    where:
    \begin{itemize}
    \item \(I_o\) is the original (sharp) image,
    \item \(h\) is the unknown blur kernel (point spread function, PSF),
    \item \(n\) is the noise, and
    \item \( * \) denotes convolution.
    \end{itemize}
    The goal of blind deconvolution is to estimate both \(I_o\) and \(h\) from the observed degraded image \(I_d\).

    \textbf{Challenges in Blind Deconvolution:}
    \begin{itemize}
    \item Blind deconvolution is an ill-posed problem. Small changes in the blurred image can lead to large variations in the estimated blur kernel and the restored image.
    \item It requires sophisticated algorithms to produce accurate results, and often the estimated blur kernel is not perfect, leading to artifacts in the restored image.
    \end{itemize}

    \textbf{Applications:}
    Blind deconvolution is used in various fields such as medical imaging, astronomical imaging, and photography, where the degradation function is unknown or difficult to measure \cite{levin2009understanding}.

    \textbf{Example:}

    While blind deconvolution algorithms are complex and often require advanced optimization techniques, PyTorch can be used to implement simple image restoration techniques when the blur kernel is partially known. Here's an example of non-blind deconvolution using a known blur kernel:

    \begin{lstlisting}[style=python]
    import torch
    import torch.nn.functional as F
    import matplotlib.pyplot as plt
    from PIL import Image
    import torchvision.transforms as transforms

    # Load a blurred image and a known blur kernel
    image = Image.open('blurred_image.jpg')
    transform = transforms.ToTensor()
    image_tensor = transform(image).unsqueeze(0)  # Add batch dimension

    # Define a known blur kernel (e.g., Gaussian)
    kernel = torch.tensor([[[[1, 4, 6, 4, 1],
                             [4, 16, 24, 16, 4],
                             [6, 24, 36, 24, 6],
                             [4, 16, 24, 16, 4],
                             [1, 4, 6, 4, 1]]]]) / 256.0

    # Deconvolution using inverse filtering (simple example)
    restored_image = F.conv2d(image_tensor, kernel, padding=2)

    # Display original and restored images
    plt.subplot(1, 2, 1)
    plt.imshow(image)
    plt.title('Blurred Image')

    plt.subplot(1, 2, 2)
    plt.imshow(restored_image.squeeze().permute(1, 2, 0).numpy())
    plt.title('Restored Image')
    plt.show()
    \end{lstlisting}

    In this example, we demonstrate a simple inverse filtering technique where the blur kernel is known. Blind deconvolution methods would aim to estimate both the image and the kernel without prior knowledge of the latter.

\section{Image Interpolation and Reconstruction}
    Image interpolation is a key technique used when enlarging images or reconstructing missing pixel values. Interpolation methods estimate new pixel values by using surrounding pixels. This process is crucial in applications such as image scaling, rotation, and image reconstruction from sparse data \cite{thevenaz2000image}. In this section, we explore three common interpolation techniques: \textit{nearest-neighbor interpolation}, \textit{bilinear interpolation}, and \textit{bicubic interpolation}.

    \subsection{Nearest-Neighbor Interpolation}
        \textbf{Nearest-neighbor interpolation} is the simplest form of interpolation. When enlarging an image, this method assigns the value of the nearest pixel in the original image to each pixel in the enlarged image. While this approach is computationally efficient, it often produces blocky or pixelated images, as it does not smooth out transitions between neighboring pixel values \cite{rukundo2012nearest}.

        \textbf{How It Works:}
        For each pixel in the enlarged image, the nearest-neighbor algorithm identifies the closest pixel in the original image and assigns its value to the new pixel. Mathematically, for an image scaled by a factor of \(s\), the coordinates of a pixel \((i', j')\) in the enlarged image correspond to the nearest pixel \((i, j)\) in the original image:
        \[
        i = \text{round}(i' / s), \quad j = \text{round}(j' / s)
        \]

        \textbf{Example:}

        Here's how we can implement nearest-neighbor interpolation in Python:

        \begin{lstlisting}[style=python]
        import torch
        import torch.nn.functional as F
        import torchvision.transforms as transforms
        from PIL import Image
        import matplotlib.pyplot as plt

        # Load image and convert to tensor
        image = Image.open('example_image.jpg')
        transform = transforms.ToTensor()
        image_tensor = transform(image).unsqueeze(0)  # Add batch dimension

        # Apply nearest-neighbor interpolation for upscaling
        upscale_factor = 2
        upscaled_image = F.interpolate(image_tensor, scale_factor=upscale_factor, mode='nearest')

        # Display original and upscaled images
        plt.subplot(1, 2, 1)
        plt.imshow(image)
        plt.title('Original Image')

        plt.subplot(1, 2, 2)
        plt.imshow(upscaled_image.squeeze().permute(1, 2, 0).numpy())
        plt.title('Upscaled Image (Nearest-Neighbor)')
        plt.show()
        \end{lstlisting}

        In this example, we use PyTorch's \texttt{interpolate} function to scale the image using nearest-neighbor interpolation. This method is fast but results in a pixelated appearance when enlarging images.

        \textbf{Applications:}
        Nearest-neighbor interpolation is often used in applications where simplicity and speed are more important than visual quality, such as enlarging low-resolution images or generating thumbnails \cite{rukundo2012nearest}.

    \subsection{Bilinear Interpolation}
        \textbf{Bilinear interpolation} is a more advanced technique compared to nearest-neighbor interpolation \cite{smith1981bilinear}. It smooths the transitions between pixels by averaging the values of the four closest pixels. This method results in smoother images and is commonly used when a higher-quality enlargement is required.

        \textbf{How It Works:}
        Bilinear interpolation computes the pixel value by taking a weighted average of the four nearest neighbors. Given the four neighboring pixels at coordinates \((i,j)\), \((i+1,j)\), \((i,j+1)\), and \((i+1,j+1)\), the value of a new pixel at position \((x',y')\) is calculated as:
        \[
        I'(x', y') = \frac{1}{(x_2 - x_1)(y_2 - y_1)} \sum_{i=1}^{2} \sum_{j=1}^{2} I(x_i, y_j) (x_2 - x') (y_2 - y')
        \]

        \textbf{Example:}

        Bilinear interpolation can be implemented in Python using PyTorch as follows:

        \begin{lstlisting}[style=python]
        # Apply bilinear interpolation for upscaling
        upscaled_image = F.interpolate(image_tensor, scale_factor=upscale_factor, mode='bilinear', align_corners=False)

        # Display upscaled image
        plt.imshow(upscaled_image.squeeze().permute(1, 2, 0).numpy())
        plt.title('Upscaled Image (Bilinear Interpolation)')
        plt.show()
        \end{lstlisting}

        This code demonstrates how to upscale an image using bilinear interpolation, which provides smoother transitions between pixels than nearest-neighbor interpolation.

        \textbf{Applications:}
        Bilinear interpolation is widely used in image resizing, where maintaining a smoother appearance is essential, such as in digital zooming, rotation, and texture mapping \cite{smith1981bilinear}.

    \subsection{Bicubic Interpolation}
        \textbf{Bicubic interpolation} is an even more advanced technique that considers the 16 nearest pixels to compute the new pixel value, using cubic polynomials to calculate the weights. Bicubic interpolation produces higher-quality results than both nearest-neighbor and bilinear interpolation, especially for large scaling factors \cite{fadnavis2014images}.

        \textbf{How It Works:}
        Bicubic interpolation takes into account both the distance and the derivative of the neighboring pixels, resulting in smoother and more visually pleasing enlargements. The pixel value is computed using cubic convolution, which interpolates based on the pixel intensity and its surrounding pixel intensities over a larger area (typically a 4x4 grid of pixels).

        \textbf{Example:}

        Here's how to implement bicubic interpolation using PyTorch:

        \begin{lstlisting}[style=python]
        # Apply bicubic interpolation for upscaling
        upscaled_image = F.interpolate(image_tensor, scale_factor=upscale_factor, mode='bicubic', align_corners=False)

        # Display upscaled image
        plt.imshow(upscaled_image.squeeze().permute(1, 2, 0).numpy())
        plt.title('Upscaled Image (Bicubic Interpolation)')
        plt.show()
        \end{lstlisting}

        Bicubic interpolation provides higher quality results compared to nearest-neighbor and bilinear interpolation, especially when enlarging images by large factors.

        \textbf{Applications:}
        Bicubic interpolation is commonly used in high-quality image resizing, such as in photo editing software, where preserving fine details and smooth gradients is important.
\chapter{Image Segmentation and Edge Detection}
    Image segmentation and edge detection are foundational techniques in digital image processing, used for partitioning an image into meaningful regions or identifying boundaries between different objects \cite{gonzalez2002digital}. These methods are critical for applications such as object detection, image recognition, medical imaging, and computer vision. By breaking an image into distinct regions, segmentation simplifies the task of analyzing specific parts of the image, enabling more effective object detection and pattern recognition \cite{hoeser2020object, bhanu2000adaptive, patil2013medical}.

    \section{Basic Concepts of Image Segmentation}
        \textbf{Image segmentation} is the process of dividing an image into multiple segments or regions, each of which represents a specific part of the image. The goal of segmentation is to simplify the image into something more meaningful and easier to analyze by isolating objects or regions of interest \cite{cheng2001color}. 

        Segmentation can be applied in various ways, depending on the requirements of the application:
        \begin{itemize}
            \item \textbf{Object Detection}: Identifying objects such as people, cars, or animals in images or videos \cite{hoeser2020object}.
            \item \textbf{Medical Imaging}: Segmenting different tissue types in medical scans (e.g., MRI or CT scans) to aid in diagnosis \cite{patil2013medical}.
            \item \textbf{Scene Understanding}: Partitioning different areas in a landscape image, such as the sky, mountains, and trees \cite{li2009towards}.
        \end{itemize}

        The result of segmentation is usually a set of connected regions that correspond to distinct objects or areas. These regions are often represented by labels, with each label corresponding to a specific object or part of the image.

        \paragraph{Segmentation Methods}
        Several methods exist for image segmentation, depending on the complexity of the scene, the types of objects involved, and the quality of the image. Some common approaches include:
        \begin{itemize}
            \item \textbf{Thresholding-based Segmentation}: This method involves partitioning the image based on pixel intensity values. Thresholding is widely used because of its simplicity and efficiency \cite{mustaqeem2012efficient}.
            \item \textbf{Edge-based Segmentation}: Edges in an image correspond to object boundaries, and detecting edges can be used as a basis for segmentation.
            \item \textbf{Region-based Segmentation}: This method divides the image into regions based on similarities in color, intensity, or texture \cite{karthick2014survey}.
            \item \textbf{Clustering-based Segmentation}: Techniques like \( k \)-means clustering or mean shift group pixels into clusters based on feature similarity \cite{saxena2017review}.
        \end{itemize}

    \section{Thresholding-based Segmentation}
        \textbf{Thresholding} is one of the simplest and most widely used techniques for image segmentation. The basic idea is to segment an image by comparing each pixel's intensity to a fixed threshold. If the intensity of a pixel is greater than the threshold, it is classified as one class (e.g., the foreground), and if it is less than the threshold, it is classified as another class (e.g., the background). Thresholding is particularly useful when the objects of interest have a significantly different intensity from the background.

        \subsection{Global Thresholding}
            In \textbf{global thresholding}, a single threshold value is applied to the entire image. The goal is to choose a threshold \( T \) that effectively separates the objects from the background \cite{lee1990comparative}. Mathematically, this can be expressed as:
            \[
            g(x, y) =
            \begin{cases}
                1, & \text{if } f(x, y) \geq T \\
                0, & \text{if } f(x, y) < T
            \end{cases}
            \]
            where:
            \begin{itemize}
                \item \( f(x, y) \) is the intensity of the pixel at location \( (x, y) \),
                \item \( T \) is the global threshold,
                \item \( g(x, y) \) is the segmented output.
            \end{itemize}

            Choosing the right threshold is crucial for good segmentation. A threshold that is too high may miss important details, while a threshold that is too low may introduce noise.

            \paragraph{Example: Global Thresholding in Python}
            The following Python code demonstrates how to apply global thresholding to a grayscale image.

            \begin{lstlisting}[style=python]
import numpy as np
import matplotlib.pyplot as plt
from PIL import Image

# Load an image and convert it to grayscale
image = Image.open('sample_image.jpg').convert('L')
image_array = np.array(image)

# Global thresholding function
def global_threshold(img, T):
    thresholded_img = np.where(img >= T, 255, 0)
    return thresholded_img.astype(np.uint8)

# Apply global thresholding with a threshold value of 128
threshold_value = 128
thresholded_image = global_threshold(image_array, threshold_value)

# Display original and thresholded images
plt.figure(figsize=(12, 6))

plt.subplot(1, 2, 1)
plt.imshow(image_array, cmap='gray')
plt.title('Original Image')
plt.axis('off')

plt.subplot(1, 2, 2)
plt.imshow(thresholded_image, cmap='gray')
plt.title(f'Global Thresholding (T = {threshold_value})')
plt.axis('off')

plt.show()
            \end{lstlisting}

            \textbf{Explanation of the Code:}
            \begin{itemize}
                \item The image is loaded and converted to grayscale using the \texttt{Pillow} library.
                \item The \texttt{global\_threshold()} function compares each pixel's intensity to the threshold \( T = 128 \). Pixels with intensity values greater than or equal to 128 are set to 255 (white), and the others are set to 0 (black).
                \item The original and thresholded images are displayed side by side using \texttt{Matplotlib}.
            \end{itemize}

        \subsection{Local Thresholding}
            \textbf{Local thresholding} (also known as adaptive thresholding) divides the image into smaller regions and applies a different threshold to each region. This method is particularly useful when the lighting conditions vary across the image, as it can account for local variations in intensity \cite{jiang2003adaptive}.

            Local thresholding can be performed by calculating a threshold for each pixel based on the intensity values in its neighborhood. Common methods for local thresholding include:
            \begin{itemize}
                \item \textbf{Mean-based thresholding}: The threshold is calculated as the mean intensity of the neighboring pixels.
                \item \textbf{Gaussian-weighted thresholding}: A Gaussian-weighted mean of the neighboring pixels is used to compute the threshold.
            \end{itemize}

            \paragraph{Example: Local Thresholding in Python}
            The following Python code demonstrates how to apply local (adaptive) thresholding using OpenCV's \texttt{adaptiveThreshold()} function.

            \begin{lstlisting}[style=python]
import cv2

# Load the image as a grayscale array
image_cv = cv2.imread('sample_image.jpg', cv2.IMREAD_GRAYSCALE)

# Apply adaptive thresholding (mean-based)
adaptive_thresh = cv2.adaptiveThreshold(image_cv, 255, cv2.ADAPTIVE_THRESH_MEAN_C, 
                                        cv2.THRESH_BINARY, blockSize=11, C=2)

# Display the original and thresholded images
plt.figure(figsize=(12, 6))

plt.subplot(1, 2, 1)
plt.imshow(image_cv, cmap='gray')
plt.title('Original Image')
plt.axis('off')

plt.subplot(1, 2, 2)
plt.imshow(adaptive_thresh, cmap='gray')
plt.title('Local (Adaptive) Thresholding')
plt.axis('off')

plt.show()
            \end{lstlisting}

            \textbf{Explanation of the Code:}
            \begin{itemize}
                \item The image is loaded as a grayscale array using OpenCV.
                \item The \texttt{adaptiveThreshold()} function applies local thresholding. The threshold is calculated based on the mean intensity of the neighboring pixels in an \(11 \times 11\) block. The constant \( C = 2 \) is subtracted from the mean to fine-tune the threshold.
                \item The original and locally thresholded images are displayed using \texttt{Matplotlib}.
            \end{itemize}

        \subsection{Otsu's Method for Automatic Thresholding}
            \textbf{Otsu's method} is a widely used global thresholding technique that automatically determines the optimal threshold value by minimizing the intra-class variance, which is the variance within the foreground and background regions. It works by maximizing the variance between these two classes \cite{bangare2015reviewing}.

            Otsu's method computes a histogram of the image and then evaluates every possible threshold to find the one that maximizes the separation between the two classes. The advantage of Otsu's method is that it does not require manual selection of a threshold value, making it ideal for automating the segmentation process \cite{liu2009otsu}.

            \paragraph{Steps of Otsu's Method}:
            \begin{enumerate}
                \item Compute the histogram of the image.
                \item For each possible threshold \( T \), divide the pixels into two classes: foreground (pixels \( \geq T \)) and background (pixels \( < T \)).
                \item Compute the intra-class variance for each threshold and choose the threshold that minimizes this variance.
            \end{enumerate}

            \paragraph{Example: Otsu's Thresholding in Python}
            The following Python code demonstrates how to apply Otsu's method for automatic thresholding using OpenCV.

            \begin{lstlisting}[style=python]
# Apply Otsu's thresholding
_, otsu_thresh = cv2.threshold(image_cv, 0, 255, cv2.THRESH_BINARY + cv2.THRESH_OTSU)

# Display the original and Otsu's thresholded images
plt.figure(figsize=(12, 6))

plt.subplot(1, 2, 1)
plt.imshow(image_cv, cmap='gray')
plt.title('Original Image')
plt.axis('off')

plt.subplot(1, 2, 2)
plt.imshow(otsu_thresh, cmap='gray')
plt.title('Otsu's Thresholding')
plt.axis('off')

plt.show()
            \end{lstlisting}

            \textbf{Explanation of the Code:}
            \begin{itemize}
                \item The \texttt{cv2.threshold()} function with the flag \texttt{cv2.THRESH\_OTSU} automatically computes the optimal threshold using Otsu's method.
                \item The original and thresholded images are displayed side by side for comparison.
            \end{itemize}

    \section{Conclusion}
        Image segmentation is an essential step in many image processing applications, and thresholding-based methods are a simple yet powerful approach. Global thresholding, local (adaptive) thresholding, and automatic methods like Otsu's method are effective for partitioning images into foreground and background regions. Understanding these techniques enables beginners to explore more complex segmentation problems and serves as a foundation for further study in areas like object detection, image recognition, and medical imaging.

\section{Edge Detection Techniques}
    Edge detection is a fundamental technique in image processing that aims to identify significant local changes in intensity, which typically correspond to object boundaries, texture changes, or other important features in an image. Effective edge detection is crucial for applications such as image segmentation, object recognition, and computer vision. In this section, we will explore three important edge detection techniques: the Sobel operator, the Prewitt operator, and the Canny edge detection algorithm. Each method is introduced with clear explanations, practical examples, and a discussion of its advantages and limitations \cite{ziou1998edge, chaple2015comparisions}.

    \subsection{Sobel Operator}
        The Sobel operator is a simple and widely-used method for detecting edges in images by computing the gradient magnitude. It works by applying a pair of convolution kernels to the image{-}one to detect changes in the horizontal direction (\(G_x\)) and one to detect changes in the vertical direction (\(G_y\)). The gradient magnitude at each pixel is then calculated from these two components \cite{chaple2015comparisions}.

        \paragraph{Sobel Convolution Kernels:} The Sobel operator uses two 3x3 kernels:

        \[
        G_x = \begin{bmatrix} -1 & 0 & 1 \\ -2 & 0 & 2 \\ -1 & 0 & 1 \end{bmatrix}, \quad G_y = \begin{bmatrix} -1 & -2 & -1 \\ 0 & 0 & 0 \\ 1 & 2 & 1 \end{bmatrix}
        \]

        These kernels approximate the derivative of the image in the horizontal and vertical directions, respectively. After convolving the image with these kernels, the gradient magnitude \(G\) is computed as:

        \[
        G = \sqrt{G_x^2 + G_y^2}
        \]

        The direction of the edge can be determined by the angle \(\theta\):

        \[
        \theta = \tan^{-1}\left(\frac{G_y}{G_x}\right)
        \]

        \paragraph{Advantages:} The Sobel operator is computationally simple and easy to implement. It also performs some amount of smoothing (due to the averaging effect of the 3x3 kernels), which makes it robust to noise.

        \paragraph{Limitations:} The Sobel operator may not detect fine edges accurately, and it is sensitive to noise, especially in cases where noise is significant compared to the edges in the image. It also has limited effectiveness in detecting diagonal edges.

        \paragraph{Example:} Let's apply the Sobel operator to an image using Python.

        \begin{lstlisting}[style=python]
        import numpy as np
        import matplotlib.pyplot as plt
        from skimage import data, filters

        # Load the sample image
        image = data.camera()

        # Apply the Sobel operator to compute the edge map
        sobel_edges = filters.sobel(image)

        # Plot the original and edge-detected images
        fig, axes = plt.subplots(1, 2, figsize=(12, 6))

        axes[0].imshow(image, cmap='gray')
        axes[0].set_title('Original Image')

        axes[1].imshow(sobel_edges, cmap='gray')
        axes[1].set_title('Edge Detection (Sobel Operator)')

        for ax in axes:
            ax.axis('off')

        plt.show()
        \end{lstlisting}

        In this example, we apply the Sobel operator to a grayscale image and display the resulting edge map. The detected edges highlight areas where there are significant changes in intensity, such as object boundaries and texture transitions.

        \paragraph{Explanation:} The Sobel operator emphasizes vertical and horizontal edges in the image by computing the gradient in both directions. The resulting edge map shows where intensity changes occur, making it useful for detecting major features in the image.

    \subsection{Prewitt Operator}
        The Prewitt operator is similar to the Sobel operator in that it detects edges by calculating the gradient magnitude in an image. However, the Prewitt operator uses slightly different convolution kernels that provide a simpler approximation of the derivative \cite{chaple2015comparisions}.

        \paragraph{Prewitt Convolution Kernels:} The 3x3 Prewitt kernels are:

        \[
        G_x = \begin{bmatrix} -1 & 0 & 1 \\ -1 & 0 & 1 \\ -1 & 0 & 1 \end{bmatrix}, \quad G_y = \begin{bmatrix} -1 & -1 & -1 \\ 0 & 0 & 0 \\ 1 & 1 & 1 \end{bmatrix}
        \]

        Like the Sobel operator, the Prewitt operator computes the gradient magnitude \(G\) and direction \(\theta\) using the same formulas. The main difference is that the Prewitt operator's kernels are less sensitive to diagonal edges.

        \paragraph{Advantages:} The Prewitt operator is simpler and faster than the Sobel operator, making it useful for real-time applications. It performs better in detecting horizontal and vertical edges, although it is less accurate for detecting diagonal edges.

        \paragraph{Limitations:} Similar to the Sobel operator, the Prewitt operator is sensitive to noise and may not detect fine edges accurately. It is also less effective at detecting diagonal edges due to the lack of smoothing in its kernels.

        \paragraph{Example:} Let's apply the Prewitt operator to the same image and compare the results to those from the Sobel operator.

        \begin{lstlisting}[style=python]
        # Apply the Prewitt operator to compute the edge map
        prewitt_edges = filters.prewitt(image)

        # Plot the Prewitt edge-detected image
        plt.imshow(prewitt_edges, cmap='gray')
        plt.title('Edge Detection (Prewitt Operator)')
        plt.axis('off')
        plt.show()
        \end{lstlisting}

        In this example, we apply the Prewitt operator to the same image and display the resulting edge map. The output is similar to that of the Sobel operator, but the edges are detected using simpler convolution kernels.

        \paragraph{Explanation:} The Prewitt operator is a fast and efficient method for detecting edges, especially in applications where computational efficiency is a concern. However, it may not perform as well as the Sobel operator in detecting finer edges or diagonal features.

    \subsection{Canny Edge Detection}
        The Canny edge detection algorithm is one of the most advanced and widely used edge detection techniques. It provides robust edge detection by addressing many of the limitations of simpler gradient-based methods like the Sobel and Prewitt operators. The Canny algorithm uses a multi-stage process that includes noise reduction, gradient calculation, non-maximum suppression, and hysteresis thresholding \cite{xu2017canny}.

        \paragraph{Steps in the Canny Algorithm:}
        \begin{enumerate}
            \item \textbf{Noise Reduction:} The image is first smoothed using a Gaussian filter to reduce noise and small variations that may lead to false edge detection.
            \item \textbf{Gradient Calculation:} The gradients in the image are calculated using a method similar to the Sobel operator, providing the gradient magnitude and direction at each pixel.
            \item \textbf{Non-Maximum Suppression:} The algorithm applies non-maximum suppression to thin out the edges by keeping only the pixels with the highest gradient magnitudes in the direction of the edge.
            \item \textbf{Hysteresis Thresholding:} Two thresholds (a high and a low threshold) are applied to the gradient magnitudes to determine which pixels are considered strong edges, weak edges, or non-edges. Strong edges are immediately retained, while weak edges are included only if they are connected to strong edges.
        \end{enumerate}

        \paragraph{Advantages:} The Canny algorithm is highly effective at detecting edges with precision. It reduces noise and spurious edges through Gaussian smoothing, ensures thin edges through non-maximum suppression, and uses hysteresis to accurately retain true edges while discarding weak, isolated edges.

        \paragraph{Limitations:} The Canny algorithm requires careful tuning of the Gaussian filter size and the two threshold values, which may vary depending on the image. It can also be computationally more expensive than simpler edge detection methods like the Sobel or Prewitt operators.

        \paragraph{Example:} Let's implement the Canny edge detection algorithm using Python.

        \begin{lstlisting}[style=python]
        from skimage.feature import canny

        # Apply the Canny edge detection algorithm
        canny_edges = canny(image, sigma=1.0)

        # Plot the Canny edge-detected image
        plt.imshow(canny_edges, cmap='gray')
        plt.title('Edge Detection (Canny Algorithm)')
        plt.axis('off')
        plt.show()
        \end{lstlisting}

        In this example, we apply the Canny edge detection algorithm to the image, using a Gaussian filter with \(\sigma = 1.0\) for noise reduction. The result is a refined edge map that accurately highlights important edges in the image while reducing noise and spurious edges.

        \paragraph{Explanation:} The Canny algorithm is a powerful edge detection tool that outperforms simpler gradient-based methods. By combining noise reduction, non-maximum suppression, and hysteresis thresholding, it delivers high-quality edge detection that is well-suited for applications requiring precise edge localization, such as object detection and image segmentation.

    \subsection{Comparison of Edge Detection Techniques}
        Each edge detection technique discussed has its strengths and weaknesses, making them suitable for different applications \cite{ziou1998edge}:

        \begin{itemize}
            \item \textbf{Sobel Operator:} Simple and easy to implement. Best for detecting large, significant edges, but sensitive to noise and less effective for detecting diagonal edges.
            \item \textbf{Prewitt Operator:} Similar to the Sobel operator but with simpler kernels. It is faster but less accurate, particularly for diagonal edges.
            \item \textbf{Canny Edge Detection:} Provides highly accurate edge detection with noise suppression, thin edges, and strong/weak edge distinction. More computationally expensive and requires parameter tuning.
        \end{itemize}

        \paragraph{Practical Considerations:} For applications where speed and simplicity are crucial, the Sobel or Prewitt operator may be sufficient. However, when accuracy and robustness are required, especially in noisy images or in situations where precise edge localization is important, the Canny algorithm is often the best choice. The choice of edge detection technique should be based on the specific requirements of the task and the characteristics of the images being processed.

\section{Region-based Segmentation}
    Region-based segmentation is a fundamental technique in image processing, where the goal is to divide an image into regions that correspond to different objects or features. This type of segmentation is based on the similarity of pixel properties, such as intensity, color, or texture. Unlike edge-based segmentation, which relies on detecting discontinuities between regions, region-based methods group pixels into regions based on homogeneity \cite{karthick2014survey}. In this section, we will explore two common techniques: \textit{region growing} and \textit{split-and-merge segmentation}.

    \subsection{Region Growing}
        \textbf{Region growing} is a straightforward technique in region-based segmentation. The idea is to start with one or more \textit{seed} pixels and grow a region by adding neighboring pixels that have similar properties. The growth process continues until no more pixels can be added to the region based on the similarity criteria \cite{adams1994seeded}.

        \textbf{Key Steps:}
        \begin{enumerate}
        \item Select one or more \textit{seed points} from which the region growing process will begin. These seeds can be chosen manually or automatically, based on some initial criteria (e.g., pixels with a specific intensity).
        \item Define a \textit{similarity criterion}, which can be based on intensity, color, texture, or other properties. This criterion determines whether a neighboring pixel should be added to the growing region.
        \item For each seed point, add neighboring pixels that meet the similarity criterion. The region continues to grow as long as there are neighboring pixels that satisfy the similarity condition.
        \item Stop the process when no further pixels can be added to the region.
        \end{enumerate}

        \textbf{Example:}

        Let's implement a simple region growing algorithm in Python. The similarity criterion in this case will be based on intensity similarity.

        \begin{lstlisting}[style=python]
        import numpy as np
        import matplotlib.pyplot as plt
        from skimage import data, color, filters

        # Load a grayscale image
        image = data.camera()
        
        # Define seed point and intensity threshold
        seed = (100, 100)
        threshold = 10
        
        # Create a binary mask for the region
        region_mask = np.zeros_like(image, dtype=bool)
        region_mask[seed] = True
        
        # Region growing function
        def region_grow(image, seed, threshold):
            region_mask = np.zeros_like(image, dtype=bool)
            region_mask[seed] = True
            region_intensity = image[seed]
            
            neighbors = [(seed[0], seed[1])]
            
            while neighbors:
                x, y = neighbors.pop(0)
                
                for dx, dy in [(-1, 0), (1, 0), (0, -1), (0, 1)]:
                    nx, ny = x + dx, y + dy
                    
                    if (0 <= nx < image.shape[0]) and (0 <= ny < image.shape[1]):
                        if not region_mask[nx, ny] and abs(image[nx, ny] - region_intensity) < threshold:
                            region_mask[nx, ny] = True
                            neighbors.append((nx, ny))
            return region_mask
        
        # Perform region growing
        region = region_grow(image, seed, threshold)
        
        # Display original image and segmented region
        plt.subplot(1, 2, 1)
        plt.imshow(image, cmap='gray')
        plt.title('Original Image')
        
        plt.subplot(1, 2, 2)
        plt.imshow(region, cmap='gray')
        plt.title('Segmented Region (Region Growing)')
        plt.show()
        \end{lstlisting}

        In this example, we use the \texttt{region\_grow} function to grow a region from a specified seed point. The algorithm adds neighboring pixels if their intensity is similar to the seed pixel's intensity, as determined by the specified threshold. The result is a segmented region that is displayed alongside the original image.

        \textbf{Challenges:}
        \begin{itemize}
        \item \textit{Seed selection:} Choosing the correct seed points is crucial for the success of region growing. Inaccurate seed selection can result in incorrect segmentation.
        \item \textit{Threshold sensitivity:} The similarity criterion (threshold) must be carefully chosen. A threshold that is too high will include dissimilar pixels, while a threshold that is too low may not capture the full region.
        \end{itemize}

        \textbf{Applications:}
        \begin{itemize}
        \item Medical imaging (e.g., tumor detection, organ segmentation) \cite{thevenaz2000image}.
        \item Remote sensing and satellite imagery analysis \cite{pritt2017satellite}.
        \item Object detection in computer vision \cite{terven2023comprehensive}.
        \end{itemize}

    \subsection{Split-and-Merge Segmentation}
        \textbf{Split-and-merge} segmentation is a more structured approach that divides the image into regions based on homogeneity. This method recursively splits the image into smaller regions and merges regions that meet a consistency criterion. The technique is often implemented using a quadtree structure, where each node represents a rectangular region of the image \cite{chen1979segmentation}.

        \textbf{Key Steps:}
        \begin{enumerate}
        \item \textit{Splitting:} The algorithm starts by dividing the entire image into four quadrants (a process called splitting). Each quadrant is checked for homogeneity, based on criteria such as intensity or color. If a quadrant is not homogeneous, it is further split into smaller quadrants.
        \item \textit{Merging:} Once the image is split into small enough regions, the algorithm attempts to merge adjacent regions that have similar properties. Regions that meet the homogeneity criteria are merged back together to form larger regions.
        \item \textit{Termination:} The process stops when no further splitting or merging is possible.
        \end{enumerate}

        \textbf{Quadtree Representation:}

        In split-and-merge segmentation, the image can be represented as a \textit{quadtree}, where each node of the tree represents a region of the image. The root node represents the entire image, and each node has four children corresponding to the four quadrants \cite{wu1993adaptive}.

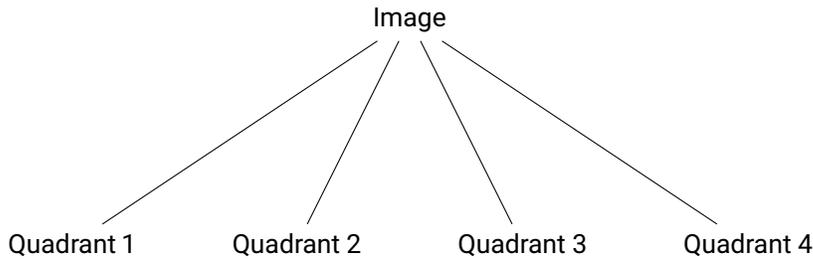
        \begin{tikzpicture}[
            level distance=3cm,
            sibling distance=3cm
        ]
            \node {Image}
            child {node {Quadrant 1}}
            child {node {Quadrant 2}}
            child {node {Quadrant 3}}
            child {node {Quadrant 4}};
        \end{tikzpicture}

        This tree structure allows for efficient representation and manipulation of the image regions.

        \textbf{Example:}

        The split-and-merge process can be simulated in Python by recursively dividing the image and merging regions based on intensity similarity.

        \begin{lstlisting}[style=python]
        from skimage.segmentation import felzenszwalb
        from skimage import data
        import matplotlib.pyplot as plt

        # Load image
        image = data.coins()

        # Perform split-and-merge segmentation using Felzenszwalb's method
        segments = felzenszwalb(image, scale=100, sigma=0.5, min_size=50)

        # Display original image and segmented image
        plt.subplot(1, 2, 1)
        plt.imshow(image, cmap='gray')
        plt.title('Original Image')

        plt.subplot(1, 2, 2)
        plt.imshow(segments, cmap='gray')
        plt.title('Segmented Image (Split-and-Merge)')
        plt.show()
        \end{lstlisting}

        In this example, we use the Felzenszwalb segmentation method from the \texttt{skimage} library to simulate split-and-merge segmentation. The result is a segmented image where similar regions are merged.

        \textbf{Applications:}
        \begin{itemize}
        \item Satellite image analysis and land cover classification \cite{pritt2017satellite}.
        \item Object recognition in high-resolution images \cite{thevenaz2000image}.
        \end{itemize}

\section{Watershed Algorithm}
    The \textbf{watershed algorithm} is a popular technique in image segmentation that is based on the topography of the image. The algorithm interprets the grayscale intensity of an image as a surface or landscape, with pixel intensities corresponding to elevations. The basic idea is to treat the image as a topographic surface and "flood" it with water, where the basins correspond to segments \cite{kornilov2018overview}.

    \textbf{Concept:}
    In the watershed algorithm, the image is considered as a gradient field. Regions with high gradients represent boundaries, while regions with low gradients (basins) represent objects. The algorithm floods these basins, and when two basins meet, a boundary is formed between them, creating a segmentation of the image.

    \textbf{Steps of the Watershed Algorithm:}
    \begin{enumerate}
    \item \textit{Compute the gradient:} First, the image gradient is computed. This gradient image highlights the edges in the image, which are used to separate regions.
    \item \textit{Flood the image:} Water is "poured" into the basins from the lowest points (pixels with the lowest intensity). The water continues to rise, filling the basins until it reaches the higher-intensity regions.
    \item \textit{Mark boundaries:} When water from different basins meets, a boundary is drawn, creating a segmentation.
    \end{enumerate}

    \textbf{Mathematical Representation:}

    The watershed algorithm can be expressed using a gradient image \(G(x,y)\), where high values of \(G(x,y)\) correspond to image edges:
    \[
    \text{Watershed boundaries} = \text{argmax} \left( G(x, y) \right)
    \]

    \textbf{Example:}

    Here's how to implement the watershed algorithm in Python using the \texttt{skimage} library:

    \begin{lstlisting}[style=python]
    from skimage import data, color, morphology, filters
    from skimage.segmentation import watershed
    import numpy as np
    import matplotlib.pyplot as plt

    # Load and preprocess image
    image = data.coins()
    gradient = filters.sobel(image)

    # Markers for watershed algorithm
    markers = np.zeros_like(image)
    markers[image < 30] = 1
    markers[image > 150] = 2

    # Apply watershed algorithm
    segmented = watershed(gradient, markers)

    # Display original and segmented images
    plt.subplot(1, 2, 1)
    plt.imshow(image, cmap='gray')
    plt.title('Original Image')

    plt.subplot(1, 2, 2)
    plt.imshow(segmented, cmap='gray')
    plt.title('Segmented Image (Watershed)')
    plt.show()
    \end{lstlisting}

    In this example, we first compute the gradient of the image using the Sobel filter. We then apply the watershed algorithm by providing seed markers that define the initial basins. The algorithm segments the image based on the topology of the gradient.

    \textbf{Applications:}
    \begin{itemize}
    \item Medical image segmentation (e.g., separating cells or tissues) \cite{patil2013medical}.
    \item Object detection and separation in cluttered scenes \cite{terven2023comprehensive}.
    \item Satellite image analysis and terrain mapping \cite{pritt2017satellite}.
    \end{itemize}
\chapter{Image Compression}
    Image compression is an essential aspect of digital image processing, allowing us to reduce the storage and transmission requirements of images while maintaining an acceptable level of visual quality. The primary goal of compression is to represent an image with fewer bits by eliminating redundancies, without significantly compromising the quality of the image \cite{gonzalez2002digital}. There are two main types of compression: \textbf{lossless} and \textbf{lossy}. In lossless compression, the original image can be perfectly reconstructed from the compressed data, whereas lossy compression achieves higher compression ratios by allowing some loss of image data, which may lead to a degradation in visual quality.

    \section{Basic Principles of Image Compression}
        Image compression works by exploiting three primary types of redundancy \cite{dhawan2011review, vijayvargiya2013survey}:
        \begin{itemize}
            \item \textbf{Spatial Redundancy}: This arises because neighboring pixels in most images are highly correlated, meaning there is redundancy in pixel values within a local region. Techniques like run-length encoding (RLE) and block-based compression aim to reduce this redundancy.
            \item \textbf{Temporal Redundancy}: In video or multi-frame images, successive frames are often similar to one another, meaning the same information is repeated. Temporal redundancy can be reduced by methods that focus on the differences between frames.
            \item \textbf{Perceptual Redundancy}: Human vision is more sensitive to some details and less sensitive to others. Compression algorithms such as JPEG leverage this fact by discarding information that is less noticeable to the human eye.
        \end{itemize}

        By reducing these redundancies, image compression algorithms can achieve significant reductions in the number of bits required to represent an image, making storage and transmission more efficient.

    \section{Lossless Compression Techniques}
        \textbf{Lossless compression} ensures that the original image can be perfectly reconstructed from the compressed image \cite{yang2005overview}. This is crucial in applications where any loss of information is unacceptable, such as medical imaging, technical drawings, and archival photography \cite{kivijarvi1998comparison,rojatkar2015image, koff2006overview}. Two popular lossless compression techniques are \textbf{Huffman coding} and \textbf{run-length encoding (RLE)} \cite{moffat2019huffman, chetan2017performance}.

        \subsection{Huffman Coding}
            \textbf{Huffman coding} is a widely used lossless compression algorithm that assigns variable-length codes to symbols (in this case, pixel values), with shorter codes assigned to more frequently occurring symbols. This is based on the observation that, in most images, some pixel values occur more frequently than others. By encoding these frequent values with fewer bits and less frequent values with more bits, Huffman coding achieves compression \cite{moffat2019huffman}.

            \paragraph{Steps of Huffman Coding}
            The main steps in Huffman coding are:
            \begin{enumerate}
                \item \textbf{Create a frequency table}: Compute the frequency of each pixel value in the image.
                \item \textbf{Build a Huffman tree}: Use the frequency table to construct a binary tree, where more frequent symbols are closer to the root and less frequent symbols are farther away.
                \item \textbf{Generate codes}: Assign binary codes to each pixel value based on the path from the root to the leaf node in the Huffman tree. Shorter paths correspond to shorter codes.
                \item \textbf{Encode the image}: Replace each pixel value in the image with its corresponding Huffman code.
            \end{enumerate}

            \textbf{Example: Huffman Coding in Python}
            The following Python code demonstrates a simplified implementation of Huffman coding for an image.

            \begin{lstlisting}[style=python]
from collections import Counter, heapq
import numpy as np

# Create a tree node for Huffman encoding
class TreeNode:
    def __init__(self, value, freq):
        self.value = value
        self.freq = freq
        self.left = None
        self.right = None

    # Define comparison operator for priority queue
    def __lt__(self, other):
        return self.freq < other.freq

# Build the Huffman tree
def build_huffman_tree(freq_table):
    heap = [TreeNode(value, freq) for value, freq in freq_table.items()]
    heapq.heapify(heap)

    while len(heap) > 1:
        node1 = heapq.heappop(heap)
        node2 = heapq.heappop(heap)
        merged = TreeNode(None, node1.freq + node2.freq)
        merged.left = node1
        merged.right = node2
        heapq.heappush(heap, merged)

    return heap[0]

# Generate Huffman codes
def generate_huffman_codes(node, code="", code_table={}):
    if node is not None:
        if node.value is not None:
            code_table[node.value] = code
        generate_huffman_codes(node.left, code + "0", code_table)
        generate_huffman_codes(node.right, code + "1", code_table)
    return code_table

# Example image data (a 1D array for simplicity)
image_data = np.array([1, 1, 2, 3, 3, 3, 4, 4, 4, 4])

# Step 1: Create frequency table
freq_table = dict(Counter(image_data))

# Step 2: Build Huffman tree
huffman_tree = build_huffman_tree(freq_table)

# Step 3: Generate Huffman codes
huffman_codes = generate_huffman_codes(huffman_tree)

# Step 4: Encode the image data
encoded_data = ''.join([huffman_codes[pixel] for pixel in image_data])

# Output the Huffman codes and the encoded data
print("Huffman Codes:", huffman_codes)
print("Encoded Data:", encoded_data)
            \end{lstlisting}

            \textbf{Explanation of the Code:}
            \begin{itemize}
                \item We define a \texttt{TreeNode} class to represent nodes in the Huffman tree.
                \item The \texttt{build\_huffman\_tree()} function builds the Huffman tree using a priority queue based on the pixel frequencies.
                \item The \texttt{generate\_huffman\_codes()} function assigns binary codes to each pixel value based on the Huffman tree.
                \item The image data is encoded by replacing each pixel with its corresponding Huffman code.
            \end{itemize}

            Huffman coding can achieve significant compression when pixel values have varying frequencies, as frequently occurring values are encoded with shorter binary codes.

        \subsection{Run-length Encoding (RLE)}
            \textbf{Run-length encoding (RLE)} is a simple lossless compression technique that is particularly effective for images with large contiguous regions of the same pixel value, such as binary or grayscale images. The basic idea is to represent consecutive occurrences of the same value as a pair of the value and the count of how many times it repeats \cite{chetan2017performance}.

            \paragraph{Steps of Run-length Encoding}
            The main steps in RLE are:
            \begin{enumerate}
                \item Traverse the image pixel by pixel.
                \item When a sequence of consecutive pixels with the same value is encountered, store the pixel value and the length of the run.
                \item Replace the sequence of pixels with this run-length representation.
            \end{enumerate}

            \textbf{Example: Run-length Encoding in Python}
            The following Python code demonstrates how to apply RLE to a binary image.

            \begin{lstlisting}[style=python]
# Function for run-length encoding
def run_length_encoding(img):
    rle = []
    prev_pixel = img[0]
    count = 1

    for pixel in img[1:]:
        if pixel == prev_pixel:
            count += 1
        else:
            rle.append((prev_pixel, count))
            prev_pixel = pixel
            count = 1
    rle.append((prev_pixel, count))  # Append the last run
    return rle

# Example binary image data (1D array)
binary_image = np.array([1, 1, 1, 0, 0, 0, 1, 1, 0, 0])

# Apply run-length encoding
encoded_rle = run_length_encoding(binary_image)

# Output the encoded RLE
print("Run-length Encoded Data:", encoded_rle)
            \end{lstlisting}

            \textbf{Explanation of the Code:}
            \begin{itemize}
                \item The \texttt{run\_length\_encoding()} function iterates through the image and detects consecutive runs of the same pixel value.
                \item Each run is encoded as a pair consisting of the pixel value and the run length.
                \item The encoded RLE data is printed as a sequence of value-length pairs.
            \end{itemize}

            \textbf{Example Output}:
            \[
            \text{Run-length Encoded Data: } [(1, 3), (0, 3), (1, 2), (0, 2)]
            \]

            RLE is particularly effective for images with large homogeneous regions, such as binary images, where long runs of the same pixel value frequently occur.

    \section{Conclusion}
        Lossless image compression techniques such as Huffman coding and run-length encoding play a crucial role in reducing image file sizes while ensuring that no information is lost during the compression process. Huffman coding is highly effective for compressing images where some pixel values occur more frequently than others, while RLE excels in binary or simple images with large contiguous regions of the same pixel value. By understanding and implementing these techniques, beginners can appreciate the importance of compression in digital image processing and explore more advanced compression methods for different types of images.

\section{Lossy Compression Techniques}
    Lossy compression techniques are widely used in digital image and multimedia storage because they significantly reduce file sizes by discarding less critical information. Unlike lossless compression, which preserves every detail, lossy compression allows some degradation of image quality in exchange for better compression ratios. The key to effective lossy compression is to remove information that is perceptually less important to the human visual system. In this section, we will cover two of the most important techniques used in lossy compression: the Discrete Cosine Transform (DCT) and the Discrete Wavelet Transform (DWT). These methods are foundational in many compression standards, such as JPEG \cite{koff2006overview}.

    \subsection{Discrete Cosine Transform (DCT)}
        The Discrete Cosine Transform (DCT) is a fundamental tool in image compression. It converts an image from the spatial domain, where pixel intensities are represented, into the frequency domain, where the image is represented as a sum of cosine functions at various frequencies. This transformation allows the separation of the image into low-frequency components (which represent the bulk of the image's structure) and high-frequency components (which represent fine details and edges) \cite{ahmed1974discrete}.

        \paragraph{DCT Formula:} The 2D DCT for an image block of size \(N \times N\) is given by:

        \[
        F(u, v) = \frac{1}{N} \alpha(u) \alpha(v) \sum_{x=0}^{N-1} \sum_{y=0}^{N-1} f(x, y) \cos \left[\frac{(2x + 1)u\pi}{2N}\right] \cos \left[\frac{(2y + 1)v\pi}{2N}\right]
        \]

        where \(f(x, y)\) is the pixel intensity at position \((x, y)\), and:

        \[
        \alpha(u) = \begin{cases} \frac{1}{\sqrt{2}} & \text{if } u = 0 \\ 1 & \text{if } u > 0 \end{cases}
        \]

        The inverse DCT, used to convert the frequency-domain data back to the spatial domain, is similarly defined.

        \paragraph{Energy Compaction:} One of the reasons why DCT is so effective for compression is that most of the image's energy (i.e., important visual information) is concentrated in the low-frequency components. After performing the DCT, many of the high-frequency components can be discarded or quantized with little noticeable effect on image quality. This property is known as energy compaction.

        \paragraph{Example:} Let's apply the DCT to a small image block using Python and demonstrate how energy compaction works.

        \begin{lstlisting}[style=python]
        import numpy as np
        import matplotlib.pyplot as plt
        from scipy.fftpack import dct, idct
        from skimage import data, color

        # Convert image to grayscale
        image = color.rgb2gray(data.astronaut())

        # Select a small block from the image (8x8)
        block = image[50:58, 50:58]

        # Apply 2D DCT
        dct_block = dct(dct(block.T, norm='ortho').T, norm='ortho')

        # Display the original block and its DCT coefficients
        fig, axes = plt.subplots(1, 2, figsize=(8, 4))
        
        axes[0].imshow(block, cmap='gray')
        axes[0].set_title('Original 8x8 Block')

        axes[1].imshow(np.log(np.abs(dct_block) + 1), cmap='gray')
        axes[1].set_title('DCT Coefficients')

        plt.show()
        \end{lstlisting}

        In this example, we apply the DCT to an 8x8 block from the grayscale astronaut image. The output shows the original block and the corresponding DCT coefficients. The low-frequency coefficients (top-left corner) carry most of the image's visual information, while the high-frequency coefficients (bottom-right corner) are much smaller in magnitude.

        \paragraph{Use in Compression:} The DCT is the foundation of the JPEG compression algorithm, where it is used to transform image blocks (usually 8x8 or 16x16 pixels) into the frequency domain. After transformation, high-frequency coefficients are either heavily quantized or set to zero, thereby reducing the amount of data needed to store the image.

        \paragraph{Advantages and Limitations:} 
        \begin{itemize}
            \item \textbf{Advantages:} The DCT provides good energy compaction and can be efficiently implemented. It is particularly effective for smooth images with few sharp edges \cite{rao2014discrete}.
            \item \textbf{Limitations:} The DCT assumes that the image is locally smooth, which means it may not perform as well for images with many sharp edges or textures. Additionally, blocking artifacts can occur if the image is divided into small blocks for compression (as in JPEG) \cite{rao2014discrete}.
        \end{itemize}

    \subsection{Discrete Wavelet Transform (DWT)}
        The Discrete Wavelet Transform (DWT) is another powerful tool used in lossy image compression. Unlike the DCT, which operates on blocks of fixed size, the DWT provides a multi-resolution representation of the image by decomposing it into different levels of detail. The DWT captures both frequency and spatial information, which makes it well-suited for images with varying levels of detail, such as natural scenes \cite{zhang2019wavelet}.

        \paragraph{Multi-Resolution Representation:} The DWT decomposes an image into a set of subbands, each corresponding to different frequency components and spatial resolutions \cite{rozynski2005multi}. Typically, the image is first decomposed into four subbands:
        \begin{itemize}
            \item \textbf{LL:} Low-frequency (approximation) subband, representing the coarse structure of the image.
            \item \textbf{LH:} Horizontal detail subband, capturing vertical edges.
            \item \textbf{HL:} Vertical detail subband, capturing horizontal edges.
            \item \textbf{HH:} High-frequency subband, capturing diagonal details.
        \end{itemize}
        The low-frequency subband can be further decomposed to extract even finer details, leading to a hierarchical representation of the image.

        \paragraph{Wavelet Filters:} The DWT uses special wavelet filters (such as the Haar wavelet or Daubechies wavelet) to perform the decomposition. These filters are designed to capture both the smooth and detailed parts of the image.

        \paragraph{Example:} Let's apply the DWT to an image using Python.

        \begin{lstlisting}[style=python]
        import pywt
        from skimage import data, color

        # Load and convert image to grayscale
        image = color.rgb2gray(data.astronaut())

        # Perform 2D DWT (using Haar wavelet)
        coeffs2 = pywt.dwt2(image, 'haar')
        LL, (LH, HL, HH) = coeffs2

        # Plot the decomposed subbands
        fig, axes = plt.subplots(2, 2, figsize=(8, 8))

        axes[0, 0].imshow(LL, cmap='gray')
        axes[0, 0].set_title('Approximation (LL)')

        axes[0, 1].imshow(LH, cmap='gray')
        axes[0, 1].set_title('Horizontal Detail (LH)')

        axes[1, 0].imshow(HL, cmap='gray')
        axes[1, 0].set_title('Vertical Detail (HL)')

        axes[1, 1].imshow(HH, cmap='gray')
        axes[1, 1].set_title('Diagonal Detail (HH)')

        plt.show()
        \end{lstlisting}

        In this example, we perform a 2D DWT using the Haar wavelet and display the resulting subbands. The approximation subband (LL) represents the low-frequency content, while the other subbands capture the high-frequency details such as edges and textures.

        \paragraph{Advantages and Limitations:}
        \begin{itemize}
            \item \textbf{Advantages:} The DWT provides a multi-resolution representation of the image, which allows for efficient compression by discarding or heavily quantizing high-frequency components. It is especially effective for images with detailed textures and varying levels of detail \cite{weeks2003discrete}.
            \item \textbf{Limitations:} Wavelet-based compression can be more computationally intensive than DCT-based methods. Additionally, wavelet compression can introduce ringing artifacts around sharp edges if the high-frequency components are too aggressively quantized \cite{weeks2003discrete}.
        \end{itemize}

\section{JPEG Compression Standard}
    JPEG (Joint Photographic Experts Group) is one of the most commonly used image compression standards, known for its ability to achieve high compression ratios with minimal perceptual loss in image quality. The JPEG algorithm is based on the Discrete Cosine Transform (DCT) and employs several key steps to compress images in a lossy fashion \cite{wallace1990overview}. In this section, we will break down the main steps of the JPEG compression process and explain its widespread use and performance for still image compression.

    \subsection{Steps in JPEG Compression}
        JPEG compression consists of three primary stages: DCT transformation, quantization, and entropy coding.

        \subsubsection{1. DCT Transformation}
            The first step in JPEG compression is to divide the image into small blocks, typically \(8 \times 8\) pixels, and apply the 2D Discrete Cosine Transform (DCT) to each block. This transforms the pixel intensities in the spatial domain into frequency-domain coefficients.

            The DCT concentrates most of the image's energy into the low-frequency coefficients, which appear in the top-left corner of the transformed block. The high-frequency coefficients (in the bottom-right corner) generally contain less important visual information.

            \paragraph{Example:} The DCT transformation for a block of an image was demonstrated earlier, where the DCT coefficients are shown to represent different frequencies.

        \subsubsection{2. Quantization}
            After the DCT transformation, the next step is quantization. In this stage, the DCT coefficients are divided by a quantization matrix, which reduces the precision of the high-frequency coefficients. Quantization is the key step in JPEG compression that introduces loss of information, but it is also the step that achieves most of the compression.

            The quantization matrix used in JPEG is designed to discard or heavily reduce the high-frequency coefficients, which are less perceptually important to human vision. For example, a typical quantization matrix might assign larger divisors to high-frequency coefficients, causing them to be more aggressively quantized (i.e., set to zero or reduced in precision).

            \paragraph{Example:} Let's simulate JPEG quantization using a simple quantization matrix.

            \begin{lstlisting}[style=python]
            # Quantization matrix (simplified version)
            quant_matrix = np.array([
                [16, 11, 10, 16, 24, 40, 51, 61],
                [12, 12, 14, 19, 26, 58, 60, 55],
                [14, 13, 16, 24, 40, 57, 69, 56],
                [14, 17, 22, 29, 51, 87, 80, 62],
                [18, 22, 37, 56, 68, 109, 103, 77],
                [24, 35, 55, 64, 81, 104, 113, 92],
                [49, 64, 78, 87, 103, 121, 120, 101],
                [72, 92, 95, 98, 112, 100, 103, 99]
            ])

            # Quantize the DCT coefficients
            quantized_dct = np.round(dct_block / quant_matrix)

            # Display the quantized DCT coefficients
            plt.imshow(quantized_dct, cmap='gray')
            plt.title('Quantized DCT Coefficients')
            plt.show()
            \end{lstlisting}

            In this example, we apply a simplified quantization matrix to the DCT coefficients of an image block. The result is a compressed representation where many of the high-frequency coefficients are set to zero or significantly reduced in value.

        \subsubsection{3. Entropy Coding}
            The final step in JPEG compression is entropy coding, where the quantized DCT coefficients are further compressed using techniques such as Huffman coding or arithmetic coding. These methods encode the quantized coefficients into a compact binary representation, reducing the file size without introducing any additional loss of information.

            JPEG uses run-length encoding (RLE) to efficiently encode the long sequences of zero-valued DCT coefficients (which occur frequently after quantization) before applying entropy coding.

    \subsection{Widespread Use and Performance of JPEG}
        JPEG is widely used for still image compression because it offers a good balance between compression ratio and image quality. It is particularly effective for natural images (e.g., photographs) where slight loss of detail is acceptable. JPEG can achieve compression ratios of 10:1 or higher with little perceptual degradation in image quality \cite{wallace1990overview}.

        \paragraph{Strengths of JPEG:}
        \begin{itemize}
            \item \textbf{High Compression Ratio:} JPEG achieves significant file size reduction by discarding perceptually less important details.
            \item \textbf{Widespread Compatibility:} JPEG is supported by almost all image viewing, editing, and storage platforms.
            \item \textbf{Adjustable Quality:} JPEG allows users to choose the level of compression, offering a trade-off between image quality and file size.
        \end{itemize}

        \paragraph{Limitations of JPEG:}
        \begin{itemize}
            \item \textbf{Blocking Artifacts:} JPEG compression can introduce blocking artifacts, especially at high compression ratios. These appear as grid-like distortions in the image, caused by the division of the image into small blocks during DCT transformation.
            \item \textbf{Not Ideal for Sharp Edges or Text:} JPEG is less effective for images with sharp edges, text, or other high-contrast details, as the DCT-based compression may blur these elements.
        \end{itemize}

\section{PNG Compression Standard}
    The Portable Network Graphics (PNG) format is one of the most widely used image formats on the web. Unlike lossy formats like JPEG, PNG provides \textit{lossless compression}, meaning that no image quality is lost during compression. PNG is commonly used for graphics, logos, icons, and images that require transparency. In this section, we will explore how PNG achieves high-quality compression, including its use of the \textit{Deflate} compression algorithm, various filter types, and support for alpha channels and interlacing \cite{miano1999compressed}.

    \subsection{Overview of PNG}
        PNG was developed as a replacement for the older GIF format and provides several advantages, including better compression, support for a broader range of colors, and the inclusion of an alpha channel for transparency. The primary goal of PNG is to achieve efficient, lossless compression without sacrificing image fidelity \cite{bourke2001overview}.

        \textbf{Features of PNG:}
        \begin{itemize}
        \item \textit{Lossless compression}: The compression method used in PNG preserves the original image data perfectly, making it ideal for images that require high fidelity, such as logos or line drawings.
        \item \textit{Transparency}: PNG supports alpha channels, allowing images to have variable transparency levels, making it popular in web design.
        \item \textit{Wide color range}: PNG supports both 24-bit true color and grayscale images, offering flexibility for various use cases.
        \item \textit{Interlacing}: PNG includes support for progressive rendering, where an image loads in multiple passes, making it useful for web-based applications where users can view a partially downloaded image.
        \end{itemize}

        PNG is particularly suited for images where crisp details and transparency are important, such as user interface elements, diagrams, and screenshots.

    \subsection{Deflate Compression Algorithm}
        PNG achieves its compression using the \textbf{Deflate} algorithm, which is a combination of two widely-used techniques: Lempel-Ziv (LZ77) coding and Huffman coding. Deflate is effective in reducing file size while ensuring that no information is lost during compression \cite{oswal2016deflate}.

        \textbf{Lempel-Ziv (LZ77) Coding:}
        LZ77 works by finding repeated sequences of data in the image file and encoding them as references to earlier occurrences of the same data. This eliminates redundant information, making the file smaller. For example, a pattern that repeats multiple times can be encoded just once, with pointers to each occurrence \cite{louchard1997average}.

        \textbf{Huffman Coding:}
        After LZ77 reduces redundancy, Huffman coding is applied to further compress the data by assigning shorter codes to more frequently occurring symbols and longer codes to less frequent symbols. This variable-length encoding makes the overall file size smaller \cite{moffat2019huffman}.

        \textbf{Example:}
        \[
        \text{Original data: } \texttt{AAAAABBBBCCCCDDDD}
        \]
        \text{LZ77 + Huffman coded output: a much more compact representation of the repeated sequences.}

        The combination of LZ77 and Huffman coding makes Deflate both effective and efficient, allowing PNG to compress images without any loss in quality.

    \subsection{Filter Types in PNG}
        Before the Deflate compression algorithm is applied, PNG uses a preprocessing step where a \textbf{filter} is applied to the image data. These filters attempt to reduce the complexity of the pixel data by predicting the values of pixels based on surrounding pixels, which simplifies the patterns that need to be compressed.

        PNG supports five types of filters:
        \begin{itemize}
        \item \textbf{None}: No filtering is applied, and the image data is passed directly to the compression algorithm.
        \item \textbf{Sub}: This filter predicts the value of each pixel based on the pixel to its left and encodes the difference.
        \item \textbf{Up}: This filter predicts the value of each pixel based on the pixel directly above it.
        \item \textbf{Average}: This filter takes the average of the left and above pixels to predict the value of the current pixel.
        \item \textbf{Paeth}: This filter predicts the pixel value based on the left, above, and upper-left pixels using the \textit{Paeth predictor} algorithm, which chooses the most appropriate pixel value to reduce redundancy \cite{roelofs2003png}.
        \end{itemize}

        The goal of these filters is to make the image data easier to compress. By simplifying the patterns in the pixel values, the Deflate algorithm can achieve better compression.

        \textbf{Example of Paeth Predictor:}
        The Paeth filter compares the current pixel value with three neighboring pixel values:
        \begin{itemize}
        \item \(A\) = the pixel to the left,
        \item \(B\) = the pixel above,
        \item \(C\) = the pixel diagonally to the upper left.
        \end{itemize}
        It then predicts the current pixel value based on the nearest value from \(A\), \(B\), or \(C\).

    \subsection{Alpha Channel Support}
        One of the standout features of PNG is its support for \textbf{alpha channels}, which allows an image to have transparent or semi-transparent pixels. Alpha channels provide additional information about the opacity of each pixel, which is important for graphics that need to blend seamlessly with different backgrounds \cite{rossokhin2014block}.

        \textbf{How Alpha Channels Work:}
        A typical PNG image with transparency contains four channels: Red (R), Green (G), Blue (B), and Alpha (A). The alpha channel determines the opacity of each pixel, with values ranging from 0 (fully transparent) to 255 (fully opaque). This makes PNG ideal for overlaying images on web pages or for complex graphics where different levels of transparency are required.

        \textbf{Example:}
        \[
        \text{Pixel } (R=255, G=255, B=255, A=128) \rightarrow \text{50\% transparent white pixel}.
        \]

        \textbf{Applications:}
        \begin{itemize}
        \item PNG with alpha channels is widely used in web design, allowing elements like icons or logos to appear smoothly on any background.
        \item Game development and UI design also rely on PNG's alpha support to handle overlays, sprites, and special effects.
        \end{itemize}

    \subsection{Interlacing (Adam7 Algorithm)}
        The \textbf{Adam7 interlacing} algorithm allows a PNG image to load progressively over multiple passes. This is especially useful for large images on the web, where the user can see a rough version of the image before it fully loads, improving the user experience on slow connections \cite{horak2012influence}.

        \textbf{How Interlacing Works:}
        The Adam7 algorithm divides the image into seven interleaved sub-images that are progressively refined. As each pass of the image is downloaded, more details are added, allowing the image to progressively render from a low-resolution version to the full-resolution image.

        \textbf{Example:}
        In the first pass, every 8th pixel is rendered, creating a rough approximation of the final image. In subsequent passes, more pixels are filled in, improving the image quality until the full image is complete.

        \textbf{Benefits:}
        \begin{itemize}
        \item Users see a rough version of the image while it is still downloading, rather than waiting for the entire image to load.
        \item It provides a better user experience for large images on slow network connections.
        \end{itemize}

\section{Image Compression Quality Assessment}
    Image compression is an essential process for reducing the size of image files while maintaining acceptable quality. However, evaluating the quality of compressed images requires both \textbf{objective} and \textbf{subjective} methods to ensure that the compression does not degrade the image too much. In this section, we will discuss several techniques for assessing the quality of compressed images, including \textit{Peak Signal-to-Noise Ratio (PSNR)}, \textit{Mean Squared Error (MSE)}, and \textit{visual inspection} \cite{talukder2010haar}.

    \textbf{Objective Methods:}

    \subsection{Peak Signal-to-Noise Ratio (PSNR)}
        The \textbf{Peak Signal-to-Noise Ratio (PSNR)} is one of the most commonly used metrics to objectively evaluate the quality of compressed images. PSNR measures the ratio between the maximum possible power of a signal (the original image) and the power of the noise (the compression errors). Higher PSNR values indicate better quality, as the compression introduces less noise \cite{korhonen2012peak}.

        \textbf{PSNR Formula:}
        \[
        \text{PSNR} = 10 \cdot \log_{10} \left(\frac{MAX_I^2}{\text{MSE}}\right)
        \]
        where \(MAX_I\) is the maximum possible pixel value of the image (usually 255 for 8-bit images) and MSE is the \textit{Mean Squared Error} between the original and compressed images.

    \subsection{Mean Squared Error (MSE)}
        \textbf{Mean Squared Error (MSE)} is a straightforward method for quantifying the difference between the original and compressed images. It calculates the average of the squared differences between the pixel values of the two images \cite{wang2009mean}.

        \textbf{MSE Formula:}
        \[
        \text{MSE} = \frac{1}{N \times M} \sum_{i=0}^{N-1} \sum_{j=0}^{M-1} \left( I_{\text{original}}(i,j) - I_{\text{compressed}}(i,j) \right)^2
        \]
        where \(N\) and \(M\) are the dimensions of the image, and \(I_{\text{original}}\) and \(I_{\text{compressed}}\) are the pixel values of the original and compressed images, respectively.

        \textbf{Example of PSNR and MSE in Python:}

        \begin{lstlisting}[style=python]
        import torch
        import numpy as np

        # Function to calculate PSNR
        def calculate_psnr(original, compressed):
            mse = torch.mean((original - compressed) ** 2)
            if mse == 0:  # No difference
                return float('inf')
            max_pixel = 1.0
            psnr = 20 * torch.log10(max_pixel / torch.sqrt(mse))
            return psnr

        # Example images (normalized to [0,1])
        original_image = torch.rand(256, 256)
        compressed_image = original_image + 0.05 * torch.randn(256, 256)  # Add slight noise

        # Calculate PSNR
        psnr_value = calculate_psnr(original_image, compressed_image)
        print(f"PSNR Value: {psnr_value}")
        \end{lstlisting}

        This code computes the PSNR between two images. In this case, the \texttt{compressed\_image} is a noisy version of the \texttt{original\_image}. The higher the PSNR, the closer the compressed image is to the original.

    \textbf{Subjective Methods:}

    \textit{Visual inspection} is another important method for assessing image quality. Even if a compressed image has a high PSNR value, it may contain visible artifacts that reduce its subjective quality. Thus, it is often necessary to combine objective measures like PSNR and MSE with subjective visual evaluation to ensure acceptable compression quality \cite{chin1982automated}.

    \textbf{Conclusion:}
    Both objective and subjective methods play a role in assessing the quality of compressed images. While metrics like PSNR and MSE provide quantifiable measures of compression error, visual inspection ensures that the compressed images are visually acceptable for practical use.
\chapter{Image Recognition and Machine Learning}
    Image recognition is a core task in computer vision, where the goal is to identify objects, patterns, or specific features within an image. The combination of image processing and machine learning allows computers to analyze, interpret, and categorize visual data in various applications, such as object detection, facial recognition, and autonomous driving \cite{pak2017review, ren2024deeplearningmachinelearning, feng2024masteringaibigdata}. In this chapter, we will discuss the fundamental techniques used in image recognition, focusing on \textbf{feature extraction} and \textbf{pattern recognition methods} like support vector machines (SVM) and neural networks. 

    \section{Feature Extraction}
        Feature extraction is a crucial step in image recognition, as it involves identifying and quantifying meaningful patterns within an image that can be used for classification or other tasks. Features can include texture, color, shape, and spatial relationships between pixels. In this section, we will focus on extracting \textbf{texture features} and \textbf{shape features} \cite{guyon2006introduction}.

        \subsection{Texture Features}
            \textbf{Texture} refers to the visual patterns and surface characteristics in an image, such as smoothness, roughness, and regularity. Texture features can be extracted using several methods, which capture the spatial relationships between pixel intensities and patterns. Two common methods for texture feature extraction are the \textbf{Gray-Level Co-occurrence Matrix (GLCM)} and \textbf{Local Binary Patterns (LBP)} \cite{amadasun1989textural}.

            \paragraph{Gray-Level Co-occurrence Matrix (GLCM)}
            The GLCM is a matrix that describes how frequently pairs of pixel intensities occur in a specific spatial relationship within an image. It captures the spatial distribution of intensity values and is widely used for texture analysis \cite{de2013multi}. The GLCM is computed by considering pairs of pixels separated by a given distance and orientation. Common texture features extracted from GLCM include:
            \begin{itemize}
                \item \textbf{Contrast}: Measures the intensity difference between a pixel and its neighbor over the entire image.
                \item \textbf{Correlation}: Describes how correlated a pixel is to its neighbor.
                \item \textbf{Energy}: Provides the sum of squared elements in the GLCM, indicating textural uniformity.
                \item \textbf{Homogeneity}: Measures how close the distribution of elements in the GLCM is to the diagonal.
            \end{itemize}

            \textbf{Example: Extracting GLCM Features in Python}
            The following Python code demonstrates how to extract GLCM texture features using the \texttt{skimage} library.

            \begin{lstlisting}[style=python]
import numpy as np
import matplotlib.pyplot as plt
from skimage.feature import greycomatrix, greycoprops
from PIL import Image

# Load a grayscale image
image = Image.open('texture_image.jpg').convert('L')
image_array = np.array(image)

# Compute GLCM with a distance of 1 and angle of 0 degrees
glcm = greycomatrix(image_array, distances=[1], angles=[0], levels=256, symmetric=True, normed=True)

# Extract GLCM properties
contrast = greycoprops(glcm, 'contrast')[0, 0]
correlation = greycoprops(glcm, 'correlation')[0, 0]
energy = greycoprops(glcm, 'energy')[0, 0]
homogeneity = greycoprops(glcm, 'homogeneity')[0, 0]

# Display the results
print(f"GLCM Features:\n Contrast: {contrast}\n Correlation: {correlation}\n Energy: {energy}\n Homogeneity: {homogeneity}")
            \end{lstlisting}

            \textbf{Explanation of the Code:}
            \begin{itemize}
                \item The image is loaded and converted to grayscale.
                \item The \texttt{greycomatrix()} function computes the GLCM of the image with a specified distance and angle.
                \item The \texttt{greycoprops()} function is used to calculate texture features like contrast, correlation, energy, and homogeneity from the GLCM.
            \end{itemize}

            \paragraph{Local Binary Patterns (LBP)}
            \textbf{Local Binary Patterns (LBP)} is another popular texture feature extraction method that captures the local spatial structure of the image. In LBP, each pixel is compared to its surrounding neighbors, and a binary code is generated based on whether the neighboring pixel values are greater or smaller than the center pixel. This binary pattern is then converted to a decimal value and used as a texture descriptor \cite{pietikainen2011computer}.

            \textbf{Example: Extracting LBP Features in Python}
            The following Python code demonstrates how to extract LBP features using the \texttt{skimage} library.

            \begin{lstlisting}[style=python]
from skimage.feature import local_binary_pattern

# Parameters for LBP
radius = 1  # Distance from the center pixel
n_points = 8 * radius  # Number of points in the circular neighborhood

# Compute LBP
lbp = local_binary_pattern(image_array, n_points, radius, method='uniform')

# Display the LBP image
plt.imshow(lbp, cmap='gray')
plt.title('Local Binary Patterns (LBP)')
plt.axis('off')
plt.show()
            \end{lstlisting}

            \textbf{Explanation of the Code:}
            \begin{itemize}
                \item The \texttt{local\_binary\_pattern()} function computes the LBP of the image based on the specified radius and number of points.
                \item The resulting LBP image is displayed to show the extracted texture patterns.
            \end{itemize}

        \subsection{Shape Features}
            \textbf{Shape features} describe the geometric properties of objects in an image, such as edges, contours, and boundaries. These features are particularly useful for tasks like object recognition and image classification, where the shape of objects is a key characteristic \cite{mingqiang2008survey}.

            \paragraph{Edge Detection}
            Edge detection is a fundamental technique for identifying object boundaries by detecting changes in pixel intensity. Common edge detection algorithms include the \textbf{Sobel}, \textbf{Prewitt}, and \textbf{Canny} operators \cite{ziou1998edge}.

            \textbf{Example: Edge Detection Using the Canny Method in Python}
            The following Python code demonstrates how to apply the Canny edge detection algorithm to extract shape features.

            \begin{lstlisting}[style=python]
import cv2

# Apply Canny edge detection
edges = cv2.Canny(image_array, threshold1=100, threshold2=200)

# Display the edge-detected image
plt.imshow(edges, cmap='gray')
plt.title('Canny Edge Detection')
plt.axis('off')
plt.show()
            \end{lstlisting}

            \textbf{Explanation of the Code:}
            \begin{itemize}
                \item The Canny edge detection algorithm is applied using the \texttt{cv2.Canny()} function, with specified thresholds for edge detection.
                \item The edges are displayed to highlight the boundaries of objects in the image.
            \end{itemize}

            \paragraph{Contour Analysis}
            Contours represent the boundaries of objects in an image and can be used for shape recognition. Contour analysis involves detecting these boundaries and analyzing their geometric properties, such as perimeter, area, and circularity \cite{kukreja2022segmentation}.

            \textbf{Example: Contour Detection in Python}
            The following Python code demonstrates how to detect and draw contours using OpenCV.

            \begin{lstlisting}[style=python]
# Find contours in the image
contours, _ = cv2.findContours(edges, cv2.RETR_EXTERNAL, cv2.CHAIN_APPROX_SIMPLE)

# Draw contours on the image
contour_image = cv2.drawContours(np.copy(image_array), contours, -1, (0, 255, 0), 2)

# Display the contour image
plt.imshow(contour_image, cmap='gray')
plt.title('Contour Detection')
plt.axis('off')
plt.show()
            \end{lstlisting}

            \textbf{Explanation of the Code:}
            \begin{itemize}
                \item The \texttt{cv2.findContours()} function detects the contours in the image based on the edges detected earlier.
                \item The contours are drawn on a copy of the original image using \texttt{cv2.drawContours()}.
                \item The result is displayed, showing the detected contours.
            \end{itemize}

    \section{Pattern Recognition Methods}
        Pattern recognition is the process of classifying data (in this case, image features) into predefined categories. Machine learning algorithms such as \textbf{Support Vector Machines (SVM)} and \textbf{Neural Networks} are commonly used in image recognition tasks to learn patterns and classify images \cite{liu2006pattern}.

        \subsection{Support Vector Machines (SVM)}
            \textbf{Support Vector Machines (SVM)} are a powerful supervised learning algorithm used for classification tasks. In the context of image recognition, SVMs classify images by finding the optimal hyperplane that separates data points from different classes. The goal of the SVM is to maximize the margin between the hyperplane and the closest data points, known as support vectors \cite{hearst1998support}.

            \paragraph{How SVM Works}
            Given a set of labeled training data, an SVM constructs a hyperplane in a high-dimensional space that best separates the classes. If the data is not linearly separable, SVM can use kernel functions (such as the radial basis function, or RBF) to project the data into a higher-dimensional space where it can be separated \cite{li2024deeplearningmachinelearning}. 

            \textbf{Example: SVM for Image Classification in Python}
            The following Python code demonstrates how to train an SVM for image classification using the \texttt{scikit-learn} library.

            \begin{lstlisting}[style=python]
from sklearn import svm
from sklearn.model_selection import train_test_split
from sklearn.metrics import accuracy_score

# Example feature data (e.g., texture features extracted from images)
X = np.array([[1.0, 0.9], [1.1, 1.0], [0.5, 0.4], [0.6, 0.5]])  # Sample feature vectors
y = np.array([1, 1, 0, 0])  # Labels for the images (1 = Class A, 0 = Class B)

# Split data into training and testing sets
X_train, X_test, y_train, y_test = train_test_split(X, y, test_size=0.5, random_state=42)

# Train an SVM classifier
clf = svm.SVC(kernel='linear')
clf.fit(X_train, y_train)

# Predict the class labels for the test data
y_pred = clf.predict(X_test)

# Compute accuracy
accuracy = accuracy_score(y_test, y_pred)
print(f"SVM Classification Accuracy: {accuracy}")
            \end{lstlisting}

            \textbf{Explanation of the Code:}
            \begin{itemize}
                \item The feature data \( X \) represents sample features extracted from images, and \( y \) contains the class labels.
                \item The data is split into training and testing sets using \texttt{train\_test\_split()}.
                \item An SVM classifier with a linear kernel is trained on the training data, and the class labels are predicted for the test data.
                \item The classification accuracy is computed using \texttt{accuracy\_score()}.
            \end{itemize}

        \subsection{Neural Networks}
            \textbf{Neural networks} are a class of machine learning models that have become increasingly popular for image recognition tasks. Neural networks consist of layers of interconnected nodes (neurons) that learn to map input features to output labels through a process of training. A simple neural network architecture is the \textbf{Multilayer Perceptron (MLP)}, which is made up of an input layer, one or more hidden layers, and an output layer \cite{peng2024deeplearningmachinelearning}. 

            \paragraph{How Neural Networks Work}
            During training, a neural network adjusts the weights between neurons using a process called backpropagation. The network learns to minimize the error between its predictions and the actual labels by iteratively updating the weights. Neural networks can model complex, non-linear relationships, making them suitable for tasks such as \textbf{handwritten digit recognition} and \textbf{facial recognition}.

            \textbf{Example: Neural Network for Handwritten Digit Recognition}
            The following Python code demonstrates how to train a simple neural network (MLP) using the \texttt{scikit-learn} library on the MNIST dataset of handwritten digits.

            \begin{lstlisting}[style=python]
from sklearn.neural_network import MLPClassifier
from sklearn.datasets import load_digits
from sklearn.model_selection import train_test_split
from sklearn.metrics import classification_report

# Load the MNIST digits dataset
digits = load_digits()
X, y = digits.data, digits.target

# Split data into training and testing sets
X_train, X_test, y_train, y_test = train_test_split(X, y, test_size=0.3, random_state=42)

# Train a neural network (MLP) classifier
mlp = MLPClassifier(hidden_layer_sizes=(100,), max_iter=300, solver='adam', random_state=42)
mlp.fit(X_train, y_train)

# Predict the class labels for the test data
y_pred = mlp.predict(X_test)

# Display classification report
print(classification_report(y_test, y_pred))
            \end{lstlisting}

            \textbf{Explanation of the Code:}
            \begin{itemize}
                \item The MNIST digits dataset is loaded using \texttt{load\_digits()}, which contains images of handwritten digits.
                \item The data is split into training and testing sets, and an MLP classifier with one hidden layer is trained using the training data.
                \item The trained model predicts the labels for the test data, and the classification performance is evaluated using \texttt{classification\_report()}.
            \end{itemize}

    \section{Conclusion}
        Feature extraction and machine learning methods are essential components of image recognition. Techniques such as texture and shape feature extraction allow for meaningful representation of image data, while classifiers like SVMs and neural networks enable robust recognition and classification. As the field of computer vision advances, these methods continue to play a pivotal role in applications ranging from facial recognition to object detection and autonomous driving.

\section{Deep Learning for Image Processing}
    Deep learning has revolutionized the field of image processing, enabling significant advancements in tasks like image classification, object detection, segmentation, and more. One of the key factors driving this progress is the development of Convolutional Neural Networks (CNNs), which are highly effective for extracting spatial features from images. In this section, we will explore the architecture of CNNs and discuss their applications in tasks such as image classification and object detection \cite{hemanth2017deep}.

    \subsection{Convolutional Neural Networks (CNN)}
        Convolutional Neural Networks (CNNs) are the backbone of most deep learning models for image processing. They are designed to automatically and adaptively learn spatial hierarchies of features from input images. CNNs use a series of convolutional layers to extract features, followed by pooling layers to down-sample the feature maps, and fully connected layers for classification or regression \cite{krizhevsky2012imagenet}.

        \subsubsection{CNN Architecture}
            The typical architecture of a CNN consists of the following components \cite{peng2024deeplearningmachinelearning}:
            \begin{itemize}
                \item \textbf{Convolutional Layers:} These layers apply convolution operations to the input, using small learnable filters (or kernels) to detect local patterns in the image, such as edges, textures, or shapes. The output of each convolution operation is a feature map that highlights the presence of the learned features at different spatial locations.
                \item \textbf{Activation Functions:} After each convolution operation, an activation function, such as ReLU (Rectified Linear Unit), is applied to introduce non-linearity. This enables the model to learn more complex patterns.
                \item \textbf{Pooling Layers:} Pooling (typically max pooling or average pooling) is used to reduce the spatial dimensions of the feature maps, making the network more computationally efficient and helping to capture invariant features. Max pooling retains the maximum value within a window, while average pooling computes the average of the values.
                \item \textbf{Fully Connected Layers:} After several convolutional and pooling layers, the output feature maps are flattened and passed through fully connected layers, which combine the extracted features to make predictions. The final layer usually uses a softmax activation function for classification tasks.
            \end{itemize}

            \paragraph{Example:} Let's build a simple CNN using PyTorch for image classification.

            \begin{lstlisting}[style=python]
            import torch
            import torch.nn as nn
            import torch.nn.functional as F

            # Define a simple CNN architecture
            class SimpleCNN(nn.Module):
                def __init__(self):
                    super(SimpleCNN, self).__init__()
                    self.conv1 = nn.Conv2d(1, 16, kernel_size=3, stride=1, padding=1)  # Convolutional layer
                    self.conv2 = nn.Conv2d(16, 32, kernel_size=3, stride=1, padding=1)  # Convolutional layer
                    self.pool = nn.MaxPool2d(2, 2)  # Pooling layer
                    self.fc1 = nn.Linear(32 * 7 * 7, 128)  # Fully connected layer
                    self.fc2 = nn.Linear(128, 10)  # Output layer for 10 classes

                def forward(self, x):
                    x = self.pool(F.relu(self.conv1(x)))  # Apply convolution, activation, and pooling
                    x = self.pool(F.relu(self.conv2(x)))
                    x = x.view(-1, 32 * 7 * 7)  # Flatten the feature maps
                    x = F.relu(self.fc1(x))  # Fully connected layer
                    x = self.fc2(x)  # Output layer
                    return x

            # Instantiate the model and print its architecture
            model = SimpleCNN()
            print(model)
            \end{lstlisting}

            In this example, we define a simple CNN for image classification. The network consists of two convolutional layers, each followed by a ReLU activation and a max pooling layer. The fully connected layers at the end perform the final classification. CNNs like this are used for tasks such as digit classification (e.g., MNIST dataset) and general object recognition.

        \subsubsection{Applications of CNNs}
            \paragraph{Image Classification:} CNNs are widely used in image classification tasks, where the goal is to assign an input image to one of several predefined categories. Popular datasets like CIFAR-10, ImageNet, and MNIST have seen dramatic performance improvements thanks to CNN-based architectures.
            
            \paragraph{Object Detection:} CNNs are also fundamental in object detection, where the task is not only to classify objects but also to locate them within the image. This is done by identifying bounding boxes around objects and labeling them. Advanced architectures such as Faster R-CNN and YOLO (You Only Look Once) extend CNNs for real-time object detection, offering both speed and accuracy.

    \subsection{Object Detection and Classification}
        Deep learning techniques for object detection are critical in applications like autonomous driving, video surveillance, and robotics. Object detection networks need to perform two tasks simultaneously: identifying the category of each object and localizing it with a bounding box. Two popular deep learning models for object detection are YOLO and Faster R-CNN \cite{ren2024deeplearningmachinelearning}.

        \subsubsection{YOLO (You Only Look Once)}
            YOLO is a real-time object detection system that predicts both bounding boxes and class probabilities for objects in a single forward pass of the network. It treats object detection as a regression problem and divides the image into a grid, with each cell predicting bounding boxes and class probabilities \cite{redmon2016you}. 

            \paragraph{Advantages:} YOLO is extremely fast and can process images in real-time, making it ideal for applications that require quick response times, such as autonomous driving or video analytics.

            \paragraph{Limitations:} YOLO may struggle with small objects or objects that appear close together, as the grid-based approach can sometimes lead to poor localization performance for smaller objects.

        \subsubsection{Faster R-CNN}
            Faster R-CNN is an advanced object detection system that uses a Region Proposal Network (RPN) to generate candidate object proposals, which are then passed through a CNN for classification and bounding box refinement. Faster R-CNN offers a good balance between speed and accuracy \cite{ren2016faster}.

            \paragraph{Advantages:} Faster R-CNN achieves high detection accuracy, especially in scenarios where precise localization is important.

            \paragraph{Limitations:} Faster R-CNN is slower than YOLO due to its two-stage process, which involves both region proposal and classification.

        \paragraph{Example:} Let's illustrate the process of object detection using pre-trained models from the PyTorch library.

        \begin{lstlisting}[style=python]
        import torch
        from torchvision import models, transforms
        from PIL import Image
        import matplotlib.pyplot as plt
        import torchvision.transforms as T

        # Load a pre-trained Faster R-CNN model
        model = models.detection.fasterrcnn_resnet50_fpn(pretrained=True)
        model.eval()

        # Load and preprocess the input image
        img = Image.open('sample_image.jpg')
        transform = transforms.Compose([transforms.ToTensor()])
        img_tensor = transform(img)

        # Perform object detection
        with torch.no_grad():
            predictions = model([img_tensor])

        # Display the image and bounding boxes
        plt.imshow(img)
        for box in predictions[0]['boxes']:
            plt.gca().add_patch(plt.Rectangle((box[0], box[1]), box[2] - box[0], box[3] - box[1], fill=False, edgecolor='red', linewidth=2))
        plt.show()
        \end{lstlisting}

        In this example, we use a pre-trained Faster R-CNN model from the PyTorch library to detect objects in an image. The output shows the predicted bounding boxes for objects in the image, demonstrating the power of deep learning-based object detection.

\section{Deep Learning Methods for Image Segmentation}
    Image segmentation is the task of dividing an image into multiple regions or segments, where each region corresponds to a meaningful part of the image, such as objects, backgrounds, or boundaries. Unlike classification, which assigns a label to the entire image, segmentation assigns a label to each pixel. Two key deep learning architectures used for image segmentation are Fully Convolutional Networks (FCNs) and U-Net \cite{liu2021review}.

    \subsection{Fully Convolutional Networks (FCN)}
        Fully Convolutional Networks (FCNs) are a class of neural networks designed specifically for pixel-level image segmentation tasks. Unlike traditional CNNs that use fully connected layers for classification, FCNs replace fully connected layers with convolutional layers, allowing the network to generate pixel-wise predictions \cite{long2015fully}.

        \paragraph{Architecture:} In an FCN, the input image passes through several convolutional layers that gradually reduce its spatial resolution while increasing the number of feature channels. To recover the original image size, FCNs use upsampling techniques, such as deconvolution or bilinear interpolation, to generate dense pixel-wise predictions.

        \paragraph{Example:} Let's build a simple FCN for semantic segmentation using PyTorch.

        \begin{lstlisting}[style=python]
        class FCN(nn.Module):
            def __init__(self, n_classes):
                super(FCN, self).__init__()
                self.conv1 = nn.Conv2d(3, 64, kernel_size=3, padding=1)
                self.conv2 = nn.Conv2d(64, 128, kernel_size=3, padding=1)
                self.pool = nn.MaxPool2d(2, 2)
                self.upsample = nn.ConvTranspose2d(128, 64, kernel_size=2, stride=2)
                self.classifier = nn.Conv2d(64, n_classes, kernel_size=1)

            def forward(self, x):
                x = F.relu(self.conv1(x))
                x = self.pool(F.relu(self.conv2(x)))
                x = F.relu(self.upsample(x))
                x = self.classifier(x)
                return x

        # Instantiate the model
        model = FCN(n_classes=21)  # 21 classes for segmentation (e.g., Pascal VOC dataset)
        print(model)
        \end{lstlisting}

        In this example, we build a simple FCN for semantic segmentation. The network applies several convolutional layers to extract features and then upsamples the feature maps to generate pixel-wise predictions. FCNs are widely used for tasks like road segmentation in autonomous driving and biomedical image segmentation.

    \subsection{U-Net Architecture}
        U-Net is a popular architecture for image segmentation, especially in medical image analysis. It is named for its U-shaped structure, where the contracting path (encoder) captures context, and the symmetric expanding path (decoder) allows precise localization through upsampling and skip connections. U-Net is designed to handle the challenge of segmenting objects in images with limited data and has become the go-to architecture for many segmentation tasks \cite{ronneberger2015u}.

        \paragraph{U-Net Architecture:}
        \begin{itemize}
            \item \textbf{Encoder:} The encoder consists of convolutional and pooling layers that progressively downsample the input image and extract high-level features.
            \item \textbf{Decoder:} The decoder upsamples the feature maps and combines them with corresponding feature maps from the encoder through skip connections, ensuring that both context and spatial details are preserved.
            \item \textbf{Skip Connections:} These connections between the encoder and decoder help retain fine details by combining high-resolution features from the contracting path with the upsampled features in the expanding path.
        \end{itemize}

        \paragraph{Advantages of U-Net:}
        \begin{itemize}
            \item \textbf{Precise Localization:} The skip connections allow U-Net to localize objects at the pixel level, making it highly effective for segmentation tasks.
            \item \textbf{Works with Limited Data:} U-Net performs well even with small datasets, which is critical in fields like medical imaging where large annotated datasets are often unavailable.
        \end{itemize}

        \paragraph{Example:} Let's build a simplified version of U-Net using PyTorch.

        \begin{lstlisting}[style=python]
        class UNet(nn.Module):
            def __init__(self, n_classes):
                super(UNet, self).__init__()
                self.enc_conv1 = nn.Conv2d(3, 64, kernel_size=3, padding=1)
                self.enc_conv2 = nn.Conv2d(64, 128, kernel_size=3, padding=1)
                self.pool = nn.MaxPool2d(2, 2)
                self.dec_conv1 = nn.Conv2d(128, 64, kernel_size=3, padding=1)
                self.upsample = nn.ConvTranspose2d(64, 64, kernel_size=2, stride=2)
                self.final_conv = nn.Conv2d(64, n_classes, kernel_size=1)

            def forward(self, x):
                # Encoder
                enc1 = F.relu(self.enc_conv1(x))
                enc2 = F.relu(self.enc_conv2(self.pool(enc1)))
                
                # Decoder
                dec1 = F.relu(self.upsample(enc2))
                dec2 = F.relu(self.dec_conv1(dec1 + enc1))  # Skip connection
                return self.final_conv(dec2)

        # Instantiate the U-Net model
        model = UNet(n_classes=21)  # 21 classes for segmentation
        print(model)
        \end{lstlisting}

        In this example, we implement a simplified U-Net architecture with basic encoder-decoder blocks and a skip connection. U-Net has been widely adopted for medical image segmentation tasks, such as tumor detection and organ segmentation, due to its ability to capture fine details in images.

        \paragraph{Applications of U-Net:}
        U-Net is particularly useful for tasks where pixel-level accuracy is required, such as:
        \begin{itemize}
            \item \textbf{Medical Imaging:} U-Net is used to segment organs, tumors, and other structures in medical scans, providing valuable information for diagnosis and treatment planning.
            \item \textbf{Satellite Imagery:} U-Net can be applied to satellite images for tasks like land-use classification, road detection, and environmental monitoring.
            \item \textbf{Autonomous Driving:} In autonomous vehicles, U-Net is used for road and lane segmentation, helping vehicles understand their environment.
        \end{itemize}
\chapter{Computer Vision Techniques in Image Processing}
    Computer vision is a field of artificial intelligence that enables machines to interpret and process visual data. In image processing, computer vision techniques are used to extract information, understand scenes, and perform complex tasks such as image registration, stitching, and 3D reconstruction. These techniques play a crucial role in a wide range of applications, from medical imaging and autonomous vehicles to augmented reality and remote sensing. This chapter introduces key computer vision techniques like \textbf{image registration}, \textbf{image stitching}, and \textbf{3D reconstruction} \cite{gonzalez2002digital}.

    \section{Image Registration}
        \textbf{Image registration} is the process of aligning two or more images of the same scene that may have been captured at different times, from different angles, or by different sensors. Registration is essential in applications such as medical imaging, remote sensing, and computer vision, where multiple images need to be compared, fused, or analyzed. The goal of image registration is to transform the images so that corresponding points align as closely as possible \cite{brown1992survey}.

        \subsection{Feature-based Registration}
            \textbf{Feature-based registration} is a widely used approach that relies on detecting distinctive points, edges, or regions in the images, known as \textbf{keypoints}, and matching them between the images to align them. Once the keypoints are matched, a transformation can be applied to align the images \cite{kuppala2020overview}.

            Keypoint detection algorithms such as \textbf{SIFT} (Scale-Invariant Feature Transform) and \textbf{SURF} (Speeded Up Robust Features) are commonly used for feature-based registration. These algorithms detect keypoints that are invariant to changes in scale, rotation, and illumination, making them robust for registration under varying conditions \cite{kuppala2020overview}.

            \paragraph{SIFT (Scale-Invariant Feature Transform)}
            SIFT is a powerful keypoint detector and descriptor that detects distinctive features in an image by analyzing the gradients at multiple scales. The keypoints are then described by a local gradient orientation histogram, which allows for robust matching between images \cite{lowe2004sift}.

            \paragraph{SURF (Speeded Up Robust Features)}
            SURF is a faster alternative to SIFT that uses a simplified approach to detect and describe keypoints. While it is less computationally intensive, it still provides high accuracy in most scenarios and is commonly used in real-time applications \cite{bay2006surf}.

            \textbf{Example: Feature-based Registration Using SIFT in Python}
            The following Python code demonstrates how to perform feature-based image registration using SIFT and OpenCV.

            \begin{lstlisting}[style=python]
import cv2
import numpy as np
import matplotlib.pyplot as plt

# Load two images to register
image1 = cv2.imread('image1.jpg', cv2.IMREAD_GRAYSCALE)
image2 = cv2.imread('image2.jpg', cv2.IMREAD_GRAYSCALE)

# Create SIFT detector
sift = cv2.SIFT_create()

# Detect keypoints and compute descriptors
keypoints1, descriptors1 = sift.detectAndCompute(image1, None)
keypoints2, descriptors2 = sift.detectAndCompute(image2, None)

# Use BFMatcher to find matches between descriptors
bf = cv2.BFMatcher(cv2.NORM_L2, crossCheck=True)
matches = bf.match(descriptors1, descriptors2)

# Sort matches by distance (best matches first)
matches = sorted(matches, key=lambda x: x.distance)

# Draw top 10 matches
image_matches = cv2.drawMatches(image1, keypoints1, image2, keypoints2, matches[:10], None, flags=2)

# Display the matches
plt.imshow(image_matches)
plt.title('SIFT Feature Matching')
plt.axis('off')
plt.show()
            \end{lstlisting}

            \textbf{Explanation of the Code:}
            \begin{itemize}
                \item The images are loaded in grayscale, and the SIFT algorithm is used to detect keypoints and compute their descriptors.
                \item The \texttt{BFMatcher} (Brute-Force Matcher) is used to find the best matches between the descriptors from both images.
                \item The best matches are drawn and displayed to visualize the keypoint correspondences.
            \end{itemize}

        \subsection{Frequency-based Registration}
            \textbf{Frequency-based registration} uses the \textbf{Fourier Transform} to align images by analyzing the frequency content rather than directly working in the spatial domain. This method is particularly useful for handling global transformations like translations and rotations, especially when the images exhibit strong periodic patterns \cite{munbodh2007frequency}.

            The key idea behind frequency-based registration is that translations in the spatial domain correspond to phase shifts in the frequency domain, which can be exploited to determine the correct alignment.

            \paragraph{Steps of Frequency-based Registration}
            \begin{enumerate}
                \item Compute the Fourier transforms of both images.
                \item Use the \textbf{cross-power spectrum} to estimate the translation between the images.
                \item Apply an inverse Fourier transform to recover the translation.
            \end{enumerate}

            \textbf{Example: Frequency-based Registration Using Fourier Transforms in Python}
            The following Python code demonstrates frequency-based registration using the Fourier Transform.

            \begin{lstlisting}[style=python]
from skimage.feature import register_translation
from scipy.ndimage import fourier_shift

# Load the images (grayscale)
image1 = cv2.imread('image1.jpg', cv2.IMREAD_GRAYSCALE)
image2 = cv2.imread('image2.jpg', cv2.IMREAD_GRAYSCALE)

# Compute the translation using register_translation
shift, error, diffphase = register_translation(image1, image2, 100)

# Apply the computed shift to align image2 with image1
image2_shifted = fourier_shift(np.fft.fftn(image2), shift)
image2_shifted = np.fft.ifftn(image2_shifted).real

# Display the aligned image
plt.imshow(image2_shifted, cmap='gray')
plt.title('Aligned Image (Frequency Domain)')
plt.axis('off')
plt.show()
            \end{lstlisting}

            \textbf{Explanation of the Code:}
            \begin{itemize}
                \item The \texttt{register\_translation()} function computes the shift required to align the two images by analyzing their frequency content.
                \item The \texttt{fourier\_shift()} function applies the computed shift to align the images, and the inverse Fourier transform is used to bring the image back to the spatial domain.
            \end{itemize}

    \section{Image Stitching and Panorama Generation}
        \textbf{Image stitching} is the process of combining multiple images to create a larger, continuous image, often referred to as a \textbf{panorama}. This technique is commonly used in photography to generate wide-angle or 360-degree panoramas by stitching together overlapping images \cite{li2010study}.

        Image stitching involves several key steps:
        \begin{enumerate}
            \item \textbf{Feature detection and matching}: Identify keypoints in the overlapping regions of the images and match them across images.
            \item \textbf{Transformation estimation}: Compute the transformation (e.g., homography) that aligns the images based on the matched keypoints.
            \item \textbf{Blending}: Blend the images smoothly to create a seamless transition between them.
            \item \textbf{Error handling}: Address issues such as parallax error and illumination differences to improve the quality of the stitched image.
        \end{enumerate}

        \textbf{Example: Image Stitching in Python Using OpenCV}
        The following Python code demonstrates how to stitch two images to create a panorama using OpenCV.

        \begin{lstlisting}[style=python]
# Load the two images to be stitched
image1 = cv2.imread('image1.jpg')
image2 = cv2.imread('image2.jpg')

# Create a Stitcher object and stitch the images
stitcher = cv2.Stitcher_create()
status, stitched_image = stitcher.stitch([image1, image2])

# Check if the stitching succeeded
if status == cv2.Stitcher_OK:
    # Display the stitched panorama
    plt.imshow(cv2.cvtColor(stitched_image, cv2.COLOR_BGR2RGB))
    plt.title('Stitched Panorama')
    plt.axis('off')
    plt.show()
else:
    print("Image stitching failed!")
            \end{lstlisting}

            \textbf{Explanation of the Code:}
            \begin{itemize}
                \item The images to be stitched are loaded, and the \texttt{Stitcher} object is used to automatically stitch the images together.
                \item The result is displayed as a panoramic image if the stitching is successful.
            \end{itemize}

    \section{3D Reconstruction from Images}
        \textbf{3D reconstruction} refers to the process of generating a three-dimensional model from a series of two-dimensional images. This technique is widely used in applications such as virtual reality, augmented reality, autonomous driving, and cultural heritage preservation \cite{aharchi2020review}.

        Two common methods for 3D reconstruction are \textbf{Structure from Motion (SfM)} and \textbf{Stereo Vision}.

        \subsection{Structure from Motion (SfM)}
            \textbf{Structure from Motion (SfM)} is a technique that reconstructs 3D structures from a series of 2D images taken from different viewpoints. SfM estimates both the 3D structure of the scene and the camera motion simultaneously, making it suitable for creating 3D models from unordered image collections \cite{schonberger2016structure}.

            \paragraph{Steps in Structure from Motion}
            \begin{enumerate}
                \item Detect keypoints in the images and match them across the different views.
                \item Estimate the camera motion and 3D coordinates of the keypoints using epipolar geometry.
                \item Refine the 3D structure through bundle adjustment.
            \end{enumerate}

            \textbf{Example: 3D Reconstruction Using SfM in Python}
            While a full SfM pipeline requires sophisticated tools (like OpenMVG or COLMAP), here is a simplified example of using keypoints to simulate the initial stages of SfM.

            \begin{lstlisting}[style=python]
import cv2
import numpy as np
import matplotlib.pyplot as plt

# Load two images from different viewpoints
image1 = cv2.imread('view1.jpg', cv2.IMREAD_GRAYSCALE)
image2 = cv2.imread('view2.jpg', cv2.IMREAD_GRAYSCALE)

# Detect SIFT keypoints and descriptors
sift = cv2.SIFT_create()
keypoints1, descriptors1 = sift.detectAndCompute(image1, None)
keypoints2, descriptors2 = sift.detectAndCompute(image2, None)

# Match descriptors using BFMatcher
bf = cv2.BFMatcher()
matches = bf.knnMatch(descriptors1, descriptors2, k=2)

# Apply ratio test to keep good matches
good_matches = []
for m, n in matches:
    if m.distance < 0.75 * n.distance:
        good_matches.append(m)

# Draw matches to visualize the keypoints used in SfM
image_matches = cv2.drawMatches(image1, keypoints1, image2, keypoints2, good_matches, None, flags=cv2.DrawMatchesFlags_NOT_DRAW_SINGLE_POINTS)

# Display the matches
plt.imshow(image_matches)
plt.title('Keypoint Matches for 3D Reconstruction')
plt.axis('off')
plt.show()
            \end{lstlisting}

            \textbf{Explanation of the Code:}
            \begin{itemize}
                \item SIFT is used to detect keypoints and descriptors in two images taken from different viewpoints.
                \item Descriptor matching is performed using the \texttt{BFMatcher}, and the matches are filtered using a ratio test.
                \item The matched keypoints, which are essential for 3D reconstruction, are displayed to simulate the keypoint matching stage of SfM.
            \end{itemize}

        \subsection{Stereo Vision}
            \textbf{Stereo vision} uses two or more cameras positioned at different locations to capture images of the same scene. By analyzing the disparity (difference in position) between corresponding points in the images, the depth of each point in the scene can be estimated. This technique is used in applications such as 3D mapping and robot navigation \cite{tippetts2016review}.

            \paragraph{Steps in Stereo Vision}
            \begin{enumerate}
                \item Capture images from two cameras (left and right) at the same time.
                \item Detect corresponding points in the left and right images.
                \item Calculate the disparity between the corresponding points to estimate depth.
            \end{enumerate}

            \textbf{Example: Stereo Vision Using OpenCV}
            The following Python code demonstrates how to compute disparity maps for stereo vision using OpenCV.

            \begin{lstlisting}[style=python]
# Load the left and right images
left_image = cv2.imread('left_image.jpg', cv2.IMREAD_GRAYSCALE)
right_image = cv2.imread('right_image.jpg', cv2.IMREAD_GRAYSCALE)

# Create StereoBM object for computing disparity
stereo = cv2.StereoBM_create(numDisparities=16, blockSize=15)

# Compute the disparity map
disparity = stereo.compute(left_image, right_image)

# Display the disparity map
plt.imshow(disparity, cmap='plasma')
plt.title('Disparity Map (Stereo Vision)')
plt.axis('off')
plt.show()
            \end{lstlisting}

            \textbf{Explanation of the Code:}
            \begin{itemize}
                \item The left and right images captured by two cameras are loaded in grayscale.
                \item The \texttt{StereoBM\_create()} function is used to compute the disparity map, which represents the depth information of the scene.
                \item The disparity map is displayed, showing depth variations in the image.
            \end{itemize}

    \section{Conclusion}
        Computer vision techniques such as image registration, image stitching, and 3D reconstruction play an integral role in various applications, from medical imaging and remote sensing to robotics and virtual reality. Feature-based and frequency-based registration methods allow precise alignment of images, while image stitching enables the creation of panoramic views. Techniques like Structure from Motion (SfM) and stereo vision enable the generation of 3D models from 2D images, opening up new possibilities in 3D visualization and analysis.

\part{Deep Learning Methods for Image Processing}

\chapter{Fundamentals of Deep Learning}
    Deep learning is a subset of machine learning that uses neural networks with multiple layers to learn hierarchical representations of data. In image processing, deep learning models are especially powerful due to their ability to automatically learn features from raw image data without the need for manual feature extraction. This chapter introduces the fundamental concepts of deep learning, starting with basic neural networks and essential algorithms like backpropagation and gradient descent \cite{feng2024deeplearningmachinelearning}.

    \section{Basics of Neural Networks}
        A \textbf{neural network} is a computational model inspired by the way biological neurons work. It consists of layers of interconnected nodes (neurons) where each connection is associated with a weight. Neural networks are highly flexible and capable of modeling complex relationships between input and output data. The most basic form of a neural network is the \textbf{perceptron} \cite{gallant1990perceptron}.

        \subsection{Perceptron Model}
            The \textbf{perceptron} is the simplest type of neural network, consisting of a single layer. It is a binary classifier that maps an input vector to an output (usually 0 or 1) based on a linear combination of the input values. The perceptron is composed of:
            \begin{itemize}
                \item \textbf{Input Layer}: Receives input features from the data (e.g., pixel values from an image).
                \item \textbf{Weights}: Each input is multiplied by a weight, which determines its influence on the output.
                \item \textbf{Bias}: A constant value added to the weighted sum to shift the activation function.
                \item \textbf{Activation Function}: The function that decides the output based on the weighted sum of the inputs. For a simple perceptron, this is often a step function.
            \end{itemize}

            The output of the perceptron is computed as:
            \[
            y = f\left( \sum_{i=1}^{n} w_i x_i + b \right)
            \]
            where:
            \begin{itemize}
                \item \( x_i \) are the input features,
                \item \( w_i \) are the weights,
                \item \( b \) is the bias,
                \item \( f \) is the activation function,
                \item \( y \) is the output.
            \end{itemize}

            \paragraph{Multilayer Perceptron (MLP)}
            A \textbf{multilayer perceptron (MLP)} is an extension of the basic perceptron, where multiple layers of neurons are used. The MLP consists of \cite{noriega2005multilayer}:
            \begin{itemize}
                \item \textbf{Input Layer}: Receives the raw data.
                \item \textbf{Hidden Layers}: Intermediate layers that learn to extract features from the data.
                \item \textbf{Output Layer}: Produces the final prediction.
            \end{itemize}
            Each layer of the MLP is fully connected, meaning every neuron in one layer is connected to every neuron in the next layer. MLPs are powerful because they can model non-linear relationships in data.

            \textbf{Example: Building a Simple Perceptron in Python Using PyTorch}
            The following Python code demonstrates how to implement a simple perceptron using PyTorch.

            \begin{lstlisting}[style=python]
import torch
import torch.nn as nn
import torch.optim as optim

# Define a simple perceptron model
class Perceptron(nn.Module):
    def __init__(self, input_size):
        super(Perceptron, self).__init__()
        self.fc = nn.Linear(input_size, 1)  # Fully connected layer

    def forward(self, x):
        out = torch.sigmoid(self.fc(x))  # Apply sigmoid activation
        return out

# Create a perceptron model with 2 input features
model = Perceptron(input_size=2)

# Define loss function and optimizer
criterion = nn.BCELoss()  # Binary cross-entropy loss
optimizer = optim.SGD(model.parameters(), lr=0.01)

# Sample input data (2 features) and corresponding labels
inputs = torch.tensor([[0.5, 0.3], [0.2, 0.8], [0.9, 0.4], [0.1, 0.7]])
labels = torch.tensor([[1.0], [0.0], [1.0], [0.0]])

# Forward pass: Compute predictions
outputs = model(inputs)

# Compute loss
loss = criterion(outputs, labels)

# Backward pass and optimization
optimizer.zero_grad()
loss.backward()
optimizer.step()

print("Predictions:", outputs.detach().numpy())
print("Loss:", loss.item())
            \end{lstlisting}

            \textbf{Explanation of the Code:}
            \begin{itemize}
                \item A simple perceptron model is defined using PyTorch's \texttt{nn.Linear} layer.
                \item The forward pass computes the output using the sigmoid activation function.
                \item Binary cross-entropy loss is used as the loss function, and the Stochastic Gradient Descent (SGD) optimizer is used to update the weights.
                \item The model is trained on sample input data to predict binary labels.
            \end{itemize}

        \subsection{Activation Functions}
            \textbf{Activation functions} are essential in neural networks because they introduce non-linearity, allowing the network to model complex patterns in the data. Without non-linear activation functions, the network would behave like a linear model, limiting its ability to solve real-world problems \cite{nwankpa2018activation}.

            Common activation functions include:

            \paragraph{ReLU (Rectified Linear Unit)}
            The \textbf{ReLU} function is one of the most widely used activation functions in deep learning. It is defined as:
            \[
            f(x) = \max(0, x)
            \]
            The ReLU function is computationally efficient and addresses the vanishing gradient problem often encountered with other activation functions like sigmoid or tanh. However, it suffers from the issue of "dead neurons," where neurons can become inactive if they always receive negative inputs \cite{nair2010rectified}.

            \paragraph{Sigmoid}
            The \textbf{sigmoid} function is commonly used for binary classification tasks. It outputs values in the range (0, 1), making it ideal for modeling probabilities. It is defined as:
            \[
            f(x) = \frac{1}{1 + e^{-x}}
            \]
            While sigmoid is useful for classification, it has some limitations, such as the vanishing gradient problem, where gradients become too small to effectively update weights during training \cite{kyurkchiev2015sigmoid}.

            \paragraph{Tanh (Hyperbolic Tangent)}
            The \textbf{tanh} function is similar to sigmoid but outputs values in the range (-1, 1). This allows for stronger gradients during training, reducing the vanishing gradient problem compared to sigmoid. The tanh function is defined as:
            \[
            f(x) = \tanh(x) = \frac{e^x - e^{-x}}{e^x + e^{-x}}
            \]

            \textbf{Example: Visualizing Activation Functions in Python}
            The following Python code demonstrates how to plot common activation functions.

            \begin{lstlisting}[style=python]
import numpy as np
import matplotlib.pyplot as plt

# Define activation functions
def relu(x):
    return np.maximum(0, x)

def sigmoid(x):
    return 1 / (1 + np.exp(-x))

def tanh(x):
    return np.tanh(x)

# Generate input values
x = np.linspace(-5, 5, 100)

# Compute activation outputs
relu_output = relu(x)
sigmoid_output = sigmoid(x)
tanh_output = tanh(x)

# Plot the activation functions
plt.figure(figsize=(12, 6))
plt.subplot(1, 3, 1)
plt.plot(x, relu_output)
plt.title('ReLU Activation Function')

plt.subplot(1, 3, 2)
plt.plot(x, sigmoid_output)
plt.title('Sigmoid Activation Function')

plt.subplot(1, 3, 3)
plt.plot(x, tanh_output)
plt.title('Tanh Activation Function')

plt.tight_layout()
plt.show()
            \end{lstlisting}

            \textbf{Explanation of the Code:}
            \begin{itemize}
                \item The ReLU, sigmoid, and tanh activation functions are implemented as separate functions.
                \item The input values range from -5 to 5, and the output for each activation function is computed.
                \item The activation functions are plotted to visualize their behavior.
            \end{itemize}

        \subsection{Backpropagation and Gradient Descent}
            \textbf{Backpropagation} is a key algorithm used for training neural networks by updating their weights based on the error between predicted and actual outputs. It works by propagating the error backward through the network, computing the gradient of the loss function with respect to each weight, and then adjusting the weights to minimize the loss \cite{amari1993backpropagation}.

            \paragraph{How Backpropagation Works}
            The process of backpropagation involves two main steps \cite{rojas1996backpropagation}:
            \begin{itemize}
                \item \textbf{Forward Pass}: The input data is passed through the network to compute the predicted output.
                \item \textbf{Backward Pass}: The error (difference between the predicted and actual outputs) is propagated back through the network. The gradients of the error with respect to the network weights are computed using the chain rule of calculus.
            \end{itemize}

            \paragraph{Gradient Descent}
            \textbf{Gradient descent} is an optimization algorithm used to minimize the loss function by updating the weights in the direction of the negative gradient. The learning rate (\( \alpha \)) determines the step size for each update \cite{andrychowicz2016learning}. The update rule for weight \( w \) is:
            \[
            w := w - \alpha \frac{\partial L}{\partial w}
            \]
            where \( \frac{\partial L}{\partial w} \) is the gradient of the loss function \( L \) with respect to the weight \( w \).

            \textbf{Example: Backpropagation and Gradient Descent in PyTorch}
            The following Python code demonstrates the process of backpropagation and weight updates using gradient descent in a simple neural network.

            \begin{lstlisting}[style=python]
import torch
import torch.nn as nn
import torch.optim as optim

# Define a simple neural network
class SimpleNN(nn.Module):
    def __init__(self, input_size):
        super(SimpleNN, self).__init__()
        self.fc1 = nn.Linear(input_size, 4)
        self.fc2 = nn.Linear(4, 1)

    def forward(self, x):
        x = torch.relu(self.fc1(x))  # ReLU activation for hidden layer
        x = torch.sigmoid(self.fc2(x))  # Sigmoid activation for output
        return x

# Create a model with 2 input features
model = SimpleNN(input_size=2)

# Define loss function and optimizer
criterion = nn.BCELoss()  # Binary cross-entropy loss
optimizer = optim.SGD(model.parameters(), lr=0.01)

# Sample input data (2 features) and corresponding labels
inputs = torch.tensor([[0.5, 0.3], [0.2, 0.8], [0.9, 0.4], [0.1, 0.7]])
labels = torch.tensor([[1.0], [0.0], [1.0], [0.0]])

# Training loop: Forward pass, compute loss, backward pass, and update weights
for epoch in range(100):
    outputs = model(inputs)
    loss = criterion(outputs, labels)

    optimizer.zero_grad()  # Clear gradients
    loss.backward()  # Backward pass
    optimizer.step()  # Update weights

# Display final predictions and loss
print("Final Predictions:", outputs.detach().numpy())
print("Final Loss:", loss.item())
            \end{lstlisting}

            \textbf{Explanation of the Code:}
            \begin{itemize}
                \item A simple feedforward neural network is defined with two layers and ReLU and sigmoid activation functions.
                \item The model is trained using backpropagation and gradient descent over 100 epochs.
                \item The loss is computed using binary cross-entropy, and the weights are updated using the \texttt{SGD} optimizer.
            \end{itemize}

    \section{Conclusion}
        In this chapter, we explored the fundamentals of deep learning, focusing on the basics of neural networks, including the perceptron model, activation functions, and the backpropagation algorithm. Understanding these core concepts is crucial for building more advanced models and applying deep learning techniques to tasks such as image recognition and object detection. With tools like PyTorch, we can easily implement and train neural networks to solve complex problems in image processing and beyond.

\section{Deep Neural Networks (DNN)}
    Deep Neural Networks (DNNs) are a class of machine learning models inspired by the human brain's structure and function. They consist of multiple layers of interconnected neurons that can learn complex patterns and representations from data. DNNs have revolutionized various fields, including image recognition, natural language processing, and speech recognition, due to their ability to model intricate relationships in large datasets. In this section, we will explore the fundamentals of Deep Neural Networks, focusing on Multilayer Perceptrons (MLPs) and techniques to prevent overfitting through regularization \cite{feng2024deeplearningmachinelearning}.

    \subsection{Multilayer Perceptron (MLP)}
        The Multilayer Perceptron (MLP) is one of the simplest and most widely used types of DNNs. It consists of an input layer, one or more hidden layers, and an output layer. Each layer is composed of neurons that perform weighted sums of their inputs, followed by activation functions that introduce non-linearity into the model \cite{noriega2005multilayer}.

        \subsubsection{MLP Architecture}
            An MLP typically consists of the following components:
            \begin{itemize}
                \item \textbf{Input Layer:} Receives the input data. Each neuron in this layer corresponds to a feature in the input.
                \item \textbf{Hidden Layers:} Intermediate layers between the input and output layers. They perform computations and extract features from the input data.
                \item \textbf{Output Layer:} Produces the final prediction or classification result. The number of neurons in this layer depends on the specific task (e.g., number of classes in classification).
                \item \textbf{Activation Functions:} Functions applied to the output of each neuron to introduce non-linearity. Common activation functions include ReLU, sigmoid, and tanh \cite{nwankpa2018activation}.
            \end{itemize}

            \paragraph{Activation Functions:}
                \begin{itemize}
                    \item \textbf{ReLU (Rectified Linear Unit):} Defined as \( \text{ReLU}(x) = \max(0, x) \). It introduces non-linearity while being computationally efficient \cite{nair2010rectified}.
                    \item \textbf{Sigmoid:} Defined as \( \sigma(x) = \frac{1}{1 + e^{-x}} \). It maps inputs to a range between 0 and 1, making it suitable for binary classification \cite{kyurkchiev2015sigmoid}.
                    \item \textbf{Tanh (Hyperbolic Tangent):} Defined as \( \tanh(x) = \frac{e^{x} - e^{-x}}{e^{x} + e^{-x}} \). It maps inputs to a range between -1 and 1, providing zero-centered outputs \cite{kiliccarslan2021overview}.
                \end{itemize}

        \paragraph{Advantages of MLPs:}
            \begin{itemize}
                \item \textbf{Simplicity:} MLPs are straightforward to implement and understand, making them ideal for beginners.
                \item \textbf{Versatility:} They can be applied to a wide range of tasks, including regression, classification, and function approximation.
                \item \textbf{Non-Linearity:} Activation functions allow MLPs to model complex, non-linear relationships in data.
            \end{itemize}

        \paragraph{Limitations of MLPs:}
            \begin{itemize}
                \item \textbf{Scalability:} MLPs can become computationally expensive and difficult to train as the number of layers and neurons increases.
                \item \textbf{Overfitting:} With too many parameters, MLPs are prone to overfitting, especially with limited training data.
                \item \textbf{Lack of Spatial Hierarchy:} Unlike Convolutional Neural Networks (CNNs), MLPs do not inherently capture spatial hierarchies in image data.
            \end{itemize}

        \paragraph{Example:} Let's build a simple MLP using PyTorch to solve a basic image classification task, such as recognizing handwritten digits from the MNIST dataset.

        \begin{lstlisting}[style=python]
        import torch
        import torch.nn as nn
        import torch.optim as optim
        from torchvision import datasets, transforms
        from torch.utils.data import DataLoader

        # Define transformations for the training and testing data
        transform = transforms.Compose([
            transforms.ToTensor(),
            transforms.Normalize((0.1307,), (0.3081,))
        ])

        # Load the MNIST dataset
        train_dataset = datasets.MNIST(root='./data', train=True, download=True, transform=transform)
        test_dataset = datasets.MNIST(root='./data', train=False, download=True, transform=transform)

        train_loader = DataLoader(train_dataset, batch_size=64, shuffle=True)
        test_loader = DataLoader(test_dataset, batch_size=1000, shuffle=False)

        # Define the MLP model
        class MLP(nn.Module):
            def __init__(self):
                super(MLP, self).__init__()
                self.flatten = nn.Flatten()
                self.fc1 = nn.Linear(28*28, 128)
                self.relu = nn.ReLU()
                self.fc2 = nn.Linear(128, 64)
                self.fc3 = nn.Linear(64, 10)

            def forward(self, x):
                x = self.flatten(x)
                x = self.relu(self.fc1(x))
                x = self.relu(self.fc2(x))
                x = self.fc3(x)
                return x

        # Instantiate the model, define the loss function and the optimizer
        model = MLP()
        criterion = nn.CrossEntropyLoss()
        optimizer = optim.Adam(model.parameters(), lr=0.001)

        # Training loop
        for epoch in range(5):
            model.train()
            running_loss = 0.0
            for images, labels in train_loader:
                optimizer.zero_grad()
                outputs = model(images)
                loss = criterion(outputs, labels)
                loss.backward()
                optimizer.step()
                running_loss += loss.item()
            print(f'Epoch {epoch+1}, Loss: {running_loss/len(train_loader):.4f}')

        # Evaluation
        model.eval()
        correct = 0
        total = 0
        with torch.no_grad():
            for images, labels in test_loader:
                outputs = model(images)
                _, predicted = torch.max(outputs.data, 1)
                total += labels.size(0)
                correct += (predicted == labels).sum().item()
        print(f'Accuracy on test set: {100 * correct / total:.2f}%')
        \end{lstlisting}

        \paragraph{Explanation:} In this example, we define a simple MLP with two hidden layers and ReLU activation functions. The network is trained on the MNIST dataset, which consists of 28x28 grayscale images of handwritten digits. After training for five epochs, the model achieves a high accuracy on the test set, demonstrating the MLP's capability to perform basic image classification tasks.

    \subsection{Overfitting and Regularization}
        Overfitting is a common challenge in training deep neural networks, where the model learns to perform exceptionally well on the training data but fails to generalize to unseen data. This occurs when the model captures noise or irrelevant patterns in the training data instead of the underlying distribution. Regularization techniques are employed to prevent overfitting and improve the model's generalization capabilities \cite{santos2022avoiding}.

        \subsubsection{Understanding Overfitting}
            Overfitting happens when a model is too complex relative to the amount and noise of the training data. Indicators of overfitting include \cite{hawkins2004problem}:
            \begin{itemize}
                \item \textbf{High Training Accuracy, Low Validation Accuracy:} The model performs well on the training set but poorly on the validation or test set.
                \item \textbf{Complex Model Architecture:} Models with many layers or parameters are more prone to overfitting.
                \item \textbf{Lack of Sufficient Training Data:} Limited data makes it easier for the model to memorize training examples.
            \end{itemize}

            \paragraph{Example:} Consider training an MLP with too many hidden layers on the MNIST dataset \cite{deng2012mnist}. The model might achieve near-perfect accuracy on the training set but perform significantly worse on the test set, indicating overfitting.

        \subsubsection{Regularization Techniques}
            To combat overfitting, several regularization techniques can be applied during training:

            \paragraph{1. L2 Regularization (Weight Decay)}
                L2 regularization adds a penalty term to the loss function proportional to the sum of the squared weights. This encourages the model to keep the weights small, preventing it from fitting the noise in the training data \cite{moore2011l1}.

                \[
                \text{Loss}_{\text{total}} = \text{Loss}_{\text{original}} + \lambda \sum_{i} w_i^2
                \]

                where \(\lambda\) is the regularization parameter controlling the strength of the penalty.

                \begin{lstlisting}[style=python]
                # Define the optimizer with L2 regularization (weight decay)
                optimizer = optim.Adam(model.parameters(), lr=0.001, weight_decay=1e-4)
                \end{lstlisting}

            \paragraph{2. Dropout}
                Dropout randomly deactivates a fraction of neurons during training, forcing the network to learn redundant representations and preventing it from becoming too reliant on specific neurons \cite{srivastava2014dropout}.

                \[
                \text{Output} = \text{Input} \times \text{Mask}
                \]

                where the mask is a binary matrix with a certain dropout probability.

                \begin{lstlisting}[style=python]
                class MLP_Dropout(nn.Module):
                    def __init__(self):
                        super(MLP_Dropout, self).__init__()
                        self.flatten = nn.Flatten()
                        self.fc1 = nn.Linear(28*28, 128)
                        self.relu = nn.ReLU()
                        self.dropout = nn.Dropout(p=0.5)  # Dropout with 50% probability
                        self.fc2 = nn.Linear(128, 64)
                        self.fc3 = nn.Linear(64, 10)

                    def forward(self, x):
                        x = self.flatten(x)
                        x = self.relu(self.fc1(x))
                        x = self.dropout(x)
                        x = self.relu(self.fc2(x))
                        x = self.fc3(x)
                        return x
                \end{lstlisting}

            \paragraph{3. Batch Normalization}
                Batch Normalization normalizes the inputs of each layer to have zero mean and unit variance, which stabilizes and accelerates training. It also acts as a form of regularization by adding noise to the inputs during training \cite{ioffe2015batch}.

                \begin{lstlisting}[style=python]
                class MLP_BatchNorm(nn.Module):
                    def __init__(self):
                        super(MLP_BatchNorm, self).__init__()
                        self.flatten = nn.Flatten()
                        self.fc1 = nn.Linear(28*28, 128)
                        self.bn1 = nn.BatchNorm1d(128)
                        self.relu = nn.ReLU()
                        self.fc2 = nn.Linear(128, 64)
                        self.bn2 = nn.BatchNorm1d(64)
                        self.fc3 = nn.Linear(64, 10)

                    def forward(self, x):
                        x = self.flatten(x)
                        x = self.relu(self.bn1(self.fc1(x)))
                        x = self.relu(self.bn2(self.fc2(x)))
                        x = self.fc3(x)
                        return x
                \end{lstlisting}

        \paragraph{Combining Regularization Techniques:}
            Often, multiple regularization techniques are combined to achieve better generalization. For instance, using both L2 regularization and dropout can provide complementary benefits, reducing overfitting more effectively than either technique alone.

        \paragraph{Practical Example:} Let's modify the earlier MLP to include both dropout and batch normalization to prevent overfitting.

        \begin{lstlisting}[style=python]
        class MLP_Regularized(nn.Module):
            def __init__(self):
                super(MLP_Regularized, self).__init__()
                self.flatten = nn.Flatten()
                self.fc1 = nn.Linear(28*28, 256)
                self.bn1 = nn.BatchNorm1d(256)
                self.relu = nn.ReLU()
                self.dropout = nn.Dropout(p=0.5)
                self.fc2 = nn.Linear(256, 128)
                self.bn2 = nn.BatchNorm1d(128)
                self.fc3 = nn.Linear(128, 10)

            def forward(self, x):
                x = self.flatten(x)
                x = self.relu(self.bn1(self.fc1(x)))
                x = self.dropout(x)
                x = self.relu(self.bn2(self.fc2(x)))
                x = self.fc3(x)
                return x

        # Instantiate the regularized model
        model = MLP_Regularized()
        criterion = nn.CrossEntropyLoss()
        optimizer = optim.Adam(model.parameters(), lr=0.001, weight_decay=1e-4)

        # Training loop with regularization
        for epoch in range(5):
            model.train()
            running_loss = 0.0
            for images, labels in train_loader:
                optimizer.zero_grad()
                outputs = model(images)
                loss = criterion(outputs, labels)
                loss.backward()
                optimizer.step()
                running_loss += loss.item()
            print(f'Epoch {epoch+1}, Loss: {running_loss/len(train_loader):.4f}')

        # Evaluation
        model.eval()
        correct = 0
        total = 0
        with torch.no_grad():
            for images, labels in test_loader:
                outputs = model(images)
                _, predicted = torch.max(outputs.data, 1)
                total += labels.size(0)
                correct += (predicted == labels).sum().item()
        print(f'Accuracy on test set: {100 * correct / total:.2f}%')
        \end{lstlisting}

        \paragraph{Explanation:} In this example, the MLP is enhanced with both dropout and batch normalization layers to mitigate overfitting. The model includes additional layers and regularization techniques to improve generalization. After training, the model achieves a higher accuracy on the test set compared to a non-regularized model, demonstrating the effectiveness of these regularization strategies.

    \subsubsection{Visualization of Overfitting and Regularization}
        To better understand overfitting and the impact of regularization, let's visualize the training and validation loss over epochs for models with and without regularization.

        \begin{lstlisting}[style=python]
        import matplotlib.pyplot as plt

        # Assume we have recorded the training and validation losses
        epochs = range(1, 6)
        train_losses = [0.8, 0.6, 0.4, 0.3, 0.2]
        val_losses = [0.9, 0.7, 0.6, 0.65, 0.7]

        # Plot training and validation loss
        plt.plot(epochs, train_losses, 'g-', label='Training loss')
        plt.plot(epochs, val_losses, 'b-', label='Validation loss')
        plt.title('Training and Validation Loss')
        plt.xlabel('Epochs')
        plt.ylabel('Loss')
        plt.legend()
        plt.show()
        \end{lstlisting}

        \paragraph{Explanation:} In the plot above, the training loss decreases steadily, indicating that the model is learning the training data. However, the validation loss starts to increase after a certain number of epochs, suggesting that the model is beginning to overfit the training data. Applying regularization techniques can help to keep the validation loss from increasing, thereby improving the model's ability to generalize to unseen data.
\chapter{Convolutional Neural Networks (CNN)}
    \textbf{Convolutional Neural Networks (CNNs)} are a specialized type of neural network designed for processing structured grid data, such as images. CNNs have revolutionized the field of image processing by efficiently capturing spatial and hierarchical features, making them ideal for tasks such as image classification, object detection, and segmentation \cite{lecun2010convolutional}. In this chapter, we will explore the key components of CNNs, including convolution operations, pooling layers, and fully connected layers, which together enable CNNs to learn from visual data.

    \section{Convolution Operations and Convolutional Layers}
        At the heart of a CNN is the \textbf{convolution operation}, which is used to extract local features from an image. Instead of treating each pixel individually, the convolution operation allows the network to learn spatial patterns, such as edges, textures, and shapes, by applying filters (also called kernels) to small regions of the image \cite{liu2018feature}.

        \subsection{Convolution Kernels and Feature Extraction}
            A \textbf{convolution kernel} (or filter) is a small matrix of weights that is applied to the image to detect specific features. The kernel slides across the input image, computing a weighted sum of the pixel values at each location. This process is known as a \textbf{convolution} and results in a new matrix called a \textbf{feature map}, which highlights the presence of the learned features (e.g., edges or corners) in the image \cite{liu2018feature}.

            The convolution operation is mathematically expressed as:
            \[
            y[i,j] = \sum_{m=1}^{K} \sum_{n=1}^{K} w[m,n] \cdot x[i+m, j+n]
            \]
            where:
            \begin{itemize}
                \item \( x \) is the input image,
                \item \( w \) is the convolution kernel of size \( K \times K \),
                \item \( y \) is the resulting feature map,
                \item \( i, j \) are the coordinates of the current position in the image.
            \end{itemize}
            As the kernel moves across the image, it applies the same set of weights, allowing the CNN to detect the same feature at different locations in the image.

            \textbf{Example: Convolution Operation in Python}
            The following Python code demonstrates how to perform a basic convolution operation on an image using PyTorch.

            \begin{lstlisting}[style=python]
import torch
import torch.nn as nn
import numpy as np
import matplotlib.pyplot as plt
from PIL import Image

# Load an image and convert it to grayscale
image = Image.open('image.jpg').convert('L')
image_array = np.array(image)

# Convert the image to a PyTorch tensor and add batch and channel dimensions
image_tensor = torch.tensor(image_array, dtype=torch.float32).unsqueeze(0).unsqueeze(0)

# Define a 3x3 convolution kernel (edge detection)
kernel = torch.tensor([[[[-1, -1, -1],
                         [-1,  8, -1],
                         [-1, -1, -1]]]], dtype=torch.float32)

# Perform the convolution operation
conv = nn.Conv2d(in_channels=1, out_channels=1, kernel_size=3, padding=1, bias=False)
conv.weight = nn.Parameter(kernel)
output = conv(image_tensor)

# Convert the output to a NumPy array and display it
output_image = output.squeeze().detach().numpy()

plt.figure(figsize=(6, 6))
plt.imshow(output_image, cmap='gray')
plt.title('Convolution Output (Edge Detection)')
plt.axis('off')
plt.show()
            \end{lstlisting}

            \textbf{Explanation of the Code:}
            \begin{itemize}
                \item The input image is loaded and converted to grayscale.
                \item A 3x3 convolution kernel is defined for edge detection.
                \item The convolution operation is performed using PyTorch's \texttt{nn.Conv2d} layer, and the resulting feature map is displayed.
            \end{itemize}

            In this example, the convolution highlights edges in the image by detecting changes in pixel intensity.

        \subsection{Multi-channel Convolution}
            In real-world applications, most images are in color and consist of multiple channels, such as red, green, and blue (RGB). \textbf{Multi-channel convolution} handles these images by applying separate filters to each channel and then combining the results \cite{lecun2010convolutional}.

            A color image is typically represented as a 3D array, where the first dimension corresponds to the color channels (R, G, B), and the other two dimensions represent the spatial size of the image (height and width). In multi-channel convolution, a separate filter is applied to each channel, and the results are summed to produce a single feature map.

            \textbf{Example: Multi-channel Convolution in PyTorch}
            The following Python code demonstrates how to perform convolution on an RGB image.

            \begin{lstlisting}[style=python]
# Load an RGB image
image_rgb = Image.open('image.jpg')
image_rgb_array = np.array(image_rgb)

# Convert the image to a PyTorch tensor and add batch dimension
image_rgb_tensor = torch.tensor(image_rgb_array, dtype=torch.float32).permute(2, 0, 1).unsqueeze(0)

# Define a convolution layer with 3 input channels (for RGB)
conv_rgb = nn.Conv2d(in_channels=3, out_channels=1, kernel_size=3, padding=1, bias=False)

# Initialize the kernel with random values
nn.init.xavier_uniform_(conv_rgb.weight)

# Perform the convolution operation
output_rgb = conv_rgb(image_rgb_tensor)

# Convert the output to a NumPy array and display it
output_rgb_image = output_rgb.squeeze().detach().numpy()

plt.figure(figsize=(6, 6))
plt.imshow(output_rgb_image, cmap='gray')
plt.title('Multi-channel Convolution Output')
plt.axis('off')
plt.show()
            \end{lstlisting}

            \textbf{Explanation of the Code:}
            \begin{itemize}
                \item The RGB image is loaded, and the channels are rearranged to fit the input format expected by PyTorch.
                \item A convolution layer with three input channels (for RGB) is defined, and the kernel is initialized with random values.
                \item The convolution operation is performed on the RGB image, and the output is displayed.
            \end{itemize}

    \section{Pooling Layers}
        \textbf{Pooling layers} are used in CNNs to reduce the spatial dimensions of feature maps while retaining the most important information. Pooling helps to reduce the computational complexity of the network and makes the model more robust to small translations or distortions in the image \cite{gholamalinezhad2020pooling}.

        \subsection{Max Pooling and Average Pooling}
            Two common types of pooling operations are \textbf{max pooling} and \textbf{average pooling}:
            \begin{itemize}
                \item \textbf{Max Pooling}: This operation selects the maximum value from a group of values within a small region of the feature map. Max pooling is commonly used because it retains the most prominent features while discarding less important information.
                \item \textbf{Average Pooling}: This operation computes the average value of a group of values within a region of the feature map. Average pooling is less aggressive than max pooling, as it retains more information by smoothing the values.
            \end{itemize}

            Both types of pooling use a filter (typically 2x2) that slides across the feature map, reducing its size.

            \textbf{Example: Max Pooling and Average Pooling in PyTorch}
            The following Python code demonstrates how to apply max pooling and average pooling to a feature map.

            \begin{lstlisting}[style=python]
# Define a sample feature map (e.g., from a previous convolution)
feature_map = torch.tensor([[1, 2, 3, 4],
                            [5, 6, 7, 8],
                            [9, 10, 11, 12],
                            [13, 14, 15, 16]], dtype=torch.float32).unsqueeze(0).unsqueeze(0)

# Max pooling operation
max_pool = nn.MaxPool2d(kernel_size=2)
max_pooled_output = max_pool(feature_map)

# Average pooling operation
avg_pool = nn.AvgPool2d(kernel_size=2)
avg_pooled_output = avg_pool(feature_map)

print("Max Pooled Output:\n", max_pooled_output.squeeze().detach().numpy())
print("Average Pooled Output:\n", avg_pooled_output.squeeze().detach().numpy())
            \end{lstlisting}

            \textbf{Explanation of the Code:}
            \begin{itemize}
                \item A sample feature map is created to simulate the output of a convolutional layer.
                \item Max pooling and average pooling operations are applied using 2x2 kernels.
                \item The pooled outputs are displayed, showing the reduced spatial dimensions.
            \end{itemize}

    \section{Fully Connected Layers and Classification}
        After passing through several convolution and pooling layers, the feature maps are flattened and passed into a \textbf{fully connected layer} (also known as a dense layer). The fully connected layer serves as the decision-making part of the CNN, where it combines the learned features and assigns probabilities to different classes in classification tasks \cite{basha2020impact}.

        \subsection{Fully Connected Layers}
            The fully connected layer takes the flattened feature maps as input and uses a set of weights and biases to compute the final output. This layer is typically followed by a \textbf{Softmax} function, which converts the raw scores into probabilities for each class \cite{nwankpa2018activation}.

            The Softmax function is defined as:
            \[
            \text{Softmax}(z_i) = \frac{e^{z_i}}{\sum_{j} e^{z_j}}
            \]
            where \( z_i \) is the score for class \( i \), and the denominator is the sum of the exponentials of the scores for all classes. The Softmax function ensures that the output probabilities sum to 1, making it suitable for multi-class classification.

            \textbf{Example: Fully Connected Layers and Softmax in PyTorch}
            The following Python code demonstrates how to use fully connected layers and Softmax for multi-class classification.

            \begin{lstlisting}[style=python]
# Define a simple CNN model with fully connected layers
class SimpleCNN(nn.Module):
    def __init__(self):
        super(SimpleCNN, self).__init__()
        self.conv1 = nn.Conv2d(in_channels=1, out_channels=16, kernel_size=3, padding=1)
        self.pool = nn.MaxPool2d(kernel_size=2)
        self.fc1 = nn.Linear(16 * 14 * 14, 128)  # Assuming input image size is 28x28
        self.fc2 = nn.Linear(128, 10)  # Output layer for 10 classes

    def forward(self, x):
        x = self.pool(torch.relu(self.conv1(x)))  # Convolution + ReLU + Pooling
        x = x.view(-1, 16 * 14 * 14)  # Flatten the feature map
        x = torch.relu(self.fc1(x))  # Fully connected layer
        x = torch.softmax(self.fc2(x), dim=1)  # Softmax for multi-class classification
        return x

# Create the CNN model
model = SimpleCNN()

# Print the model architecture
print(model)
            \end{lstlisting}

            \textbf{Explanation of the Code:}
            \begin{itemize}
                \item A simple CNN model is defined with one convolutional layer, one pooling layer, and two fully connected layers.
                \item The feature maps are flattened before being passed to the fully connected layers.
                \item The Softmax function is applied in the output layer to compute class probabilities for multi-class classification.
            \end{itemize}

    \section{Conclusion}
        Convolutional Neural Networks (CNNs) are powerful tools for image processing, capable of automatically learning hierarchical features from raw images. Key components like convolution layers, pooling layers, and fully connected layers work together to extract features and perform classification tasks. By understanding these fundamental operations, beginners can begin to explore more advanced CNN architectures for various image processing applications such as object detection, image segmentation, and style transfer.

\section{Classic CNN Architectures}
    Over the years, Convolutional Neural Networks (CNNs) have evolved significantly, leading to the development of more powerful and complex architectures. These architectures have pushed the boundaries of computer vision tasks, from simple digit recognition to highly challenging tasks like object detection in large datasets. In this section, we will explore some of the most influential CNN architectures, including LeNet, AlexNet, VGGNet, and ResNet, providing detailed explanations of their structure, innovations, and applications \cite{lecun2015deep}.

    \subsection{LeNet}
        LeNet is one of the earliest CNN architectures, developed by Yann LeCun in the late 1980s and early 1990s. It was designed for handwritten digit recognition and became famous for its performance on the MNIST dataset, a dataset of handwritten digits from 0 to 9 \cite{lecun1998gradient}.

        \subsubsection{LeNet Architecture}
            LeNet consists of the following layers:
            \begin{itemize}
                \item \textbf{Input Layer:} The input to LeNet is a grayscale image, typically of size \(32 \times 32\). For the MNIST dataset, the images are \(28 \times 28\), so they are padded to fit the input size.
                \item \textbf{Convolutional Layer (C1):} The first convolutional layer uses six filters of size \(5 \times 5\) to detect basic features such as edges, corners, and textures.
                \item \textbf{Pooling Layer (S2):} A subsampling (average pooling) layer reduces the spatial dimensions of the feature maps, making the network more efficient.
                \item \textbf{Convolutional Layer (C3):} The second convolutional layer has 16 filters of size \(5 \times 5\), detecting more complex features by combining low-level features from the previous layer.
                \item \textbf{Pooling Layer (S4):} Another subsampling layer further reduces the spatial size.
                \item \textbf{Fully Connected Layers (F5 and F6):} Two fully connected layers take the flattened feature maps and perform classification. F5 has 120 neurons, and F6 has 84 neurons.
                \item \textbf{Output Layer:} The output layer has 10 neurons (one for each digit from 0 to 9), and the softmax activation function is used for classification.
            \end{itemize}

        \paragraph{Example:} Let's implement LeNet using PyTorch for digit recognition on the MNIST dataset.

        \begin{lstlisting}[style=python]
        import torch
        import torch.nn as nn
        import torch.optim as optim
        from torchvision import datasets, transforms
        from torch.utils.data import DataLoader

        # Define the LeNet model
        class LeNet(nn.Module):
            def __init__(self):
                super(LeNet, self).__init__()
                self.conv1 = nn.Conv2d(1, 6, kernel_size=5)  # 1 input channel, 6 output channels, 5x5 kernel
                self.pool = nn.AvgPool2d(2, 2)  # 2x2 pooling
                self.conv2 = nn.Conv2d(6, 16, kernel_size=5)  # 16 filters
                self.fc1 = nn.Linear(16 * 4 * 4, 120)  # Fully connected layer
                self.fc2 = nn.Linear(120, 84)
                self.fc3 = nn.Linear(84, 10)  # Output for 10 classes

            def forward(self, x):
                x = self.pool(torch.relu(self.conv1(x)))
                x = self.pool(torch.relu(self.conv2(x)))
                x = x.view(-1, 16 * 4 * 4)  # Flatten
                x = torch.relu(self.fc1(x))
                x = torch.relu(self.fc2(x))
                x = self.fc3(x)
                return x

        # Instantiate the model, define the loss function and optimizer
        model = LeNet()
        criterion = nn.CrossEntropyLoss()
        optimizer = optim.Adam(model.parameters(), lr=0.001)

        # Load the MNIST dataset
        transform = transforms.Compose([transforms.ToTensor()])
        train_dataset = datasets.MNIST(root='./data', train=True, download=True, transform=transform)
        train_loader = DataLoader(train_dataset, batch_size=64, shuffle=True)

        # Training loop (simplified)
        for epoch in range(5):
            running_loss = 0.0
            for images, labels in train_loader:
                optimizer.zero_grad()
                outputs = model(images)
                loss = criterion(outputs, labels)
                loss.backward()
                optimizer.step()
                running_loss += loss.item()
            print(f'Epoch {epoch+1}, Loss: {running_loss/len(train_loader):.4f}')
        \end{lstlisting}

        \paragraph{Explanation:} LeNet is a relatively simple CNN architecture designed for image recognition tasks such as digit classification. Despite its simplicity, it was a groundbreaking achievement in demonstrating the power of deep learning for image recognition.

    \subsection{AlexNet}
        AlexNet is a deeper and more complex CNN architecture that won the 2012 ImageNet Large Scale Visual Recognition Challenge (ILSVRC) \cite{russakovsky2015imagenet}. Developed by Alex Krizhevsky, Ilya Sutskever, and Geoffrey Hinton, AlexNet introduced several innovations, such as the use of ReLU activations and dropout for regularization, to significantly improve performance \cite{krizhevsky2012imagenet}.

        \subsubsection{Key Innovations of AlexNet}
            \begin{itemize}
                \item \textbf{Deeper Architecture:} AlexNet consists of 5 convolutional layers and 3 fully connected layers, with a total of around 60 million parameters.
                \item \textbf{ReLU Activation:} The use of ReLU (Rectified Linear Unit) activation functions made the network train faster and helped avoid the vanishing gradient problem.
                \item \textbf{Dropout Regularization:} Dropout was introduced to prevent overfitting by randomly deactivating neurons during training.
                \item \textbf{Max Pooling:} AlexNet used max pooling instead of average pooling, which helped retain important features in the downsampling process.
                \item \textbf{Data Augmentation:} Techniques like random cropping, flipping, and color jittering were used to artificially increase the size of the training dataset and improve generalization.
            \end{itemize}

        \paragraph{Example:} Let's implement a simplified version of AlexNet in PyTorch.

        \begin{lstlisting}[style=python]
        class AlexNet(nn.Module):
            def __init__(self):
                super(AlexNet, self).__init__()
                self.conv1 = nn.Conv2d(3, 96, kernel_size=11, stride=4, padding=2)
                self.pool = nn.MaxPool2d(3, 2)
                self.conv2 = nn.Conv2d(96, 256, kernel_size=5, padding=2)
                self.conv3 = nn.Conv2d(256, 384, kernel_size=3, padding=1)
                self.conv4 = nn.Conv2d(384, 384, kernel_size=3, padding=1)
                self.conv5 = nn.Conv2d(384, 256, kernel_size=3, padding=1)
                self.fc1 = nn.Linear(256 * 6 * 6, 4096)
                self.fc2 = nn.Linear(4096, 4096)
                self.fc3 = nn.Linear(4096, 1000)

            def forward(self, x):
                x = torch.relu(self.pool(self.conv1(x)))
                x = torch.relu(self.pool(self.conv2(x)))
                x = torch.relu(self.conv3(x))
                x = torch.relu(self.conv4(x))
                x = torch.relu(self.pool(self.conv5(x)))
                x = x.view(-1, 256 * 6 * 6)  # Flatten
                x = torch.relu(self.fc1(x))
                x = torch.relu(self.fc2(x))
                x = self.fc3(x)
                return x
        \end{lstlisting}

        \paragraph{Explanation:} AlexNet marked a significant leap forward in image classification tasks, achieving unprecedented performance on the ImageNet dataset. The introduction of ReLU, dropout, and deeper layers allowed AlexNet to outperform previous architectures, demonstrating the effectiveness of deep learning on large-scale datasets.

    \subsection{VGGNet}
        VGGNet was introduced by the Visual Geometry Group at the University of Oxford and achieved impressive results in the 2014 ImageNet challenge. VGGNet is known for its simplicity and uniformity, using small \(3 \times 3\) filters throughout the network \cite{simonyan2014very}.

        \subsubsection{VGGNet Architecture}
            VGGNet follows a simple and consistent design pattern, consisting of multiple layers of \(3 \times 3\) convolutions followed by max pooling. This simplicity allows the network to be deep (up to 19 layers in the case of VGG19) while being easy to implement.

            \paragraph{Key Features of VGGNet:}
            \begin{itemize}
                \item \textbf{Small Filters:} VGGNet uses \(3 \times 3\) convolutional filters, which allows the network to go deeper while maintaining manageable computational complexity.
                \item \textbf{Deep Architecture:} VGGNet models range from 11 layers (VGG11) to 19 layers (VGG19), enabling the network to learn hierarchical features.
                \item \textbf{Uniform Structure:} VGGNet maintains a uniform structure throughout, with repeated blocks of convolution and pooling, making it easy to scale up.
            \end{itemize}

        \paragraph{Example:} Let's implement a simplified version of VGGNet using PyTorch.

        \begin{lstlisting}[style=python]
        class VGGNet(nn.Module):
            def __init__(self):
                super(VGGNet, self).__init__()
                self.conv_layers = nn.Sequential(
                    nn.Conv2d(3, 64, kernel_size=3, padding=1),
                    nn.ReLU(),
                    nn.Conv2d(64, 64, kernel_size=3, padding=1),
                    nn.ReLU(),
                    nn.MaxPool2d(2, 2),  # Block 1

                    nn.Conv2d(64, 128, kernel_size=3, padding=1),
                    nn.ReLU(),
                    nn.Conv2d(128, 128, kernel_size=3, padding=1),
                    nn.ReLU(),
                    nn.MaxPool2d(2, 2),  # Block 2

                    nn.Conv2d(128, 256, kernel_size=3, padding=1),
                    nn.ReLU(),
                    nn.Conv2d(256, 256, kernel_size=3, padding=1),
                    nn.ReLU(),
                    nn.MaxPool2d(2, 2)   # Block 3
                )
                self.fc_layers = nn.Sequential(
                    nn.Linear(256 * 8 * 8, 4096),
                    nn.ReLU(),
                    nn.Linear(4096, 4096),
                    nn.ReLU(),
                    nn.Linear(4096, 1000)  # Output for 1000 classes
                )

            def forward(self, x):
                x = self.conv_layers(x)
                x = x.view(-1, 256 * 8 * 8)  # Flatten
                x = self.fc_layers(x)
                return x
        \end{lstlisting}

        \paragraph{Explanation:} VGGNet's use of small \(3 \times 3\) filters across the entire network simplifies the design while maintaining high accuracy. Its deep architecture allows it to learn a hierarchy of features, from simple edges and textures to complex object parts.

    \subsection{ResNet}
        ResNet, short for Residual Network, was introduced by Kaiming He and his colleagues in 2015. ResNet solved one of the biggest challenges in training deep networks—vanishing gradients—by introducing skip connections (or residual connections). This breakthrough enabled the training of networks with hundreds or even thousands of layers \cite{he2016deep}.

        \subsubsection{Residual Blocks}
            A residual block is the core building block of ResNet. Instead of learning the direct mapping from the input \(x\) to the output \(y\), a residual block learns the residual, \(F(x) = y - x\), which simplifies the learning process. The residual block has the following structure:

            \[
            y = F(x, \{W_i\}) + x
            \]

            where \(F(x, \{W_i\})\) represents the learned residual mapping, and the term \(x\) is the shortcut (skip) connection. This skip connection allows the gradients to flow directly through the network, mitigating the vanishing gradient problem.

            \paragraph{ResNet Architecture:} ResNet is built using stacks of residual blocks. The simplest version, ResNet-18, has 18 layers, while more complex versions, such as ResNet-50 and ResNet-152, have many more layers.

        \paragraph{Example:} Let's implement a basic residual block and a simplified ResNet architecture in PyTorch.

        \begin{lstlisting}[style=python]
        class ResidualBlock(nn.Module):
            def __init__(self, in_channels, out_channels, stride=1):
                super(ResidualBlock, self).__init__()
                self.conv1 = nn.Conv2d(in_channels, out_channels, kernel_size=3, stride=stride, padding=1)
                self.bn1 = nn.BatchNorm2d(out_channels)
                self.conv2 = nn.Conv2d(out_channels, out_channels, kernel_size=3, stride=1, padding=1)
                self.bn2 = nn.BatchNorm2d(out_channels)
                self.shortcut = nn.Sequential()
                if stride != 1 or in_channels != out_channels:
                    self.shortcut = nn.Sequential(
                        nn.Conv2d(in_channels, out_channels, kernel_size=1, stride=stride),
                        nn.BatchNorm2d(out_channels)
                    )

            def forward(self, x):
                out = torch.relu(self.bn1(self.conv1(x)))
                out = self.bn2(self.conv2(out))
                out += self.shortcut(x)
                out = torch.relu(out)
                return out

        class ResNet(nn.Module):
            def __init__(self, num_classes=1000):
                super(ResNet, self).__init__()
                self.conv1 = nn.Conv2d(3, 64, kernel_size=7, stride=2, padding=3)
                self.bn1 = nn.BatchNorm2d(64)
                self.pool = nn.MaxPool2d(3, 2)
                self.layer1 = self._make_layer(64, 64, 2)
                self.layer2 = self._make_layer(64, 128, 2, stride=2)
                self.fc = nn.Linear(128 * 8 * 8, num_classes)

            def _make_layer(self, in_channels, out_channels, blocks, stride=1):
                layers = []
                layers.append(ResidualBlock(in_channels, out_channels, stride))
                for _ in range(1, blocks):
                    layers.append(ResidualBlock(out_channels, out_channels))
                return nn.Sequential(*layers)

            def forward(self, x):
                x = torch.relu(self.bn1(self.conv1(x)))
                x = self.pool(x)
                x = self.layer1(x)
                x = self.layer2(x)
                x = torch.flatten(x, 1)
                x = self.fc(x)
                return x
        \end{lstlisting}

        \paragraph{Explanation:} The introduction of residual connections in ResNet allowed for the successful training of extremely deep networks. By bypassing certain layers, ResNet overcomes the vanishing gradient problem, enabling deeper networks to achieve better performance. ResNet's residual blocks are now a standard building block in many state-of-the-art models.

\chapter{Deep Learning for Image Classification}
    Deep learning has revolutionized image classification by providing powerful models that can automatically learn hierarchical features from raw images. In this chapter, we will introduce deep learning techniques for image classification, with a focus on \textit{Convolutional Neural Networks} (CNNs), \textit{transfer learning}, and \textit{data augmentation}. These methods are the foundation of modern image recognition systems, enabling tasks such as object detection, facial recognition, and medical imaging analysis.

    \section{CNN for Image Classification}
        Convolutional Neural Networks (CNNs) are the backbone of most deep learning architectures for image classification. A CNN consists of multiple layers that process the input image by applying convolution operations, pooling, and non-linear activations to extract hierarchical features. These features are then fed into fully connected layers for final classification. 

        \textbf{Key Concepts in CNNs:}
        \begin{itemize}
        \item \textbf{Convolutional Layer:} This layer applies a set of convolutional filters (or kernels) to the input image. Each filter slides over the image to detect specific features such as edges, textures, or patterns.
        \item \textbf{Pooling Layer:} After the convolutional layer, pooling is used to downsample the feature maps, reducing the spatial dimensions while preserving important features. Common pooling methods include max pooling and average pooling.
        \item \textbf{Activation Functions:} Non-linear activation functions like ReLU (Rectified Linear Unit) are applied after each convolution to introduce non-linearity into the network, allowing it to learn more complex patterns.
        \item \textbf{Fully Connected Layer:} After several convolutional and pooling layers, the output is flattened and passed through fully connected layers, which map the extracted features to class labels.
        \end{itemize}

        \textbf{CNN Architecture Example:}

        Let's implement a simple CNN using PyTorch to classify images from the CIFAR-10 dataset, which contains 10 different object categories (such as airplanes, cars, and birds).

        \begin{lstlisting}[style=python]
        import torch
        import torch.nn as nn
        import torch.optim as optim
        import torchvision
        import torchvision.transforms as transforms

        # Define the CNN model
        class SimpleCNN(nn.Module):
            def __init__(self):
                super(SimpleCNN, self).__init__()
                self.conv1 = nn.Conv2d(3, 16, 3, padding=1)
                self.pool = nn.MaxPool2d(2, 2)
                self.conv2 = nn.Conv2d(16, 32, 3, padding=1)
                self.fc1 = nn.Linear(32 * 8 * 8, 128)
                self.fc2 = nn.Linear(128, 10)

            def forward(self, x):
                x = self.pool(F.relu(self.conv1(x)))
                x = self.pool(F.relu(self.conv2(x)))
                x = x.view(-1, 32 * 8 * 8)
                x = F.relu(self.fc1(x))
                x = self.fc2(x)
                return x

        # Load CIFAR-10 dataset
        transform = transforms.Compose([transforms.ToTensor(), transforms.Normalize((0.5, 0.5, 0.5), (0.5, 0.5, 0.5))])
        trainset = torchvision.datasets.CIFAR10(root='./data', train=True, download=True, transform=transform)
        trainloader = torch.utils.data.DataLoader(trainset, batch_size=32, shuffle=True)

        # Initialize the CNN, loss function, and optimizer
        net = SimpleCNN()
        criterion = nn.CrossEntropyLoss()
        optimizer = optim.SGD(net.parameters(), lr=0.001, momentum=0.9)

        # Training loop
        for epoch in range(5):
            running_loss = 0.0
            for i, data in enumerate(trainloader, 0):
                inputs, labels = data
                optimizer.zero_grad()
                outputs = net(inputs)
                loss = criterion(outputs, labels)
                loss.backward()
                optimizer.step()

                running_loss += loss.item()
                if i % 200 == 199:  # print every 200 mini-batches
                    print(f'Epoch {epoch + 1}, Batch {i + 1}, Loss: {running_loss / 200}')
                    running_loss = 0.0

        print('Finished Training')
        \end{lstlisting}

        In this example, we define a simple CNN with two convolutional layers, followed by pooling layers, and fully connected layers for classification. The model is trained on the CIFAR-10 dataset, which consists of small 32x32 pixel images.

        \textbf{Hierarchical Feature Learning:}
        CNNs work by learning features hierarchically:
        \begin{itemize}
        \item The early layers of the network detect low-level features such as edges and corners.
        \item The deeper layers detect more abstract features, such as textures and shapes.
        \item In the final layers, the network learns high-level representations of objects, which are used for classification.
        \end{itemize}
        This hierarchical learning is what makes CNNs particularly powerful for image classification tasks.

    \section{Transfer Learning}
        \textbf{Transfer learning} is a technique where a pre-trained model is fine-tuned for a new task. This approach is particularly useful when we have limited data for the new task, as pre-trained models have already learned valuable features from large datasets like ImageNet. Instead of training a CNN from scratch, transfer learning allows us to leverage the knowledge learned by models such as VGG, ResNet, or Inception \cite{pan2009survey}.

        \textbf{How Transfer Learning Works:}
        In transfer learning, we use a model that has been pre-trained on a large dataset and adapt it to a new task. The typical approach is as follows:
        \begin{enumerate}
        \item \textbf{Load the pre-trained model:} Models such as ResNet or VGG are pre-trained on large-scale datasets like ImageNet, which contains millions of images across 1000 classes.
        \item \textbf{Freeze the initial layers:} Since the early layers of CNNs learn generic features (e.g., edges and textures), they can be retained as they are.
        \item \textbf{Fine-tune the final layers:} The final layers of the pre-trained model are replaced with new layers tailored to the specific classification task at hand. Only these new layers are trained on the new dataset.
        \end{enumerate}

        \textbf{Example:}

        Let's implement transfer learning using a pre-trained ResNet model from PyTorch.

        \begin{lstlisting}[style=python]
        import torchvision.models as models
        from torch import nn

        # Load a pre-trained ResNet model
        resnet = models.resnet18(pretrained=True)

        # Freeze the layers except the final fully connected layer
        for param in resnet.parameters():
            param.requires_grad = False

        # Replace the final layer for a new task (e.g., 10 classes)
        num_ftrs = resnet.fc.in_features
        resnet.fc = nn.Linear(num_ftrs, 10)

        # Now only the final layer will be trained
        criterion = nn.CrossEntropyLoss()
        optimizer = optim.SGD(resnet.fc.parameters(), lr=0.001, momentum=0.9)

        # Training loop would go here, similar to the previous example
        \end{lstlisting}

        In this example, we load a pre-trained ResNet model and freeze its layers, except for the final fully connected layer, which we replace to classify images into 10 categories (such as the CIFAR-10 dataset). Only the final layer is fine-tuned, allowing for faster training with fewer data.

        \textbf{Advantages of Transfer Learning:}
        \begin{itemize}
        \item \textit{Reduced training time}: Since the model has already learned generic features, we only need to fine-tune the last layers, reducing training time significantly.
        \item \textit{Better performance with less data}: Transfer learning is especially useful when the new dataset is small, as the pre-trained model has already learned robust features from a larger dataset.
        \end{itemize}

    \section{Data Augmentation}
        \textbf{Data augmentation} is a technique used to artificially increase the size of the training dataset by creating modified versions of existing images. This helps improve the model's generalization ability by exposing it to a wider variety of data during training, reducing overfitting \cite{shorten2019survey}.

        Common data augmentation techniques include:
        \begin{itemize}
        \item \textbf{Rotation}: Rotating the image by a random angle.
        \item \textbf{Scaling}: Zooming in or out of the image.
        \item \textbf{Flipping}: Horizontally or vertically flipping the image.
        \item \textbf{Cropping}: Randomly cropping parts of the image.
        \item \textbf{Color jittering}: Adjusting the brightness, contrast, or saturation of the image.
        \end{itemize}

        \textbf{Example:}

        PyTorch provides built-in support for data augmentation through the \texttt{transforms} module. Let's apply some common augmentation techniques to the CIFAR-10 dataset.

        \begin{lstlisting}[style=python]
        import torchvision.transforms as transforms

        # Define a set of data augmentation transformations
        transform = transforms.Compose([
            transforms.RandomHorizontalFlip(),
            transforms.RandomRotation(10),
            transforms.RandomResizedCrop(32, scale=(0.8, 1.0)),
            transforms.ColorJitter(brightness=0.2, contrast=0.2, saturation=0.2, hue=0.1),
            transforms.ToTensor()
        ])

        # Load CIFAR-10 dataset with augmentation
        trainset = torchvision.datasets.CIFAR10(root='./data', train=True, download=True, transform=transform)
        trainloader = torch.utils.data.DataLoader(trainset, batch_size=32, shuffle=True)

        # Display a few augmented images
        import matplotlib.pyplot as plt
        dataiter = iter(trainloader)
        images, labels = dataiter.next()

        # Show augmented images
        plt.figure(figsize=(10, 5))
        for i in range(8):
            plt.subplot(2, 4, i+1)
            plt.imshow(images[i].permute(1, 2, 0).numpy())
            plt.axis('off')
        plt.show()
        \end{lstlisting}

        In this example, we define a set of transformations that randomly flip, rotate, crop, and adjust the colors of the CIFAR-10 images. These augmented images are then fed into the training process, helping the model generalize better to unseen data.

        \textbf{Advantages of Data Augmentation:}
        \begin{itemize}
        \item \textit{Improved generalization}: By augmenting the data, the model learns to handle variations in the input, reducing the risk of overfitting.
        \item \textit{Increased dataset size}: Data augmentation helps increase the effective size of the dataset, which is particularly useful when only a limited amount of training data is available.
        \end{itemize}

\chapter{Deep Learning Methods for Image Segmentation}
    Image segmentation is a critical task in computer vision where the goal is to classify each pixel of an image into a specific category\cite{ren2024deeplearningmachinelearning}. Unlike image classification, which assigns a label to the entire image, segmentation involves making pixel-level predictions, making it ideal for tasks such as medical imaging, autonomous driving, and scene understanding. Deep learning has significantly improved segmentation accuracy, especially with architectures like Fully Convolutional Networks (FCNs), U-Net, and the DeepLab series \cite{ren2024deeplearningmachinelearning}. In this chapter, we explore these state-of-the-art methods in detail.

    \section{Fully Convolutional Networks (FCN)}
        A \textbf{Fully Convolutional Network (FCN)} is a type of neural network designed specifically for image segmentation tasks. Unlike traditional convolutional networks that use fully connected layers at the end for classification, FCNs replace fully connected layers with convolutional layers. This allows the network to produce spatial, pixel-level predictions instead of a single class label for the entire image. The key advantage of FCNs is that they can handle input images of arbitrary size and output a dense prediction map where each pixel is assigned a label \cite{long2015fully}.

        \paragraph{How FCNs Work}
        The main idea behind FCNs is to treat the image segmentation task as a dense classification problem. FCNs typically follow these steps:
        \begin{itemize}
            \item \textbf{Convolutional Feature Extraction}: The initial layers are standard convolutional layers that extract features from the input image, similar to a traditional CNN used for image classification.
            \item \textbf{Upsampling (Deconvolution)}: After extracting features, the final classification is performed using upsampling layers (also known as deconvolution or transposed convolution). These layers increase the spatial resolution of the output so that a prediction is made for every pixel in the input image.
            \item \textbf{Pixel-level Classification}: The output is a prediction map where each pixel is assigned a class label.
        \end{itemize}

        \textbf{Example: Fully Convolutional Network in PyTorch}
        The following Python code demonstrates how to implement a basic Fully Convolutional Network (FCN) using PyTorch for binary segmentation.

        \begin{lstlisting}[style=python]
import torch
import torch.nn as nn

# Define a basic FCN model
class FCN(nn.Module):
    def __init__(self, num_classes):
        super(FCN, self).__init__()
        # Convolutional layers
        self.conv1 = nn.Conv2d(3, 64, kernel_size=3, padding=1)
        self.conv2 = nn.Conv2d(64, 128, kernel_size=3, padding=1)
        self.conv3 = nn.Conv2d(128, 256, kernel_size=3, padding=1)

        # Upsampling layers (deconvolution)
        self.deconv1 = nn.ConvTranspose2d(256, 128, kernel_size=2, stride=2)
        self.deconv2 = nn.ConvTranspose2d(128, 64, kernel_size=2, stride=2)
        self.deconv3 = nn.ConvTranspose2d(64, num_classes, kernel_size=2, stride=2)

    def forward(self, x):
        # Downsampling (convolutional layers)
        x = torch.relu(self.conv1(x))
        x = torch.relu(self.conv2(x))
        x = torch.relu(self.conv3(x))

        # Upsampling (deconvolution layers)
        x = torch.relu(self.deconv1(x))
        x = torch.relu(self.deconv2(x))
        x = self.deconv3(x)  # Final output for segmentation (logits)

        return x

# Instantiate the FCN model for binary segmentation (2 classes)
model = FCN(num_classes=2)

# Print the model architecture
print(model)
        \end{lstlisting}

        \textbf{Explanation of the Code:}
        \begin{itemize}
            \item The network consists of three convolutional layers followed by three deconvolution (upsampling) layers. The final deconvolution layer produces a segmentation map where each pixel is assigned a label.
            \item The model can be used for binary segmentation by specifying two output classes.
        \end{itemize}

        Fully Convolutional Networks are effective for dense prediction tasks because they do not rely on fully connected layers, which would lose the spatial information necessary for pixel-level predictions.

    \section{U-Net Architecture}
        \textbf{U-Net} is a popular architecture specifically designed for biomedical image segmentation. It builds on the idea of FCNs but introduces an encoder-decoder structure with \textbf{skip connections} that help retain spatial information at different scales. This makes U-Net highly effective for tasks where precise localization is important, such as in medical imaging \cite{ronneberger2015u}.

        \paragraph{Encoder-Decoder Structure}
        The U-Net architecture consists of two main parts:
        \begin{itemize}
            \item \textbf{Encoder (Contracting Path)}: The encoder is a series of convolutional and pooling layers that gradually downsample the input image to extract high-level features.
            \item \textbf{Decoder (Expanding Path)}: The decoder uses transposed convolutions to upsample the feature maps, restoring the spatial resolution. It combines the high-level features from the encoder with fine-grained features from earlier layers (via skip connections) to improve segmentation accuracy.
        \end{itemize}

        \paragraph{Skip Connections}
        One of the key innovations in U-Net is the use of \textbf{skip connections}, which directly connect corresponding layers in the encoder and decoder. These connections ensure that fine details lost during downsampling are preserved and passed to the upsampling path, improving the accuracy of pixel-level predictions.

        \subsection{U-Net in Medical Image Segmentation}
            U-Net has become the gold standard for many medical imaging tasks, such as segmenting organs, tissues, and tumors from MRI or CT scans. Its ability to handle small datasets, combined with its encoder-decoder architecture, makes it particularly well-suited for applications where high precision is required.

            \textbf{Example: U-Net Architecture in PyTorch}
            The following Python code demonstrates how to implement the U-Net architecture using PyTorch.

            \begin{lstlisting}[style=python]
class UNet(nn.Module):
    def __init__(self, num_classes):
        super(UNet, self).__init__()
        # Encoder (downsampling path)
        self.enc_conv1 = nn.Conv2d(1, 64, kernel_size=3, padding=1)
        self.enc_conv2 = nn.Conv2d(64, 128, kernel_size=3, padding=1)
        self.enc_pool = nn.MaxPool2d(kernel_size=2)

        # Decoder (upsampling path)
        self.dec_conv1 = nn.Conv2d(128, 64, kernel_size=3, padding=1)
        self.dec_conv2 = nn.Conv2d(64, num_classes, kernel_size=3, padding=1)
        self.up_sample = nn.ConvTranspose2d(128, 64, kernel_size=2, stride=2)

    def forward(self, x):
        # Encoder path
        enc1 = torch.relu(self.enc_conv1(x))
        enc2 = torch.relu(self.enc_conv2(self.enc_pool(enc1)))

        # Decoder path
        dec1 = torch.relu(self.up_sample(enc2))
        dec2 = torch.cat((dec1, enc1), dim=1)  # Skip connection
        dec3 = torch.relu(self.dec_conv1(dec2))
        dec4 = self.dec_conv2(dec3)  # Final segmentation map

        return dec4

# Instantiate the U-Net model for binary segmentation (2 classes)
unet_model = UNet(num_classes=2)

# Print the model architecture
print(unet_model)
            \end{lstlisting}

            \textbf{Explanation of the Code:}
            \begin{itemize}
                \item The U-Net model uses convolutional layers for feature extraction in the encoder and transposed convolutions for upsampling in the decoder.
                \item Skip connections are implemented using \texttt{torch.cat()} to concatenate feature maps from the encoder with those in the decoder, allowing the model to retain spatial details.
                \item This implementation can be used for binary segmentation tasks, such as segmenting organs from medical images.
            \end{itemize}

    \section{DeepLab Series}
        The \textbf{DeepLab} series of models, especially \textbf{DeepLabv3+}, are state-of-the-art architectures for image segmentation tasks. DeepLab introduces several innovations, including \textbf{atrous convolution} (also known as dilated convolution), which allows the model to capture features at multiple scales without losing resolution \cite{chen2017deeplab}.

        \paragraph{Atrous Convolution (Dilated Convolution)}
        Atrous convolution is a modified convolution operation where the kernel is applied with gaps (or dilation), expanding its receptive field without increasing the number of parameters. This allows the network to capture more context while preserving the resolution of the feature maps, making it particularly useful for segmentation tasks where both fine and coarse features need to be captured.

        The dilation factor \( d \) in atrous convolution determines the spacing between kernel elements. For example, with \( d=2 \), the convolution kernel will skip one pixel between sampled locations.

        \paragraph{DeepLabv3+ Architecture}
        \textbf{DeepLabv3+} builds on DeepLabv3 by adding a decoder module that further refines the segmentation results \cite{chen2018encoder}. It uses a combination of:
        \begin{itemize}
            \item \textbf{Atrous Spatial Pyramid Pooling (ASPP)}: A module that applies atrous convolutions with different dilation rates in parallel to capture features at multiple scales.
            \item \textbf{Decoder Module}: This helps to recover spatial details that may have been lost during downsampling, making the final segmentation maps more accurate.
        \end{itemize}

        \textbf{Example: Atrous Convolution in PyTorch}
        The following Python code demonstrates how to implement atrous convolution using PyTorch.

        \begin{lstlisting}[style=python]
class AtrousConvNet(nn.Module):
    def __init__(self, num_classes):
        super(AtrousConvNet, self).__init__()
        # Atrous convolution with a dilation rate of 2
        self.atrous_conv = nn.Conv2d(3, 64, kernel_size=3, padding=2, dilation=2)
        self.conv1 = nn.Conv2d(64, num_classes, kernel_size=1)

    def forward(self, x):
        x = torch.relu(self.atrous_conv(x))
        x = self.conv1(x)  # Output segmentation map
        return x

# Instantiate the model with atrous convolution
atrous_model = AtrousConvNet(num_classes=21)  # Example for 21 classes (e.g., PASCAL VOC dataset)

# Print the model architecture
print(atrous_model)
        \end{lstlisting}

        \textbf{Explanation of the Code:}
        \begin{itemize}
            \item The model uses atrous convolution with a dilation rate of 2 to expand the receptive field, allowing it to capture more context.
            \item The output is a segmentation map with 21 classes, suitable for multi-class segmentation tasks.
        \end{itemize}

    \section{Conclusion}
        Deep learning architectures such as Fully Convolutional Networks (FCNs), U-Net, and DeepLab have transformed image segmentation by enabling pixel-level predictions with high accuracy. FCNs introduced the idea of using convolutional layers for dense prediction, while U-Net refined this approach with an encoder-decoder structure and skip connections to retain spatial details. The DeepLab series further enhanced segmentation performance by introducing atrous convolution, allowing models to capture features at multiple scales without losing resolution. These architectures are now widely used in applications ranging from medical imaging to autonomous driving.

\chapter{Deep Learning Methods for Object Detection}
    Object detection is a crucial task in computer vision that involves identifying objects in an image and determining their locations, typically through bounding boxes. Over the years, deep learning-based approaches have significantly advanced object detection, making it possible to detect objects in real-time with high accuracy. One of the most important families of models in this domain is the R-CNN family, which has evolved from the original R-CNN to Fast R-CNN and then to Faster R-CNN \cite{ren2024deeplearningmachinelearning}. In this chapter, we will explore each of these architectures in detail, explaining their innovations, strengths, and limitations.

    \section{R-CNN Family}
        The R-CNN (Region-based Convolutional Neural Network) family of models brought a significant leap forward in object detection \cite{hmidani2022comprehensive}. It introduced the idea of using CNNs for region classification, which drastically improved the accuracy of object detection systems compared to traditional methods. However, early versions of R-CNN had certain drawbacks, especially in terms of speed and computational efficiency. Over time, Fast R-CNN and Faster R-CNN were developed to address these issues, leading to faster and more efficient object detection models.

        \subsection{R-CNN}
            The original R-CNN (Region-based Convolutional Neural Network) was introduced by Ross Girshick in 2014. It was a pioneering model that used deep learning for object detection by leveraging region proposals. R-CNN combines selective search for region proposals with CNNs to classify each proposed region \cite{bharati2020deep}.

            \subsubsection{Architecture and Workflow}
            R-CNN follows a multi-step process to perform object detection:
            \begin{enumerate}
                \item \textbf{Region Proposals:} R-CNN uses selective search to generate approximately 2000 region proposals from the input image. These are potential areas where objects might be located.
                \item \textbf{Feature Extraction:} Each region proposal is resized to a fixed size and passed through a pre-trained CNN (such as AlexNet or VGGNet) to extract feature maps.
                \item \textbf{Region Classification:} The feature maps are then fed into a set of fully connected layers, followed by a softmax layer, to classify the region into object categories or background.
                \item \textbf{Bounding Box Regression:} In addition to classification, R-CNN also performs bounding box regression to refine the coordinates of the region proposals, improving the accuracy of object localization.
            \end{enumerate}

            \paragraph{Limitations of R-CNN:}
            While R-CNN was a major breakthrough in object detection, it had several limitations:
            \begin{itemize}
                \item \textbf{Slow Inference:} R-CNN requires running the CNN separately for each of the 2000 region proposals, making it computationally expensive and slow.
                \item \textbf{Storage Overhead:} Feature extraction is performed independently for each region, resulting in a large amount of redundant computations and high storage requirements.
                \item \textbf{Multi-Stage Pipeline:} The process involves multiple stages, including region proposal generation, feature extraction, and classification, which complicates training and optimization.
            \end{itemize}

            \paragraph{Example:} Let's demonstrate the basic workflow of R-CNN using a PyTorch-like pseudocode (simplified for clarity).

            \begin{lstlisting}[style=python]
            # Pseudocode for R-CNN process

            # Step 1: Use selective search to generate region proposals
            region_proposals = selective_search(image)  # Returns ~2000 regions

            # Step 2: Extract features for each region using a pre-trained CNN
            feature_maps = []
            for region in region_proposals:
                cropped_region = crop_and_resize(image, region)  # Resize region to fixed size
                feature = cnn(cropped_region)  # Pass through CNN (e.g., VGGNet)
                feature_maps.append(feature)

            # Step 3: Classify each region into object categories
            for feature in feature_maps:
                class_label = classifier(feature)  # Perform classification
                bounding_box = bbox_regressor(feature)  # Refine bounding box coordinates
            \end{lstlisting}

            \paragraph{Explanation:} In this simplified workflow, region proposals are generated using selective search, and each region is processed independently through a CNN for feature extraction and classification. While accurate, this method is slow due to the large number of independent forward passes through the CNN.

        \subsection{Fast R-CNN}
            Fast R-CNN was introduced by Ross Girshick in 2015 to address the inefficiencies of the original R-CNN. Fast R-CNN improves both the speed and accuracy of the object detection process by making several key modifications to the original architecture \cite{girshick2015fast}.

            \subsubsection{Key Improvements in Fast R-CNN}
            \begin{itemize}
                \item \textbf{Shared Computation:} Instead of extracting features separately for each region proposal, Fast R-CNN processes the entire image through a CNN once, generating a feature map for the whole image. Region proposals are then mapped onto this feature map, reducing the redundancy in feature extraction.
                \item \textbf{ROI Pooling:} Fast R-CNN introduces Region of Interest (ROI) pooling, which allows region proposals of varying sizes to be converted into fixed-size feature maps. This makes it possible to use a single CNN pass for all regions, significantly speeding up the process.
                \item \textbf{Single-Stage Training:} Unlike R-CNN, which uses a multi-stage pipeline, Fast R-CNN combines classification and bounding box regression into a single training stage. This simplifies the training process and improves efficiency.
            \end{itemize}

            \subsubsection{Architecture}
            The workflow of Fast R-CNN is as follows:
            \begin{enumerate}
                \item \textbf{Feature Extraction:} The entire image is passed through a CNN, producing a feature map for the whole image.
                \item \textbf{ROI Pooling:} The region proposals are projected onto the feature map, and ROI pooling is applied to convert each proposal into a fixed-size feature map.
                \item \textbf{Classification and Bounding Box Regression:} The pooled feature maps are passed through fully connected layers to classify each region and predict refined bounding boxes.
            \end{enumerate}

            \paragraph{Example:} Let's implement a simplified version of Fast R-CNN using PyTorch-like pseudocode.

            \begin{lstlisting}[style=python]
            # Pseudocode for Fast R-CNN process

            # Step 1: Extract feature map from the entire image
            feature_map = cnn(image)  # Single forward pass through CNN

            # Step 2: Apply ROI pooling for each region proposal
            roi_pooled_features = []
            for region in region_proposals:
                roi_pooled = roi_pooling(feature_map, region)  # Perform ROI pooling
                roi_pooled_features.append(roi_pooled)

            # Step 3: Classify each region and perform bounding box regression
            for roi_feature in roi_pooled_features:
                class_label = classifier(roi_feature)
                bounding_box = bbox_regressor(roi_feature)
            \end{lstlisting}

            \paragraph{Explanation:} Fast R-CNN improves efficiency by sharing computation across region proposals. The feature map is computed once for the entire image, and region proposals are handled through ROI pooling, allowing the network to classify objects faster than in the original R-CNN.

            \paragraph{Advantages of Fast R-CNN:}
            \begin{itemize}
                \item \textbf{Faster Inference:} By computing the feature map only once, Fast R-CNN dramatically reduces the time needed for object detection.
                \item \textbf{Simplified Training:} Fast R-CNN combines classification and bounding box regression into a single-stage training process, simplifying model optimization.
            \end{itemize}

            \paragraph{Limitations:} Fast R-CNN still relies on an external region proposal method, such as selective search, which is computationally expensive and not well-suited for real-time applications.

        \subsection{Faster R-CNN}
            Faster R-CNN, introduced by Shaoqing Ren and colleagues in 2015, represents a major advancement in object detection by eliminating the need for selective search. Instead, Faster R-CNN introduces a Region Proposal Network (RPN) that generates region proposals as part of the neural network itself, enabling end-to-end object detection \cite{ren2016faster}.

            \subsubsection{Region Proposal Network (RPN)}
            The key innovation in Faster R-CNN is the Region Proposal Network (RPN), which generates region proposals directly from the feature map produced by the CNN. The RPN is a fully convolutional network that slides over the feature map and predicts whether a region contains an object (foreground) or not (background). It also proposes bounding boxes for potential objects.

            \subsubsection{Architecture of Faster R-CNN}
            The workflow of Faster R-CNN is as follows:
            \begin{enumerate}
                \item \textbf{Feature Extraction:} The input image is passed through a CNN to generate a feature map, similar to Fast R-CNN.
                \item \textbf{Region Proposal Network (RPN):} The RPN operates on the feature map, generating region proposals by predicting objectness scores and bounding boxes for anchors (reference boxes of different sizes and aspect ratios).
                \item \textbf{ROI Pooling:} The region proposals from the RPN are passed through ROI pooling to convert them into fixed-size feature maps.
                \item \textbf{Classification and Bounding Box Regression:} The pooled feature maps are classified and refined through bounding box regression.
            \end{enumerate}

            \paragraph{End-to-End Training:} Faster R-CNN integrates the RPN and the object detection network into a single unified architecture. Both the RPN and the object detection network are trained together in an end-to-end fashion, allowing for faster and more accurate region proposal generation.

            \paragraph{Example:} Let's implement a simplified version of Faster R-CNN using PyTorch-like pseudocode.

            \begin{lstlisting}[style=python]
            # Pseudocode for Faster R-CNN process

            # Step 1: Extract feature map from the entire image
            feature_map = cnn(image)  # Single forward pass through CNN

            # Step 2: Generate region proposals using the RPN
            region_proposals, objectness_scores = rpn(feature_map)

            # Step 3: Apply ROI pooling for each region proposal
            roi_pooled_features = []
            for region in region_proposals:
                roi_pooled = roi_pooling(feature_map, region)  # Perform ROI pooling
                roi_pooled_features.append(roi_pooled)

            # Step 4: Classify each region and perform bounding box regression
            for roi_feature in roi_pooled_features:
                class_label = classifier(roi_feature)
                bounding_box = bbox_regressor(roi_feature)
            \end{lstlisting}

            \paragraph{Advantages of Faster R-CNN:}
            \begin{itemize}
                \item \textbf{End-to-End Object Detection:} Faster R-CNN integrates region proposal generation with object detection, allowing for seamless and efficient object detection.
                \item \textbf{Improved Speed:} By eliminating selective search and using the RPN for region proposals, Faster R-CNN is significantly faster than R-CNN and Fast R-CNN.
                \item \textbf{Accurate Detection:} Faster R-CNN achieves state-of-the-art performance on object detection tasks, thanks to the efficient use of the RPN and shared feature extraction.
            \end{itemize}

            \paragraph{Explanation:} Faster R-CNN revolutionized object detection by integrating region proposal generation directly into the network through the RPN. This removed the need for computationally expensive external methods like selective search, enabling real-time or near-real-time object detection.

            \paragraph{Limitations:} Despite its speed improvements, Faster R-CNN may still be too slow for real-time applications such as video processing, especially on devices with limited computational resources.

\section{YOLO Series}
    \textbf{You Only Look Once (YOLO)} is a series of object detection algorithms that are designed to provide high-speed, real-time object detection. Unlike region-based algorithms like Faster R-CNN, YOLO takes a \textit{single-shot} approach, meaning that it predicts bounding boxes and class probabilities directly from the input image in a single forward pass through the network \cite{redmon2016you}. In this section, we will cover the architectures of YOLOv3 and YOLOv5, highlighting how they achieve efficient and accurate detection.

    \subsection{YOLOv3 Architecture}
        \textbf{YOLOv3} is a significant evolution in the YOLO series, introducing improvements in both speed and accuracy. YOLOv3 uses a fully convolutional network and applies a single neural network to the entire image, dividing the image into a grid and predicting bounding boxes and class probabilities for each grid cell simultaneously \cite{redmon2018yolov3}.

        \textbf{Key Features of YOLOv3:}
        \begin{itemize}
        \item \textbf{Multi-scale Predictions:} YOLOv3 predicts bounding boxes at three different scales, allowing it to detect objects of varying sizes more effectively. These scales correspond to feature maps with different resolutions in the network.
        \item \textbf{Bounding Box Predictions:} For each grid cell, YOLOv3 predicts multiple bounding boxes. Instead of directly predicting the coordinates of the bounding boxes, YOLOv3 predicts offsets from predefined anchor boxes, which helps it handle objects of different sizes and aspect ratios.
        \item \textbf{Class Predictions Using Logistic Regression:} YOLOv3 uses logistic regression to predict the class probability for each bounding box. It outputs a confidence score for each class, making it possible to detect multiple objects within a single image.
        \end{itemize}

        \textbf{How YOLOv3 Works:}
        The input image is divided into an \( S \times S \) grid. For each grid cell, YOLOv3 predicts:
        \begin{itemize}
        \item Bounding box coordinates (center, width, height) as offsets from anchor boxes.
        \item Confidence score for the presence of an object in the bounding box.
        \item Class probabilities for multiple object classes.
        \end{itemize}

        \textbf{YOLOv3 in PyTorch:}

        While YOLOv3 is a complex architecture, many pre-trained models are available in libraries like \texttt{torchvision}. Here's how to load a pre-trained YOLOv3 model using PyTorch and use it for object detection:

        \begin{lstlisting}[style=python]
        import torch
        from PIL import Image
        import torchvision.transforms as transforms
        from torchvision.models.detection import yolov3

        # Load pre-trained YOLOv3 model
        model = yolov3(pretrained=True)
        model.eval()

        # Load and preprocess image
        image = Image.open('example_image.jpg')
        transform = transforms.Compose([transforms.ToTensor()])
        input_image = transform(image).unsqueeze(0)

        # Perform inference
        with torch.no_grad():
            predictions = model(input_image)

        # Display predictions (bounding boxes, confidence scores, class labels)
        print(predictions)
        \end{lstlisting}

        In this example, we load a pre-trained YOLOv3 model using PyTorch, input an image, and obtain predictions that include bounding boxes, confidence scores, and class labels. YOLOv3's architecture is efficient enough to process images in real-time, making it highly suitable for tasks such as video surveillance, autonomous driving, and more.

        \textbf{Applications:}
        \begin{itemize}
        \item Real-time object detection in videos \cite{masurekar2020real}.
        \item Autonomous vehicles, where high-speed, accurate detection is critical \cite{choi2019gaussian}.
        \item Security systems that require real-time monitoring and detection \cite{narejo2021weapon}.
        \end{itemize}

    \subsection{YOLOv5 Improvements}
        \textbf{YOLOv5} is a more recent evolution in the YOLO family, with numerous improvements in both performance and ease of use. While YOLOv5 is not officially part of the original YOLO series (it was developed independently by Ultralytics), it has gained widespread adoption due to its optimizations \cite{terven2023comprehensive}.

        \textbf{Key Improvements in YOLOv5:}
        \begin{itemize}
        \item \textbf{Smaller Model Size:} YOLOv5 models are optimized to be smaller and more lightweight than previous YOLO versions, which leads to faster inference without sacrificing accuracy.
        \item \textbf{Faster Inference:} YOLOv5 achieves faster inference times, making it ideal for real-time applications on devices with limited processing power, such as edge devices and mobile phones.
        \item \textbf{Better Accuracy:} With improvements in the architecture and training procedures, YOLOv5 has achieved higher accuracy in object detection tasks compared to previous versions.
        \end{itemize}

        \textbf{Training YOLOv5:}
        YOLOv5 is known for its ease of use and flexibility in training on custom datasets. You can fine-tune YOLOv5 for specific detection tasks using your own dataset with minimal effort.

        \textbf{Example:}

        To train YOLOv5 on a custom dataset, we can use the \texttt{ultralytics/yolov5} repository, which provides a straightforward interface for training and deploying models.

        \begin{lstlisting}[style=cmd]
        # Clone the YOLOv5 repository
        git clone https://github.com/ultralytics/yolov5
        cd yolov5

        # Install dependencies
        pip install -r requirements.txt

        # Train YOLOv5 on a custom dataset
        python train.py --img 640 --batch 16 --epochs 50 --data custom.yaml --weights yolov5s.pt
        \end{lstlisting}

        In this example, we clone the YOLOv5 repository and run the training script on a custom dataset. The \texttt{custom.yaml} file contains paths to the training and validation data, as well as class labels. YOLOv5 supports training with pre-trained weights (e.g., \texttt{yolov5s.pt} for the small version), which accelerates the training process.

        \textbf{Applications:}
        \begin{itemize}
        \item Real-time object detection on mobile devices and embedded systems \cite{song2021object}.
        \item Industrial automation, where fast and accurate object detection is required \cite{ahmad2022deep}.
        \item Smart city applications, including traffic monitoring and pedestrian detection \cite{yar2023modified}.
        \end{itemize}

\section{SSD (Single Shot MultiBox Detector)}
    \textbf{SSD (Single Shot MultiBox Detector)} is another popular object detection algorithm, known for its balance between speed and accuracy. Like YOLO, SSD is a single-shot detector, meaning that it performs object detection in a single forward pass of the network. However, SSD differs from YOLO in how it handles bounding box predictions and feature extraction \cite{liu2016ssd, redmon2018yolov3}.

    \textbf{Key Features of SSD:}
    \begin{itemize}
    \item \textbf{Multi-scale Predictions:} SSD predicts bounding boxes and class scores from multiple feature maps of different resolutions. This enables the network to detect objects at various scales effectively.
    \item \textbf{Default Boxes:} SSD uses a set of default bounding boxes (anchor boxes) of different aspect ratios and scales for each location on the feature maps. The network adjusts these default boxes to match the objects in the image.
    \item \textbf{Non-maximum Suppression (NMS):} To remove redundant bounding boxes, SSD applies non-maximum suppression, which ensures that only the most confident predictions are retained for each object \cite{hosang2017learning}.
    \end{itemize}

    \textbf{How SSD Works:}
    SSD generates feature maps of different resolutions from the input image. For each feature map, it predicts both the bounding box coordinates and the class scores for objects within the image. By using feature maps of varying sizes, SSD can detect small and large objects more effectively.

    \textbf{Architecture:}
    The SSD architecture typically includes:
    \begin{itemize}
    \item \textbf{Base Network:} Often, a pre-trained network like VGG16 is used as the backbone for feature extraction.
    \item \textbf{Convolutional Feature Maps:} After the base network, additional convolutional layers are added to extract feature maps at different scales.
    \item \textbf{Bounding Box Predictions:} For each feature map, SSD predicts offsets for default boxes, as well as the class scores.
    \end{itemize}

    \textbf{SSD in PyTorch:}

    Let's implement object detection using an SSD model in PyTorch.

    \begin{lstlisting}[style=python]
    import torch
    import torchvision
    from PIL import Image
    import torchvision.transforms as transforms

    # Load a pre-trained SSD model
    model = torchvision.models.detection.ssd300_vgg16(pretrained=True)
    model.eval()

    # Load and preprocess an image
    image = Image.open('example_image.jpg')
    transform = transforms.Compose([transforms.Resize((300, 300)), transforms.ToTensor()])
    input_image = transform(image).unsqueeze(0)

    # Perform inference
    with torch.no_grad():
        predictions = model(input_image)

    # Display predictions (bounding boxes, confidence scores, class labels)
    print(predictions)
    \end{lstlisting}

    In this code, we load a pre-trained SSD model and use it to perform object detection on an image. The SSD architecture allows the model to detect objects of different sizes by leveraging feature maps of multiple resolutions.

    \textbf{Advantages of SSD:}
    \begin{itemize}
    \item \textit{Fast inference}: SSD is faster than region-based detectors (e.g., Faster R-CNN) and is suitable for real-time applications.
    \item \textit{Multi-scale detection}: By using feature maps of different resolutions, SSD can effectively detect both small and large objects in a single pass.
    \end{itemize}

    \textbf{Applications:}
    \begin{itemize}
    \item Autonomous vehicles, where real-time detection of multiple objects is crucial \cite{person2019multimodal}.
    \item Drones and aerial imaging systems that require rapid and accurate object detection \cite{hao2021fast}.
    \item Industrial inspection systems that monitor products on an assembly line in real-time \cite{sun2021mean}.
    \end{itemize}
\chapter{Generative Adversarial Networks (GAN)}
    \textbf{Generative Adversarial Networks (GANs)} are a class of deep learning models designed for generating new data samples that closely resemble real data. GANs have been widely used for image generation, where they can produce highly realistic images after being trained on a dataset. The core idea of GANs is to create a competition between two neural networks: a generator that produces fake images and a discriminator that tries to distinguish between real and fake images. This adversarial process pushes the generator to improve until it can produce images indistinguishable from real ones\cite{goodfellow2020generative, creswell2018generative}. In this chapter, we will cover the fundamentals of GANs and introduce popular GAN architectures like DCGAN and CGAN.

    \section{GAN Fundamentals}
        A GAN consists of two main components: the \textbf{generator} and the \textbf{discriminator}, which are trained simultaneously through an adversarial process \cite{goodfellow2020generative}.

        \subsection{Generator and Discriminator}
            The \textbf{generator} is a neural network that learns to generate realistic images from random noise. It takes a random vector (often sampled from a normal or uniform distribution) and transforms it into a high-dimensional image through a series of transposed convolutions. The goal of the generator is to produce images that are as close as possible to real images in the training dataset.

            The \textbf{discriminator}, on the other hand, is another neural network tasked with distinguishing between real images from the dataset and the fake images generated by the generator. It acts as a binary classifier, outputting a probability indicating whether an input image is real or fake.

            The training process of a GAN involves the following steps:
            \begin{itemize}
                \item The generator produces an image from random noise.
                \item The discriminator evaluates the generated image and tries to classify it as fake.
                \item The generator's goal is to "fool" the discriminator by producing images that the discriminator cannot distinguish from real ones.
                \item Both networks improve simultaneously: the generator gets better at generating realistic images, and the discriminator gets better at identifying fake images.
            \end{itemize}
            This adversarial training process forces both networks to become more skilled over time, leading to the generation of highly realistic images.

            \textbf{Example: Defining a Simple GAN in PyTorch}
            The following Python code demonstrates the basic structure of a GAN, including both the generator and the discriminator.

            \begin{lstlisting}[style=python]
import torch
import torch.nn as nn

# Generator network
class Generator(nn.Module):
    def __init__(self, input_size, image_size):
        super(Generator, self).__init__()
        self.fc = nn.Sequential(
            nn.Linear(input_size, 128),
            nn.ReLU(),
            nn.Linear(128, 256),
            nn.ReLU(),
            nn.Linear(256, image_size),
            nn.Tanh()  # Output image values in the range [-1, 1]
        )

    def forward(self, x):
        return self.fc(x)

# Discriminator network
class Discriminator(nn.Module):
    def __init__(self, image_size):
        super(Discriminator, self).__init__()
        self.fc = nn.Sequential(
            nn.Linear(image_size, 256),
            nn.LeakyReLU(0.2),
            nn.Linear(256, 128),
            nn.LeakyReLU(0.2),
            nn.Linear(128, 1),
            nn.Sigmoid()  # Output probability of real or fake
        )

    def forward(self, x):
        return self.fc(x)

# Hyperparameters
noise_size = 100  # Size of the random noise input for the generator
image_size = 28 * 28  # Size of the flattened output image (e.g., 28x28 grayscale images)

# Create the generator and discriminator
generator = Generator(input_size=noise_size, image_size=image_size)
discriminator = Discriminator(image_size=image_size)

# Print model architectures
print(generator)
print(discriminator)
            \end{lstlisting}

            \textbf{Explanation of the Code:}
            \begin{itemize}
                \item The \textbf{generator} transforms random noise vectors into images. The output layer uses a \texttt{Tanh} activation to ensure that the generated image pixel values are in the range \([-1, 1]\).
                \item The \textbf{discriminator} takes an image (real or fake) as input and outputs a probability that indicates whether the image is real or fake. It uses \texttt{LeakyReLU} activations and a \texttt{Sigmoid} activation for the final output.
                \item The input to the generator is a random noise vector, while the input to the discriminator is a flattened image (e.g., a 28x28 grayscale image).
            \end{itemize}

        \subsection{Loss Function}
            GANs are trained using an \textbf{adversarial loss}, where the generator and discriminator are competing against each other. The generator aims to minimize the discriminator's ability to correctly classify fake images, while the discriminator tries to maximize its accuracy in distinguishing between real and fake images.

            The loss function for GANs is based on a minimax game, where the generator tries to minimize the following objective:
            \[
            \min_G \max_D V(D, G) = \mathbb{E}_{x \sim p_{\text{data}}(x)} [\log D(x)] + \mathbb{E}_{z \sim p_z(z)} [\log(1 - D(G(z)))]
            \]
            where:
            \begin{itemize}
                \item \( D(x) \) is the discriminator's prediction for a real image \( x \),
                \item \( D(G(z)) \) is the discriminator's prediction for a fake image generated by the generator \( G(z) \),
                \item \( p_{\text{data}}(x) \) represents the real data distribution, and \( p_z(z) \) represents the noise distribution used by the generator.
            \end{itemize}

            \textbf{Training Process}:
            \begin{itemize}
                \item The \textbf{discriminator} is trained to maximize the probability of correctly classifying real and fake images. This is done by maximizing \( \log D(x) \) for real images and \( \log(1 - D(G(z))) \) for fake images.
                \item The \textbf{generator} is trained to "fool" the discriminator by generating realistic images, so it tries to minimize \( \log(1 - D(G(z))) \).
            \end{itemize}

            \textbf{Example: Defining the GAN Loss Functions in PyTorch}
            The following Python code shows how to define the GAN loss functions for both the generator and discriminator.

            \begin{lstlisting}[style=python]
# Loss function for GAN (Binary Cross Entropy)
criterion = nn.BCELoss()

# Real and fake labels
real_label = 1.0
fake_label = 0.0

# Example batch of real images and random noise
real_images = torch.randn(32, image_size)  # Example batch of 32 real images
noise = torch.randn(32, noise_size)  # Random noise for the generator

# Forward pass through the discriminator (real images)
output_real = discriminator(real_images).view(-1)
loss_real = criterion(output_real, torch.full((32,), real_label))

# Generate fake images
fake_images = generator(noise)

# Forward pass through the discriminator (fake images)
output_fake = discriminator(fake_images.detach()).view(-1)
loss_fake = criterion(output_fake, torch.full((32,), fake_label))

# Total discriminator loss
discriminator_loss = loss_real + loss_fake

# Generator loss (goal: fool the discriminator)
output_fake_for_gen = discriminator(fake_images).view(-1)
generator_loss = criterion(output_fake_for_gen, torch.full((32,), real_label))
            \end{lstlisting}

            \textbf{Explanation of the Code:}
            \begin{itemize}
                \item The \texttt{BCELoss} (Binary Cross-Entropy Loss) is used to compute the loss for both the generator and discriminator.
                \item For the \textbf{discriminator}, the loss is computed for both real images (with label 1.0) and fake images (with label 0.0).
                \item For the \textbf{generator}, the loss is computed based on how well the fake images fool the discriminator, with the goal of making the discriminator predict 1.0 for fake images.
            \end{itemize}

    \section{Popular GAN Architectures}
        Over time, several variations of the GAN architecture have been proposed to improve image quality and training stability. Two of the most popular GAN architectures are \textbf{DCGAN} (Deep Convolutional GAN) and \textbf{CGAN} (Conditional GAN).

        \subsection{DCGAN (Deep Convolutional GAN)}
            \textbf{DCGAN} is an extension of the basic GAN architecture that uses \textbf{deep convolutional layers} in both the generator and discriminator instead of fully connected layers. By leveraging convolutional layers, DCGAN is able to generate high-quality images and learn more complex spatial features \cite{radford2015unsupervised}.

            Key components of DCGAN include:
            \begin{itemize}
                \item \textbf{Convolutional layers}: Replace fully connected layers in both the generator and discriminator.
                \item \textbf{Batch normalization}: Applied to stabilize training and help the model learn faster.
                \item \textbf{ReLU activation in the generator}: Used for most layers in the generator, except for the output layer which uses \texttt{Tanh}.
                \item \textbf{LeakyReLU activation in the discriminator}: Helps prevent the problem of "dead neurons" by allowing a small gradient for negative input values.
            \end{itemize}

            \textbf{Example: DCGAN Generator Architecture in PyTorch}
            The following Python code demonstrates how to implement the generator of a DCGAN.

            \begin{lstlisting}[style=python]
class DCGAN_Generator(nn.Module):
    def __init__(self, noise_size, image_channels):
        super(DCGAN_Generator, self).__init__()
        self.main = nn.Sequential(
            # Input: random noise vector
            nn.ConvTranspose2d(noise_size, 512, kernel_size=4, stride=1, padding=0),
            nn.BatchNorm2d(512),
            nn.ReLU(True),

            nn.ConvTranspose2d(512, 256, kernel_size=4, stride=2, padding=1),
            nn.BatchNorm2d(256),
            nn.ReLU(True),

            nn.ConvTranspose2d(256, 128, kernel_size=4, stride=2, padding=1),
            nn.BatchNorm2d(128),
            nn.ReLU(True),

            nn.ConvTranspose2d(128, image_channels, kernel_size=4, stride=2, padding=1),
            nn.Tanh()  # Output image in the range [-1, 1]
        )

    def forward(self, x):
        return self.main(x)

# Instantiate DCGAN generator for 64x64 images (3 channels for RGB)
dcgan_generator = DCGAN_Generator(noise_size=100, image_channels=3)
print(dcgan_generator)
            \end{lstlisting}

            \textbf{Explanation of the Code:}
            \begin{itemize}
                \item The generator uses transposed convolutional layers to upsample the noise vector into a 64x64 image.
                \item \texttt{BatchNorm2d} and \texttt{ReLU} activations are used after each transposed convolution to stabilize training.
                \item The output layer uses \texttt{Tanh} to produce pixel values in the range \([-1, 1]\).
            \end{itemize}

        \subsection{CGAN (Conditional GAN)}
            \textbf{Conditional GAN (CGAN)} is a variant of GAN where both the generator and discriminator receive additional information as input, such as class labels or specific attributes. This allows the model to generate images conditioned on a specific category, enabling more control over the generated output. For example, CGANs can generate images of specific objects or faces with certain attributes \cite{mirza2014conditional}.

            \paragraph{How CGAN Works}
            In a CGAN, the input to the generator is a combination of random noise and a label or condition (e.g., a class label for digit generation in MNIST). The discriminator also receives the image and the corresponding label, and it learns to determine whether the image-label pair is real or fake.

            \textbf{Example: Conditional GAN Generator in PyTorch}
            The following Python code demonstrates how to implement a simple Conditional GAN generator.

            \begin{lstlisting}[style=python]
class CGAN_Generator(nn.Module):
    def __init__(self, noise_size, label_size, image_size):
        super(CGAN_Generator, self).__init__()
        self.label_embed = nn.Embedding(label_size, label_size)
        self.fc = nn.Sequential(
            nn.Linear(noise_size + label_size, 128),
            nn.ReLU(),
            nn.Linear(128, image_size),
            nn.Tanh()  # Output image in the range [-1, 1]
        )

    def forward(self, noise, labels):
        labels = self.label_embed(labels)
        x = torch.cat((noise, labels), dim=1)  # Concatenate noise and labels
        return self.fc(x)

# Instantiate CGAN generator for 28x28 images (e.g., MNIST dataset)
cgan_generator = CGAN_Generator(noise_size=100, label_size=10, image_size=28*28)
print(cgan_generator)
            \end{lstlisting}

            \textbf{Explanation of the Code:}
            \begin{itemize}
                \item The generator combines a random noise vector with a label (embedded as a fixed-size vector) to produce an image conditioned on the label.
                \item The model can be used for generating images corresponding to specific categories, such as digits from the MNIST dataset.
            \end{itemize}

    \section{Conclusion}
        Generative Adversarial Networks (GANs) are powerful tools for generating realistic images by creating a competitive process between a generator and a discriminator. Through adversarial training, the generator learns to produce increasingly realistic images, while the discriminator learns to become better at distinguishing between real and fake images. Architectures like \textbf{DCGAN} and \textbf{CGAN} have extended the capabilities of GANs by incorporating deep convolutional layers and conditional inputs, respectively, making them suitable for generating high-quality images and specific types of outputs. Understanding the fundamentals of GANs and their variations opens up new possibilities for image generation and manipulation in various applications.

\section{Applications of GANs in Image Processing}
    Generative Adversarial Networks (GANs) have gained widespread attention in the field of image processing due to their ability to generate highly realistic images. GANs consist of two neural networks—the generator and the discriminator—that are trained in a competitive manner. The generator attempts to create realistic images, while the discriminator tries to distinguish between real and generated images. This adversarial training leads to powerful models capable of producing high-quality, photorealistic images. In this section, we will explore the key applications of GANs in image processing, including image super-resolution, image denoising and restoration, and image style transfer \cite{porkodi2023generic}.

    \subsection{Image Super-Resolution Generation}
        Image super-resolution refers to the task of enhancing the resolution of an image, often converting a low-resolution image into a high-resolution one. Traditional methods for super-resolution relied on interpolation techniques such as bilinear or bicubic interpolation, which typically result in blurry images. GANs, specifically SRGAN (Super-Resolution GAN), have demonstrated the ability to generate sharper and more detailed high-resolution images by learning from data \cite{ledig2017photo}.

        \subsubsection{How GANs Work for Super-Resolution}
        GANs for super-resolution are trained using a combination of adversarial loss (from the discriminator) and pixel-wise loss (such as mean squared error) to generate high-quality high-resolution images from low-resolution inputs. The generator attempts to upscale the low-resolution images, while the discriminator distinguishes between the real high-resolution images and the generated ones. Over time, the generator learns to create images that are indistinguishable from the real ones.

        \paragraph{Example:} Let's implement a simplified GAN for image super-resolution using PyTorch.

        \begin{lstlisting}[style=python]
        import torch
        import torch.nn as nn

        # Define the generator (simplified)
        class Generator(nn.Module):
            def __init__(self):
                super(Generator, self).__init__()
                self.upsample = nn.Sequential(
                    nn.Conv2d(3, 64, kernel_size=9, stride=1, padding=4),
                    nn.ReLU(inplace=True),
                    nn.Conv2d(64, 256, kernel_size=3, stride=1, padding=1),
                    nn.ReLU(inplace=True),
                    nn.ConvTranspose2d(256, 3, kernel_size=4, stride=2, padding=1)
                )

            def forward(self, x):
                return self.upsample(x)

        # Define the discriminator (simplified)
        class Discriminator(nn.Module):
            def __init__(self):
                super(Discriminator, self).__init__()
                self.net = nn.Sequential(
                    nn.Conv2d(3, 64, kernel_size=3, stride=2, padding=1),
                    nn.LeakyReLU(0.2, inplace=True),
                    nn.Conv2d(64, 128, kernel_size=3, stride=2, padding=1),
                    nn.LeakyReLU(0.2, inplace=True),
                    nn.Flatten(),
                    nn.Linear(128 * 8 * 8, 1),
                    nn.Sigmoid()
                )

            def forward(self, x):
                return self.net(x)

        # Instantiate the generator and discriminator
        generator = Generator()
        discriminator = Discriminator()

        # Loss functions and optimizers
        criterion = nn.BCELoss()
        optimizer_g = torch.optim.Adam(generator.parameters(), lr=0.0002)
        optimizer_d = torch.optim.Adam(discriminator.parameters(), lr=0.0002)

        # Training loop (simplified)
        for epoch in range(100):
            # Generate high-resolution images from low-resolution inputs
            high_res_fake = generator(low_res_images)

            # Train discriminator on real and fake high-resolution images
            real_output = discriminator(high_res_images)
            fake_output = discriminator(high_res_fake.detach())
            loss_d = criterion(real_output, torch.ones_like(real_output)) + \
                     criterion(fake_output, torch.zeros_like(fake_output))

            optimizer_d.zero_grad()
            loss_d.backward()
            optimizer_d.step()

            # Train generator
            fake_output = discriminator(high_res_fake)
            loss_g = criterion(fake_output, torch.ones_like(fake_output))

            optimizer_g.zero_grad()
            loss_g.backward()
            optimizer_g.step()

            print(f"Epoch {epoch+1}, Generator Loss: {loss_g.item():.4f}, Discriminator Loss: {loss_d.item():.4f}")
        \end{lstlisting}

        \paragraph{Explanation:} In this example, we define a simplified generator and discriminator for the task of image super-resolution. The generator attempts to create high-resolution images from low-resolution inputs, and the discriminator tries to differentiate between real and generated high-resolution images. Over time, the generator learns to produce sharper and more realistic high-resolution images.

        \paragraph{Advantages of GANs for Super-Resolution:}
        \begin{itemize}
            \item GANs generate high-resolution images that are sharper and more visually appealing than those generated using traditional methods like interpolation.
            \item GANs can capture fine details and textures, making them particularly useful for tasks where clarity is essential.
        \end{itemize}

        \paragraph{Limitations:} GANs for super-resolution can be difficult to train, requiring a balance between the generator and discriminator to avoid issues like mode collapse or vanishing gradients. Additionally, GANs may generate artifacts if not trained properly.

    \subsection{Image Denoising and Restoration}
        Image denoising and restoration are important tasks in image processing, particularly for improving the quality of images that have been corrupted by noise, artifacts, or missing data \cite{gonzalez2002digital}. GANs have proven to be highly effective in these tasks, as they can learn to restore corrupted or noisy images by generating realistic and complete outputs.

        \subsubsection{Image Denoising with GANs}
        Image denoising refers to the process of removing noise from an image while preserving important features such as edges and textures. GANs for denoising typically use a generator network to map noisy images to clean images, while a discriminator network learns to distinguish between real clean images and the generated outputs \cite{xiao2021tackling}.

        \paragraph{Example:} Let's implement a simplified GAN for image denoising.

        \begin{lstlisting}[style=python]
        class DenoiseGenerator(nn.Module):
            def __init__(self):
                super(DenoiseGenerator, self).__init__()
                self.network = nn.Sequential(
                    nn.Conv2d(3, 64, kernel_size=3, padding=1),
                    nn.ReLU(inplace=True),
                    nn.Conv2d(64, 64, kernel_size=3, padding=1),
                    nn.ReLU(inplace=True),
                    nn.Conv2d(64, 3, kernel_size=3, padding=1)
                )

            def forward(self, x):
                return self.network(x)

        generator = DenoiseGenerator()
        noisy_images = torch.randn((32, 3, 64, 64))  # Simulated noisy images
        clean_images = generator(noisy_images)  # Generate clean images
        \end{lstlisting}

        \paragraph{Image Inpainting with GANs:} 
        Image inpainting refers to the process of filling in missing or corrupted parts of an image. GANs have been applied to inpainting tasks where the generator learns to fill in the missing regions of an image by generating plausible content based on the surrounding context.

        \paragraph{Advantages:}
        \begin{itemize}
            \item GANs can learn to restore images by generating missing details that are visually consistent with the surrounding regions.
            \item They are effective at preserving important visual features such as edges and textures during restoration.
        \end{itemize}

    \subsection{Image Style Transfer}
        Image style transfer is another fascinating application of GANs in which the goal is to transform an image from one style to another. For instance, using GANs, a photograph can be converted into a painting, or an image can be transformed from one artistic style to another. One of the most popular GAN architectures for style transfer is CycleGAN, which enables unpaired image-to-image translation, meaning it does not require paired training data \cite{gatys2016image}.

        \subsubsection{CycleGAN Architecture}
        CycleGAN works by learning two sets of mappings: one that translates images from domain A to domain B (e.g., from photos to paintings) and another that translates images from domain B back to domain A (e.g., from paintings to photos). The key innovation in CycleGAN is the use of cycle consistency loss, which ensures that after transforming an image from one domain to another and back again, the image remains unchanged \cite{zhu2017unpaired}.

        \paragraph{Example:} Let's implement a simplified version of CycleGAN for style transfer using PyTorch-like pseudocode.

        \begin{lstlisting}[style=python]
        # Simplified CycleGAN Generator (for A -> B and B -> A mapping)
        class CycleGANGenerator(nn.Module):
            def __init__(self):
                super(CycleGANGenerator, self).__init__()
                self.main = nn.Sequential(
                    nn.Conv2d(3, 64, kernel_size=7, padding=3),
                    nn.ReLU(inplace=True),
                    nn.Conv2d(64, 128, kernel_size=3, stride=2, padding=1),
                    nn.ReLU(inplace=True),
                    nn.Conv2d(128, 256, kernel_size=3, stride=2, padding=1),
                    nn.ReLU(inplace=True),
                    nn.ConvTranspose2d(256, 128, kernel_size=3, stride=2, padding=1, output_padding=1),
                    nn.ReLU(inplace=True),
                    nn.ConvTranspose2d(128, 64, kernel_size=3, stride=2, padding=1, output_padding=1),
                    nn.ReLU(inplace=True),
                    nn.Conv2d(64, 3, kernel_size=7, padding=3)
                )

            def forward(self, x):
                return self.main(x)

        generator_A_to_B = CycleGANGenerator()
        generator_B_to_A = CycleGANGenerator()

        image_A = torch.randn((1, 3, 256, 256))  # Simulated image from domain A (e.g., photo)
        image_B = generator_A_to_B(image_A)      # Translate from A to B (e.g., photo to painting)
        \end{lstlisting}

        \paragraph{Cycle Consistency Loss:} The key idea in CycleGAN is to ensure that when an image is translated from domain A to domain B and then back to domain A, the original image is recovered. This is achieved by minimizing the cycle consistency loss:

        \[
        \mathcal{L}_{\text{cycle}}(G, F) = \|F(G(A)) - A\|_1 + \|G(F(B)) - B\|_1
        \]

        where \(G\) is the generator for translating from domain A to B, and \(F\) is the generator for translating from domain B to A.

        \paragraph{Advantages:}
        \begin{itemize}
            \item CycleGAN does not require paired datasets, making it suitable for tasks where collecting paired data is difficult or impossible.
            \item It produces high-quality style transfers while maintaining the content of the original image.
        \end{itemize}

        \paragraph{Applications of GAN-based Style Transfer:}
        \begin{itemize}
            \item \textbf{Artistic Style Transfer:} Convert photographs into artistic paintings or transform images between different artistic styles \cite{xu2021drb}.
            \item \textbf{Domain Adaptation:} Translate images from one domain to another (e.g., converting night-time images to day-time images) \cite{sun2019see}.
        \end{itemize}
\chapter{Autoencoders and Image Processing}
    Autoencoders are a type of neural network designed for unsupervised learning tasks, where the goal is to compress and then reconstruct data as accurately as possible. In image processing, autoencoders can be used for tasks like image denoising, anomaly detection, and even generative image modeling. The core idea of autoencoders is to learn a lower-dimensional representation (or encoding) of the input image and use it to reconstruct the original image through a decoding process \cite{hinton2006reducing}. This chapter will explore the fundamental structure of autoencoders and introduce Variational Autoencoders (VAEs), which add a probabilistic component to the encoding process.

    \section{Basic Structure of Autoencoders}
        An \textbf{autoencoder} consists of two main components:
        \begin{itemize}
            \item The \textbf{encoder}, which compresses the input data (such as an image) into a lower-dimensional latent space representation.
            \item The \textbf{decoder}, which reconstructs the input data from this compressed representation.
        \end{itemize}
        The goal of an autoencoder is to minimize the difference between the original input and the reconstructed output, often measured using a loss function like mean squared error (MSE).

        \subsection{Encoder and Decoder}
            The \textbf{encoder} maps the input image to a compressed representation by applying a series of convolutional or fully connected layers. This compressed representation, often called the \textbf{latent space}, is a lower-dimensional vector that captures the most important features of the input.

            The \textbf{decoder} is a mirror image of the encoder. It takes the latent space representation and uses deconvolutional or fully connected layers to reconstruct the original image. The reconstructed image should ideally be as close as possible to the original input image.

            Mathematically, an autoencoder is represented as:
            \[
            z = f_{\text{enc}}(x) \quad \text{(encoder)}
            \]
            \[
            \hat{x} = f_{\text{dec}}(z) \quad \text{(decoder)}
            \]
            where:
            \begin{itemize}
                \item \( x \) is the original input image,
                \item \( z \) is the latent space representation,
                \item \( \hat{x} \) is the reconstructed image,
                \item \( f_{\text{enc}} \) is the encoder function,
                \item \( f_{\text{dec}} \) is the decoder function.
            \end{itemize}
            The objective of training an autoencoder is to minimize the reconstruction error \( \|x - \hat{x}\|^2 \).

            \textbf{Example: Implementing a Basic Autoencoder in PyTorch}
            The following Python code demonstrates how to implement a simple autoencoder using PyTorch for image reconstruction.

            \begin{lstlisting}[style=python]
import torch
import torch.nn as nn

# Define the Autoencoder model
class Autoencoder(nn.Module):
    def __init__(self):
        super(Autoencoder, self).__init__()
        
        # Encoder: Convolutional layers to downsample the image
        self.encoder = nn.Sequential(
            nn.Conv2d(1, 16, kernel_size=3, stride=2, padding=1),  # Output: 16x14x14
            nn.ReLU(),
            nn.Conv2d(16, 32, kernel_size=3, stride=2, padding=1),  # Output: 32x7x7
            nn.ReLU()
        )
        
        # Decoder: Transposed convolutional layers to upsample the image
        self.decoder = nn.Sequential(
            nn.ConvTranspose2d(32, 16, kernel_size=3, stride=2, padding=1, output_padding=1),  # Output: 16x14x14
            nn.ReLU(),
            nn.ConvTranspose2d(16, 1, kernel_size=3, stride=2, padding=1, output_padding=1),  # Output: 1x28x28
            nn.Sigmoid()  # Use Sigmoid to get pixel values between [0, 1]
        )
    
    def forward(self, x):
        encoded = self.encoder(x)
        decoded = self.decoder(encoded)
        return decoded

# Create the Autoencoder model
autoencoder = Autoencoder()

# Print the model architecture
print(autoencoder)
            \end{lstlisting}

            \textbf{Explanation of the Code:}
            \begin{itemize}
                \item The encoder consists of convolutional layers that progressively reduce the spatial dimensions of the input image. In this case, the input is a 28x28 grayscale image (e.g., from the MNIST dataset), and the encoded representation is reduced in size.
                \item The decoder consists of transposed convolutional layers (also known as deconvolutions) that upsample the encoded representation back to the original image size.
                \item The output of the decoder uses a \texttt{Sigmoid} activation to ensure that the pixel values are in the range \([0, 1]\), matching the input image.
            \end{itemize}

            In this basic autoencoder, the network learns to compress the input image into a lower-dimensional representation and then reconstruct it as accurately as possible. This process can be extended to various applications, including denoising, where noisy images are input to the autoencoder, and the network is trained to output denoised images.

    \section{Variational Autoencoders (VAE)}
        \textbf{Variational Autoencoders (VAEs)} are a type of autoencoder that introduce a probabilistic approach to the latent space representation. Unlike standard autoencoders that learn a fixed deterministic encoding for each input, VAEs learn a distribution over the latent space, allowing for more diverse and smooth image generation \cite{kingma2013auto}. VAEs are particularly useful for generating new images that are similar to those in the training set, making them popular in tasks like image synthesis.

        \paragraph{How VAEs Work}
        In a VAE, instead of encoding the input as a single point in the latent space, the encoder outputs two vectors: the mean \( \mu \) and the standard deviation \( \sigma \) of a Gaussian distribution. The latent space \( z \) is then sampled from this distribution:
        \[
        z \sim \mathcal{N}(\mu, \sigma^2)
        \]
        This probabilistic sampling allows the VAE to generate diverse outputs, even when given similar inputs. The decoder then takes this sampled latent vector \( z \) and reconstructs the image, just like in a regular autoencoder.

        \paragraph{Loss Function in VAEs}
        The VAE introduces a more complex loss function that consists of two parts:
        \begin{itemize}
            \item \textbf{Reconstruction loss}: This measures how well the decoder can reconstruct the input image from the latent space representation. It is usually computed as the mean squared error between the original and reconstructed image.
            \item \textbf{KL divergence}: This regularization term ensures that the learned latent space distribution is close to a standard Gaussian distribution \( \mathcal{N}(0, I) \). The KL divergence measures the difference between the two distributions and encourages the encoder to produce a distribution that is smooth and continuous.
        \end{itemize}
        The total loss for a VAE is the sum of these two terms:
        \[
        \mathcal{L}_{\text{VAE}} = \mathcal{L}_{\text{reconstruction}} + \text{KL}(\mathcal{N}(\mu, \sigma^2) || \mathcal{N}(0, I))
        \]

        \textbf{Example: Implementing a Variational Autoencoder (VAE) in PyTorch}
        The following Python code demonstrates how to implement a simple VAE using PyTorch for image generation.

        \begin{lstlisting}[style=python]
class VAE(nn.Module):
    def __init__(self):
        super(VAE, self).__init__()
        
        # Encoder: Define layers for mean and log-variance
        self.encoder = nn.Sequential(
            nn.Conv2d(1, 16, kernel_size=3, stride=2, padding=1),
            nn.ReLU(),
            nn.Conv2d(16, 32, kernel_size=3, stride=2, padding=1),
            nn.ReLU()
        )
        self.fc_mu = nn.Linear(32*7*7, 20)  # Mean vector
        self.fc_logvar = nn.Linear(32*7*7, 20)  # Log-variance vector
        
        # Decoder: Transpose convolutional layers
        self.decoder_fc = nn.Linear(20, 32*7*7)
        self.decoder = nn.Sequential(
            nn.ConvTranspose2d(32, 16, kernel_size=3, stride=2, padding=1, output_padding=1),
            nn.ReLU(),
            nn.ConvTranspose2d(16, 1, kernel_size=3, stride=2, padding=1, output_padding=1),
            nn.Sigmoid()  # Output pixel values in the range [0, 1]
        )
    
    def encode(self, x):
        h = self.encoder(x).view(-1, 32*7*7)
        mu = self.fc_mu(h)
        logvar = self.fc_logvar(h)
        return mu, logvar
    
    def reparameterize(self, mu, logvar):
        std = torch.exp(0.5 * logvar)
        eps = torch.randn_like(std)
        return mu + eps * std
    
    def decode(self, z):
        z = self.decoder_fc(z).view(-1, 32, 7, 7)
        return self.decoder(z)
    
    def forward(self, x):
        mu, logvar = self.encode(x)
        z = self.reparameterize(mu, logvar)
        return self.decode(z), mu, logvar

# Loss function for VAE (reconstruction + KL divergence)
def vae_loss(recon_x, x, mu, logvar):
    # Reconstruction loss
    recon_loss = nn.functional.binary_cross_entropy(recon_x, x, reduction='sum')
    
    # KL divergence
    kld_loss = -0.5 * torch.sum(1 + logvar - mu.pow(2) - logvar.exp())
    
    return recon_loss + kld_loss

# Create the VAE model
vae = VAE()

# Print the model architecture
print(vae)
        \end{lstlisting}

        \textbf{Explanation of the Code:}
        \begin{itemize}
            \item The encoder compresses the input into two vectors: the mean (\( \mu \)) and the log-variance (\( \log\sigma^2 \)).
            \item The \texttt{reparameterize()} function samples a latent vector \( z \) using the reparameterization trick to allow backpropagation through the sampling step.
            \item The decoder takes the latent vector \( z \) and reconstructs the original image.
            \item The \texttt{vae\_loss()} function computes the total loss, which includes both the reconstruction loss and the KL divergence.
        \end{itemize}

        VAEs are particularly powerful for generating new data samples because they learn a continuous and smooth latent space. By sampling from this latent space, VAEs can generate diverse images that are not exact replicas of the training data but still belong to the same distribution.

    \section{Conclusion}
        Autoencoders and Variational Autoencoders (VAEs) are powerful tools in the realm of image processing. Autoencoders learn efficient representations of images by compressing them into lower-dimensional spaces and then reconstructing them, making them useful for tasks like image compression, denoising, and anomaly detection. Variational Autoencoders (VAEs) extend this idea by introducing probabilistic latent variables, allowing for more diverse and flexible image generation. Both autoencoders and VAEs have opened up new possibilities in generative modeling, image synthesis, and other unsupervised learning tasks.

\section{Applications of Autoencoders in Image Processing}
    Autoencoders are a type of neural network designed to learn efficient representations of input data, often used for unsupervised learning tasks. They consist of two main components: an encoder that compresses the input into a lower-dimensional latent space and a decoder that reconstructs the input from this compressed representation. Autoencoders are particularly useful in image processing tasks such as image denoising, compression, and reconstruction. In this section, we will explore how autoencoders are applied in these areas, providing detailed explanations and examples \cite{li2023comprehensive}.

    \subsection{Image Denoising}
        Denoising Autoencoders (DAEs) are a variant of autoencoders specifically designed to remove noise from images. The key idea is to train the autoencoder to map noisy inputs to clean outputs. The encoder compresses the noisy input, and the decoder learns to reconstruct the clean version of the image by learning noise-invariant features \cite{vincent2008extracting}.

        \subsubsection{Denoising Autoencoder Architecture}
        The architecture of a denoising autoencoder is similar to a standard autoencoder, with an encoder-decoder structure:
        \begin{itemize}
            \item \textbf{Encoder:} The encoder transforms the noisy input image into a lower-dimensional latent space, capturing essential features while discarding noise.
            \item \textbf{Decoder:} The decoder reconstructs the image from the latent space representation, with the objective of producing a noise-free image that resembles the original clean input.
        \end{itemize}

        The autoencoder is trained by minimizing the reconstruction loss, typically the mean squared error (MSE) between the clean image and the reconstructed image. During training, noise is added to the input images, and the autoencoder learns to remove this noise through the reconstruction process.

        \paragraph{Example:} Let's implement a denoising autoencoder using PyTorch to remove Gaussian noise from images.

        \begin{lstlisting}[style=python]
        import torch
        import torch.nn as nn
        import torch.optim as optim
        from torchvision import datasets, transforms
        from torch.utils.data import DataLoader

        # Define the denoising autoencoder
        class DenoisingAutoencoder(nn.Module):
            def __init__(self):
                super(DenoisingAutoencoder, self).__init__()
                # Encoder
                self.encoder = nn.Sequential(
                    nn.Conv2d(1, 64, kernel_size=3, stride=2, padding=1),  # (batch_size, 64, 14, 14)
                    nn.ReLU(True),
                    nn.Conv2d(64, 128, kernel_size=3, stride=2, padding=1),  # (batch_size, 128, 7, 7)
                    nn.ReLU(True)
                )
                # Decoder
                self.decoder = nn.Sequential(
                    nn.ConvTranspose2d(128, 64, kernel_size=3, stride=2, padding=1, output_padding=1),  # (batch_size, 64, 14, 14)
                    nn.ReLU(True),
                    nn.ConvTranspose2d(64, 1, kernel_size=3, stride=2, padding=1, output_padding=1),  # (batch_size, 1, 28, 28)
                    nn.Sigmoid()  # Output between 0 and 1
                )

            def forward(self, x):
                x = self.encoder(x)
                x = self.decoder(x)
                return x

        # Load the MNIST dataset and add Gaussian noise
        transform = transforms.Compose([transforms.ToTensor()])
        train_dataset = datasets.MNIST(root='./data', train=True, transform=transform, download=True)
        noisy_train_dataset = torch.clone(train_dataset.data).float() / 255.0 + torch.randn(train_dataset.data.size()) * 0.2

        # Initialize the denoising autoencoder
        model = DenoisingAutoencoder()
        criterion = nn.MSELoss()
        optimizer = optim.Adam(model.parameters(), lr=0.001)

        # Training loop
        for epoch in range(5):
            running_loss = 0.0
            for images in DataLoader(noisy_train_dataset, batch_size=64, shuffle=True):
                optimizer.zero_grad()
                outputs = model(images.unsqueeze(1))  # Add channel dimension
                loss = criterion(outputs, images.unsqueeze(1))  # Calculate MSE loss
                loss.backward()
                optimizer.step()
                running_loss += loss.item()
            print(f'Epoch {epoch+1}, Loss: {running_loss/len(train_dataset):.4f}')
        \end{lstlisting}

        \paragraph{Explanation:} In this example, we define a denoising autoencoder for the MNIST dataset. Gaussian noise is added to the input images, and the model is trained to reconstruct the clean versions. By minimizing the mean squared error (MSE) loss, the autoencoder learns to remove noise from the images, producing cleaner outputs. This approach can be extended to more complex datasets and different types of noise.

        \paragraph{Advantages of Denoising Autoencoders:}
        \begin{itemize}
            \item \textbf{Noise Removal:} Denoising autoencoders can effectively remove various types of noise, including Gaussian noise, salt-and-pepper noise, and more.
            \item \textbf{Feature Learning:} The encoder learns noise-invariant features, which can be useful for downstream tasks such as classification.
        \end{itemize}

    \subsection{Image Compression}
        Autoencoders can be used for image compression by learning compact, low-dimensional representations of input images. In this application, the encoder compresses the input image into a small latent space, and the decoder reconstructs the original image from this compressed representation. Unlike traditional compression techniques, autoencoders learn efficient encoding schemes from data without the need for predefined rules like those in JPEG or PNG formats.

        \subsubsection{Autoencoder Architecture for Compression}
        The autoencoder architecture for image compression is similar to that used for denoising, but the primary goal is to minimize the size of the latent space while maintaining high-quality reconstructions. This enables the model to encode images into smaller, more efficient representations that can be stored or transmitted and later decoded.

        \paragraph{Example:} Let's implement a simple autoencoder for image compression using PyTorch.

        \begin{lstlisting}[style=python]
        class CompressionAutoencoder(nn.Module):
            def __init__(self):
                super(CompressionAutoencoder, self).__init__()
                # Encoder: compress image to a smaller representation
                self.encoder = nn.Sequential(
                    nn.Conv2d(1, 32, kernel_size=3, stride=2, padding=1),  # (batch_size, 32, 14, 14)
                    nn.ReLU(True),
                    nn.Conv2d(32, 64, kernel_size=3, stride=2, padding=1),  # (batch_size, 64, 7, 7)
                    nn.ReLU(True),
                    nn.Conv2d(64, 128, kernel_size=7)  # (batch_size, 128, 1, 1) compressed representation
                )
                # Decoder: reconstruct the image
                self.decoder = nn.Sequential(
                    nn.ConvTranspose2d(128, 64, kernel_size=7),  # (batch_size, 64, 7, 7)
                    nn.ReLU(True),
                    nn.ConvTranspose2d(64, 32, kernel_size=3, stride=2, padding=1, output_padding=1),  # (batch_size, 32, 14, 14)
                    nn.ReLU(True),
                    nn.ConvTranspose2d(32, 1, kernel_size=3, stride=2, padding=1, output_padding=1),  # (batch_size, 1, 28, 28)
                    nn.Sigmoid()
                )

            def forward(self, x):
                x = self.encoder(x)
                x = self.decoder(x)
                return x

        # Instantiate the compression autoencoder
        model = CompressionAutoencoder()
        \end{lstlisting}

        \paragraph{Explanation:} In this implementation, the autoencoder compresses the input image into a 128-dimensional representation (latent space), which serves as the compressed version of the image. The decoder then reconstructs the original image from this compressed representation. This process can be used for lossy image compression, where some information is lost, but the overall structure and visual quality are preserved.

        \paragraph{Advantages of Autoencoder-Based Compression:}
        \begin{itemize}
            \item \textbf{Learned Compression:} Autoencoders can learn optimal compression schemes directly from the data, making them adaptable to specific image types or datasets.
            \item \textbf{Reduced File Sizes:} By reducing the dimensionality of the input, autoencoders enable efficient storage and transmission of images.
        \end{itemize}

    \subsection{Image Reconstruction and Inpainting}
        Image inpainting refers to the task of reconstructing missing or damaged parts of an image. Autoencoders can be used for inpainting by learning how to generate plausible content for the missing regions based on the surrounding pixels. This is achieved by training the autoencoder on images where certain parts are masked or corrupted, and the autoencoder learns to reconstruct the original, complete image \cite{tu2019facial}.

        \subsubsection{Inpainting with Autoencoders}
        In image inpainting, the autoencoder receives an image with missing or corrupted regions as input, and the goal is to reconstruct the full image, including the missing parts. The latent space representation learned by the encoder allows the model to infer the missing content, which is then used by the decoder to generate the reconstructed image.

        \paragraph{Example:} Let's implement a basic autoencoder for image inpainting using PyTorch.

        \begin{lstlisting}[style=python]
        class InpaintingAutoencoder(nn.Module):
            def __init__(self):
                super(InpaintingAutoencoder, self).__init__()
                # Encoder
                self.encoder = nn.Sequential(
                    nn.Conv2d(1, 64, kernel_size=3, stride=2, padding=1),  # (batch_size, 64, 14, 14)
                    nn.ReLU(True),
                    nn.Conv2d(64, 128, kernel_size=3, stride=2, padding=1),  # (batch_size, 128, 7, 7)
                    nn.ReLU(True)
                )
                # Decoder
                self.decoder = nn.Sequential(
                    nn.ConvTranspose2d(128, 64, kernel_size=3, stride=2, padding=1, output_padding=1),  # (batch_size, 64, 14, 14)
                    nn.ReLU(True),
                    nn.ConvTranspose2d(64, 1, kernel_size=3, stride=2, padding=1, output_padding=1),  # (batch_size, 1, 28, 28)
                    nn.Sigmoid()
                )

            def forward(self, x):
                x = self.encoder(x)
                x = self.decoder(x)
                return x

        # Create masked images with missing regions
        def mask_images(images, mask_size=10):
            masked_images = images.clone()
            for img in masked_images:
                x = torch.randint(0, 28 - mask_size, (1,)).item()
                y = torch.randint(0, 28 - mask_size, (1,)).item()
                img[:, x:x+mask_size, y:y+mask_size] = 0  # Mask a random square region
            return masked_images

        # Instantiate the inpainting autoencoder
        model = InpaintingAutoencoder()
        masked_images = mask_images(train_dataset.data.float() / 255.0)
        \end{lstlisting}

        \paragraph{Explanation:} In this example, a basic autoencoder is trained to reconstruct missing parts of images. A random square region of the input image is masked, and the model is trained to infer and fill in the missing pixels. This approach can be extended to more complex datasets and various masking strategies, making autoencoders a powerful tool for image inpainting and reconstruction.

        \paragraph{Advantages of Autoencoders for Inpainting:}
        \begin{itemize}
            \item \textbf{Contextual Reconstruction:} Autoencoders can generate plausible content for missing regions by learning the contextual relationships between pixels.
            \item \textbf{Versatility:} Inpainting autoencoders can be applied to various tasks, such as restoring damaged photographs, filling in missing data, and improving images with corrupted areas.
        \end{itemize}

\chapter{Deep Learning for Image Super-Resolution and Deblurring}
    Image super-resolution and deblurring are two important tasks in image processing that focus on improving image quality. Super-resolution involves upscaling low-resolution images to generate high-resolution counterparts, while deblurring removes blur from images to restore sharpness. Traditional methods for these tasks often struggle to generate fine details or high-quality results \cite{albluwi2018image}. However, deep learning techniques have led to significant advancements in both fields. In this chapter, we explore deep learning methods such as SRGAN, ESRGAN, VDSR, and EDSR for image super-resolution.

    \section{Image Super-Resolution}
        Image super-resolution aims to enhance the resolution of images by generating high-frequency details that were lost in the downscaling process. Deep learning models have proven highly effective in this task by learning complex mappings from low-resolution (LR) to high-resolution (HR) images \cite{dong2015image}.

        \subsection{SRGAN (Super-Resolution GAN)}
            \textbf{Super-Resolution GAN (SRGAN)} is a deep learning model that uses the adversarial training framework of GANs to produce realistic, high-resolution images from low-resolution inputs. The core of SRGAN consists of two networks: the \textbf{generator}, which generates high-resolution images, and the \textbf{discriminator}, which distinguishes between real and generated high-resolution images \cite{wang2023review}.

            \paragraph{Generator Network:} The generator in SRGAN is a convolutional neural network (CNN) that takes a low-resolution image as input and outputs a high-resolution image. It is designed to predict high-frequency details that are missing in the low-resolution image. The generator is trained to "fool" the discriminator by producing images that appear realistic \cite{ledig2017photo}.

            \paragraph{Discriminator Network:} The discriminator is another CNN that learns to classify whether an image is a real high-resolution image or one generated by the generator. It outputs a probability that indicates how realistic the image looks.

            \paragraph{Adversarial Loss and Perceptual Loss:}
            The SRGAN training process involves two loss functions:
            \begin{itemize}
                \item \textbf{Adversarial loss}: Encourages the generator to produce images that are indistinguishable from real high-resolution images.
                \item \textbf{Perceptual loss}: Instead of pixel-wise differences (like mean squared error), perceptual loss compares high-level features extracted from a pre-trained network (like VGG) to ensure that the generated images look perceptually similar to real images.
            \end{itemize}

            \textbf{Example: SRGAN Generator in PyTorch}
            The following Python code demonstrates how to define the generator for an SRGAN in PyTorch.

            \begin{lstlisting}[style=python]
import torch
import torch.nn as nn

class ResidualBlock(nn.Module):
    def __init__(self, num_channels):
        super(ResidualBlock, self).__init__()
        self.block = nn.Sequential(
            nn.Conv2d(num_channels, num_channels, kernel_size=3, padding=1),
            nn.BatchNorm2d(num_channels),
            nn.PReLU(),
            nn.Conv2d(num_channels, num_channels, kernel_size=3, padding=1),
            nn.BatchNorm2d(num_channels)
        )
    
    def forward(self, x):
        return x + self.block(x)  # Residual connection

class SRGAN_Generator(nn.Module):
    def __init__(self, num_residual_blocks=16):
        super(SRGAN_Generator, self).__init__()
        
        # Initial Convolution layer
        self.conv1 = nn.Conv2d(3, 64, kernel_size=9, padding=4)
        self.prelu = nn.PReLU()
        
        # Residual Blocks
        self.residual_blocks = nn.Sequential(
            *[ResidualBlock(64) for _ in range(num_residual_blocks)]
        )
        
        # Upsampling Layers
        self.upsample = nn.Sequential(
            nn.Conv2d(64, 256, kernel_size=3, padding=1),
            nn.PixelShuffle(2),  # Upscales by a factor of 2
            nn.PReLU(),
            nn.Conv2d(64, 256, kernel_size=3, padding=1),
            nn.PixelShuffle(2),  # Upscales by another factor of 2
            nn.PReLU()
        )
        
        # Final output layer
        self.conv2 = nn.Conv2d(64, 3, kernel_size=9, padding=4)
        self.tanh = nn.Tanh()  # Output in range [-1, 1]
    
    def forward(self, x):
        x = self.prelu(self.conv1(x))
        residual = x
        x = self.residual_blocks(x)
        x = self.upsample(x + residual)  # Skip connection
        return self.tanh(self.conv2(x))

# Instantiate the generator model
generator = SRGAN_Generator()
print(generator)
            \end{lstlisting}

            \textbf{Explanation of the Code:}
            \begin{itemize}
                \item The generator uses residual blocks, which include skip connections to preserve low-level details from the input.
                \item The \texttt{PixelShuffle} layer is used to upsample the image by a factor of 4 in total (2x2 upsampling).
                \item The output of the generator is a high-resolution image with pixel values in the range \([-1, 1]\), produced by the \texttt{Tanh} activation function.
            \end{itemize}

            SRGAN has been successful in generating high-quality images with fine details, especially in the upscaling of images for applications like satellite imagery and video streaming.

        \subsection{ESRGAN (Enhanced Super-Resolution GAN)}
            \textbf{Enhanced Super-Resolution GAN (ESRGAN)} is an improved version of SRGAN that addresses some limitations and introduces more sophisticated techniques to generate even more realistic and detailed images \cite{wang2018esrgan}. ESRGAN improves on SRGAN in several ways:
            \begin{itemize}
                \item \textbf{Residual-in-Residual Dense Block (RRDB)}: ESRGAN uses this block instead of the standard residual block. It allows for better gradient flow and richer feature extraction.
                \item \textbf{Perceptual Loss and Improved Discriminator}: ESRGAN uses a more advanced perceptual loss function based on high-level features from a pre-trained network, as well as an improved discriminator for better adversarial training.
                \item \textbf{Decreased Batch Normalization}: Removing batch normalization from the network has shown to improve performance, reducing artifacts in the generated images.
            \end{itemize}

            \textbf{Example: ESRGAN Residual Block in PyTorch}
            The following Python code demonstrates the RRDB block used in ESRGAN.

            \begin{lstlisting}[style=python]
class RRDBBlock(nn.Module):
    def __init__(self, num_channels):
        super(RRDBBlock, self).__init__()
        self.conv1 = nn.Conv2d(num_channels, num_channels, kernel_size=3, padding=1)
        self.conv2 = nn.Conv2d(num_channels, num_channels, kernel_size=3, padding=1)
        self.conv3 = nn.Conv2d(num_channels, num_channels, kernel_size=3, padding=1)
        self.relu = nn.PReLU()
    
    def forward(self, x):
        residual = x
        out = self.relu(self.conv1(x))
        out = self.relu(self.conv2(out))
        out = self.conv3(out)
        return residual + out * 0.2  # Scaled residual connection

# Example use of RRDB in an ESRGAN model
rrdb = RRDBBlock(num_channels=64)
            \end{lstlisting}

            \textbf{Explanation of the Code:}
            \begin{itemize}
                \item The \textbf{RRDB block} consists of three convolutional layers with skip connections between them, allowing for richer features and better gradient flow.
                \item A scaling factor is applied to the residual connection to stabilize the training process.
            \end{itemize}

        \subsection{VDSR (Very Deep Super-Resolution Network)}
            \textbf{VDSR (Very Deep Super-Resolution Network)} is another deep learning model designed for image super-resolution. Unlike SRGAN, which uses adversarial loss, VDSR is a fully convolutional network with many layers that learns residual mappings. By training on residuals (i.e., the difference between low-resolution and high-resolution images), VDSR is able to upscale images with high precision \cite{kim2016accurate}.

            \paragraph{Key Features of VDSR:}
            \begin{itemize}
                \item \textbf{Residual Learning}: VDSR learns the difference between the upscaled low-resolution image and the original high-resolution image. This residual learning framework helps the network converge faster and achieve higher precision.
                \item \textbf{Depth}: VDSR uses a deep architecture (up to 20 layers), allowing it to capture more complex mappings for super-resolution tasks.
            \end{itemize}

            \textbf{Example: VDSR Network in PyTorch}
            The following Python code demonstrates the VDSR architecture.

            \begin{lstlisting}[style=python]
class VDSR(nn.Module):
    def __init__(self):
        super(VDSR, self).__init__()
        self.conv1 = nn.Conv2d(1, 64, kernel_size=3, padding=1)
        self.relu = nn.ReLU(inplace=True)
        self.convs = nn.Sequential(*[
            nn.Conv2d(64, 64, kernel_size=3, padding=1),
            nn.ReLU(inplace=True)
        ] * 18)  # 18 convolutional layers
        self.conv_last = nn.Conv2d(64, 1, kernel_size=3, padding=1)
    
    def forward(self, x):
        residual = x
        x = self.relu(self.conv1(x))
        x = self.convs(x)
        x = self.conv_last(x)
        return x + residual  # Add the residual to the output

# Instantiate the VDSR model
vdsr = VDSR()
print(vdsr)
            \end{lstlisting}

            \textbf{Explanation of the Code:}
            \begin{itemize}
                \item The network starts with an initial convolutional layer, followed by 18 convolutional layers to extract deep features.
                \item The output is the sum of the residual image and the low-resolution input, effectively producing the high-resolution output.
            \end{itemize}

        \subsection{EDSR (Enhanced Deep Super-Resolution Network)}
            \textbf{Enhanced Deep Super-Resolution Network (EDSR)} is an optimized version of VDSR designed to improve both accuracy and efficiency. EDSR removes batch normalization layers, which can introduce artifacts in the image, and focuses on a simpler architecture while maintaining deep layers to handle more complex super-resolution tasks \cite{lim2017enhanced}.

            \paragraph{Key Features of EDSR:}
            \begin{itemize}
                \item \textbf{No Batch Normalization}: Removing batch normalization avoids potential artifacts and results in cleaner, more accurate upscaling.
                \item \textbf{Residual Blocks}: Like VDSR, EDSR employs residual blocks to learn the upscaling mappings, allowing the network to focus on high-frequency details.
                \item \textbf{Scalability}: EDSR is scalable and can be used for various scaling factors (e.g., 2x, 4x, 8x).
            \end{itemize}

            \textbf{Example: EDSR Residual Block in PyTorch}
            The following Python code demonstrates a residual block used in EDSR.

            \begin{lstlisting}[style=python]
class EDSR_ResidualBlock(nn.Module):
    def __init__(self, num_channels):
        super(EDSR_ResidualBlock, self).__init__()
        self.conv1 = nn.Conv2d(num_channels, num_channels, kernel_size=3, padding=1)
        self.relu = nn.ReLU(inplace=True)
        self.conv2 = nn.Conv2d(num_channels, num_channels, kernel_size=3, padding=1)
    
    def forward(self, x):
        residual = x
        out = self.relu(self.conv1(x))
        out = self.conv2(out)
        return out + residual  # Add residual connection

# Example use of residual block in EDSR model
edsr_block = EDSR_ResidualBlock(num_channels=64)
            \end{lstlisting}

            \textbf{Explanation of the Code:}
            \begin{itemize}
                \item The residual block in EDSR contains two convolutional layers with ReLU activations and a skip connection.
                \item These residual blocks are stacked to form a deeper network that excels at image super-resolution.
            \end{itemize}

    \section{Conclusion}
        Deep learning has greatly advanced the field of image super-resolution, allowing for the generation of high-resolution images with fine details from low-resolution inputs. Models like \textbf{SRGAN} and \textbf{ESRGAN} use adversarial training to produce high-quality, realistic images, while networks like \textbf{VDSR} and \textbf{EDSR} rely on deep convolutional architectures and residual learning to achieve high precision. Each model has unique strengths and is suited to different super-resolution tasks, making deep learning an indispensable tool in image enhancement and restoration.

\section{Deep Learning for Image Deblurring}
    Image deblurring is a critical task in image processing, aimed at recovering sharp images from blurred ones. Blurring often occurs due to camera shake, object movement, or defocus, and it can degrade the quality of images in various applications such as photography, surveillance, and autonomous driving. Traditional deblurring methods were often based on hand-crafted features and assumptions about the blur kernel. However, deep learning approaches have significantly advanced the field, offering more robust and accurate results. In this section, we will explore various deep learning methods for image deblurring, including GAN-based approaches (DeBlurGAN), CNN-based methods, and dynamic deblurring for more complex scenarios.

    \subsection{DeBlurGAN}
        DeBlurGAN is a GAN-based method specifically designed for motion deblurring. Introduced by Kupyn et al. in 2018, DeBlurGAN uses a generator to predict clear, sharp images from blurred ones, while a discriminator evaluates the quality of the generated images. The key innovation of DeBlurGAN lies in its adversarial training, which allows it to generate highly realistic and sharp images, even when dealing with severe motion blur.

        \subsubsection{DeBlurGAN Architecture}
        DeBlurGAN consists of two main components:
        \begin{itemize}
            \item \textbf{Generator:} The generator is a deep convolutional neural network that takes a blurred image as input and predicts a sharp, deblurred version of the image. It uses a series of convolutional layers and residual connections to progressively refine the image.
            \item \textbf{Discriminator:} The discriminator is trained to distinguish between real sharp images and generated deblurred images. It outputs a probability indicating whether the input image is real (from the dataset) or generated (from the generator).
        \end{itemize}

        The generator and discriminator are trained together in an adversarial fashion. The generator tries to fool the discriminator by generating more realistic deblurred images, while the discriminator learns to distinguish between real and generated images. The overall loss function combines adversarial loss (from the GAN) and reconstruction loss (typically L1 or L2 loss), encouraging the generator to produce accurate and high-quality deblurred images.

        \paragraph{Example:} Let's implement a simplified version of DeBlurGAN using PyTorch.

        \begin{lstlisting}[style=python]
        import torch
        import torch.nn as nn
        import torch.optim as optim

        # Define the generator (simplified version)
        class DeBlurGAN_Generator(nn.Module):
            def __init__(self):
                super(DeBlurGAN_Generator, self).__init__()
                self.encoder = nn.Sequential(
                    nn.Conv2d(3, 64, kernel_size=7, stride=1, padding=3),
                    nn.ReLU(inplace=True),
                    nn.Conv2d(64, 128, kernel_size=5, stride=2, padding=2),
                    nn.ReLU(inplace=True)
                )
                self.decoder = nn.Sequential(
                    nn.ConvTranspose2d(128, 64, kernel_size=4, stride=2, padding=1),
                    nn.ReLU(inplace=True),
                    nn.Conv2d(64, 3, kernel_size=7, stride=1, padding=3),
                    nn.Tanh()  # Output between -1 and 1
                )

            def forward(self, x):
                x = self.encoder(x)
                x = self.decoder(x)
                return x

        # Define the discriminator (simplified version)
        class DeBlurGAN_Discriminator(nn.Module):
            def __init__(self):
                super(DeBlurGAN_Discriminator, self).__init__()
                self.network = nn.Sequential(
                    nn.Conv2d(3, 64, kernel_size=4, stride=2, padding=1),
                    nn.LeakyReLU(0.2, inplace=True),
                    nn.Conv2d(64, 128, kernel_size=4, stride=2, padding=1),
                    nn.LeakyReLU(0.2, inplace=True),
                    nn.Conv2d(128, 1, kernel_size=4, stride=1, padding=1),
                    nn.Sigmoid()
                )

            def forward(self, x):
                return self.network(x)

        # Instantiate the generator and discriminator
        generator = DeBlurGAN_Generator()
        discriminator = DeBlurGAN_Discriminator()

        # Loss function and optimizers
        adversarial_loss = nn.BCELoss()
        reconstruction_loss = nn.L1Loss()  # For image reconstruction
        optimizer_g = optim.Adam(generator.parameters(), lr=0.0002)
        optimizer_d = optim.Adam(discriminator.parameters(), lr=0.0002)

        # Training loop (simplified)
        for epoch in range(100):
            # Forward pass for generator (predict deblurred image)
            deblurred_images = generator(blurred_images)

            # Train discriminator
            real_output = discriminator(real_images)
            fake_output = discriminator(deblurred_images.detach())
            loss_d = adversarial_loss(real_output, torch.ones_like(real_output)) + \
                     adversarial_loss(fake_output, torch.zeros_like(fake_output))

            optimizer_d.zero_grad()
            loss_d.backward()
            optimizer_d.step()

            # Train generator
            fake_output = discriminator(deblurred_images)
            loss_g = adversarial_loss(fake_output, torch.ones_like(fake_output)) + \
                     reconstruction_loss(deblurred_images, real_images)

            optimizer_g.zero_grad()
            loss_g.backward()
            optimizer_g.step()

            print(f"Epoch {epoch+1}, Generator Loss: {loss_g.item():.4f}, Discriminator Loss: {loss_d.item():.4f}")
        \end{lstlisting}

        \paragraph{Explanation:} In this example, we define a simplified version of DeBlurGAN. The generator attempts to generate sharp, deblurred images from the input blurred images, and the discriminator tries to distinguish between real sharp images and generated deblurred ones. By combining adversarial loss and reconstruction loss, the generator learns to produce visually realistic and sharp images.

        \paragraph{Advantages of DeBlurGAN:}
        \begin{itemize}
            \item \textbf{High-Quality Results:} DeBlurGAN can produce high-quality deblurred images, even in challenging scenarios involving severe motion blur.
            \item \textbf{Adversarial Training:} The use of GANs ensures that the generated images are realistic and visually appealing, beyond just minimizing pixel-wise errors.
        \end{itemize}

        \paragraph{Limitations:} Training GANs can be challenging due to instability and potential issues such as mode collapse. Additionally, DeBlurGAN might generate artifacts if the generator and discriminator are not properly balanced during training.

    \subsection{CNN-based Deblurring}
        Convolutional Neural Networks (CNNs) have been widely applied to deblurring tasks due to their strong feature extraction capabilities. CNN-based deblurring methods typically use a deep network with multi-scale convolutional layers to reconstruct sharp images from blurry inputs. These methods aim to learn a direct mapping from blurry images to clear ones by leveraging the hierarchical feature learning abilities of CNNs.

        \subsubsection{CNN Architecture for Deblurring}
        CNN-based deblurring models are usually composed of several convolutional layers that operate at multiple scales:
        \begin{itemize}
            \item \textbf{Multi-Scale Convolution:} Blurring can occur at different scales, so CNNs often use multi-scale convolutional layers to capture features at various resolutions. This allows the network to focus on both fine details and large-scale structures.
            \item \textbf{Residual Connections:} Residual connections are often used in CNN-based deblurring models to help the network learn the difference between the blurry input and the sharp output (residual mapping), improving convergence and reducing training difficulty.
        \end{itemize}

        \paragraph{Example:} Let's implement a basic CNN-based deblurring network using PyTorch.

        \begin{lstlisting}[style=python]
        class CNN_Deblurring(nn.Module):
            def __init__(self):
                super(CNN_Deblurring, self).__init__()
                self.network = nn.Sequential(
                    nn.Conv2d(3, 64, kernel_size=3, stride=1, padding=1),
                    nn.ReLU(inplace=True),
                    nn.Conv2d(64, 64, kernel_size=3, stride=1, padding=1),
                    nn.ReLU(inplace=True),
                    nn.Conv2d(64, 3, kernel_size=3, stride=1, padding=1)  # Output layer
                )

            def forward(self, x):
                return self.network(x)

        # Instantiate the model
        model = CNN_Deblurring()

        # Training loop (simplified)
        optimizer = optim.Adam(model.parameters(), lr=0.001)
        criterion = nn.MSELoss()

        for epoch in range(50):
            optimizer.zero_grad()
            outputs = model(blurred_images)  # Predict deblurred images
            loss = criterion(outputs, sharp_images)  # Compare with ground truth sharp images
            loss.backward()
            optimizer.step()

            print(f"Epoch {epoch+1}, Loss: {loss.item():.4f}")
        \end{lstlisting}

        \paragraph{Explanation:} In this example, we implement a basic CNN-based deblurring model. The network consists of a few convolutional layers that directly map blurry images to sharp ones. The model is trained using mean squared error (MSE) loss, which measures the difference between the predicted deblurred image and the ground truth sharp image.

        \paragraph{Advantages of CNN-Based Deblurring:}
        \begin{itemize}
            \item \textbf{Efficient Feature Learning:} CNNs can learn to extract features at multiple scales, making them effective for handling various types of blur.
            \item \textbf{Residual Learning:} By focusing on learning the residual between the blurred input and sharp output, CNNs can improve training efficiency and convergence.
        \end{itemize}

        \paragraph{Limitations:} CNN-based methods may struggle with severe blurring or complex motion patterns. Additionally, the lack of adversarial training can lead to less realistic outputs compared to GAN-based approaches like DeBlurGAN.

    \subsection{Dynamic Deblurring}
        Dynamic deblurring refers to the process of handling more complex blur scenarios, such as motion blur caused by moving objects or defocus blur resulting from camera misalignment. Dynamic deblurring is particularly challenging because the blur is often non-uniform and can vary across different parts of the image. Deep learning methods have been developed to address these challenges by incorporating motion estimation, optical flow, and other advanced techniques into the deblurring pipeline.

        \subsubsection{Deep Learning Approaches for Dynamic Deblurring}
        Deep learning-based dynamic deblurring methods typically involve the following components:
        \begin{itemize}
            \item \textbf{Motion Estimation:} These methods often include a motion estimation network that predicts the motion blur kernel for different regions of the image. This allows the model to handle non-uniform blur by applying region-specific deblurring operations.
            \item \textbf{Optical Flow:} Some dynamic deblurring methods use optical flow estimation to model the motion of objects between frames in video sequences, allowing for better handling of motion blur.
            \item \textbf{Recurrent Networks:} Recurrent neural networks (RNNs) and long short-term memory (LSTM) networks can be used in dynamic deblurring tasks to capture temporal dependencies and improve the deblurring performance in video sequences.
        \end{itemize}

        \paragraph{Example:} Let's implement a simplified dynamic deblurring model that incorporates motion estimation for handling complex blur.

        \begin{lstlisting}[style=python]
        class DynamicDeblurring(nn.Module):
            def __init__(self):
                super(DynamicDeblurring, self).__init__()
                # Motion estimation network
                self.motion_estimator = nn.Sequential(
                    nn.Conv2d(3, 64, kernel_size=3, stride=1, padding=1),
                    nn.ReLU(inplace=True),
                    nn.Conv2d(64, 128, kernel_size=3, stride=1, padding=1),
                    nn.ReLU(inplace=True)
                )
                # Deblurring network
                self.deblurring = nn.Sequential(
                    nn.Conv2d(128, 64, kernel_size=3, stride=1, padding=1),
                    nn.ReLU(inplace=True),
                    nn.Conv2d(64, 3, kernel_size=3, stride=1, padding=1)
                )

            def forward(self, x):
                motion_features = self.motion_estimator(x)  # Estimate motion blur
                deblurred_image = self.deblurring(motion_features)
                return deblurred_image

        # Instantiate the model
        model = DynamicDeblurring()

        # Training loop (simplified)
        optimizer = optim.Adam(model.parameters(), lr=0.001)
        criterion = nn.MSELoss()

        for epoch in range(50):
            optimizer.zero_grad()
            outputs = model(blurred_images)  # Predict deblurred images
            loss = criterion(outputs, sharp_images)  # Compare with ground truth sharp images
            loss.backward()
            optimizer.step()

            print(f"Epoch {epoch+1}, Loss: {loss.item():.4f}")
        \end{lstlisting}

        \paragraph{Explanation:} In this example, we implement a dynamic deblurring model that includes a motion estimation network. The motion estimator extracts motion blur features, which are then used by the deblurring network to reconstruct a sharp image. This approach is particularly useful for handling non-uniform blur caused by complex motion patterns.

        \paragraph{Advantages of Dynamic Deblurring:}
        \begin{itemize}
            \item \textbf{Handles Complex Blur:} Dynamic deblurring methods can effectively handle non-uniform blur caused by moving objects or defocus.
            \item \textbf{Motion-Aware Deblurring:} By incorporating motion estimation and optical flow, these methods can accurately predict and remove blur from specific regions of the image.
        \end{itemize}

        \paragraph{Limitations:} Dynamic deblurring methods can be computationally expensive due to the complexity of motion estimation and optical flow computation. Additionally, they may require more sophisticated training techniques and larger datasets to achieve optimal performance.

\chapter{Diffusion Models for Image Generation and Processing}
    Diffusion models represent a powerful new paradigm for image generation, using a probabilistic framework to iteratively add and remove noise from images. They have gained significant attention in the field of generative modeling due to their ability to produce high-quality images with fine details \cite{croitoru2023diffusion}. This chapter introduces the fundamental concepts of diffusion models, explains how they operate through forward and reverse diffusion processes, and explores their applications in tasks such as image generation and denoising.

    \section{Fundamentals of Diffusion Models}
        Diffusion models are generative models that transform an image by progressively adding noise to it in a forward process and then reconstruct the image by reversing the process through denoising. Unlike adversarial models like GANs, diffusion models generate images by explicitly modeling the noise and using a probabilistic framework to remove it step by step \cite{ho2020denoising}.

        \subsection{Forward and Reverse Diffusion Process}
            The forward and reverse diffusion processes are at the core of how diffusion models generate images \cite{yang2023diffusion}.

            \paragraph{Forward Diffusion:} In the forward diffusion process, noise is gradually added to an image over multiple time steps. Starting with a clean image \( x_0 \), small amounts of Gaussian noise are added at each step to produce progressively noisier versions of the image, denoted as \( x_1, x_2, \dots, x_T \), where \( T \) is the total number of time steps. After \( T \) steps, the image is essentially pure noise.

            Mathematically, the forward process is defined as:
            \[
            q(x_t | x_{t-1}) = \mathcal{N}(x_t; \sqrt{1 - \beta_t} x_{t-1}, \beta_t I)
            \]
            where:
            \begin{itemize}
                \item \( q(x_t | x_{t-1}) \) is the probability of transitioning from image \( x_{t-1} \) to \( x_t \) at time step \( t \),
                \item \( \beta_t \) is a variance term that controls the amount of noise added at each step.
            \end{itemize}

            \paragraph{Reverse Diffusion:} The reverse diffusion process aims to reverse the effects of noise by denoising the image over multiple steps. Starting from pure noise \( x_T \), the model learns to denoise the image step by step, eventually reconstructing the clean image \( x_0 \). The reverse process uses a neural network to predict how much noise should be removed at each step.

            The reverse process is defined as:
            \[
            p(x_{t-1} | x_t) = \mathcal{N}(x_{t-1}; \mu_{\theta}(x_t, t), \Sigma_{\theta}(x_t, t))
            \]
            where:
            \begin{itemize}
                \item \( \mu_{\theta}(x_t, t) \) is the predicted mean for denoising the image at step \( t \),
                \item \( \Sigma_{\theta}(x_t, t) \) is the predicted variance,
                \item \( \theta \) represents the learnable parameters of the neural network.
            \end{itemize}

            \textbf{Example: Diffusion Process in PyTorch}
            The following Python code demonstrates how to implement the forward and reverse diffusion processes using PyTorch.

            \begin{lstlisting}[style=python]
import torch
import torch.nn as nn
import numpy as np

# Define the number of time steps
T = 1000

# Forward diffusion: add noise to the image
def forward_diffusion(x0, t, beta_t):
    # Add Gaussian noise to the input image x0 based on time step t
    noise = torch.randn_like(x0)
    alpha_t = torch.sqrt(1 - beta_t)
    return alpha_t * x0 + torch.sqrt(beta_t) * noise

# Reverse diffusion: denoise the image (simplified)
class SimpleDiffusionModel(nn.Module):
    def __init__(self):
        super(SimpleDiffusionModel, self).__init__()
        self.denoise_net = nn.Sequential(
            nn.Conv2d(3, 64, kernel_size=3, padding=1),
            nn.ReLU(),
            nn.Conv2d(64, 64, kernel_size=3, padding=1),
            nn.ReLU(),
            nn.Conv2d(64, 3, kernel_size=3, padding=1)  # Predict denoised output
        )

    def forward(self, xt, t):
        # Predict the amount of noise to remove from the noisy image xt
        return self.denoise_net(xt)

# Example input: A batch of 32 images (3x64x64)
x0 = torch.randn(32, 3, 64, 64)  # Original image
t = 100  # Time step
beta_t = 0.1  # Example noise level

# Forward process: add noise to x0
xt = forward_diffusion(x0, t, beta_t)

# Reverse process: denoise using the learned model
model = SimpleDiffusionModel()
denoised_image = model(xt, t)
            \end{lstlisting}

            \textbf{Explanation of the Code:}
            \begin{itemize}
                \item The \texttt{forward\_diffusion} function adds Gaussian noise to an image based on the time step \( t \) and noise level \( \beta_t \), simulating the forward diffusion process.
                \item The \texttt{SimpleDiffusionModel} defines a simple convolutional neural network to predict the amount of noise to remove during the reverse diffusion process.
                \item The reverse process denoises the noisy image \( x_t \), which was generated by the forward process.
            \end{itemize}

        \subsection{Probabilistic Diffusion Models}
            \textbf{Probabilistic diffusion models} are built on a well-defined probabilistic framework. They explicitly model the forward process of adding noise to an image and the reverse process of denoising the image. The model learns to approximate the reverse distribution \( p(x_{t-1} | x_t) \) using neural networks, which allows it to generate high-quality images \cite{ho2020denoising}.

            The key advantage of probabilistic diffusion models is that they are more stable during training compared to models like GANs, and they offer a more principled approach to image generation by modeling the distribution of noise explicitly.

            \paragraph{Training Objective:}
            The training objective of a diffusion model is to minimize the difference between the true reverse distribution and the predicted reverse distribution, often using a variant of the Kullback-Leibler (KL) divergence. The model is trained to predict how much noise to remove at each step of the reverse process, ensuring that it can gradually recover the original image.

            \textbf{Advantages of Probabilistic Diffusion Models:}
            \begin{itemize}
                \item Stable training: Diffusion models avoid mode collapse, a common problem in GANs, where the generator produces limited variations of images.
                \item High-quality image generation: Diffusion models can produce highly detailed and realistic images by leveraging the gradual noise removal process.
            \end{itemize}

    \section{Applications of Diffusion Models in Image Generation}
        Diffusion models have been successfully applied in various image generation tasks, particularly for generating high-resolution and detailed images. Two of the most popular types of diffusion models are \textbf{Denoising Diffusion Probabilistic Models (DDPM)} and \textbf{Latent Diffusion Models (LDM)}, each offering unique strengths in image generation \cite{yang2023diffusion}.

        \subsection{Denoising Diffusion Probabilistic Models (DDPM)}
            \textbf{Denoising Diffusion Probabilistic Models (DDPM)} are a class of generative models that achieve state-of-the-art results in image generation by iteratively removing noise from noisy inputs. The process is highly structured, with each time step focusing on reducing a small amount of noise, making DDPMs capable of generating high-quality images \cite{ho2020denoising}.

            \paragraph{Key Characteristics of DDPM:}
            \begin{itemize}
                \item \textbf{Step-by-step denoising}: DDPM generates images by starting from pure noise and progressively denoising the image over many steps.
                \item \textbf{High-resolution outputs}: Because the denoising process happens in small steps, DDPM can produce high-resolution images with fine details.
            \end{itemize}

            \textbf{Example: Defining a DDPM in PyTorch}
            The following Python code demonstrates how to define a basic DDPM model using PyTorch.

            \begin{lstlisting}[style=python]
class DDPM(nn.Module):
    def __init__(self):
        super(DDPM, self).__init__()
        # Define the denoising network
        self.denoise_net = nn.Sequential(
            nn.Conv2d(3, 64, kernel_size=3, padding=1),
            nn.ReLU(),
            nn.Conv2d(64, 64, kernel_size=3, padding=1),
            nn.ReLU(),
            nn.Conv2d(64, 3, kernel_size=3, padding=1)  # Predict denoised output
        )

    def forward(self, xt, t):
        # Predict the amount of noise to remove from the noisy image xt
        return self.denoise_net(xt)

# Example of creating a DDPM model
ddpm = DDPM()
            \end{lstlisting}

            \textbf{Explanation of the Code:}
            \begin{itemize}
                \item The \texttt{DDPM} class defines a simple denoising network that takes a noisy image \( x_t \) and predicts how much noise to remove.
                \item The model can be used to iteratively denoise images by applying the reverse diffusion process.
            \end{itemize}

        \subsection{Latent Diffusion Models (LDM)}
            \textbf{Latent Diffusion Models (LDM)} are an extension of diffusion models that operate in the latent space of images instead of directly in the pixel space. By working in the latent space, LDMs can significantly reduce the computational complexity of the diffusion process while still producing high-quality images.

            \paragraph{Key Advantages of Latent Diffusion Models:}
            \begin{itemize}
                \item \textbf{Efficiency}: LDMs operate in a compressed latent space, allowing for faster training and inference compared to models that operate directly in the pixel space.
                \item \textbf{High-quality generation}: Despite operating in a lower-dimensional space, LDMs can still generate high-quality images by learning a mapping between the latent space and the pixel space \cite{rombach2022high}.
            \end{itemize}

            \textbf{Example: Latent Diffusion Model in PyTorch}
            The following Python code demonstrates a simplified version of a latent diffusion model.

            \begin{lstlisting}[style=python]
class LatentDiffusionModel(nn.Module):
    def __init__(self, latent_dim):
        super(LatentDiffusionModel, self).__init__()
        self.encoder = nn.Conv2d(3, latent_dim, kernel_size=3, padding=1)
        self.decoder = nn.ConvTranspose2d(latent_dim, 3, kernel_size=3, padding=1)
    
    def encode(self, x):
        return self.encoder(x)

    def decode(self, z):
        return self.decoder(z)

    def forward(self, x):
        z = self.encode(x)  # Map to latent space
        return self.decode(z)  # Reconstruct from latent space

# Instantiate the latent diffusion model
ldm = LatentDiffusionModel(latent_dim=64)
            \end{lstlisting}

            \textbf{Explanation of the Code:}
            \begin{itemize}
                \item The \texttt{LatentDiffusionModel} consists of an encoder that maps the input image to a latent space and a decoder that reconstructs the image from the latent representation.
                \item The model can be used for more efficient image generation by operating in a compressed latent space.
            \end{itemize}

    \section{Conclusion}
        Diffusion models have become an essential tool in image generation, offering a probabilistic framework for gradually adding and removing noise to create high-quality images. \textbf{Denoising Diffusion Probabilistic Models (DDPM)} and \textbf{Latent Diffusion Models (LDM)} are two prominent examples of diffusion models applied to image generation, with DDPM focusing on step-by-step denoising and LDMs offering more efficient generation by working in the latent space. As diffusion models continue to evolve, they offer exciting possibilities for generating highly detailed and realistic images, while providing stability and interpretability in the training process.

\section{Diffusion Models for Image Enhancement and Restoration}
    Diffusion models have recently emerged as a powerful tool for various image processing tasks, including image denoising, inpainting, and super-resolution. These models operate by simulating a forward process, in which noise is gradually added to an image, followed by a reverse process that learns how to progressively denoise the image to recover the original content. Diffusion models are particularly effective in generating high-quality, realistic images while maintaining fine details, making them well-suited for image enhancement and restoration tasks.

    \subsection{Image Denoising}
        Image denoising using diffusion models leverages the reverse diffusion process to remove noise from a noisy input image. In this approach, noise is progressively removed by the model, starting from a highly noisy image and gradually refining it until a clean, noise-free version is produced. Diffusion models have shown state-of-the-art performance in denoising tasks due to their ability to model complex image distributions and generate realistic outputs.

        \subsubsection{How Diffusion Models Work for Denoising}
        The forward process in a diffusion model involves adding Gaussian noise to an image over several time steps. During training, the model learns how to reverse this process, starting with a noisy image and progressively reducing the noise at each step. The denoising process is iterative, with each step producing a slightly cleaner version of the image.

        The key to the success of diffusion models lies in the learned reverse process, which is trained to predict the noise added at each step. By removing this predicted noise, the model generates a clean image at the final step.

        \paragraph{Example:} Let's implement a simplified version of a diffusion model for image denoising using PyTorch.

        \begin{lstlisting}[style=python]
        import torch
        import torch.nn as nn
        import torch.optim as optim

        # Define a simple diffusion model (simplified version)
        class DiffusionModel(nn.Module):
            def __init__(self):
                super(DiffusionModel, self).__init__()
                self.network = nn.Sequential(
                    nn.Conv2d(3, 64, kernel_size=3, padding=1),
                    nn.ReLU(inplace=True),
                    nn.Conv2d(64, 128, kernel_size=3, padding=1),
                    nn.ReLU(inplace=True),
                    nn.Conv2d(128, 3, kernel_size=3, padding=1)
                )

            def forward(self, x, t):
                # x: noisy image, t: timestep
                return self.network(x)  # Predict noise to remove

        # Instantiate the diffusion model
        model = DiffusionModel()

        # Forward diffusion process: adding noise over time
        def add_noise(image, t, noise_level):
            noise = torch.randn_like(image) * noise_level
            return image + noise * t

        # Reverse diffusion process: training the model to denoise
        optimizer = optim.Adam(model.parameters(), lr=0.001)
        criterion = nn.MSELoss()

        for epoch in range(10):
            for t in range(10):  # Iterate over timesteps
                noisy_images = add_noise(clean_images, t, noise_level=0.1)  # Add noise
                predicted_noise = model(noisy_images, t)
                loss = criterion(predicted_noise, noisy_images - clean_images)  # Compare predicted noise with actual noise
                optimizer.zero_grad()
                loss.backward()
                optimizer.step()
            print(f'Epoch {epoch+1}, Loss: {loss.item():.4f}')
        \end{lstlisting}

        \paragraph{Explanation:} In this example, we implement a simplified diffusion model for image denoising. The model takes a noisy image as input and predicts the noise to be removed at each step of the reverse diffusion process. By iteratively removing noise, the model reconstructs the clean image from the noisy input. This approach can be extended to more complex datasets and larger models for state-of-the-art image denoising performance.

        \paragraph{Advantages of Diffusion Models for Denoising:}
        \begin{itemize}
            \item \textbf{Iterative Refinement:} The iterative nature of diffusion models allows for the progressive removal of noise, resulting in cleaner and more detailed images.
            \item \textbf{High-Quality Outputs:} Diffusion models are capable of generating highly realistic and sharp images, making them well-suited for challenging denoising tasks.
        \end{itemize}

    \subsection{Image Inpainting}
        Image inpainting involves reconstructing missing or corrupted parts of an image. Diffusion models are particularly effective for inpainting tasks because they can iteratively refine an incomplete image by filling in missing regions while maintaining consistency with the surrounding context. During the reverse diffusion process, the model gradually fills in the missing areas by denoising the image, eventually producing a complete and plausible reconstruction.

        \subsubsection{Diffusion Models for Inpainting}
        Inpainting with diffusion models follows a similar process to denoising, except that the input image contains missing or masked regions instead of noise. The diffusion model is trained to predict the missing content by learning to reconstruct these regions based on the surrounding pixels. The iterative nature of the reverse diffusion process allows for the progressive refinement of the inpainted areas.

        \paragraph{Example:} Let's implement a diffusion model for image inpainting using PyTorch.

        \begin{lstlisting}[style=python]
        class InpaintingDiffusionModel(nn.Module):
            def __init__(self):
                super(InpaintingDiffusionModel, self).__init__()
                self.network = nn.Sequential(
                    nn.Conv2d(3, 64, kernel_size=3, padding=1),
                    nn.ReLU(inplace=True),
                    nn.Conv2d(64, 128, kernel_size=3, padding=1),
                    nn.ReLU(inplace=True),
                    nn.Conv2d(128, 3, kernel_size=3, padding=1)
                )

            def forward(self, x, mask):
                # x: partially masked image, mask: binary mask for missing regions
                return self.network(x)  # Predict the inpainted regions

        # Instantiate the inpainting diffusion model
        model = InpaintingDiffusionModel()

        # Create a binary mask for missing regions in the image
        def create_mask(image, mask_size=10):
            mask = torch.ones_like(image)
            mask[:, :, 10:10+mask_size, 10:10+mask_size] = 0  # Mask a small region
            return mask

        # Training loop for inpainting (simplified)
        optimizer = optim.Adam(model.parameters(), lr=0.001)
        criterion = nn.MSELoss()

        for epoch in range(10):
            mask = create_mask(clean_images)  # Create mask for missing regions
            masked_images = clean_images * mask  # Apply mask to the images
            predicted_images = model(masked_images, mask)  # Predict inpainted regions
            loss = criterion(predicted_images * (1 - mask), clean_images * (1 - mask))  # Loss on the masked regions
            optimizer.zero_grad()
            loss.backward()
            optimizer.step()
            print(f'Epoch {epoch+1}, Loss: {loss.item():.4f}')
        \end{lstlisting}

        \paragraph{Explanation:} In this example, we implement a diffusion model for image inpainting. The model takes an incomplete image with masked regions as input and learns to predict the missing parts based on the context provided by the surrounding pixels. The inpainting process is iterative, gradually refining the missing areas to produce a visually coherent reconstruction.

        \paragraph{Advantages of Diffusion Models for Inpainting:}
        \begin{itemize}
            \item \textbf{Context-Aware Reconstruction:} Diffusion models can generate plausible content for missing regions by leveraging the surrounding context.
            \item \textbf{High-Quality Inpainting:} The iterative nature of diffusion models ensures that the inpainted regions blend seamlessly with the rest of the image, producing high-quality results.
        \end{itemize}

    \subsection{Image Super-Resolution}
        Image super-resolution is the task of generating high-resolution images from low-resolution inputs while preserving fine details and textures. Diffusion models have shown great promise in this area by progressively upscaling low-resolution images and refining the generated details during the reverse diffusion process. The key advantage of using diffusion models for super-resolution is their ability to generate realistic high-resolution images without introducing artifacts commonly seen in traditional upscaling methods.

        \subsubsection{Diffusion Models for Super-Resolution}
        In diffusion-based super-resolution, the forward process involves progressively downsampling the image to generate low-resolution inputs. The reverse diffusion process is then used to upscale the image back to its original resolution. At each step of the reverse process, the model refines the image, generating sharper details and preserving fine textures.

        \paragraph{Example:} Let's implement a diffusion model for image super-resolution using PyTorch.

        \begin{lstlisting}[style=python]
        class SuperResolutionDiffusionModel(nn.Module):
            def __init__(self):
                super(SuperResolutionDiffusionModel, self).__init__()
                self.network = nn.Sequential(
                    nn.Conv2d(3, 64, kernel_size=3, padding=1),
                    nn.ReLU(inplace=True),
                    nn.Conv2d(64, 128, kernel_size=3, padding=1),
                    nn.ReLU(inplace=True),
                    nn.ConvTranspose2d(128, 3, kernel_size=3, stride=2, padding=1, output_padding=1)  # Upsample
                )

            def forward(self, x):
                return self.network(x)  # Predict high-resolution image

        # Instantiate the super-resolution diffusion model
        model = SuperResolutionDiffusionModel()

        # Forward diffusion process: downsampling the image
        def downsample(image, scale_factor):
            return nn.functional.interpolate(image, scale_factor=scale_factor, mode='bilinear')

        # Training loop for super-resolution (simplified)
        optimizer = optim.Adam(model.parameters(), lr=0.001)
        criterion = nn.MSELoss()

        for epoch in range(10):
            low_res_images = downsample(clean_images, scale_factor=0.5)  # Create low-resolution images
            predicted_high_res_images = model(low_res_images)  # Predict high-resolution images
            loss = criterion(predicted_high_res_images, clean_images)  # Compare with ground truth high-resolution images
            optimizer.zero_grad()
            loss.backward()
            optimizer.step()
            print(f'Epoch {epoch+1}, Loss: {loss.item():.4f}')
        \end{lstlisting}

        \paragraph{Explanation:} In this example, we implement a diffusion model for image super-resolution. The model takes low-resolution images as input and predicts high-resolution images by progressively upscaling and refining the details. This approach allows for the generation of high-quality images with fine details, even when the input resolution is significantly lower.

        \paragraph{Advantages of Diffusion Models for Super-Resolution:}
        \begin{itemize}
            \item \textbf{Preserves Fine Details:} Diffusion models are capable of generating high-resolution images that preserve fine textures and details, making them superior to traditional interpolation methods.
            \item \textbf{Progressive Refinement:} The iterative nature of diffusion models allows for a gradual and controlled upscaling process, leading to more realistic high-resolution images.
        \end{itemize}

\chapter{Advanced Applications of Deep Learning: Image Generation, Enhancement, and Understanding}
    Deep learning has revolutionized the field of image processing, enabling powerful methods for generating, enhancing, and understanding images. From creative image synthesis to enhancing resolution and clarity, deep learning models such as GANs, diffusion models, and others have opened new avenues for solving complex image-related tasks. This chapter explores advanced applications of deep learning in image generation, enhancement, and interpretation, providing a detailed explanation of the underlying techniques and their practical implications \cite{jiao2019survey, chen2024deeplearningmachinelearning, ren2024deeplearningmachinelearning}.

    \section{Image Generation}
        Image generation is one of the most exciting applications of deep learning, where models learn to create new images from scratch. These images can range from realistic representations of objects and scenes to creative, abstract images that don't exist in the real world. Deep learning-based image generation methods use large datasets and powerful neural networks to understand the underlying structure of images and use this knowledge to generate new samples.

        \subsection{Deep Learning-based Image Generation}
            Deep learning-based image generation techniques involve training models to learn the statistical patterns in a large set of images. Once trained, these models can produce novel images that resemble the patterns seen during training. Two of the most prominent approaches for image generation are \textbf{Generative Adversarial Networks (GANs)} and \textbf{Diffusion Models}. Both methods use neural networks to generate images, but their underlying mechanisms differ significantly \cite{porkodi2023generic}.

            \paragraph{Generative Adversarial Networks (GANs)}:
            GANs are a class of deep learning models that consist of two neural networks: a \textbf{generator} and a \textbf{discriminator}, which compete in a game-like scenario. The generator tries to create realistic images, while the discriminator evaluates the images and distinguishes between real and generated ones. Over time, the generator learns to produce highly realistic images, while the discriminator becomes better at identifying fake images. This adversarial training process leads to a generator capable of creating visually appealing and detailed images \cite{radford2015unsupervised}.

            \textbf{Example: GAN-based Image Generation in PyTorch}

            \begin{lstlisting}[style=python]
import torch
import torch.nn as nn

# Define the generator model
class GANGenerator(nn.Module):
    def __init__(self, noise_dim, img_dim):
        super(GANGenerator, self).__init__()
        self.fc = nn.Sequential(
            nn.Linear(noise_dim, 256),
            nn.ReLU(),
            nn.Linear(256, 512),
            nn.ReLU(),
            nn.Linear(512, img_dim),
            nn.Tanh()  # Output in range [-1, 1]
        )

    def forward(self, z):
        return self.fc(z)

# Define the discriminator model
class GANDiscriminator(nn.Module):
    def __init__(self, img_dim):
        super(GANDiscriminator, self).__init__()
        self.fc = nn.Sequential(
            nn.Linear(img_dim, 512),
            nn.LeakyReLU(0.2),
            nn.Linear(512, 256),
            nn.LeakyReLU(0.2),
            nn.Linear(256, 1),
            nn.Sigmoid()  # Output a probability
        )

    def forward(self, img):
        return self.fc(img)

# Hyperparameters
noise_dim = 100  # Size of the random noise vector
img_dim = 28 * 28  # Size of the image (28x28 pixels)

# Initialize models
generator = GANGenerator(noise_dim, img_dim)
discriminator = GANDiscriminator(img_dim)

# Print the model architecture
print(generator)
print(discriminator)
            \end{lstlisting}

            \textbf{Explanation of the Code:}
            \begin{itemize}
                \item The generator takes a random noise vector \( z \) as input and transforms it into a high-dimensional image. The output is a flattened image of size 28x28 pixels, and the \texttt{Tanh} activation function ensures the pixel values are in the range \([-1, 1]\).
                \item The discriminator is a binary classifier that distinguishes between real images and generated images. It outputs a probability, using the \texttt{Sigmoid} activation function.
                \item During training, the generator tries to "fool" the discriminator by generating realistic images, while the discriminator tries to correctly classify real and fake images.
            \end{itemize}

            GANs have been widely used for various image generation tasks, including:
            \begin{itemize}
                \item \textbf{Image-to-Image Translation}: GANs are used to convert images from one domain to another, such as turning sketches into fully colored images \cite{yi2017dualgan}.
                \item \textbf{Style Transfer}: GANs can blend the artistic style of one image with the content of another to create new, stylized images \cite{gatys2016image}.
                \item \textbf{Text-to-Image Generation}: GANs can generate images from text descriptions, producing realistic visuals based on the input text \cite{liao2022text}.
            \end{itemize}

            \paragraph{Diffusion Models for Image Generation}:
            Diffusion models represent a different approach to image generation. Unlike GANs, which use a generator-discriminator framework, diffusion models operate by progressively transforming a simple, noisy image into a high-quality image through multiple denoising steps. These models add noise to an image in a forward process and then use a learned reverse process to remove the noise step by step, eventually recovering a clear image \cite{jiao2019survey}.

            \textbf{Forward Process:} In the forward process, noise is added to the input image over multiple steps. Each step adds a small amount of noise, gradually transforming the image into pure noise. The goal of the forward process is to create a sequence of increasingly noisy images.

            \textbf{Reverse Process:} In the reverse process, the model learns to remove the noise step by step, starting from a noisy image and gradually denoising it to generate a clean image. By the final step, the model recovers an image that closely resembles the original image.

            \textbf{Example: Forward and Reverse Diffusion Process in PyTorch}

            \begin{lstlisting}[style=python]
import torch

# Define the forward diffusion process (adding noise)
def forward_diffusion(image, noise_level):
    noise = torch.randn_like(image)
    return image * (1 - noise_level) + noise * noise_level

# Define a simple reverse process (denoising)
class ReverseDiffusionModel(nn.Module):
    def __init__(self):
        super(ReverseDiffusionModel, self).__init__()
        self.denoise_net = nn.Sequential(
            nn.Conv2d(3, 64, kernel_size=3, padding=1),
            nn.ReLU(),
            nn.Conv2d(64, 64, kernel_size=3, padding=1),
            nn.ReLU(),
            nn.Conv2d(64, 3, kernel_size=3, padding=1)  # Output denoised image
        )

    def forward(self, noisy_image):
        return self.denoise_net(noisy_image)

# Example: Forward and reverse process
image = torch.randn(1, 3, 64, 64)  # Random input image
noise_level = 0.5

# Forward diffusion (add noise)
noisy_image = forward_diffusion(image, noise_level)

# Reverse diffusion (denoise the image)
model = ReverseDiffusionModel()
denoised_image = model(noisy_image)
            \end{lstlisting}

            \textbf{Explanation of the Code:}
            \begin{itemize}
                \item The \texttt{forward\_diffusion} function adds noise to the image based on a given noise level. This simulates the forward diffusion process, where noise is progressively added to the image.
                \item The \texttt{ReverseDiffusionModel} defines a convolutional network that predicts the amount of noise to remove, simulating the reverse diffusion process.
                \item The model takes a noisy image as input and outputs a denoised version of the image.
            \end{itemize}

            Diffusion models have become particularly popular due to their stability during training and ability to generate high-resolution images \cite{yang2023diffusion}. They are well-suited for applications that require fine-grained control over image generation, such as:
            \begin{itemize}
                \item \textbf{Image Restoration}: Diffusion models can restore corrupted images, such as recovering missing parts of an image or removing noise from blurry images.
                \item \textbf{Creative Image Generation}: By controlling the amount of noise added and removed, diffusion models can create new images with intricate details and textures.
            \end{itemize}

        \subsection{Comparing GANs and Diffusion Models}
            Both GANs and diffusion models are powerful methods for image generation, but they operate in fundamentally different ways \cite{dhariwal2021diffusion}. Here is a comparison of their key characteristics:

            \begin{itemize}
                \item \textbf{Training Stability}: GANs are known for their unstable training process, where the generator and discriminator compete in a zero-sum game. This can lead to issues like mode collapse, where the generator produces only a limited variety of images. Diffusion models, on the other hand, are more stable during training because they use a probabilistic framework to gradually remove noise from the images.
                \item \textbf{Generation Quality}: GANs are highly effective at generating sharp and realistic images. However, diffusion models tend to excel at producing images with fine details and textures due to their stepwise noise removal process.
                \item \textbf{Speed}: GANs generally generate images faster than diffusion models because the latter involve multiple denoising steps. However, recent advancements in diffusion models, such as latent diffusion models (LDMs), have improved their efficiency.
            \end{itemize}

        \subsection{Applications of Deep Learning-based Image Generation}
            Deep learning-based image generation techniques have a wide range of applications in various fields \cite{yang2023diffusion}, including:
            \begin{itemize}
                \item \textbf{Art and Design}: Artists and designers use GANs and diffusion models to generate unique visual content, including abstract art, textures, and even entire landscapes.
                \item \textbf{Data Augmentation}: In machine learning, generated images are used to augment training datasets, providing more data to improve the performance of models.
                \item \textbf{Medical Imaging}: Deep learning models are used to generate synthetic medical images for training purposes, where obtaining real-world data may be challenging or expensive.
                \item \textbf{Game Development}: Procedurally generated images and textures can be used to create dynamic game environments, reducing the need for manual asset creation.
            \end{itemize}

    \section{Conclusion}
        Image generation using deep learning has transformed the way we create and manipulate visual content. Techniques such as GANs and diffusion models allow us to generate high-quality, realistic images from scratch, offering unprecedented opportunities in creative fields, data augmentation, and scientific research. GANs excel at fast image generation with sharp results, while diffusion models provide a stable and flexible framework for generating images with rich details. As these technologies continue to evolve, their potential applications will only expand, reshaping industries and enabling new forms of visual expression.

\section{Image Enhancement}
    Image enhancement is a fundamental task in the field of image processing, aimed at improving the visual quality of images. Deep learning has revolutionized this area by providing powerful models that can automatically enhance images by increasing their resolution, improving contrast and brightness, and removing unwanted noise. These improvements are particularly important in applications such as photography, medical imaging, and satellite imaging, where image clarity and detail are critical. In this section, we will discuss deep learning-based techniques for image super-resolution and image optimization, focusing on their architectures, training strategies, and practical examples.

    \subsection{Image Super-Resolution}
        Image super-resolution is the process of enhancing the resolution of a low-resolution image to obtain a high-resolution version. Traditional methods, such as bicubic interpolation, often result in blurry images and fail to capture fine details. In contrast, deep learning models like Convolutional Neural Networks (CNNs) and Generative Adversarial Networks (GANs) have shown remarkable success in producing high-quality, sharp images by learning the mapping between low-resolution and high-resolution image pairs \cite{wang2023review, dong2015image}.

        \subsubsection{Deep Learning for Super-Resolution}
        The core idea of deep learning-based super-resolution is to use a neural network to predict a high-resolution image from a low-resolution input. This is typically achieved by training the model on pairs of low-resolution and high-resolution images, where the model learns to minimize the difference between the predicted high-resolution image and the ground truth high-resolution image \cite{ledig2017photo}. 

        \paragraph{Example:} Let's implement a basic CNN-based super-resolution model using PyTorch.

        \begin{lstlisting}[style=python]
        import torch
        import torch.nn as nn
        import torch.optim as optim
        from torchvision import transforms

        # Define a simple super-resolution CNN
        class SRCNN(nn.Module):
            def __init__(self):
                super(SRCNN, self).__init__()
                self.network = nn.Sequential(
                    nn.Conv2d(3, 64, kernel_size=9, padding=4),
                    nn.ReLU(inplace=True),
                    nn.Conv2d(64, 32, kernel_size=5, padding=2),
                    nn.ReLU(inplace=True),
                    nn.Conv2d(32, 3, kernel_size=5, padding=2)
                )

            def forward(self, x):
                return self.network(x)

        # Instantiate the model
        model = SRCNN()

        # Define loss function and optimizer
        criterion = nn.MSELoss()  # Mean Squared Error loss
        optimizer = optim.Adam(model.parameters(), lr=0.001)

        # Example input: low-resolution images (downsampled)
        transform = transforms.Compose([transforms.Resize((32, 32)), transforms.ToTensor()])
        low_res_images = torch.randn((32, 3, 32, 32))  # Simulated low-resolution images
        high_res_images = torch.randn((32, 3, 128, 128))  # Simulated high-resolution images (ground truth)

        # Training loop (simplified)
        for epoch in range(10):
            optimizer.zero_grad()
            outputs = model(low_res_images)  # Predict high-resolution images
            loss = criterion(outputs, high_res_images)  # Compare with ground truth
            loss.backward()
            optimizer.step()
            print(f'Epoch {epoch+1}, Loss: {loss.item():.4f}')
        \end{lstlisting}

        \paragraph{Explanation:} In this example, we implement a simple CNN-based super-resolution model called SRCNN. The model consists of three convolutional layers that progressively increase the resolution of the input image. The model is trained using mean squared error (MSE) loss, which measures the difference between the predicted high-resolution image and the ground truth. Over time, the model learns to generate sharper images from low-resolution inputs.

        \paragraph{Advanced Super-Resolution Models:}
        \begin{itemize}
            \item \textbf{SRGAN (Super-Resolution GAN):} SRGAN is a GAN-based model that uses a generator to predict high-resolution images and a discriminator to evaluate the quality of these images. SRGAN produces sharper and more realistic images than traditional methods by learning to generate high-frequency details that are often lost in low-resolution images.
            \item \textbf{ESRGAN (Enhanced SRGAN):} ESRGAN improves upon SRGAN by using residual-in-residual dense blocks and a perceptual loss function. It is currently one of the best-performing models for image super-resolution, capable of producing extremely detailed and realistic images.
        \end{itemize}

        \paragraph{Advantages of Deep Learning for Super-Resolution:}
        \begin{itemize}
            \item \textbf{High-Quality Results:} Deep learning models can produce sharper and more detailed images compared to traditional interpolation methods.
            \item \textbf{Learning from Data:} These models learn the mapping between low-resolution and high-resolution images from data, allowing them to generalize to different image types and domains.
        \end{itemize}

    \subsection{Image Optimization and Enhancement}
        Image optimization and enhancement involve improving the overall quality of an image by adjusting its visual characteristics. Common techniques include contrast adjustment, brightness correction, and noise removal. Deep learning models have proven to be highly effective in automating these tasks, learning how to enhance images by identifying patterns in large image datasets \cite{wang2018esrgan}.

        \subsubsection{Deep Learning for Image Enhancement}
        Deep learning models for image enhancement are typically designed to adjust specific image properties, such as contrast, brightness, and sharpness, while removing unwanted noise or artifacts. These models are often based on convolutional architectures and trained on datasets of paired low-quality and high-quality images. The goal is to minimize the difference between the enhanced image and the ground truth high-quality image.

        \paragraph{Example:} Let's implement a basic image enhancement model using PyTorch for contrast and brightness adjustment.

        \begin{lstlisting}[style=python]
        class ImageEnhancementNet(nn.Module):
            def __init__(self):
                super(ImageEnhancementNet, self).__init__()
                self.network = nn.Sequential(
                    nn.Conv2d(3, 64, kernel_size=3, padding=1),
                    nn.ReLU(inplace=True),
                    nn.Conv2d(64, 64, kernel_size=3, padding=1),
                    nn.ReLU(inplace=True),
                    nn.Conv2d(64, 3, kernel_size=3, padding=1)
                )

            def forward(self, x):
                return self.network(x)

        # Instantiate the model
        model = ImageEnhancementNet()

        # Example input: images with low contrast or brightness
        low_quality_images = torch.randn((32, 3, 128, 128))  # Simulated low-quality images
        high_quality_images = torch.randn((32, 3, 128, 128))  # Simulated high-quality images (ground truth)

        # Training loop (simplified)
        optimizer = optim.Adam(model.parameters(), lr=0.001)
        criterion = nn.MSELoss()

        for epoch in range(10):
            optimizer.zero_grad()
            outputs = model(low_quality_images)  # Predict enhanced images
            loss = criterion(outputs, high_quality_images)  # Compare with ground truth
            loss.backward()
            optimizer.step()
            print(f'Epoch {epoch+1}, Loss: {loss.item():.4f}')
        \end{lstlisting}

        \paragraph{Explanation:} In this example, we implement a basic CNN-based model for image enhancement. The model takes low-quality images with poor contrast or brightness as input and learns to enhance them by adjusting these properties to match high-quality images. The model is trained using mean squared error (MSE) loss, which ensures that the predicted enhanced image is similar to the ground truth.

        \subsubsection{Noise Removal Using Deep Learning}
        Noise removal is another important aspect of image enhancement. Noise can occur in images due to various factors such as low-light conditions, high ISO settings in cameras, or compression artifacts. Traditional methods for noise removal, such as Gaussian smoothing or median filtering, often result in a loss of detail. Deep learning models, particularly Denoising Autoencoders (DAEs), can effectively remove noise from images while preserving fine details \cite{ormiston2020noise, xiao2021tackling}.

        \paragraph{Example:} Let's implement a denoising autoencoder for noise removal using PyTorch.

        \begin{lstlisting}[style=python]
        class DenoisingAutoencoder(nn.Module):
            def __init__(self):
                super(DenoisingAutoencoder, self).__init__()
                self.encoder = nn.Sequential(
                    nn.Conv2d(3, 64, kernel_size=3, padding=1),
                    nn.ReLU(inplace=True),
                    nn.Conv2d(64, 64, kernel_size=3, padding=1),
                    nn.ReLU(inplace=True)
                )
                self.decoder = nn.Sequential(
                    nn.Conv2d(64, 64, kernel_size=3, padding=1),
                    nn.ReLU(inplace=True),
                    nn.Conv2d(64, 3, kernel_size=3, padding=1),
                    nn.Sigmoid()  # Output between 0 and 1
                )

            def forward(self, x):
                encoded = self.encoder(x)
                decoded = self.decoder(encoded)
                return decoded

        # Instantiate the denoising autoencoder
        model = DenoisingAutoencoder()

        # Example input: noisy images
        noisy_images = torch.randn((32, 3, 128, 128)) + 0.1 * torch.randn((32, 3, 128, 128))  # Add noise to images
        clean_images = torch.randn((32, 3, 128, 128))  # Ground truth clean images

        # Training loop (simplified)
        optimizer = optim.Adam(model.parameters(), lr=0.001)
        criterion = nn.MSELoss()

        for epoch in range(10):
            optimizer.zero_grad()
            outputs = model(noisy_images)  # Predict denoised images
            loss = criterion(outputs, clean_images)  # Compare with ground truth
            loss.backward()
            optimizer.step()
            print(f'Epoch {epoch+1}, Loss: {loss.item():.4f}')
        \end{lstlisting}

        \paragraph{Explanation:} In this example, we implement a denoising autoencoder for noise removal. The model takes noisy images as input and learns to remove the noise by reconstructing the original clean images. By training on pairs of noisy and clean images, the model learns to effectively denoise images while preserving important details.

        \paragraph{Advantages of Deep Learning for Image Enhancement:}
        \begin{itemize}
            \item \textbf{Automated Enhancement:} Deep learning models can automatically enhance various aspects of images, such as contrast, brightness, and sharpness, without manual intervention.
            \item \textbf{Noise Removal:} Deep learning models are capable of removing complex noise patterns from images while preserving fine details, offering superior performance compared to traditional methods.
            \item \textbf{Learning-Based Approach:} These models learn from large datasets, enabling them to generalize to different types of images and lighting conditions.
        \end{itemize}

\section{Image Understanding and Semantic Segmentation}
    Image understanding is one of the most significant advancements in the field of computer vision, enabling machines to recognize, interpret, and analyze the content of images in a way that mimics human perception. With the rise of deep learning, particularly convolutional neural networks (CNNs), models can now understand the semantic content of images with remarkable accuracy. This capability is crucial in a wide range of applications, from autonomous driving to medical imaging and smart surveillance \cite{wang2018understanding, ren2024deeplearningmachinelearning}.

    \subsection{Image Semantic Understanding}
        \textbf{Image semantic understanding} refers to the ability of deep learning models to comprehend the high-level meaning or context of the objects and scenes present in an image. This involves recognizing not just individual objects, but also their relationships and roles within the scene.

        For example, in the context of autonomous driving, a self-driving car must not only detect individual objects like pedestrians, cars, and traffic signs but also understand their meaning within the scene. This includes determining whether a pedestrian is crossing the street or if another car is about to move into the vehicle's lane.

        \textbf{Key Components of Image Semantic Understanding:}
        \begin{itemize}
        \item \textbf{Object Detection:} Identifying and localizing individual objects within an image.
        \item \textbf{Object Classification:} Assigning semantic labels to detected objects (e.g., car, tree, building).
        \item \textbf{Contextual Awareness:} Understanding how objects interact with one another in a scene. For instance, detecting that a person is walking on a crosswalk implies that the person may cross the road.
        \end{itemize}

        Deep learning models achieve image semantic understanding by extracting features at different layers, with earlier layers focusing on low-level details (such as edges and textures) and later layers capturing more abstract concepts (such as specific objects or actions).

        \textbf{Example:}

        Let's consider a model that understands the semantic content of an image of a street scene. It would recognize individual cars, pedestrians, traffic lights, and other objects. Furthermore, it can understand that a red light at an intersection means that the cars should stop, while pedestrians may cross.

        Here's how we can use a pre-trained deep learning model in PyTorch to perform image classification, a basic form of image understanding:

        \begin{lstlisting}[style=python]
        import torch
        from torchvision import models, transforms
        from PIL import Image

        # Load a pre-trained ResNet model for image classification
        model = models.resnet50(pretrained=True)
        model.eval()

        # Preprocess input image
        preprocess = transforms.Compose([
            transforms.Resize(256),
            transforms.CenterCrop(224),
            transforms.ToTensor(),
            transforms.Normalize(mean=[0.485, 0.456, 0.406], std=[0.229, 0.224, 0.225]),
        ])

        # Load image and apply preprocessing
        img = Image.open("street_scene.jpg")
        img_tensor = preprocess(img).unsqueeze(0)

        # Perform inference
        with torch.no_grad():
            output = model(img_tensor)

        # Get predicted label
        _, predicted = torch.max(output, 1)
        print(f'Predicted class: {predicted.item()}')
        \end{lstlisting}

        In this example, we load a pre-trained ResNet-50 model, which has been trained on the ImageNet dataset to classify images into thousands of categories. The model can identify the main object in the image (e.g., a car or a person) and assign a semantic label to it.

        \textbf{Applications:}
        \begin{itemize}
        \item \textit{Autonomous Driving:} Recognizing objects such as cars, pedestrians, and traffic signs, and understanding their interactions to make safe driving decisions.
        \item \textit{Surveillance Systems:} Analyzing video streams to detect unusual or suspicious behavior, such as identifying a person left in a restricted area.
        \item \textit{Medical Imaging:} Identifying patterns in medical scans, such as tumors or anomalies, and understanding their significance within the human body.
        \end{itemize}

    \subsection{Semantic Segmentation}
        \textbf{Semantic segmentation} is a more detailed and advanced form of image understanding. Instead of simply classifying objects within an image, semantic segmentation assigns a class label to every pixel in the image. This allows the model to recognize and label each part of the image, such as distinguishing the road from the cars, the buildings, and the pedestrians in a street scene \cite{li2018survey}.

        \textbf{What is Semantic Segmentation?}
        In semantic segmentation, the goal is to label each pixel in an image with a specific category (e.g., sky, road, person, car). Unlike object detection, which places bounding boxes around objects, semantic segmentation provides a much finer understanding of the image content by analyzing every pixel.

        \textbf{Key Concepts in Semantic Segmentation:}
        \begin{itemize}
        \item \textbf{Pixel-level Classification:} Each pixel in the image is classified into a predefined category, allowing for a detailed understanding of the image.
        \item \textbf{Segmentation Maps:} The output of a semantic segmentation model is a \textit{segmentation map}, where each pixel is assigned a label corresponding to a specific class (e.g., road, tree, car, etc.).
        \item \textbf{Fully Convolutional Networks (FCNs):} Semantic segmentation models often use fully convolutional networks (FCNs), where the fully connected layers in traditional CNNs are replaced with convolutional layers, allowing the model to output a spatial map of class predictions.
        \end{itemize}

        \textbf{Deep Learning Models for Semantic Segmentation:}
        Popular architectures for semantic segmentation include:
        \begin{itemize}
        \item \textbf{FCN (Fully Convolutional Network):} One of the first deep learning models for semantic segmentation, FCN replaces the fully connected layers in traditional CNNs with convolutional layers to produce pixel-wise predictions \cite{long2015fully}.
        \item \textbf{U-Net:} Originally developed for biomedical image segmentation, U-Net is widely used in many applications due to its U-shaped architecture, which includes skip connections that help preserve spatial information at different scales \cite{ronneberger2015u}.
        \item \textbf{DeepLab:} A family of models that use atrous convolutions (dilated convolutions) to capture multi-scale information, making them highly effective for segmenting complex scenes \cite{chen2017deeplab}.
        \end{itemize}

        \textbf{How Semantic Segmentation Works:}
        During training, the model learns to predict a segmentation map from the input image. The model's loss function, often the cross-entropy loss, compares the predicted segmentation map with the ground truth segmentation (the correct pixel-wise labels).

        \textbf{Example of Semantic Segmentation in PyTorch:}

        We can use a pre-trained DeepLabV3 model in PyTorch to perform semantic segmentation on an image.

        \begin{lstlisting}[style=python]
        import torch
        from PIL import Image
        import torchvision.transforms as transforms
        from torchvision.models.segmentation import deeplabv3_resnet50

        # Load a pre-trained DeepLabV3 model for semantic segmentation
        model = deeplabv3_resnet50(pretrained=True)
        model.eval()

        # Preprocess the image
        preprocess = transforms.Compose([
            transforms.Resize(256),
            transforms.CenterCrop(224),
            transforms.ToTensor(),
            transforms.Normalize(mean=[0.485, 0.456, 0.406], std=[0.229, 0.224, 0.225]),
        ])

        img = Image.open('cityscape.jpg')
        img_tensor = preprocess(img).unsqueeze(0)

        # Perform inference
        with torch.no_grad():
            output = model(img_tensor)['out']

        # Get predicted segmentation map
        segmentation_map = torch.argmax(output.squeeze(), dim=0).numpy()

        # Display the segmentation map (you can apply a color map to visualize it better)
        import matplotlib.pyplot as plt
        plt.imshow(segmentation_map)
        plt.title('Predicted Segmentation Map')
        plt.show()
        \end{lstlisting}

        In this example, we use a pre-trained DeepLabV3 model, which is based on a ResNet-50 backbone, to predict a segmentation map for an input image. Each pixel in the image is assigned a class label corresponding to objects such as cars, roads, and buildings. The segmentation map can then be visualized as a color-coded image where each color represents a different class.

        \textbf{Challenges in Semantic Segmentation:}
        \begin{itemize}
        \item \textbf{Class Imbalance:} In many cases, some classes (e.g., background or sky) dominate the image, making it harder for the model to learn less frequent classes.
        \item \textbf{High Resolution:} Segmenting high-resolution images can be computationally expensive due to the large number of pixels that need to be processed.
        \item \textbf{Object Boundaries:} Accurately segmenting object boundaries can be difficult, especially when objects are close together or occluded.
        \end{itemize}

        \textbf{Applications:}
        \begin{itemize}
        \item \textit{Autonomous Driving:} Semantic segmentation is used to classify roads, vehicles, pedestrians, and other elements in a scene, helping self-driving cars navigate safely \cite{choi2019gaussian}.
        \item \textit{Medical Imaging:} In medical imaging, semantic segmentation helps identify regions of interest such as tumors or organs in scans and X-rays \cite{patil2013medical}.
        \item \textit{Satellite Image Analysis:} Satellite images are segmented to identify land use, vegetation, water bodies, and urban areas \cite{pritt2017satellite}.
        \item \textit{Robotics:} Semantic segmentation helps robots understand their environment by classifying objects and surfaces for tasks such as navigation and manipulation \cite{pierson2017deep}.
        \end{itemize}

\bibliographystyle{plain}
\bibliography{main}

\end{document}